%% file: Main.tex
\documentclass[preprint,12pt,authoryear]{elsarticle}
\usepackage{amssymb}
\usepackage{hyperref} 
\usepackage{lineno}
\usepackage{soul}
\usepackage{setspace}
\usepackage{subfig}
\usepackage{bm}
\usepackage{amssymb}
\usepackage{amsmath}
\usepackage{upgreek}
\usepackage{bigints}
\usepackage{algorithm2e}
\usepackage{algpseudocode}
\usepackage{url}
\usepackage{hyphenat}
\RestyleAlgo{ruled}
\usepackage{float}
\usepackage[left=1.7cm, right=1.7cm, top=3.0cm, bottom=3.0cm]{geometry}

\journal{Journal of Computational Physics}

\begin{document}

\begin{frontmatter}

\title{Recurrent Transformer U-Net Surrogate for Flow Modeling and Data Assimilation in Subsurface Formations with Faults}

\author[inst1]{Yifu Han}
\affiliation[inst1]{organization={Department of Energy Science and Engineering}, 
            addressline={Stanford University}, 
            city={Stanford},
            state={CA},
            postcode={94305}, 
            country={USA}}

\author[inst1]{Louis J.~Durlofsky}

\begin{abstract}
Many subsurface formations, including some of those under consideration for large-scale geological carbon storage, include extensive faults that can strongly impact fluid flow. In this study, we develop a new recurrent transformer U-Net surrogate model to provide very fast predictions for pressure and CO$_2$ saturation in realistic faulted subsurface aquifer systems. The geomodel includes a target aquifer (into which supercritical CO$_2$ is injected), surrounding regions, caprock, two extensive faults, and two overlying aquifers. The faults can act as leakage pathways between the three aquifers. The heterogeneous property fields in the target aquifer are characterized by hierarchical uncertainty, meaning both the geological metaparameters (e.g., mean and standard deviation of log-permeability) and the detailed cell properties of each realization, are uncertain. Fault permeabilities are also treated as uncertain. The model is trained with simulation results for (up to) 4000 randomly sampled realizations. Error assessments show that this model is more accurate than a previous recurrent residual U-Net, and that it maintains accuracy for qualitatively different leakage scenarios. The new surrogate is then used for global sensitivity analysis and data assimilation. A hierarchical Markov chain Monte Carlo data assimilation procedure is applied. Different monitoring strategies, corresponding to different amounts and types of observed data collected at monitoring wells, are considered for three synthetic true models. Detailed results demonstrate the degree of uncertainty reduction achieved with the various monitoring strategies. Posterior results for 3D saturation plumes and leakage volumes indicate the benefits of measuring pressure and saturation in all three aquifers.
\end{abstract}

\begin{keyword}
geological carbon storage, faulted aquifer system, CO$_2$ leakage, recurrent transformer U-Net, hierarchical MCMC data assimilation, history matching
\end{keyword}

\end{frontmatter}

\section{Introduction}
\label{Introduction}
\input{Introduction.tex}

\section{Geomodel and Simulation Setup}
\label{Geomodel}
\input{Section_2.tex}

\section{Recurrent Transformer U-Net Surrogate Model}
\label{Surrogate Model}
\input{Section_3.tex}

\section{Surrogate Model Evaluation and Global Sensitivity Analysis} 
\label{Surrogate Evaluation}
\input{Section_4.tex}

\section{Data Assimilation using Surrogate Model} 
\label{History Matching}
\input{Section_5.tex}

\section{Concluding Remarks} 
\label{Conclusions}
\input{Section_6.tex}

\section*{CRediT authorship contribution statement}
\textbf{Yifu Han}: Conceptualization, Methodology, Software, Visualization, Formal analysis, Writing -- original draft. \textbf{Louis J. Durlofsky}: Supervision, Conceptualization, Resources, Formal analysis, Writing -- review \& editing.

\section*{Declaration of competing interest}
The authors declare that they have no known competing financial interests or personal relationships that could have appeared to influence the work reported in this paper.

\section*{Data availability}
The code used in this study will be made available on GitHub when this paper is published. Please contact Yifu Han (yifu@stanford.edu) for earlier access.

\section*{Acknowledgments} 
We thank the Stanford Center for Carbon Storage and Stanford Smart Fields Consortium for funding. We also acknowledge Jian Huang, Dick Kachuma, and Herve Gross (TotalEnergies) for providing the SEAM CO$_2$ geomodel, GEOS simulation input files, and output data processing code, and Oleg Volkov (Stanford University) for his help with 3D plume visualization. We are grateful to the SDSS Center for Computation for HPC resources, and to developers at Lawrence Livermore National Laboratory, Stanford University, and TotalEnergies for assistance with GEOS.

\section*{Supplementary Information} 

\section*{SI 1. SI overview}
\renewcommand{\thesection}{SI \arabic{section}}
Data assimilation results were presented for true model~1 in the main text. Here we present abbreviated sets of results for two additional synthetic true models. These cases display qualitatively different flow and fault-leakage behaviors than true model~1. The problem setup and data assimilation procedures for true models~2 and~3 are identical to those used for true model~1.

\section*{SI 2. True model~2 results}\label{true-model-2}
\renewcommand{\thesection}{SI \arabic{section}} 
\input{SI_Section_2.tex}

\section*{SI 3. True model~3 results}\label{true-model-3}
\renewcommand{\thesection}{SI \arabic{section}} 
\input{SI_Section_3.tex}

\bibliographystyle{Main} 
\bibliography{ref}

\end{document}

%% file: Introduction.tex
Large-scale faults and fractures can impact subsurface flow operations, such as geological carbon storage, geothermal energy production, and hydrocarbon production, in a variety of ways. Faults can lead to qualitatively different flow behaviors as they can act as either conduits or barriers. They can also slip (shift) if the stress regime is altered by fluid injection. Fault behavior is strongly impacted by subsurface formation and fault properties such as permeability and porosity. In practice, these properties can be highly uncertain and are estimated through calibration against observed data. The advent of deep learning techniques that can be applied in place of expensive numerical simulation models has enabled the use of sophisticated Monte Carlo-based data assimilation frameworks. Although these procedures have been used for a range of geomodels, they have not been widely applied for realistic 3D systems with faults. The development of an effective surrogate model, capable of predicting qualitatively different flow behaviors in faulted systems, would thus be highly useful for comprehensive data assimilation.

Our focus in this work is on developing a surrogate model to capture the flow aspects of faults (geomechanical effects are not included). We consider a synthetic but realistic 3D faulted aquifer system, referred to as the modified SEAM (Society of Exploration Geophysicists Advanced Modeling) CO$_2$ geomodel. The overall geomodel involves a target aquifer with a surrounding region, caprock, and two overlying aquifers that are potentially in communication with the target aquifer through two extensive faults. We introduce hierarchical uncertainty into this geomodel, i.e., uncertainty in both geological scenario parameters (or metaparameters) and in cell-by-cell properties in the target aquifer. We then develop and train a new recurrent transformer U-Net surrogate model, which includes an attention mechanism, to provide fast flow predictions. The surrogate model is applied in a hierarchical Markov chain Monte Carlo (MCMC) data assimilation procedure under different monitoring strategies.

Within the context of hydrocarbon recovery, large-scale faults can significantly impact pressure communication, reservoir compartmentalization, and connectivity, and thus overall production~\citep{jolley2007faulting}. Field observations indicate that faults can provide pathways over large distances, which can lead to flow between otherwise separate formations and early water breakthrough during production~\citep{wibberley2017faults}. Faults and fractures are also key features in geothermal energy production, and enhanced geothermal technologies fundamentally entail flow through fractured systems. Many studies have investigated the impact of fractures on heat extraction efficiency, reservoir sustainability, and the risk of induced seismicity~\citep{zinsalo2020injection, gao2022impact, cao2023effect}. For direct-use geothermal systems, \citet{daniilidis2020interdependencies} showed that fault properties significantly affect the shape and extent of the cooled-fluid region.

The target application area in this study is geological carbon storage (GCS). Faulted aquifer systems, such as those in the North Sea or the Gulf of Mexico, have been investigated by a number of researchers as potential GCS sites~\citep{mulrooney2020structural, holden2024implications}. Several studies have focused on geomechanical issues, including fault reactivation and induced seismicity~\citep{dana2022two, yoon2024assessing}. In particular, \citet{rahman2021probabilistic} performed a probabilistic assessment of fault stability for a faulted aquifer in the North Sea and found a very low likelihood of fault reactivation during injection. \citet{silva2023assessing} investigated the impact of large-scale injection on a Gulf of Mexico geomodel containing faults, and found that well placement can reduce the risks associated with induced seismicity. 

There have also been a number of investigations on CO$_2$ leakage through pre-existing faults, as is addressed in this study. \citet{chang2009effect} considered the impact of fault properties on CO$_2$ migration, plume geometry, residual trapping, and countercurrent flow. \citet{rinaldi2014geomechanical} evaluated the impact of injection rate, fault permeability, and fault geometry on the leakage rate to an overlying aquifer. \citet{zhang2018mechanism} performed a sensitivity analysis on fault permeability, aquifer permeability, and fault width to investigate vertical fluid exchange during CO$_2$ leakage along a fault. They considered CO$_2$ and brine leaking upward through one fault, and freshwater in an upper aquifer migrating downward through another fault. \citet{sorgi2024new} introduced a new quantitative risk assessment approach to estimate CO$_2$ leakage risk and severity through pre-existing faults. \citet{pettersson2025copula} developed an uncertainty quantification method to investigate the impact of uncertain fault properties on CO$_2$ leakage volume for a North Sea geomodel.

Deep-learning-based surrogate models have been applied for a wide range of subsurface flow problems~\citep{mo2019deep, tang2020deep, yan2022gradient, wang2022deep, wen2023real}. Some recent studies have considered GCS and geothermal systems involving faults or fractures. In a geothermal setting, \citet{qin2023efficient} developed a surrogate model integrating convolutional and recurrent neural networks to predict production enthalpy (a scalar quantity) in a realistic faulted geothermal reservoir, with the goal of optimizing power generation. \citet{yan2024physics} developed a Fourier neural operator (FNO) surrogate model to predict the temperature field and the temperature of produced fluid in idealized geomodels. The surrogate model was then incorporated into an optimization workflow. For GCS, \citet{sun2021optimization} used a neural network-based surrogate model to predict the net present environmental value (a scalar quantity) for a synthetic but realistic system based on a faulted North Sea geomodel. The injection well location and controls were treated as (variable) input. \citet{tariq2023deep} applied an FNO-based surrogate model to predict CO$_2$ saturation plumes in 2D heterogeneous, naturally fractured geomodels. The input to the surrogate model included permeability and porosity fields constructed from discrete fracture networks, along with operational settings. \citet{ju2024learning} developed a recurrent graph neural network for idealized 2D geomodels with unstructured simulation grids and two impermeable faults. This model enabled saturation and pressure field predictions for geomodels with different injection well locations and grid configurations.

A key goal in this study is to develop a surrogate model to predict the temporal evolution of saturation and pressure fields in realistic faulted subsurface systems. Our application involves GCS, but our procedures are also relevant to other systems. Most of the surrogate modeling studies for faulted or fractured systems discussed above either focused on predicting system states in idealized (often 2D) geomodels or, if realistic 3D geomodels were considered, the surrogate model predictions were for a single quantity of interest. Here we predict CO$_2$ saturation and pressure fields in complicated geomodels derived from the SEAM CO$_2$ model~\citep{stefani2010seam, yoon2024assessing}, in which qualitatively different flow behaviors and fault leakage scenarios can occur. A new recurrent transformer U-Net model that incorporates an attention mechanism into the network architecture is developed for this purpose. The attention mechanism enables the model to better capture the spatial dependencies and specific flow behaviors that are driven by the faults. The new surrogate model is shown to provide improved performance relative to an existing recurrent R-U-Net architecture~\citep{han2023surrogate, han2025accelerated}.

Our additional goals in this work are to use the surrogate model for a global sensitivity analysis and for data assimilation. In the global sensitivity analysis, we quantify the importance of the geological metaparameters and geomodel realizations on the CO$_2$ footprint in the target aquifer. A hierarchical MCMC-based data assimilation procedure~\citep{han2023surrogate} is applied to evaluate the degree of uncertainty reduction achieved for various quantities of interest. These include the CO$_2$ saturation and pressure distributions in the target aquifer and CO$_2$ leakage volumes into the two overlying aquifers. Different monitoring strategies, which entail monitoring in all three aquifers versus monitoring only in the overlying formations, and the use of only pressure data versus both pressure and saturation data, are assessed. 

This paper proceeds as follows. In Section~\ref{Geomodel}, we describe the modified SEAM CO$_2$ geomodel, including the hierarchical uncertainties considered and the simulation setup. The design and architecture of the new recurrent transformer U-Net surrogate model are presented in Section~\ref{Surrogate Model}. The accuracy of the surrogate model, for different numbers of training samples, is assessed and compared with an existing surrogate in Section~\ref{Surrogate Evaluation}. Detailed predictions from the surrogate model for distinct flow behaviors and leakage scenarios, along with global sensitivity results, are also provided. In Section~\ref{History Matching}, we present data assimilation results using the hierarchical MCMC procedure for a synthetic true model. Conclusions and suggestions for future work appear in Section~\ref{Conclusions}. In the online Supplemental Information (SI), we provide data assimilation results for two additional synthetic true models corresponding to varying amounts of CO$_2$ leakage through the faults.

%% file: Section_2.tex
In this section, we first discuss the modified SEAM CO$_2$ geomodel applied in this study. The flow simulation setup is then described.

\subsection{Geomodels with hierarchical uncertainty}
\label{sec:SEAM_realizations}

The original SEAM Gulf of Mexico tertiary salt models, developed by \citet{stefani2010seam} and \citet{oristaglio2016seam}, represent a turbidite depositional environment involving a shaly sandstone aquifer with two overlying aquifers above the caprock and intersecting faults. \citet{fehler2024seam} and \citet{yoon2024assessing} adapted the original model for the SEAM CO$_2$ project. The main goal of the SEAM CO$_2$ project is to construct benchmark geomodels that can be simulated and used to evaluate different monitoring approaches and geophysical tools. For the SEAM CO$_2$ project, the overall domain was discretized into three different grid resolutions of about $0.3\times10^6$, $2.9\times10^6$, and $22.6\times10^6$ cells.

\begin{figure}[!ht]
\centering   
{\includegraphics[width = 165mm]{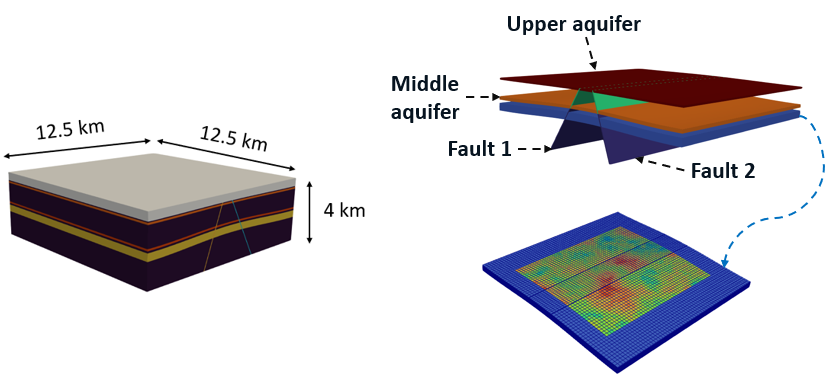}}
\caption{Overall domain of the modified SEAM CO$_2$ geomodel (left), and the two faults, upper, middle, and lower (target) aquifer, and surrounding region (right).}
\label{SEAM_Model}
\end{figure}

In this study, we further modify the geomodel constructed for the SEAM CO$_2$ project. The overall domain of our modified SEAM CO$_2$ geomodel is shown in Fig.~\ref{SEAM_Model}. The model involves three stacked formations, which are separated by low-permeability caprock and potentially connected via two extensive faults. Supercritical CO$_2$ is injected into the deepest formation, referred to as the target aquifer, and it can leak into the other two formations if the faults are sufficiently permeable. Consistent with the SEAM CO$_2$ model, the modified geomodel also includes overburden and underburden rock.

The physical dimensions of the target aquifer are generally consistent with those in the SEAM CO$_2$ project, and the level of resolution used here is comparable to that of the coarsest SEAM geomodel. The target aquifer, of dimensions 8.1~km $\times$ 9.7~km $\times$ 425~m, is represented on a structured grid containing 48 $\times$ 48 $\times$ 25 cells. The average cell size in the target aquifer is about 168~m $\times$ 202~m $\times$ 17~m. The average depth to the top of the target aquifer is about 1854~m (top depth varies with location since the target aquifer is folded). The original SEAM CO$_2$ geomodel is deterministic, though in this work we characterize the target aquifer in terms of multi-Gaussian log-permeability and porosity fields with uncertain metaparameters, as described below.

The middle aquifer, of dimensions 12.5~km $\times$ 12.5~km $\times$ 102.5~m and average depth of 1717~m, is represented on a grid of 69 $\times$ 62 $\times$ 6 cells. The upper aquifer, at an average depth of 563~m, is of comparable size and is modeled at a similar grid resolution. The three formations are fully intersected by two north-south-striking faults, which are about 2~km apart within the target aquifer. Fault~1, shown in Fig.~\ref{SEAM_Model}, has a 70$^{\circ}$ westward dip, while Fault~2 dips 60$^{\circ}$ eastward. Both faults start at the top of the upper aquifer (just below the overburden) and end at the bottom of the underburden rock. The faults are 60~m across and are represented by two grid cells. The fault-induced offsets, treated by \citet{yoon2024assessing}, are not considered in this study. This is because they  complicate the hierarchical geomodeling while having relatively little effect on flow (the main effect is from the properties of the faults themselves). The overall numerical models, including the three aquifers, two faults, and surrounding regions, contain 69 $\times$ 62 $\times$ 78 cells (total of 330,000 cells).

In this work, we construct geomodel realizations of the overall domain based on a set of geological scenario parameters for the target aquifer along with permeabilities for the middle and upper aquifers and faults. The full set of these parameters, denoted $\boldsymbol{\uptheta}_{\mathrm{meta}} \in \mathbb{R}^{11}$ and collectively
referred to as metaparameters, are 
\begin{equation} \label{metaparameters}
\boldsymbol{\uptheta}_{\mathrm{meta}} = [\mu_{\log k}, \sigma_{\log k}, a_r, d, e, k_m, k_u, k_{f1}^{tm}, k_{f1}^{mu}, k_{f2}^{tm}, k_{f2}^{mu}].
\end{equation}
\noindent Here $\mu_{\log k}$ and $\sigma_{\log k}$ are the mean and standard deviation of the multi-Gaussian log-permeability field in the target aquifer, $a_r$ is permeability anisotropy ratio ($k_v/k_h$, where $k_v$ and $k_h$ are the vertical and horizontal permeabilities), $d$ and $e$ are coefficients relating porosity to log-permeability in the target aquifer (via the equation $\phi = d \cdot \log k_h + e$), and $k_m$ and $k_u$ are the (constant and isotropic) permeabilities in the middle and upper aquifers. The parameters $k_{f1}^{tm}$ and $k_{f1}^{mu}$ represent the permeabilities for Fault~1, with $k_{f1}^{tm}$ characterizing the fault segments between the target and middle aquifers, and $k_{f1}^{mu}$ defining Fault~1 permeability between the middle and upper aquifers. The parameters $k_{f2}^{tm}$ and $k_{f2}^{mu}$ denote the analogous quantities for Fault~2. Note that this representation allows for the variation of fault permeability with depth, which is consistent with fault properties in realistic settings. This variation can result from a combination of effects, which include overburden stress and fracturing~\citep{faulkner2010review} as well as cementation and diagenesis~\citep{bense2006faults}.

The 11 metaparameters are all taken to be uncertain. Their prior ranges, along with the values specified for other (fixed) parameters, are provided in Table~\ref{rock physics}. For a given set of metaparameters $\boldsymbol{\uptheta}_{\mathrm{meta}}$, an infinite number of target aquifer realizations can be generated. This leads to a hierarchy of uncertainty since both metaparameters and the cell-by-cell permeability and porosity values for each realization are treated as random variables.

\begin{table}[!ht]
\begin{center}
\footnotesize
\caption{Parameters used in the flow simulations}
\label{rock physics}
\renewcommand{\arraystretch}{1.3}
\begin{tabular}{ c @{\hskip 3em} c } 
\hline
\textbf{Overburden/Underburden/Caprock} & \textbf{Values} \\ 
\hline
Permeability & 0.0001 md \\ 
Porosity & 0.01 \\ 
Permeability anisotropy ratio & 0.1 \\
\hline
\textbf{Upper/Middle aquifer} & \textbf{Values or ranges} \\ 
\hline
Permeability ($k_u$, $k_m$) & $k_u, k_m \sim$ $U$(50, 500) md \\ 
Porosity & 0.3 \\ 
Permeability anisotropy ratio  & 0.1 \\
\hline
\textbf{Target aquifer} & \textbf{Value or range} \\ 
\hline
Correlation length in $x$ ($l_x$) & 10~cells (1.7~km) \\
Correlation length in $y$ ($l_y$) & 10~cells (2~km) \\
Correlation length in $z$ ($l_z$) & 3~cells (51~m) \\
Mean of log-permeability ($\mu_{\log k}$)  & $\mu_{\log k} \sim$ $U$(4, 6) $\Leftrightarrow$ (54.6, 403.4)~md \\ 
Standard deviation of log-permeability ($\sigma_{\log k}$) & $\sigma_{\log k} \sim$ $U$(1, 1.5) \\ 
Parameter $d$ in $\log k$--$\phi$ correlation & $d \sim U(0.02, 0.04)$ \\ 
Parameter $e$ in $\log k$--$\phi$ correlation & $e \sim U(0.06, 0.08)$ \\ 
Permeability anisotropy ratio ($a_r$)  & $\log_{10}(a_r)$ $\sim$ $U$(-1.3, -0.7) \\
\hline
\textbf{Surrounding region} & \textbf{Values} \\ 
\hline 
Permeability & 5 md \\ 
Porosity & 0.05 \\ 
Permeability anisotropy ratio & 0.1 \\
\hline
\textbf{Faults} & \textbf{Values or ranges} \\ 
\hline
Target-middle permeability ($k_{f1}^{tm}$, $k_{f2}^{tm}$) & $\log_{10}(k_{f1}^{tm}), \log_{10}(k_{f2}^{tm}) \sim$ $U$(-1, 2.5)\\ 
Middle-upper permeability ($k_{f1}^{mu}$, $k_{f2}^{mu}$) & $\log_{10}(k_{f1}^{mu}), \log_{10}(k_{f2}^{mu}) \sim$ $U$(-1, 2.5) \\ 
Porosity & 0.2 \\ 
Permeability anisotropy ratio  & 1 \\
\hline
\end{tabular}
\end{center}
\end{table}

We use principal component analysis (PCA) along with the metaparameters $\mu_{\log k}$ and $\sigma_{\log k}$ to generate geological realizations in the target aquifer. This treatment is consistent with that applied in previous studies~\citep{han2023surrogate, han2025accelerated}. The PCA basis matrix is constructed from a set of 1000 unconditioned realizations generated using the geological modeling software SGeMS~\citep{remy2009applied}. Each realization satisfies the correlation structure defined by $l_x$, $l_y$, and $l_z$ in Table~\ref{rock physics}. These correlation lengths were estimated based on the permeability field in the SEAM CO$_2$ project~\citep{yoon2024assessing}. A spherical variogram model was applied. The SGeMS models (and thus the PCA representation) are constructed for a rectangular grid containing 48 $\times$ 48 $\times$ 25 brick-shaped cells that cover the target aquifer region. The properties for the actual (nonrectangular) cells are then assigned through a nearest-neighbor mapping. The PCA representation is given by
\begin{equation} \label{pca}
\textbf{y}^{\mathrm{pca}} = \Phi \bm{\xi} + \bar{\textbf{y}}, 
\end{equation} 
where $\textbf{y}^{\mathrm{pca}} \in \mathbb{R}^{n_{s}}$ is the PCA estimate of the spatially correlated standard multi-Gaussian field, $n_{s}=57,600$ is the number of cells in the target aquifer (48 $\times$ 48 $\times$ 25 cells), $\Phi \in \mathbb{R}^{n_{s} \times n_d}$ is the PCA basis matrix truncated to $n_d$ columns ($n_d$ = 820 in this study), $\bar{\textbf{y}} \in \mathbb{R}^{n_s}$ is the mean of the realizations used to constructed the basis matrix, and $\bm{\xi} \in \mathbb{R}^{n_d}$ is the low-dimensional (latent) variable. The matrix $\Phi$ is constructed through application of singular value decomposition to a matrix containing, as its columns, the centered unconditioned realizations from SGeMS. The use of PCA provides substantial computational speedup relative to applying sequential Gaussian simulation (within SGeMS) during the online MCMC computations. Please see \citet{han2023surrogate} for further discussion of this issue.

The metaparameters $\mu_{\log k}$ and $\sigma_{\log k}$ shift and rescale the standard multi-Gaussian fields generated by PCA. The horizontal ($x$ and $y$ direction) permeability for cell $i$ ($i = 1, 2, \dots, n_s$) in the target aquifer is given by $(k_{s})_i = \exp \left( \sigma_{\log k} \cdot (y^{\mathrm{pca}})_i + \mu_{\log k} \right)$. Vertical permeability is then computed via $(k_z)_i = a_r \cdot (k_s)_i$. Porosity $(\phi_s)_i$ is related to permeability through the expression $(\phi_s)_i = d \cdot \log(k_s)_i + e$, for $i = 1, 2, \dots, n_s$. The permeability (isotropic) for the fault cells is then replaced by the sampled metaparameters (e.g., $k_{f1}^{tm}$, $k_{f1}^{mu}$, $k_{f2}^{tm}$, and $k_{f2}^{mu}$), and porosity is set to a constant value of 0.2.

\subsection{Simulation setup}
\label{sec:simulation_setup}

\begin{figure}[!ht]
\centering   
{\includegraphics[width = 105mm]{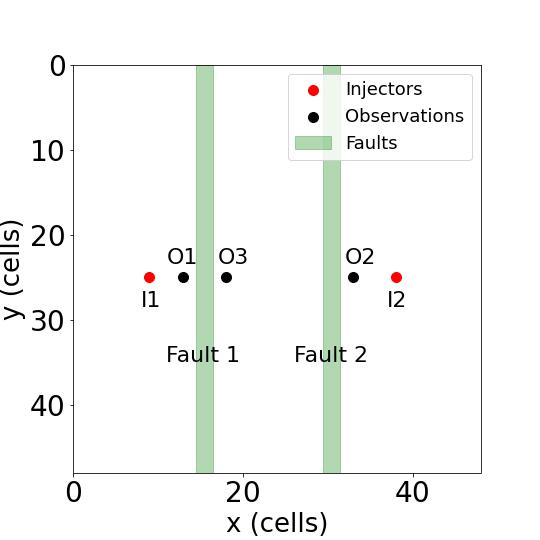}}
\caption{Areal view of the target aquifer, showing the locations of the two faults, two injection wells (I1 and I2), and three observation wells (O1--O3).}
\label{well_location}
\end{figure}

The governing equations for the flow problem solved in this work are given in \citet{tang2022deep}. The injection of CO$_2$ is accomplished via two vertical wells that are completed (open to flow) over the full thickness of the target aquifer. There is no injection into the middle or upper aquifers. Each well injects at a rate of 1~Mt~CO$_2$/year for a 50-year period, resulting in a total of 100~Mt of CO$_2$ injected. The locations of the two vertical injection wells (denoted I1 and I2), along with the two faults and three vertical observation wells (which provide data for history matching and are denoted O1--O3), are indicated in Fig.~\ref{well_location}. The distances between the injection wells and the faults at the top of the target aquifer are about 1.2~km (I1 to Fault~1) and 1.3~km (I2 to Fault~2). All three observation wells are located near the faults, with O3 placed between the faults. This placement allows the wells to detect pressure communication across faults and cross-fault flow. The distances from I1 to O1 and from I2 to O2 are about 800~m and 1000~m, respectively.

\begin{figure}[!ht] 
\centering
\subfloat[Relative permeability curves]{\label{relative_perm}\includegraphics[width = 84mm]{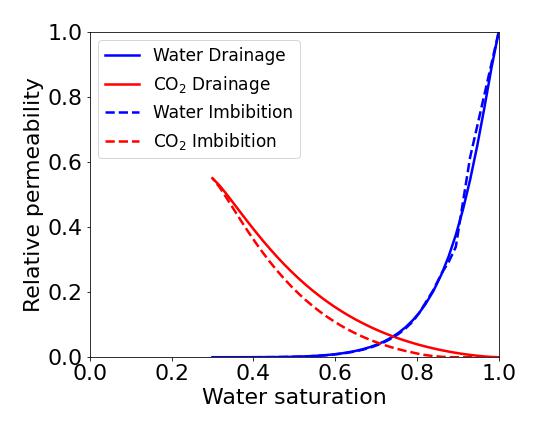}}
\hspace{4mm}
\subfloat[Capillary pressure curve]{\includegraphics[width = 84mm]{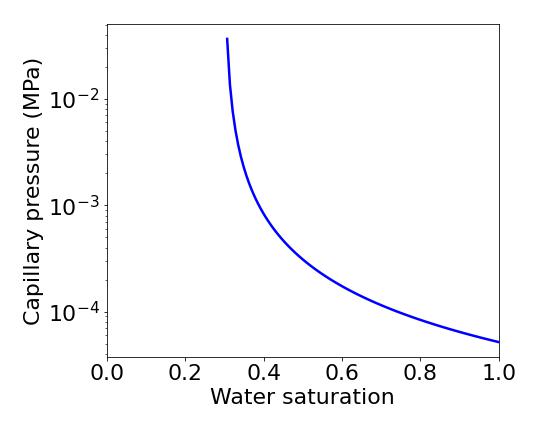}}
\caption{CO$_2$-brine relative permeability with hysteresis effects and capillary pressure curves used in all simulations. Capillary pressure curve in (b) is for a cell with porosity of 0.2 and permeability of 50~md.}
\label{fig:relper_pc}
\end{figure}

The relative permeability and capillary pressure curves used in the simulations are shown in Fig.~\ref{fig:relper_pc}. The relative permeability curves are from the simulation setup used in the SEAM CO$_2$ project. Hysteresis between drainage and imbibition is considered, as illustrated in Fig.~\ref{fig:relper_pc}a. Capillary pressure curves used for our flow simulations are based on measurements in cores from the Gulf of Mexico~\citep{trevino2017geological}. The Brooks-Corey~\citep{saadatpoor2010new} model, with an exponent ($\lambda$) of 0.7, is used to generate the capillary pressure curves. The capillary pressure curve for each cell is computed using the Leverett J-function, which involves the cell porosity and permeability. The same J-function is used for fault cells as for other cells in the domain (different J-functions could be used if fault data are available). The capillary pressure curve for a block with a porosity of 0.2 and permeability of 50~md is shown in Fig.~\ref{fig:relper_pc}b.

No-flow boundary conditions are applied at the boundaries of the full domain. The initial hydrostatic pressure and temperature at a depth of 1880~m in the target aquifer are set to 18~MPa and 67~$^{\circ}$C. These values are based on the simulation setup used in the SEAM CO$_2$ project. Pressure and temperature at the average depth of the middle aquifer (1717~m) are 16.4~MPa and 63~$^{\circ}$C, and those in the upper aquifer (average depth of 563~m) are 5.2~MPa and 34~$^{\circ}$C. The injected CO$_2$ is in a supercritical state in the target and middle aquifers, but it transitions to the gas phase if it leaks into the upper aquifer. The typical densities of CO$_2$ are about 624~kg/m$^3$, 612~kg/m$^3$, and 128~kg/m$^3$ in the three aquifers at initial conditions. 

In this work we perform flow-only simulations using the compositional formulation in GEOS~\citep{huang2023fully}. In these flow-only simulations, the impact of pressure on porosity is modeled using an effective rock compressibility. This effective rock compressibility is determined from the geomechanical properties. As was shown in \citet{han2025accelerated}, in certain settings, even when geomechanical effects are not modeled explicitly, this treatment can provide results for pressure and saturation that are in close agreement with those obtained from coupled flow-geomechanics simulations. The effective rock compressibility, denoted by $c$, is taken to be constant over the full domain for all realizations. This quantity is given by \citep{kim2010sequential}
\begin{equation} \label{compressibility}
c = \frac{1 - 2v}{\bar{\phi} E} \left( b^2 \frac{1 + v}{1 - v} + 3(b - \bar{\phi})(1 - b) \right),
\end{equation}
where $\bar{\phi} = 0.19$ is the volumetric average porosity, and $b = 1$, $E = 7.74$~GPa, and $v = 0.23$ are the volumetric average Biot coefficient, Young’s modulus, and Poisson’s ratio, respectively. These average quantities are computed over the upper aquifer, middle aquifer, and target aquifer plus the surrounding region. The impermeable caprock between aquifers, and the overburden and underburden rock, are not included in these averages. The values of the geomechanical properties used in this computation are from the SEAM CO$_2$  geomodel~\citep{fehler2024seam, yoon2024assessing}. The resulting effective rock compressibility is $c = 4 \times 10^{-6}$~$\text{psi}^{-1}$, which is the value used for all flow simulations. A typical GEOS run with this treatment, for the setup described in this section, requires about 10~minutes using 32 AMD EPYC-7543 CPU cores in parallel.

%% file: Section_3.tex
In this section, we describe the design and architecture of the new recurrent transformer U-Net surrogate model. The training procedure is then discussed.

\subsection{Deep-learning-based surrogate model}
\label{sec:surrogate}

The geomodel realizations for the full domain are denoted $\mathbf{m}_f \in \mathbb{R}^{n_f \times n_p}$, where $n_f$ = 333,000 is the number of cells in the full domain, and $n_p$ = 3 is the number of rock properties in each cell. These include horizontal and vertical permeability and porosity. The flow simulations on the full domain can be expressed as
\begin{equation} \label{flow forward}
\left[{\textbf{p}}_f, \enspace {\textbf{S}}_f\right] = f\left(\mathbf{m}_f\right),
\end{equation}
where $f$ represents the high-fidelity flow-only simulation, and ${\textbf{p}}_f \in \mathbb{R}^{n_f \times n_{ts}}$ and ${\textbf{S}}_f \in \mathbb{R}^{n_f \times n_{ts}}$ are pressure and saturation in the full domain at $n_{ts}$ simulation time steps.  

The surrogate model is constructed to provide predictions for key quantities in the domain of interest at a set of specific time steps. This domain of interest comprises the 48 $\times$ 48 $\times$ 48 cell region that covers the target aquifer, the two overlying aquifers, the two faults, and the caprock between the aquifers. We denote the geomodel for this domain as $\mathbf{m}_d \in \mathbb{R}^{n_d \times n_p}$, where $n_d$ = 110,592 is the total number of cells within the domain. The overburden and underburden and the region surrounding the target aquifer are not included. Two separate networks are trained to predict pressure and saturation at 10 discrete time steps, specifically $t$ = 2, 6, 10, 14, 20, 26, 32, 38, 44, and 50~years after the start of injection. Consistent with the notation used in \citet{han2023surrogate, han2025accelerated}, the surrogate model predictions for pressure and saturation fields in the domain of interest are expressed as 
\begin{equation} \label{flow surrogate}
\left[\hat{\textbf{p}}_{d}, \enspace \hat{\textbf{S}}_{d}\right] = \hat{f}\left(\mathbf{m}_d; \textbf{w}\right),
\end{equation}
\noindent where $\hat{f}$ denotes the surrogate model, $\hat{\textbf{p}}_d \in \mathbb{R}^{n_d \times n_{t}}$ and $\hat{\textbf{S}}_d \in \mathbb{R}^{n_d \times n_{t}}$ are surrogate model predictions for pressure and saturation in the domain of interest, $n_t = 10$ is the number of surrogate model time steps (note $n_t < n_{ts}$), and $\textbf{w}$ represents the neural network parameters for the surrogate models determined during training. 

The original recurrent R-U-Net surrogate model for CO$_2$ storage was developed by \citet{tang2022deep} and was extended to treat geomodels drawn from multiple geological scenarios by \citet{han2023surrogate}. Please consult these papers for details. The recurrent R-U-Net model involves a residual U-Net with an encoder and a decoder, and a recurrent neural network. In the encoder, input geomodels are transformed by a series of convolutional and residual blocks to latent feature maps at multiple levels of fidelity. The decoder provides pressure or saturation fields at a particular time step by combining upsampled feature maps in the decoder with the latent features from the encoder. The recurrent network enables the model to capture temporal dynamics. The extended recurrent R-U-Net model was applied to geomodels involving a single aquifer without faults~\citep{han2023surrogate, han2025accelerated}. As we will see in Section~\ref{Surrogate Evaluation}, this model can provide saturation and pressure predictions in faulted systems that are of reasonable accuracy in many cases, though in some instances it may miss qualitative behaviors. This motivated the development of an enhanced capability.

\begin{figure}[!ht]
\centering   
\subfloat[Transformer U-Net]{\includegraphics[width = 165mm]{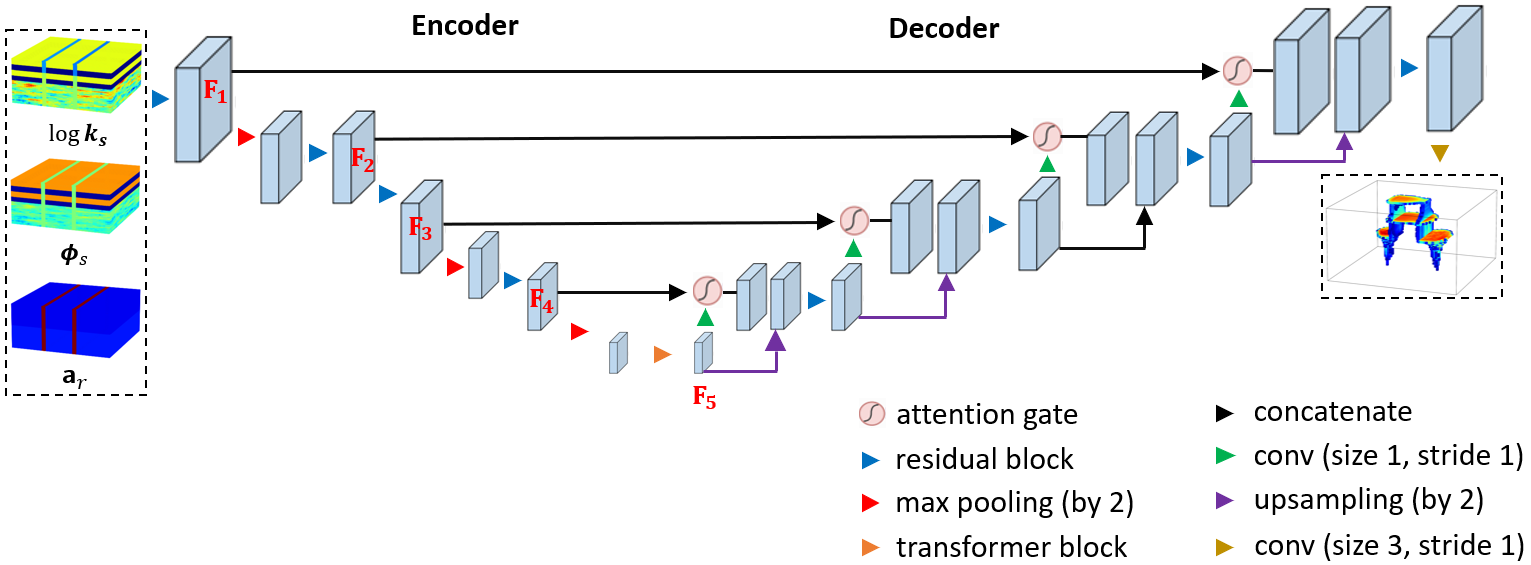}}
\\[2ex]
\subfloat[Recurrent transformer U-Net]
{\includegraphics[width = 115mm]{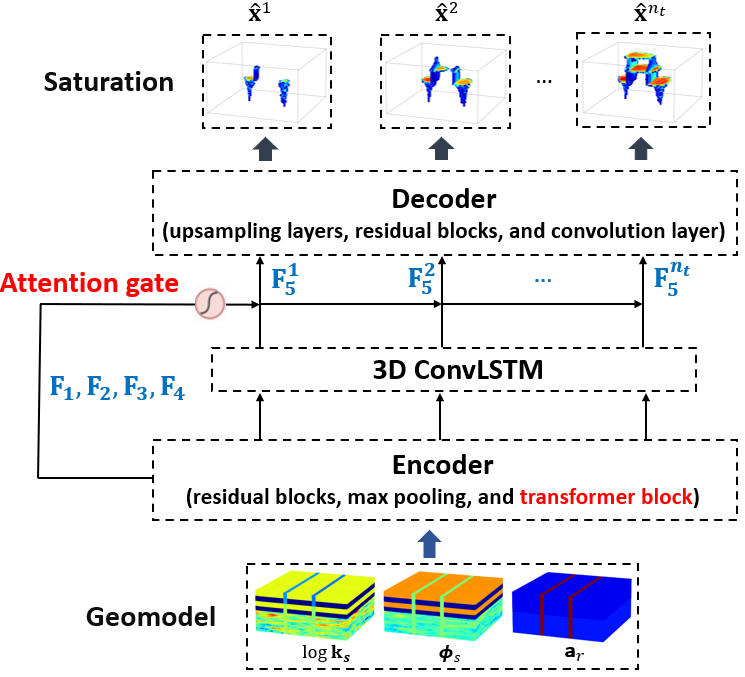}}
\caption{Schematic of the recurrent transformer U-Net model architecture. Transformer U-Net architecture for pressure and saturation field predictions in the domain of interest at a specific time step is shown in (a). The incorporation of a convolutional long short-term memory recurrent neural network to predict temporal evolution is shown in (b).}
\label{nn}
\end{figure}

The recurrent transformer U-Net surrogate model, designed to predict accurate CO$_2$ saturation and pressure fields in realistic faulted subsurface systems, is shown in Fig.~\ref{nn}. The new surrogate model involves an attention mechanism within a residual U-Net architecture to predict qualitatively different flow behaviors, which may occur given the wide range of fault properties considered. Specifically, this model incorporates an attention gate~\citep{oktay2018attention} and transformer layers~\citep{vaswani2017attention} into the model architecture. These components enable the model to effectively combine low-dimensional latent feature maps at different levels of fidelity and to better capture the spatial dependencies and flow responses due to the faults. As in the recurrent R-U-Net, a 3D convolutional long short-term memory recurrent network is again incorporated to capture temporal evolution. 

\begin{figure}[!ht]
\centering   
\subfloat[Attention gate]{\includegraphics[width = 168mm]{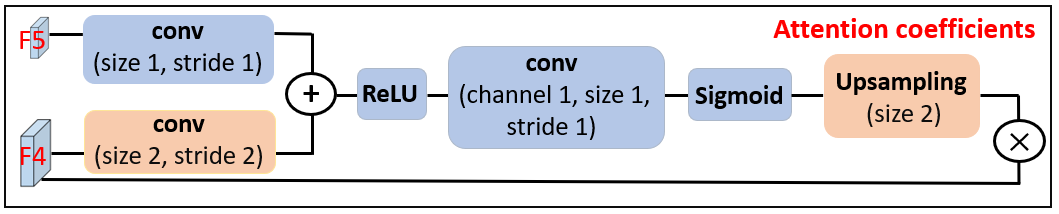}}
\\[4ex]
\subfloat[Transformer block]
{\includegraphics[width = 122mm]{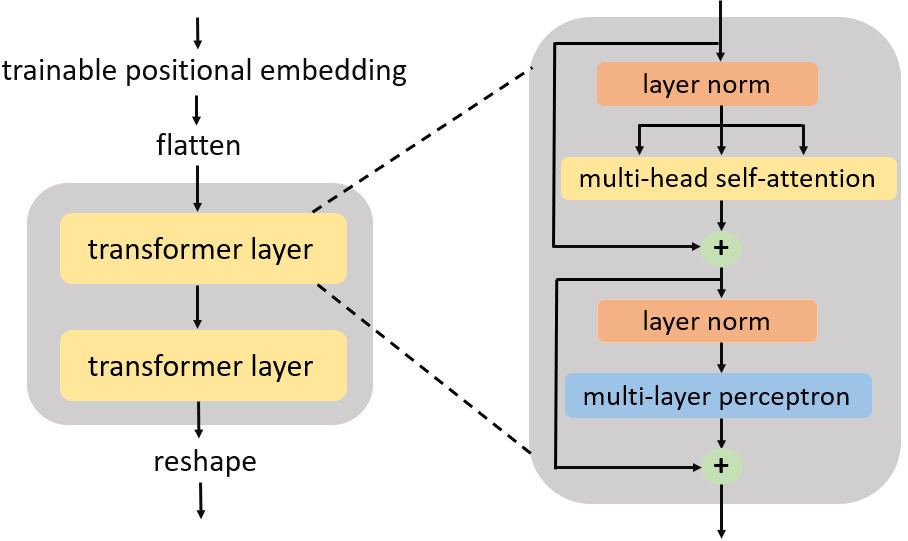}}
\caption{Schematic of attention gate and transformer block incorporated in the recurrent transformer U-Net model.}
\label{attention}
\end{figure}

Details of the recurrent transformer U-Net architecture are provided in Table~\ref{architecture}. This transformer U-Net architecture, shown in Fig.~\ref{nn}(a), shares similarities with the network used by \citet{jamali2023transu}. In the encoder, the input geomodels are processed by residual blocks and max pooling layers to generate a series of low-dimensional latent feature maps at three different levels of fidelity ($\textbf{F}_1$ at $48 \times 48 \times 48$, $\textbf{F}_2$ and $\textbf{F}_3$ at $24 \times 24 \times 24$, $\textbf{F}_4$ at $12 \times 12 \times 12$, $\textbf{F}_5$ at $6 \times 6 \times 6$). Rather than using a skip-connection or concatenation between the encoder and decoder (as in the residual U-Net), the attention gate is used to combine latent feature maps from the encoder and decoder. These can be at different levels of fidelity. Specifications on the attention gate, which shows the combination of latent feature maps at two different levels of fidelity (e.g., $\textbf{F}_4$ and $\textbf{F}_5$), are given in Fig.~\ref{attention}(a). The latent feature maps $\textbf{F}_4$ from the encoder have higher fidelity and capture finer spatial details, while the latent feature maps $\textbf{F}_5$ in the decoder contain more global information but at lower fidelity. The attention gate computes attention coefficients by combining these two latent feature maps. The attention coefficients indicate which parts of the latent feature maps from the encoder are given higher attention weights~\citep{oktay2018attention}. This component enables the model to focus on more relevant spatial features and finer details in important regions, such as around the faults.

A transformer block, consisting of two transformer layers, is incorporated at the end of the encoder, similar to the network used by \citet{chen2021transunet}. Details of the transformer block are provided in Fig.\ref{attention}(b). A trainable positional embedding is added to the latent feature maps before the transformer layers to encode spatial position information. Each transformer layer contains a multi-head self-attention and a multi-layer perceptron submodule. The multi-head self-attention submodule uses four attention heads, and the multi-layer perceptron submodule contains two dense layers with 64~units. The transformer block enables the surrogate model to capture long-range spatial dependencies~\citep{chen2021transunet}, which allows it to relate (associate) spatially distant regions within the input geomodel. As a result, the encoder is able to construct more comprehensive global latent feature representations. This capability can be beneficial in  faulted subsurface systems, where flow in the three aquifers is impacted by the fault properties. The recurrent transformer U-Net surrogate model developed in this work integrates residual blocks that capture local features in the geomodel with a transformer block that captures global features with long-range spatial dependencies.

As noted earlier, separate networks are trained for pressure and saturation. Each surrogate model involves three input channels, which correspond to fields for log-permeability, porosity, and permeability anisotropy ratio in the domain of interest. In the saturation network, rectified linear unit (ReLU) activation is applied at the last convolution block in Table~\ref{architecture}. No activation is applied at the last convolution block in the pressure network.

\subsection{Surrogate model training}
\label{sec:training}

The training procedure is as follows. Normalization is applied for the input channels characterizing geomodels and for the pressure fields prior to the training process. The saturation fields do not require normalization because saturation is already between 0 and 1. The input log-permeability, porosity and anisotropy ratio fields for the domain of interest are normalized by their respective maximum values. A time-dependent normalization is applied for pressure, specifically,
\begin{equation} \label{normalization}
(\bar{p}_s)_{i,j}^t = \frac{(p_s)_{i,j}^t - \mathrm{mean}{((p_s)_{i,j})}}{\mathrm{std}{((p_s)_{i,j})}}, 
\end{equation}
\noindent where $(p_s)_{i,j}^t$ and $(\bar{p}_s)_{i,j}^t$ are the pressure and normalized pressure in cell $j$ in geomodel $i$, at time step $t$, and $\mathrm{mean}{((p_s)_{i,j})}$ and $\mathrm{std}{((p_s)_{i,j})}$ represent the mean and standard deviation of pressure, computed over all pressure fields for the domain of interest at all time steps and all training runs.

In the training procedure, the $L_2$ loss function, computed based on simulated and predicted saturation or normalized pressure, is minimized. This is expressed as 
\begin{equation} \label{minimization_sp}
\textbf{w}^{\ast} = \operatorname*{argmin}_{\textbf{w}} L = \operatorname*{argmin}_{\textbf{w}} \frac{1}{n_{f}} \frac{1}{n_t} \sum_{i=1}^{n_{f}}\sum_{t=1}^{n_t}\|\hat{\textbf{x}}_{i}^{t} - \textbf{x}_{i}^{t}\|_2^2,
\end{equation}
\noindent where $\textbf{w}^{\ast}$ denotes the optimized network parameters, $L$ is the loss function, $n_{f}$ is the number of training samples, and $\textbf{x}_{i}^{t}$ and $\hat{\textbf{x}}_{i}^{t}$ represent the normalized pressure or saturation fields from simulation and the surrogate model, respectively, at time step $t$ for training sample $i$. 

For both networks, the initial learning rate is set to 0.0005. The learning rate is reduced by a factor of 2 when the performance does not improve (training losses does not decrease) for 10~epochs, with a prescribed minimum of $10^{-7}$. We use a batch size of 4 for the saturation network and 8 for the pressure network. Each network is trained for 300~epochs. The maximum number of training samples considered is $n_f$ = 4000. This training requires about 24~hours for each network (48~hours total) using a Nvidia A100 GPU. The training procedure can, however, be terminated after about 180~epochs (corresponding to 14.5~hours per network), since the test losses plateau at this point. Each GEOS training run requires about 10~minutes (with 32 AMD EPYC-7543 CPU cores), so 4000 training runs corresponds to a serial time of about 667~hours. The elapsed time for these simulations is much less because they can be run in parallel. In Section~\ref{Surrogate Evaluation}, the performance of the networks with different numbers of training samples will be considered. The timings for training and simulation for cases with $n_f < 4000$ are less than those reported here. 

\begin{table}[!ht]
\footnotesize
\begin{center}
\caption{Architecture of the recurrent transformer surrogate model}
\label{architecture}
\renewcommand{\arraystretch}{1.15}
\begin{tabular}{ c @{\hspace{12mm}} c @{\hspace{12mm}} c } 
\hline
\textbf{Network} & \textbf{Layer}  & \textbf{Output Shape} \\ 
\hline
Encoder & input & ($n_x$, $n_y$, $n_z$, 3) \\ 
 & residual block, 16 filters of size $3$ & \textbf{F}$_1$: ($n_x$, $n_y$, $n_z$, 16) \\
 & max pooling, of size $2$ & ($\frac{n_x}{2}$, $\frac{n_y}{2}$, $\frac{n_z}{2}$, 16) \\
 & residual block, 32 filters of size $3$ & \textbf{F}$_2$: ($\frac{n_x}{2}$, $\frac{n_y}{2}$, $\frac{n_z}{2}$, 32) \\
 & residual block, 32 filters of size $3$ & \textbf{F}$_3$: ($\frac{n_x}{2}$, $\frac{n_y}{2}$, $\frac{n_z}{2}$, 32) \\
 & max pooling, of size $2$ & ($\frac{n_x}{4}$, $\frac{n_y}{4}$, $\frac{n_z}{4}$, 32) \\
 & residual block, 64 filters of size $3$ & \textbf{F}$_4$: ($\frac{n_x}{4}$, $\frac{n_y}{4}$, $\frac{n_z}{4}$, 64) \\
 & max pooling, of size $2$ & ($\frac{n_x}{8}$, $\frac{n_y}{8}$, $\frac{n_z}{8}$, 64) \\
 & transformer block & \textbf{F}$_5$: ($\frac{n_x}{8}$, $\frac{n_y}{8}$, $\frac{n_z}{8}$, 64) \\
\hline
Recurrent & convLSTM, 64 filters of size $3$ & ($\frac{n_x}{4}, \frac{n_y}{4}, \frac{n_z}{4}$, 64, n$_{t}$) \\
\hline
Decoder & attention gate, of size 2 and stride 2 & ($n_{t}$, $\frac{n_x}{8}$, $\frac{n_y}{8}$, $\frac{n_z}{8}$, 64) \\
 & upsample, of size $2$ & ($n_{t}$, $\frac{n_x}{4}$, $\frac{n_y}{4}$, $\frac{n_z}{4}$, 64) \\
 & residual block, 64 filters of size $3$ & ($n_{t}$, $\frac{n_x}{4}$, $\frac{n_y}{4}$, $\frac{n_z}{4}$, 64) \\
 & attention gate, of size 2 and stride 2 & ($n_{t}$, $\frac{n_x}{4}$, $\frac{n_y}{4}$, $\frac{n_z}{4}$, 64) \\
 & upsample, of size $2$ & ($n_{t}$, $\frac{n_x}{2}$, $\frac{n_y}{2}$, $\frac{n_z}{2}$, 64) \\
 & residual block, 32 filters of size $3$ & ($n_{t}$, $\frac{n_x}{2}$, $\frac{n_y}{2}$, $\frac{n_z}{2}$, 32) \\
 & attention gate, of size 1 and stride 1 & ($n_{t}$, $\frac{n_x}{2}$, $\frac{n_y}{2}$, $\frac{n_z}{2}$, 32) \\
 & residual block, 32 filters of size $3$ & ($n_{t}$, $\frac{n_x}{2}$, $\frac{n_y}{2}$, $\frac{n_z}{2}$, 32) \\
 & attention gate, of size 2 and stride 2 & ($n_{t}$, $\frac{n_x}{2}$, $\frac{n_y}{2}$, $\frac{n_z}{2}$, 32) \\
 & upsample, of size $2$ & ($n_{t}$, $n_x$, $n_y$, $n_z$, 32) \\
 & residual block, 16 filters of size $3$ & ($n_{t}$, $n_x$, $n_y$, $n_z$, 16) \\
\hline
Output & conv block, 1 filter of size $3$ & ($n_{t}$, $n_x$, $n_y$, $n_z$, 1) \\
\hline
\end{tabular}
\end{center}
\end{table}

\subsection{Surrogate model limitations and potential extensions}
\label{sec:limitations}

As is typically the case with deep learning-based surrogate models, our recurrent transformer U-Net model is only applicable for cases `covered' by the training runs. In the current setting, all training runs include a fixed number of injection wells at prescribed locations, with each well injecting at a constant rate of 1~Mt/year for 50~years. The physical dimensions of the geomodel, grid structure, boundary conditions, relative permeability and capillary pressure curves are also specified. With the current model, this setup cannot be altered during online evaluations. The network is trained for geological models that vary over the ranges described in Section~\ref{sec:SEAM_realizations}. Any new realizations considered in subsequent applications (e.g., during data assimilation) should, therefore, be within these ranges. This means that if the prior parameter ranges are reconsidered during history matching, the surrogate model must be retrained to reflect the new ranges. We note that this may not require a full retraining, since approaches such as transfer learning can be applied.

Although the large-scale geomodel considered in this study is faulted, the corresponding simulation grid is logically rectangular even though the cells themselves are nonrectangular. This means internal cell $(i,j,k)$ always connects to cells $(i\pm 1,j,k)$ in the $x$ direction, to cells $(i,j\pm 1,k)$ in $y$, and to cells $(i,j,k\pm 1)$ in $z$. Faulted models with significant displacement are often represented on grids with (what are referred to as) non-neighbor connections. In this case cell $(i,j,k)$ could connect to, e.g., cell $(i+1,j,k+\delta_k)$, where $\delta_k$ is an integer that captures the vertical displacement due to the fault. Our current model is not immediately applicable for this situation, but we believe it could be readily adapted to treat these non-neighbor connections. This is because we do not explicitly rely on any particular grid connectivity structure, so the model can be trained to learn the patterns that appear in the training runs, even if these derive from models with non-neighbor connections. In such cases it may be beneficial to include additional input channels, including one defining the depth to each cell center. The treatment of such models is beyond the scope of this paper, but it would be an interesting topic for future investigations.

%% file: Section_4.tex
In this section, we first compare error statistics, for varying numbers of training samples, for the new recurrent transformer U-Net surrogate model to that of an existing recurrent R-U-Net model. Detailed CO$_2$ saturation and pressure fields from surrogate predictions are then compared to the corresponding simulation results. Qualitatively different fault leakage scenarios are considered. Finally, the importance of the metaparameters and geomodel realizations on the CO$_2$ footprint in the target aquifer is quantified using the surrogate model in combination with a global sensitivity analysis.

\subsection{Surrogate model evaluation and comparison}
\label{sec:surrogate_comparisons}

The first step in this assessment entails the generation of 4000 geomodel realizations of the overall domain using the procedure described in Section~\ref{sec:SEAM_realizations}. Flow simulation is then performed for each realization with the effective rock compressibility treatment discussed in Section~\ref{sec:simulation_setup}. The pressure and saturation fields in the domain of interest are collected at $n_t$ = 10 discrete time steps, and these fields are used to train surrogate models. In this study, we consider the use of 1000, 2000, and 4000 such simulation runs as training samples. Both the new recurrent transformer U-Net model and our previous recurrent R-U-Net model are trained with these samples. A new test set of $n_e$ = 500 geomodel realizations of the overall domain is then generated and simulated. These results are used to evaluate the performance of the surrogate models. 

Because of their different scales and characteristics, we use different error definitions for the CO$_2$ saturation and pressure surrogate predictions. For saturation, consistent with the approach of \citet{tang2025graph}, the mean absolute error (MAE) is computed for saturation only over cells in the plume region, i.e., where either the simulated saturation or the surrogate prediction exceeds a threshold value $\epsilon$. This treatment is appropriate because the domain includes many cells with zero saturation, especially at early time, before the plume has spread. Including these zero saturation cells can lead to an under-estimation of saturation error. The MAE in CO$_2$ saturation, for test sample $i$ ($i = 1, \ldots, 500$), is denoted $\delta_S^i$ and computed as 
\begin{equation} \label{surr_error_s}
\delta_S^i = \frac{1}{n_i n_t} \sum_{j=1}^{n_i} \sum_{t=1}^{n_t} \left|(\hat{S}_d)_{i,j}^t - (S_d)_{i,j}^t \right|, \quad \text{for } (S_d)_{i,j}^t > \epsilon \ \ \text{or} \ \ (\hat{S}_d)_{i,j}^t > \epsilon.
\end{equation}
Here $(S_d)_{i,j}^t$ and $(\hat{S}_d)_{i,j}^t$ are CO$_2$ saturation predictions from the flow simulation and surrogate model, for test case $i$, cell $j$, and surrogate model time step $t$, and $n_i$ is the number of cells in the domain for which $(S_d)_{i,j}^t > \epsilon$ or $(\hat{S}_d)_{i,j}^t > \epsilon$. Note that $n_i$ varies for different samples and time steps. We set $\epsilon=0.02$, which is consistent with the standard deviation of saturation measurement error considered in Section~\ref{History Matching}. \citet{tang2025graph} showed that saturation error results computed in this way (for their graph neural network surrogate) were not very sensitive to the specific value of $\epsilon$.

The pressure field varies from its initial state throughout the domain even at early time, so this error is computed for all cells. Consistent with \citet{han2023surrogate, han2025accelerated}, relative pressure error for test sample $i$, denoted $\delta_p^i$, is calculated as 
\begin{equation} \label{surr_error_p}
\delta_p^i = \frac{1}{n_dn_t} \sum_{j=1}^{n_d} \sum_{t=1}^{n_t} \frac{| (\hat{p}_d)_{i,j}^t - (p_d)_{i,j}^t |}{(p_d)_{i,\mathrm{max}}^t - (p_d)_{i,\mathrm{min}}^t}.
\end{equation}
Here $(p_d)_{i,j}^t$ and $(\hat{p}_d)_{i,j}^t$ are pressure results from flow simulation and the surrogate model for test case $i$, cell $j$, and surrogate output time step $t$, and $(p_d)_{i,\mathrm{max}}^t$ and $(p_d)_{i,\mathrm{min}}^t$ are the maximum and minimum simulated pressure for all cells in test case $i$ at time step $t$. Note that normalization of the error by the simulated pressure itself will result in artificially small pressure errors, since the initial pressure is already 18~MPa in the target aquifer, and pressure generally increases as the simulation proceeds. 

\begin{figure}[!ht]
\centering   
\subfloat[Saturation mean absolute errors]{\label{error_s}\includegraphics[width = 175mm]{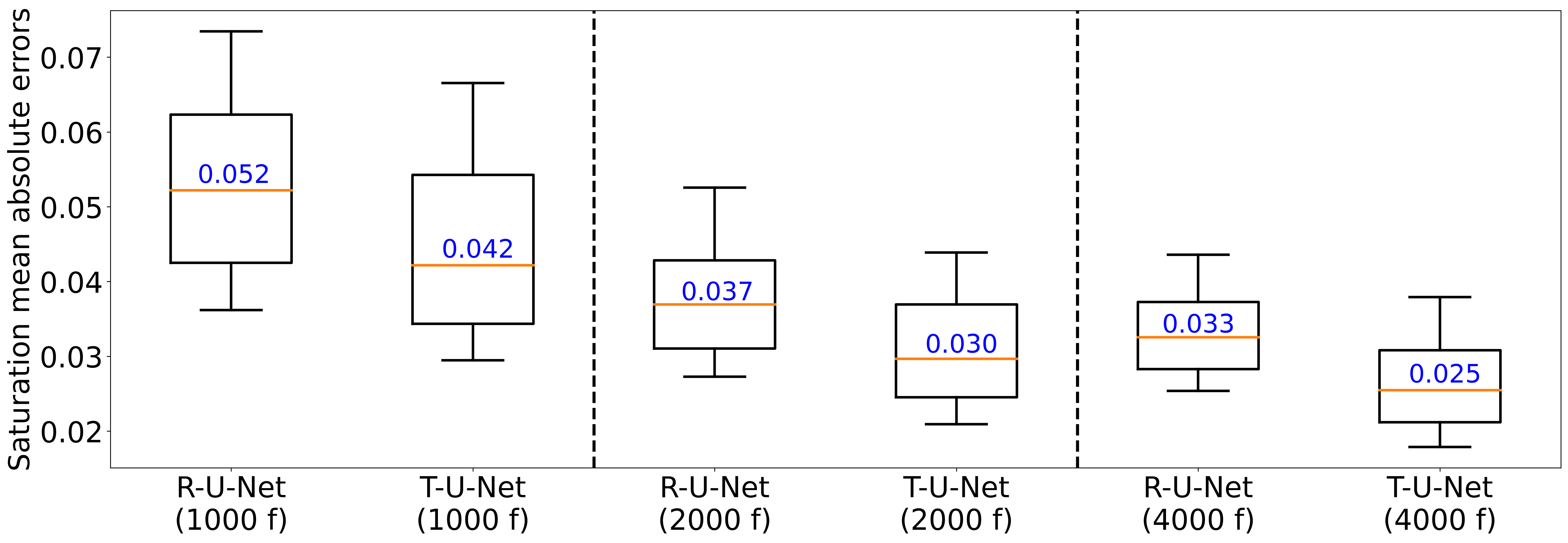}}
\\[1ex]
\subfloat[Pressure relative errors]{\label{error_p}\includegraphics[width = 175mm]{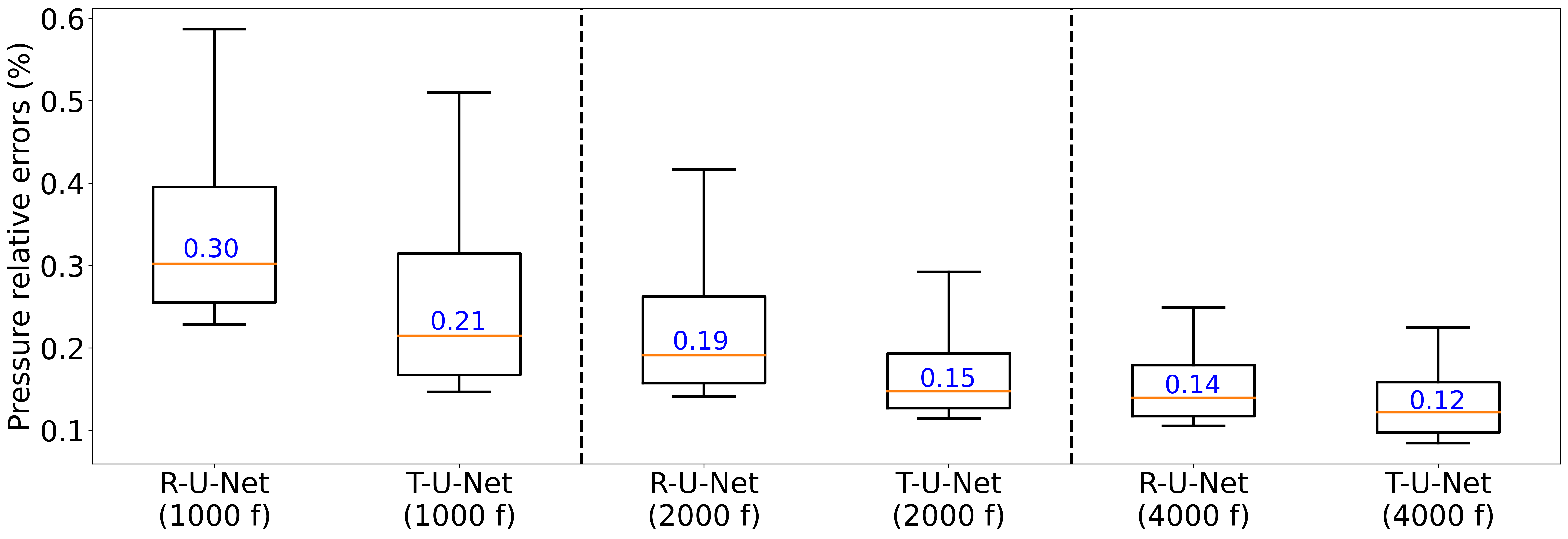}} \\
\caption{Comparison between the recurrent residual U-Net (R-U-Net) model and recurrent transformer U-Net (T-U-Net) model for (a) saturation mean absolute errors and (b) pressure relative errors, evaluated over a test set of 500 new geomodel realizations. Results are shown for both surrogate models trained with 1000 flow simulations (1000~f), 2000 flow simulations (2000~f), and 4000 flow simulations (4000~f). Boxes display P$_{90}$, P$_{75}$, P$_{50}$, P$_{25}$ and P$_{10}$ errors.}
\label{errors_box}
\end{figure}

The saturation and pressure errors for both the existing recurrent residual U-Net model and the new recurrent transformer U-Net model, for varying numbers of training samples, are displayed in Fig.~\ref{errors_box}. The box plots show the error distributions, with the lower and upper edges of each box corresponding to the 25th (P$_{25}$) and 75th (P$_{75}$) percentile errors. The median (P$_{50}$) error is indicated by the solid red line within each box, while the `whiskers' outside the boxes indicate the 10th (P$_{10}$) and 90th (P$_{90}$) percentile errors. These error quantities are computed over cells in the target, middle and upper aquifers and in the faults. Cells in the caprock between aquifers are not included in these calculations. The recurrent transformer U-Net model is seen to consistently provide predictions with smaller saturation MAE (Fig.~\ref{errors_box}(a)) and smaller pressure relative error (Fig.~\ref{errors_box}(b)). This behavior is observed for all training sample sizes. We also see that error consistently decreases with increasing numbers of training samples for both models, as would be expected. 

With 1000 training samples, the median saturation MAE and pressure relative error using the recurrent transformer U-Net model are 0.042 and 0.21\%, respectively. These error levels may be acceptable for many applications, but since we are interested here in detailed fault leakage effects, we will use the recurrent transformer U-Net trained with 4000 samples in the results presented later in this paper. These include the detailed surrogate model predictions in Section~\ref{sec:surrogate_individuals}, the global sensitivity analysis in Section~\ref{sec:gsa}, and data assimilation results in Section~\ref{History Matching} and SI.

\begin{figure}[!ht]
\centering   
{\includegraphics[width = 175mm]{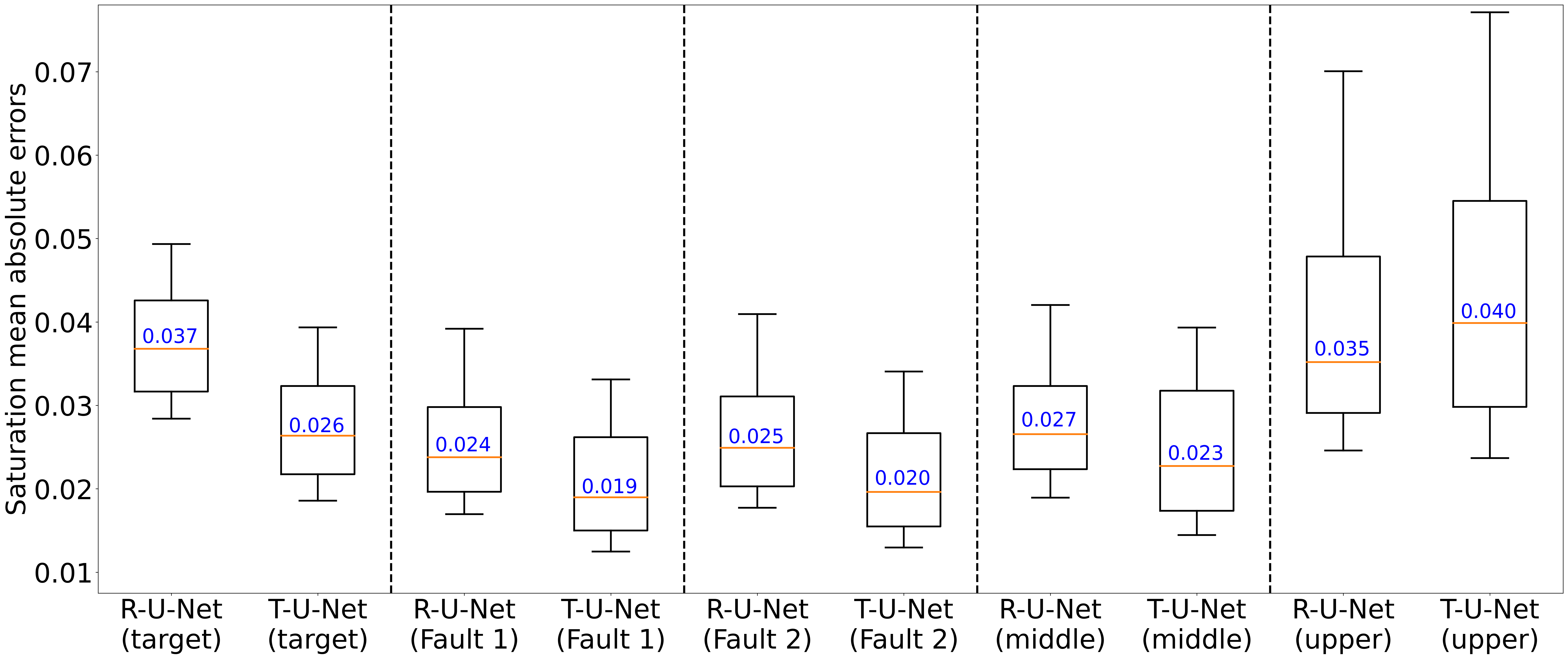}}
\caption{Comparison between the recurrent residual U-Net (R-U-Net) model and the recurrent transformer U-Net (T-U-Net) model for saturation mean absolute errors evaluated over different model domains. Both models are trained with 4000 flow simulation runs and evaluated with a test set of 500 new geomodel realizations. Boxes display P$_{90}$, P$_{75}$, P$_{50}$, P$_{25}$ and P$_{10}$ errors.}
\label{errors_box_domain}
\end{figure}

Saturation errors for individual model domains are shown in Fig.~\ref{errors_box_domain}. For these results, Eq.~\ref{surr_error_s} is evaluated separately for each region noted in the figure. These results are all for the case of 4000 training samples. Consistent with the aggregate results in Fig.~\ref{errors_box}(a), the recurrent transformer U-Net model is seen to outperform the recurrent residual U-Net in all regions except the upper aquifer. The most performance improvement is observed in the target aquifer region, which is where the actual injection occurs. In this model domain, the median saturation MAE is reduced from 0.037 with the recurrent residual U-Net to 0.026 with the recurrent transformer U-Net. The recurrent transformer U-Net is less accurate than the recurrent residual U-Net in the upper aquifer region. Only about one half of the models have nonzero saturation values in the upper aquifer region, so this portion of the domain is not heavily weighted in the training loss function. The recurrent transformer U-Net results for this region could be improved, if necessary, by appropriately weighting the loss function. 

\begin{figure}[!ht]
\centering   
\subfloat[Realization 1 (simulation)]{\label{sim_1}\includegraphics[width=48mm]{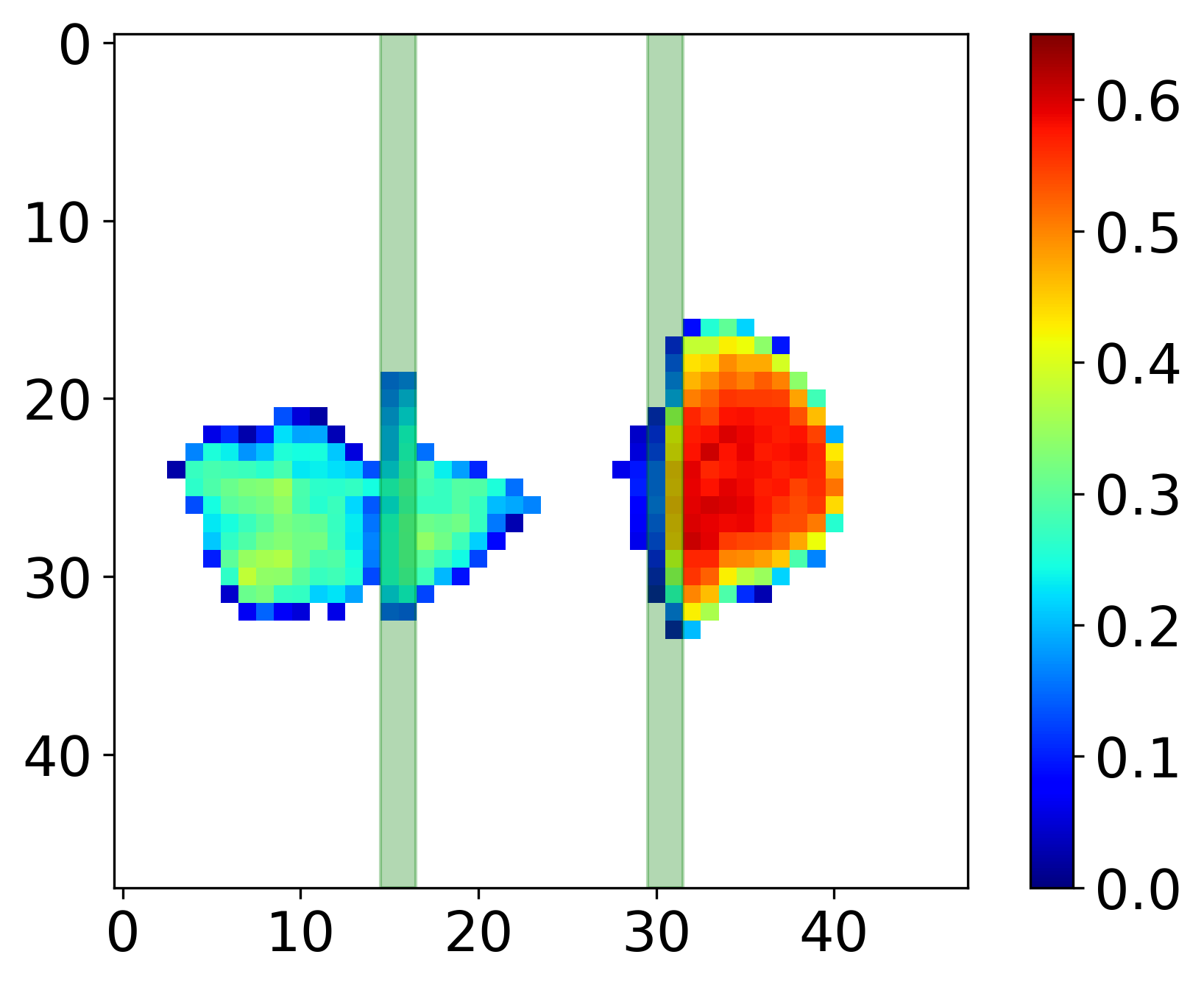}}
\hspace{9mm}
\subfloat[Realization 1 (T-U-Net)]{\label{surr_trans_1}\includegraphics[width=48mm]{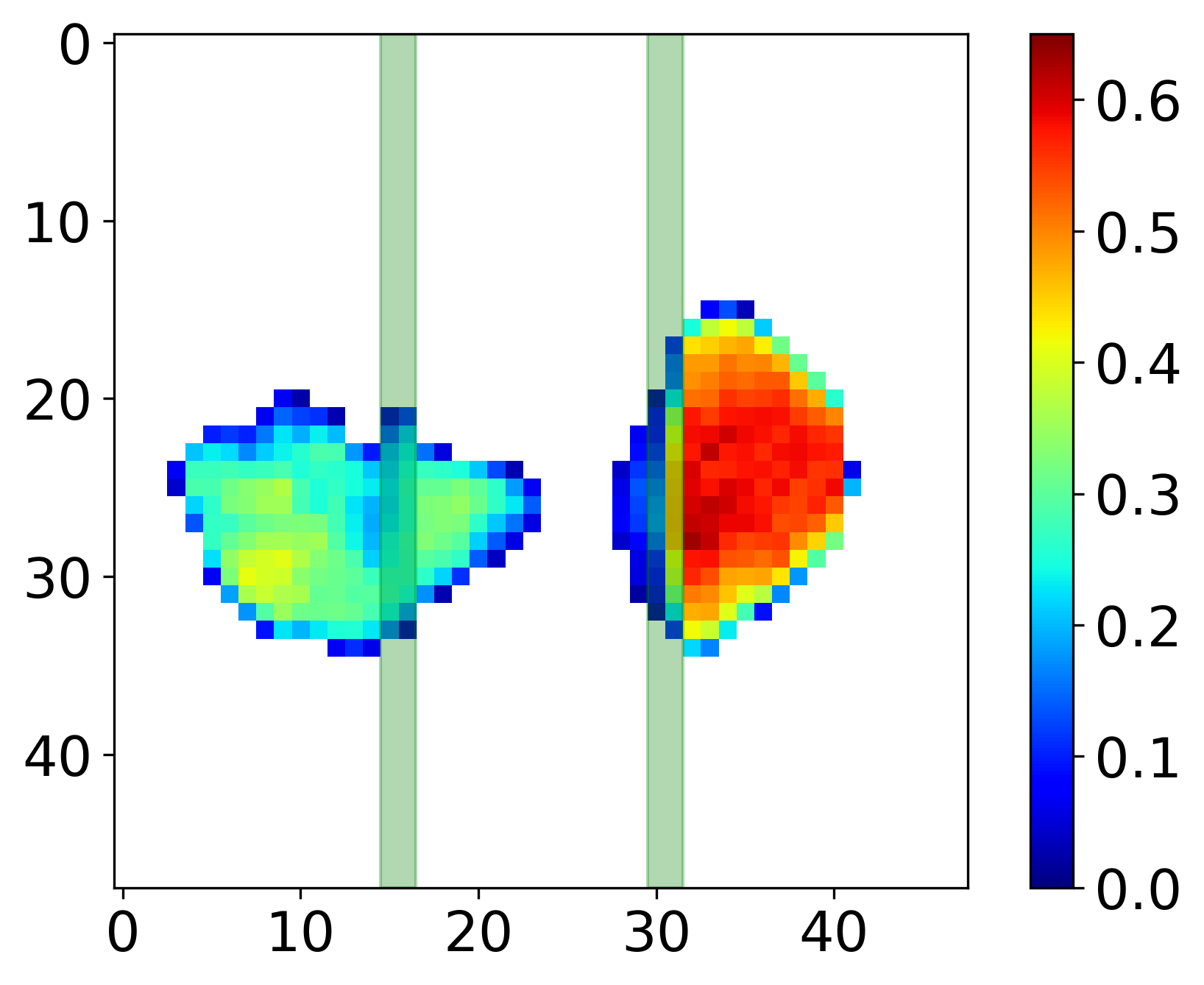}}
\hspace{9mm}
\subfloat[Realization 1 (R-U-Net)]{\label{surr_res_1}\includegraphics[width=48mm]{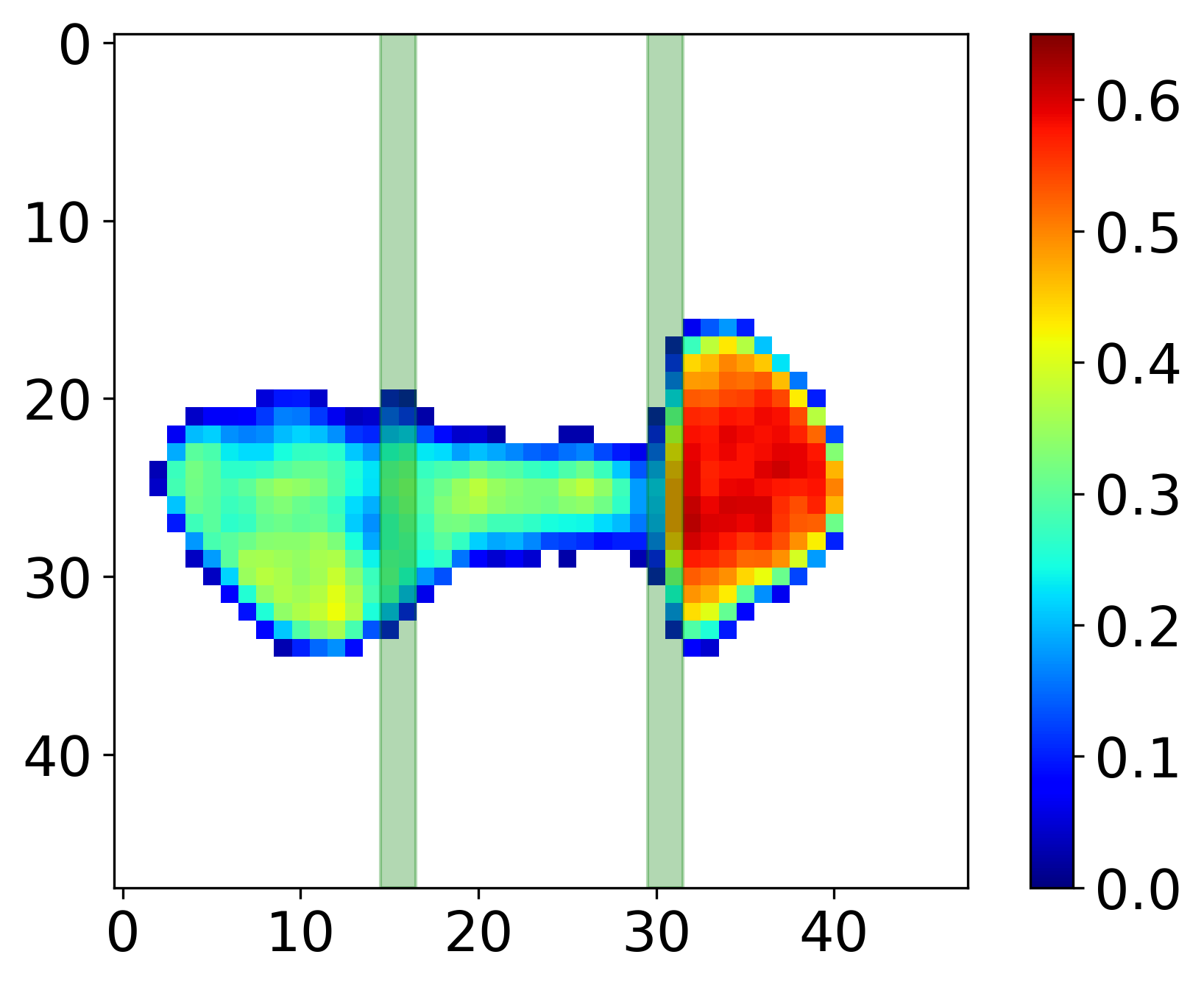}} \\[1ex]
\subfloat[Realization 2 (simulation)]{\label{sim_2}\includegraphics[width=48mm]{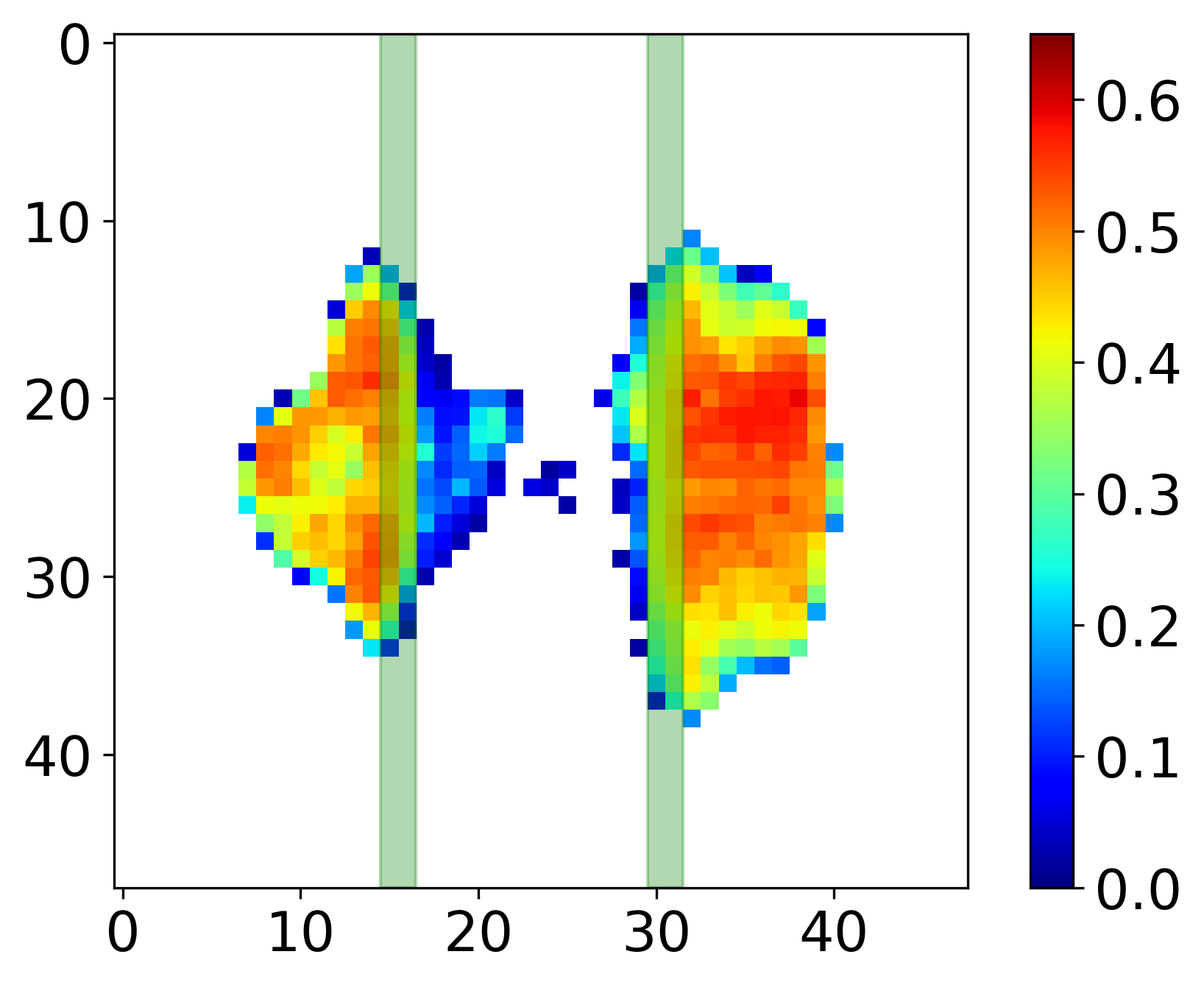}}
\hspace{9mm}
\subfloat[Realization 2 (T-U-Net)]{\label{surr_trans_2}\includegraphics[width=48mm]{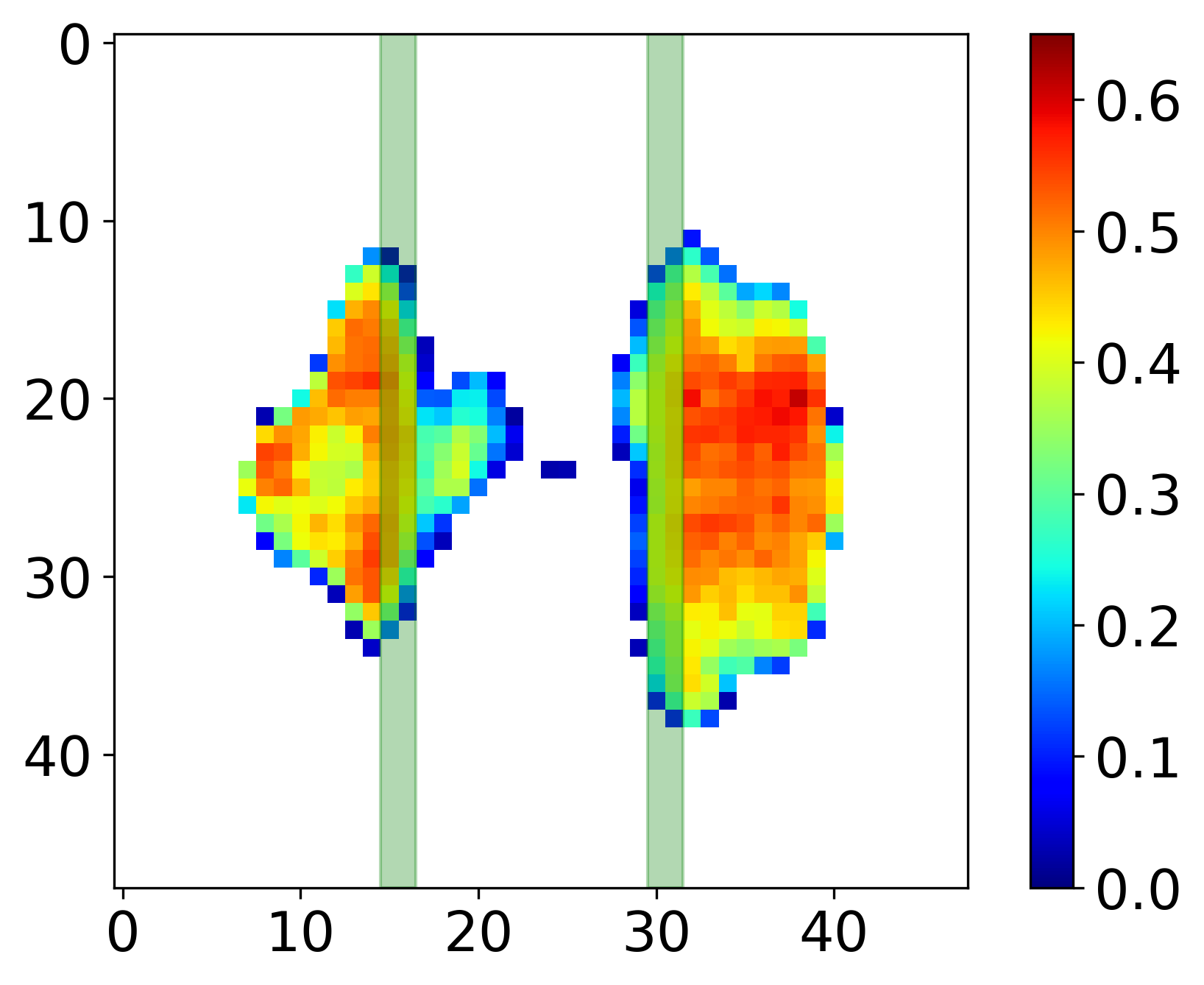}}
\hspace{9mm}
\subfloat[Realization 2 (R-U-Net)]{\label{surr_res_2}\includegraphics[width=48mm]{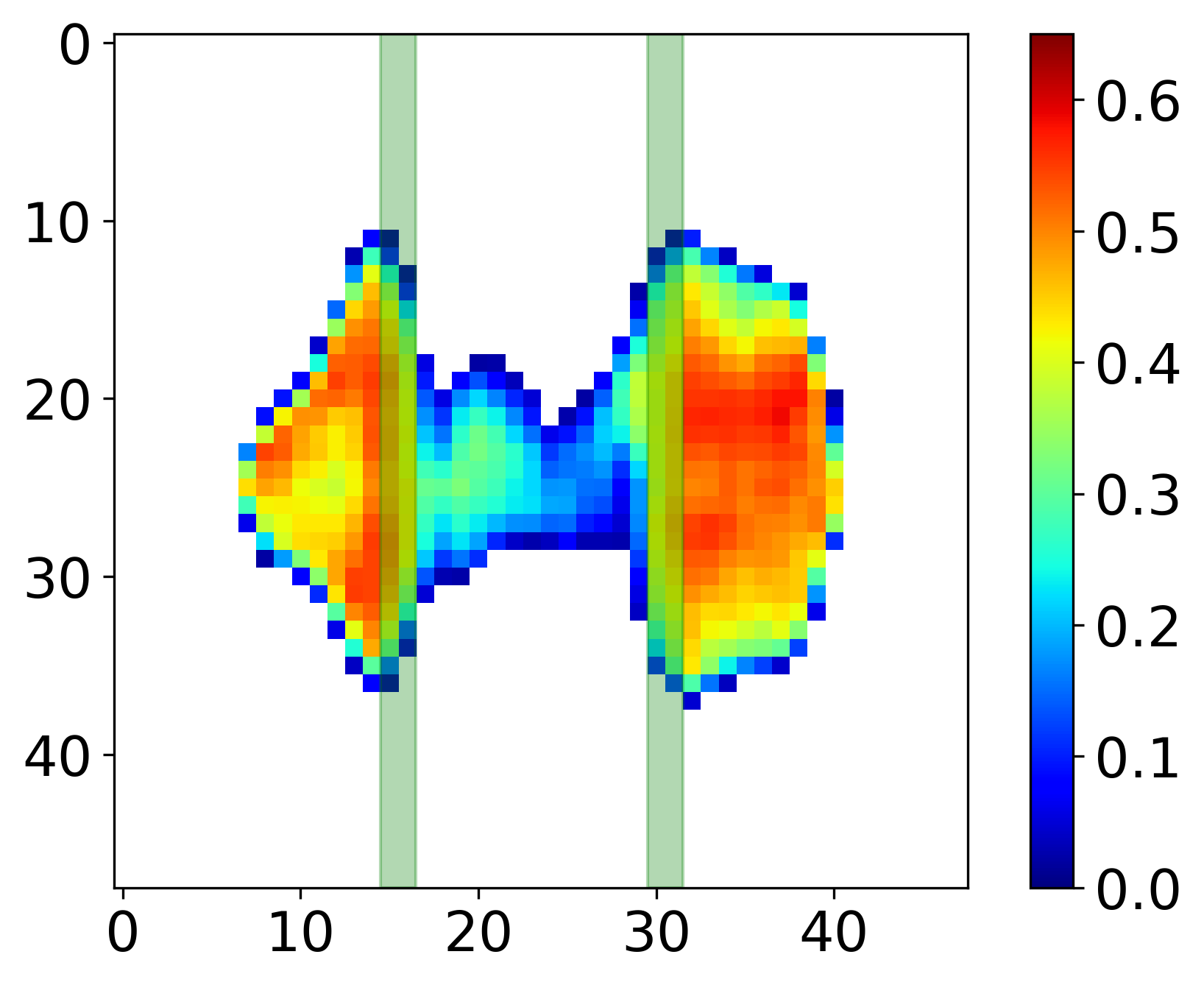}} \\[1ex]
\subfloat[Realization 3 (simulation)]{\label{sim_3}\includegraphics[width=48mm]{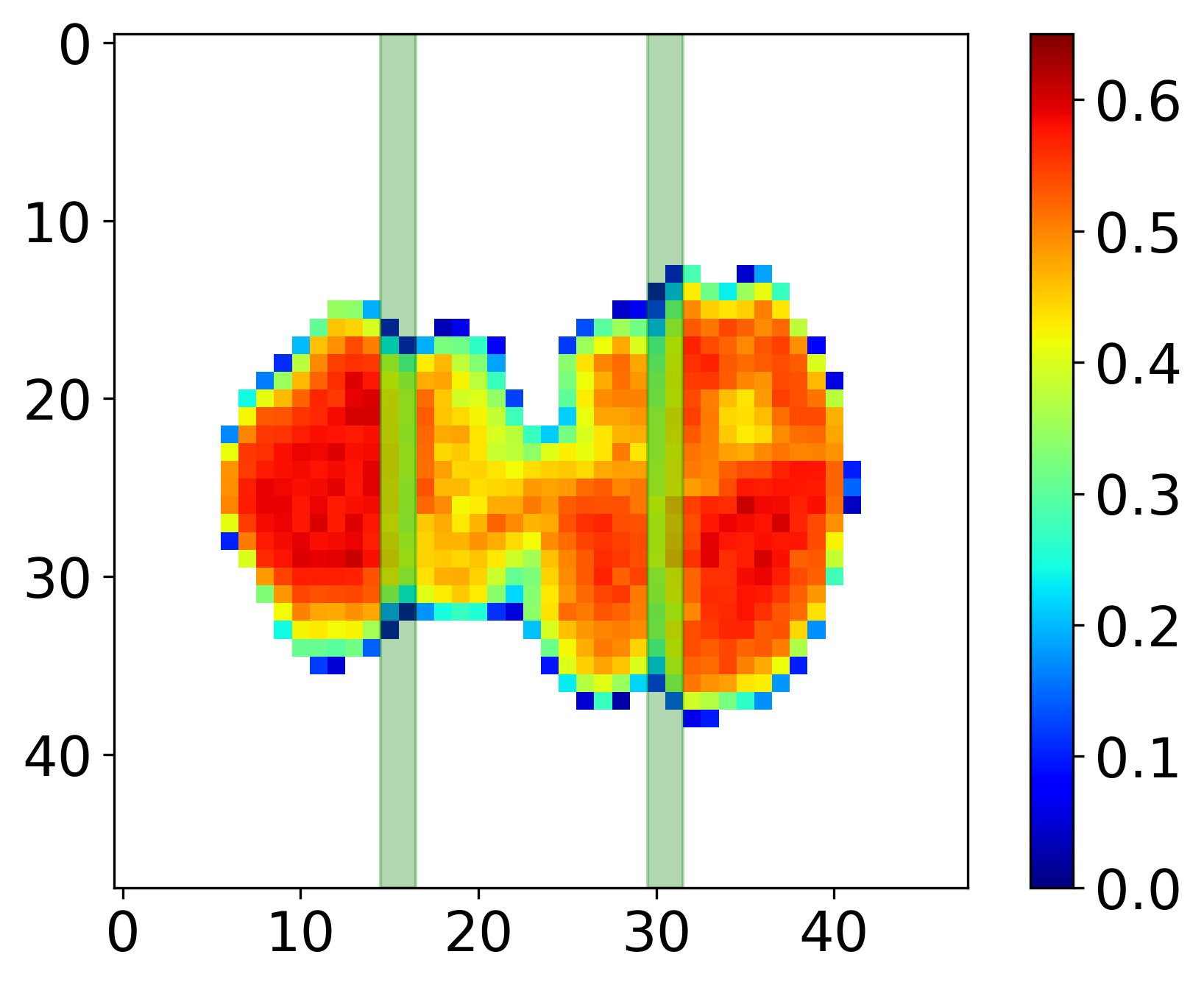}}
\hspace{9mm}
\subfloat[Realization 3 (T-U-Net)]{\label{surr_trans_3}\includegraphics[width=48mm]{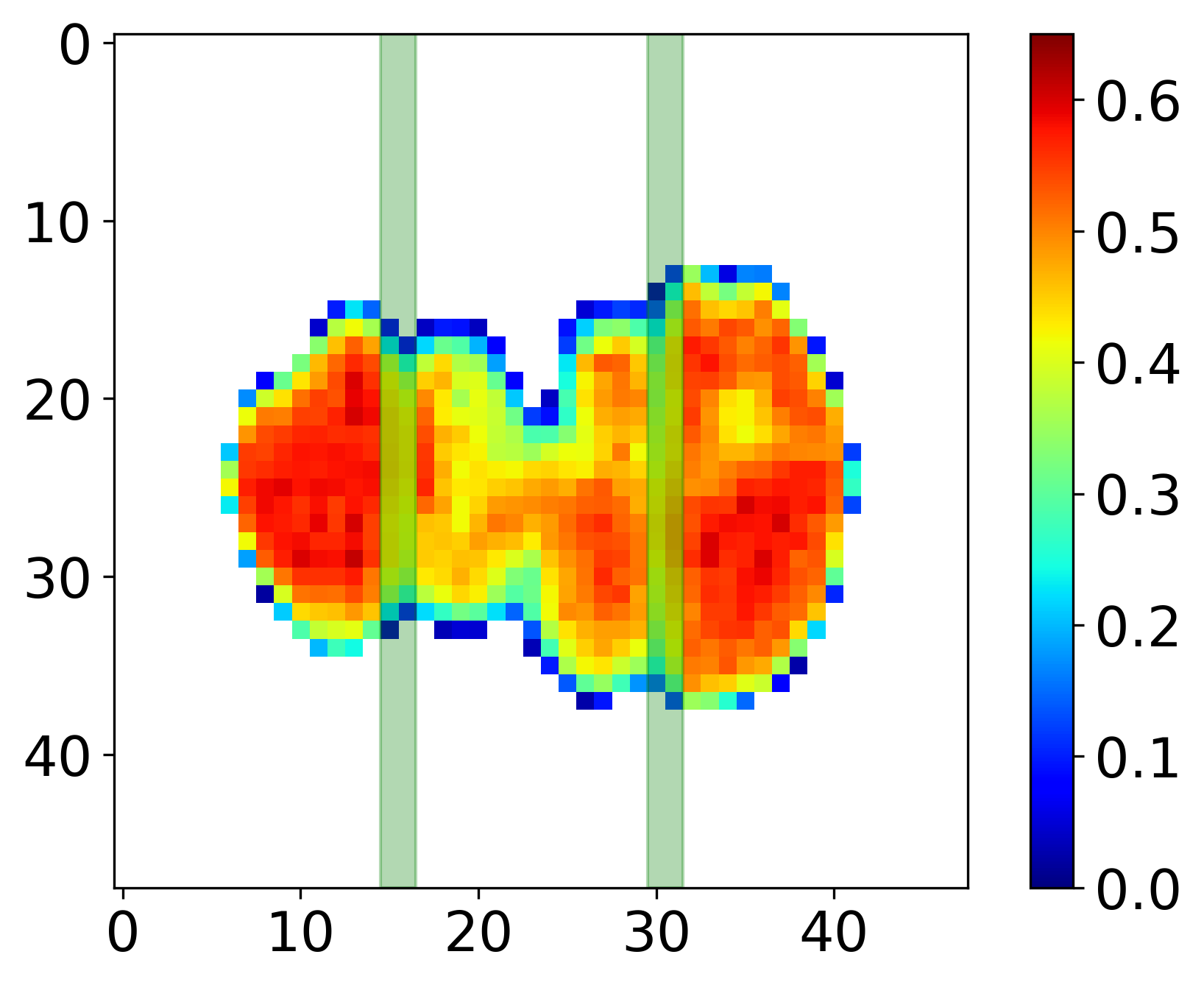}}
\hspace{9mm}
\subfloat[Realization 3 (R-U-Net)]{\label{surr_res_3}\includegraphics[width=48mm]{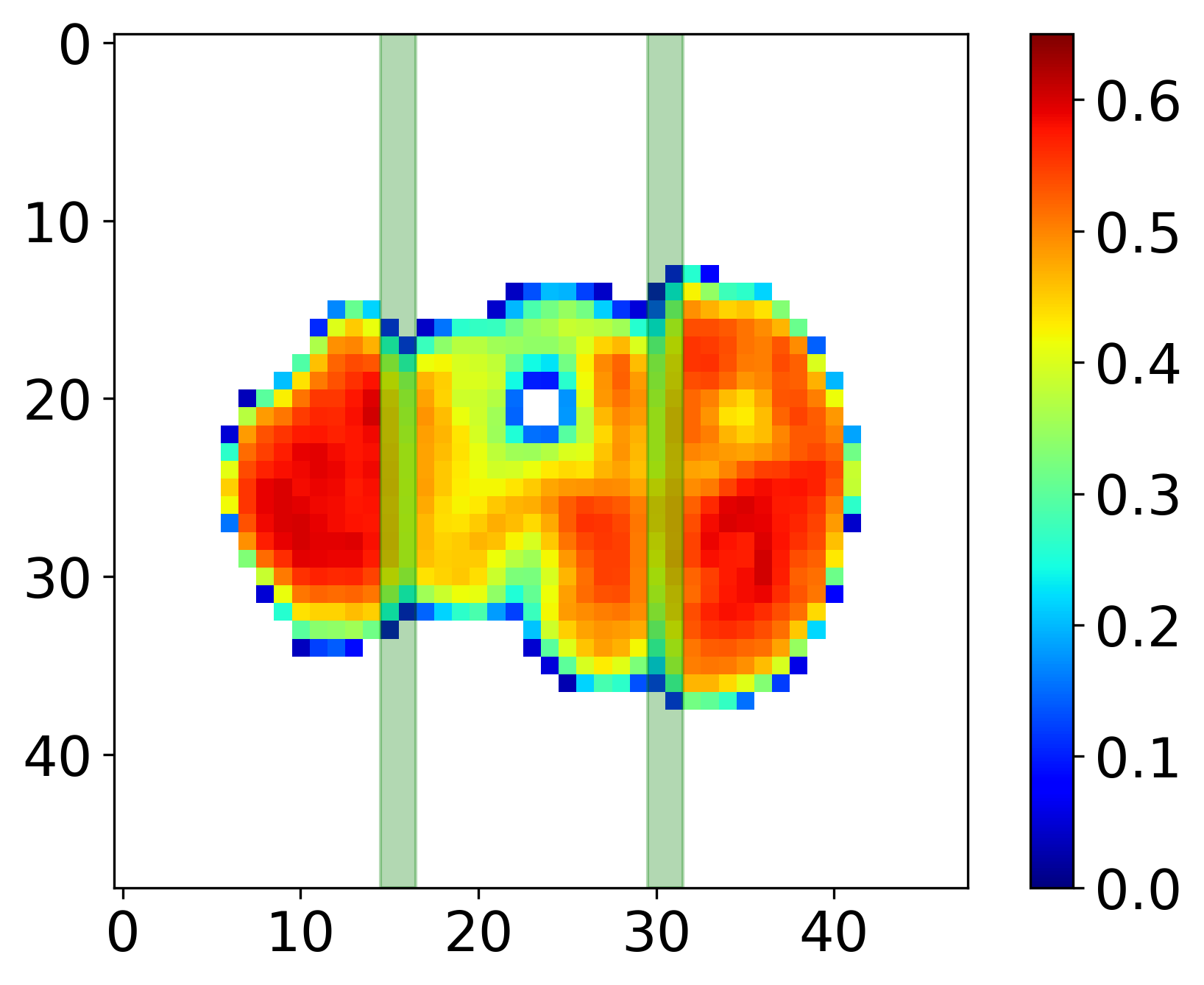}} \\[1ex]
\caption{CO$_2$ saturation for the top layers of the target aquifer region at the end of injection (50~years) from flow simulation (first column), recurrent transformer U-Net surrogate model (second column), and recurrent residual U-Net surrogate model (third column) for three geomodel realizations.}
\label{S:saturation_comparisons}
\end{figure}

The new surrogate model shows other advantages relative to our existing recurrent residual U-Net model. Namely, for some challenging cases involving flow across faults, it can provide accurate results when the earlier model gives saturation predictions with qualitative error. Three examples are shown in Fig.~\ref{S:saturation_comparisons}. Each row corresponds to results for a particular realization. The first column shows the reference flow simulation saturation field, and the second and third columns display results from the two surrogate models. The saturation field in the top layers of the target aquifer, at 50~years, is shown in all cases. For realizations~1 and 2, the reference simulation result shows some CO$_2$ between the two faults, but the plume does not span from one fault to the other. This behavior is captured by the recurrent transformer U-Net, but not by the recurrent residual U-Net. For realization~3, consistent with the simulation result, the plume spans the region between the faults in both surrogate models. The plume shape is, however, captured more accurately by the new surrogate model.

As discussed in Section~\ref{sec:SEAM_realizations}, the training realizations considered in this work include variation in fault permeability with depth (for both faults). The test-set results until now also included this variation. In many geomodels, however, faults are assigned a single permeability value, so it useful to evaluate the accuracy of our surrogate model for such cases. To conduct this assessment, we generate and simulate $n_r = 50$ geomodel realizations with depth-invariant fault permeability (i.e., $k_{f1}^{tm} = k_{f1}^{mu}$ and $k_{f2}^{tm} = k_{f2}^{mu}$). We then apply the previously trained recurrent transformer U-Net surrogate model (which was trained using 4000 flow simulation runs with $k_{f1}^{tm} \neq k_{f1}^{mu}$ and $k_{f2}^{tm} \neq k_{f2}^{mu}$) for these $n_r = 50$ new geomodel realizations. The median MAE saturation error for this new test set is 0.027, and the median pressure relative error is 0.14\%. These errors are similar to those for the original test set shown in Fig.~\ref{errors_box}, where the median errors are 0.025 (saturation) and 0.12\% (pressure). This indicates that the surrogate model is robust in terms of our treatment of fault permeability variation with depth. 

\subsection{Surrogate model predictions for different leakage scenarios}
\label{sec:surrogate_individuals}

\begin{figure}[!ht]
\centering   
\subfloat[Realization 4 (sim)]{\label{sim_1_20_years}\includegraphics[width=50mm]{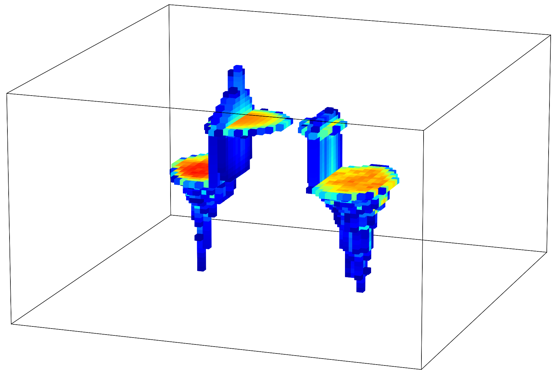}}
\hspace{4mm}
\subfloat[Realization 5 (sim)]{\label{sim_2_20_years}\includegraphics[width=50mm]{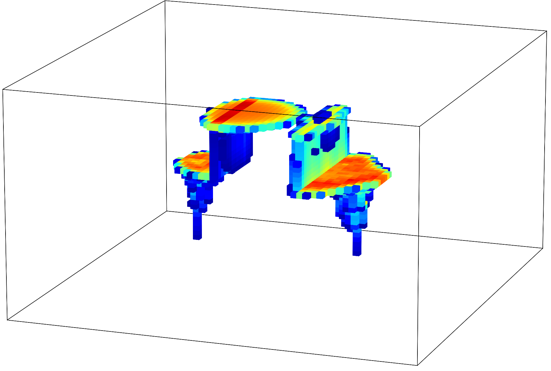}}
\hspace{4mm}
\subfloat[Realization 6 (sim)]{\label{sim_3_20_years}\includegraphics[width=50mm]{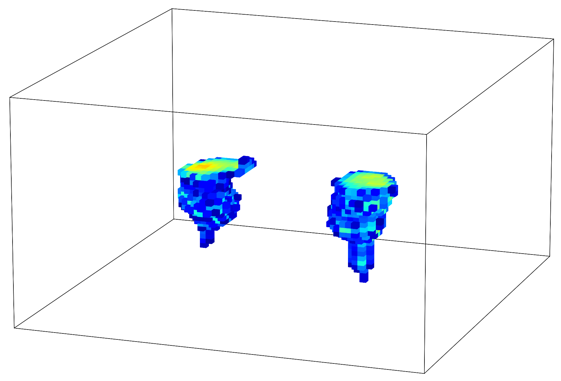}}
\hspace{1mm}
\includegraphics[width=8mm]{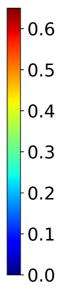}\\[1ex]
\subfloat[Realization 4 (surr)]{\label{surr_1_20_years}\includegraphics[width=50mm]{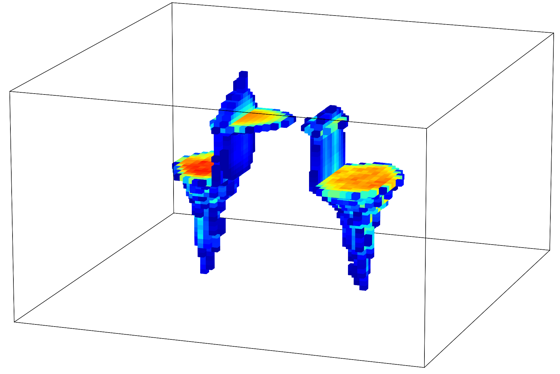}}
\hspace{4mm}
\subfloat[Realization 5 (surr)]{\label{surr_2_20_years}\includegraphics[width=50mm]{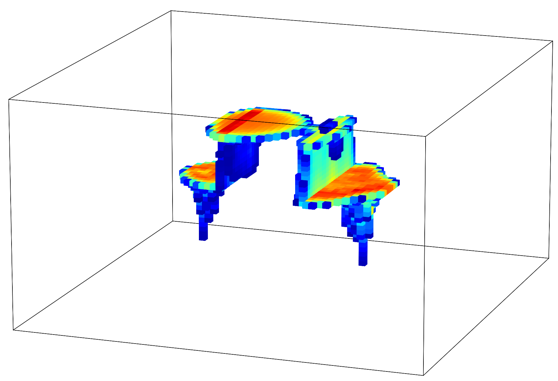}}
\hspace{4mm}
\subfloat[Realization 6 (surr)]{\label{surr_3_20_years}\includegraphics[width=50mm]{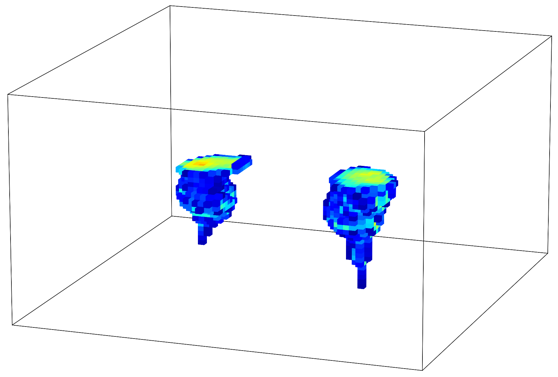}}
\hspace{1mm}
\includegraphics[width=8mm]{Sw_Scale.png}\\[1ex]
\caption{CO$_2$ saturation in the full domain at 20~years from flow simulation (upper row) and the recurrent transformer U-Net surrogate model (lower row) for three new geomodel realizations.}
\label{S:saturation_20_years}
\end{figure}

\begin{figure}[!ht]
\centering   
\subfloat[Realization 4 (sim)]{\label{sim_1_50_years}\includegraphics[width=50mm]{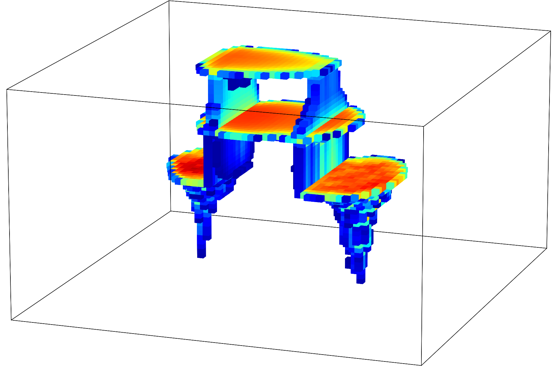}}
\hspace{4mm}
\subfloat[Realization 5 (sim)]{\label{sim_2_50_years}\includegraphics[width=50mm]{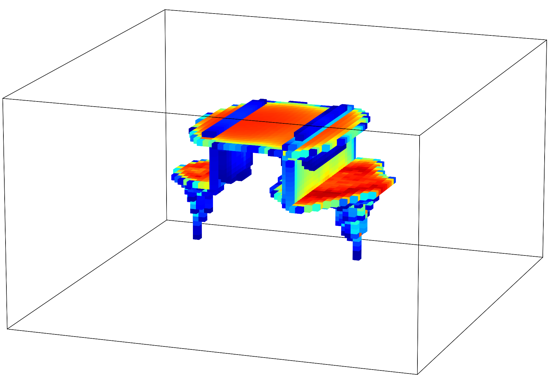}}
\hspace{4mm}
\subfloat[Realization 6 (sim)]{\label{sim_3_50_years}\includegraphics[width=50mm]{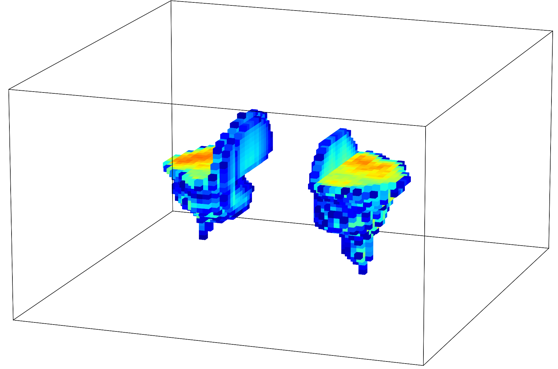}}
\hspace{1mm}
\includegraphics[width=8mm]{Sw_Scale.png}\\[1ex]
\subfloat[Realization 4 (surr)]{\label{surr_1_50_years}\includegraphics[width=50mm]{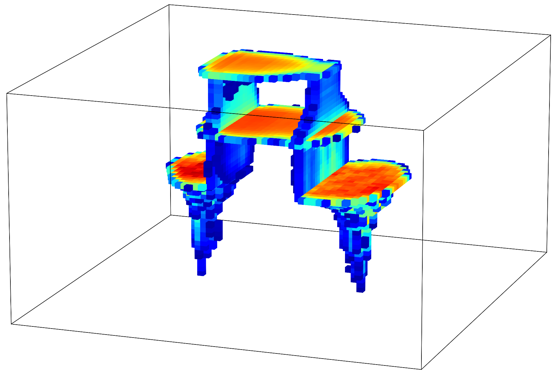}}
\hspace{4mm}
\subfloat[Realization 5 (surr)]{\label{surr_2_50_years}\includegraphics[width=50mm]{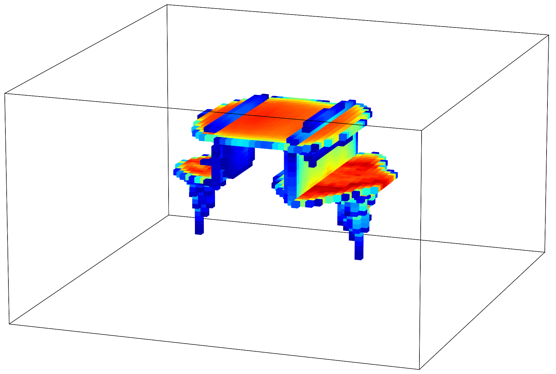}}
\hspace{4mm}
\subfloat[Realization 6 (surr)]{\label{surr_3_50_years}\includegraphics[width=50mm]{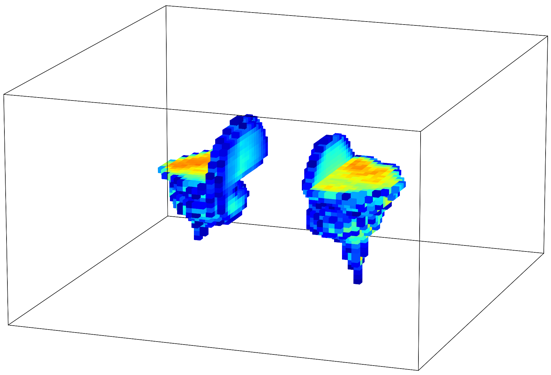}}
\hspace{1mm}
\includegraphics[width=8mm]{Sw_Scale.png}\\[1ex]
\caption{CO$_2$ saturation in the full domain at the end of injection (50~years) from flow simulation (upper row) and the recurrent transformer U-Net surrogate model (lower row) for three new geomodel realizations.}
\label{S:saturation_50_years}
\end{figure}

We now present detailed saturation and pressure fields for three test-case geomodels (realizations~4, 5 and 6). These cases correspond to a range of flow and fault leakage behaviors. The full 3D CO$_2$ saturation fields in the full domain at 20~years and at the end of the injection period (50~years) for the three models are shown in Figs.~\ref{S:saturation_20_years} and \ref{S:saturation_50_years}. The upper rows in the figures represent the reference simulation results and the lower rows show surrogate model predictions. The two sets of results show close visual correspondence for all three cases at both time steps. The saturation MAEs for realizations~4, 5 and 6 are 0.033, 0.022 and 0.029, respectively, which are all relatively near the median MAE of 0.025. Thus, in terms of accuracy, these results are representative of the full test set.

It is noteworthy that the new surrogate model is able to capture the various leakage scenarios, as this capability will be important for history matching. In realization~4, the faults are relatively permeable ($k_{f1}^{tm} = 160.2$~md, $k_{f1}^{mu} = 12.1$~md, $k_{f2}^{tm} = 116.6$~md, $k_{f2}^{mu} = 9.5$~md; recall superscript `tm' indicates the portion of the fault between the target and middle aquifers and `mu' the part between the middle and upper aquifers), so leakage into both the middle and upper aquifers occurs. For realization~5, the lower portions of the faults are relatively permeable but the upper portions are nearly impermeable ($k_{f1}^{tm} = 266.3$~md, $k_{f1}^{mu} = 0.15$~md, $k_{f2}^{tm} = 18.8$~md, $k_{f2}^{mu} = 0.57$~md). As a result of the low $k_{f1}^{mu}$ and $k_{f2}^{mu}$, we do not see leakage from the middle to upper aquifers in this case. For realization~6, the faults are of low permeability ($k_{f1}^{tm} = k_{f1}^{mu} = 0.56$~md and $k_{f2}^{tm} = k_{f2}^{mu} = 0.45$~md). As a result, no leakage is observed into the middle and upper aquifer, and the plumes in the target aquifer are large. All these qualitatively different flow behaviors are accurately captured by the surrogate model.

\begin{figure}[!ht]
\centering   
\subfloat[Realization 4 (sim)]{\label{sim_p_1_50_years}\includegraphics[width=58mm]{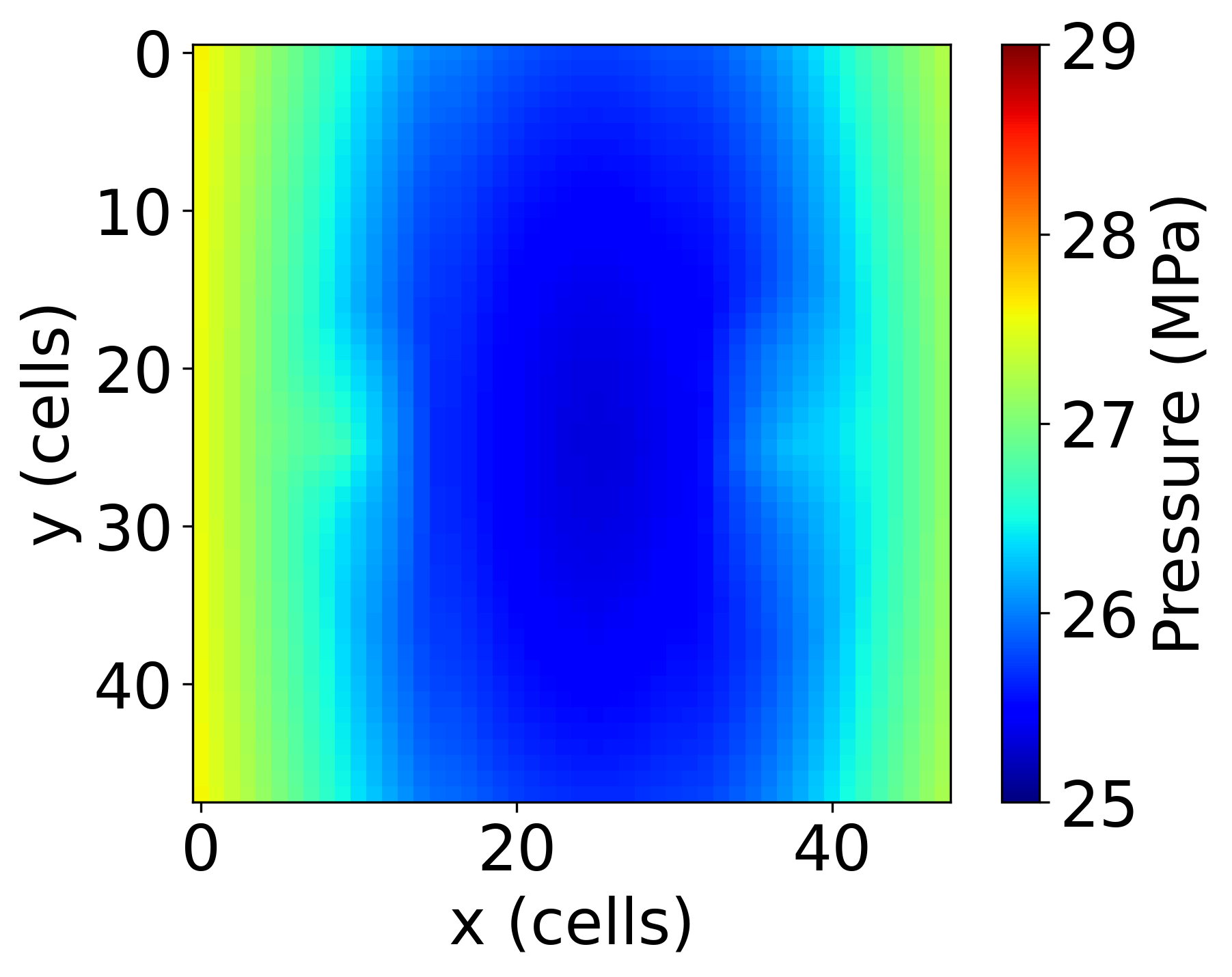}}
\hspace{2mm}
\subfloat[Realization 5 (sim)]{\label{sim_p_2_50_years}\includegraphics[width=58mm]{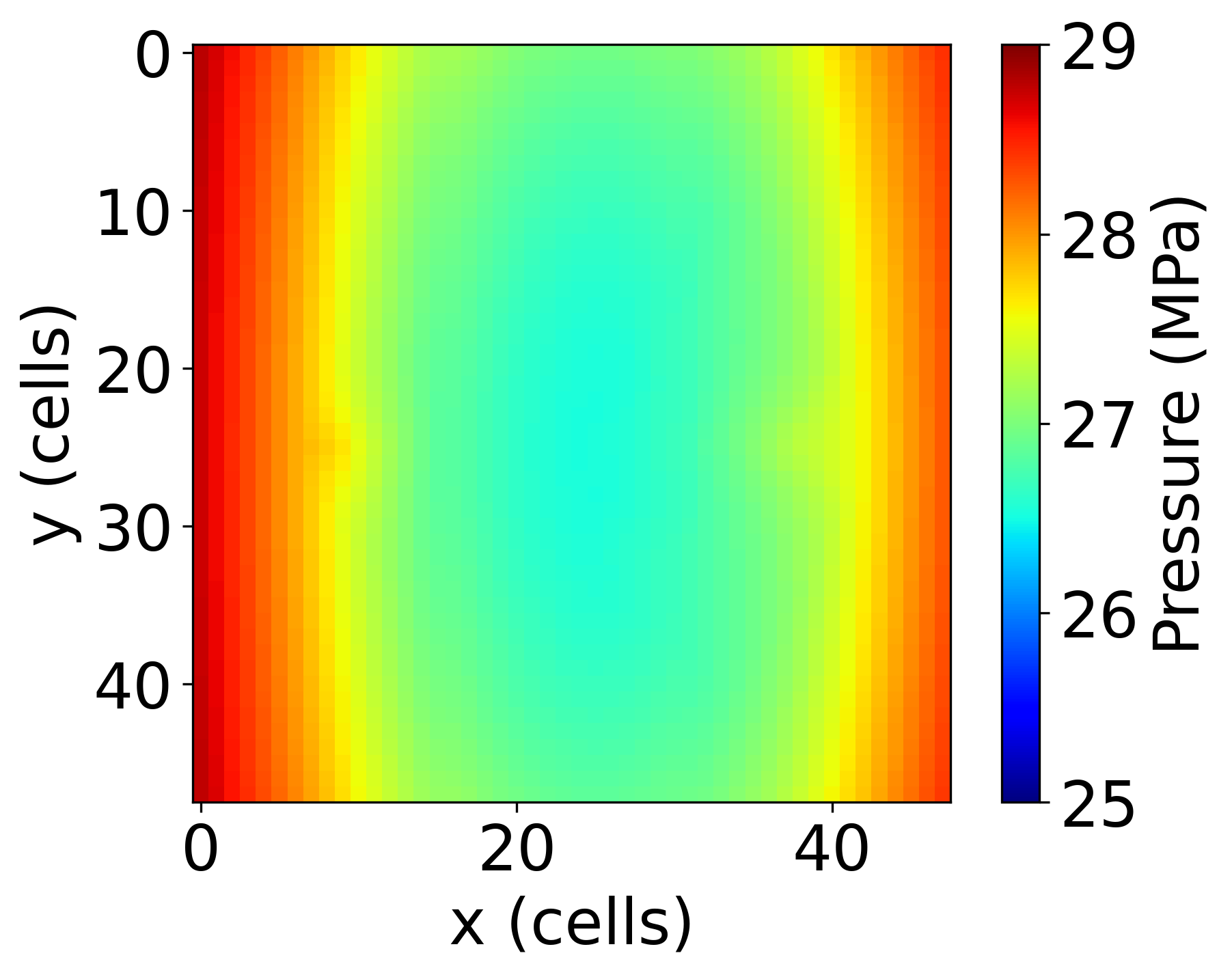}}
\hspace{2mm}
\subfloat[Realization 6 (sim)]{\label{sim_p_3_50_years}\includegraphics[width=58mm]{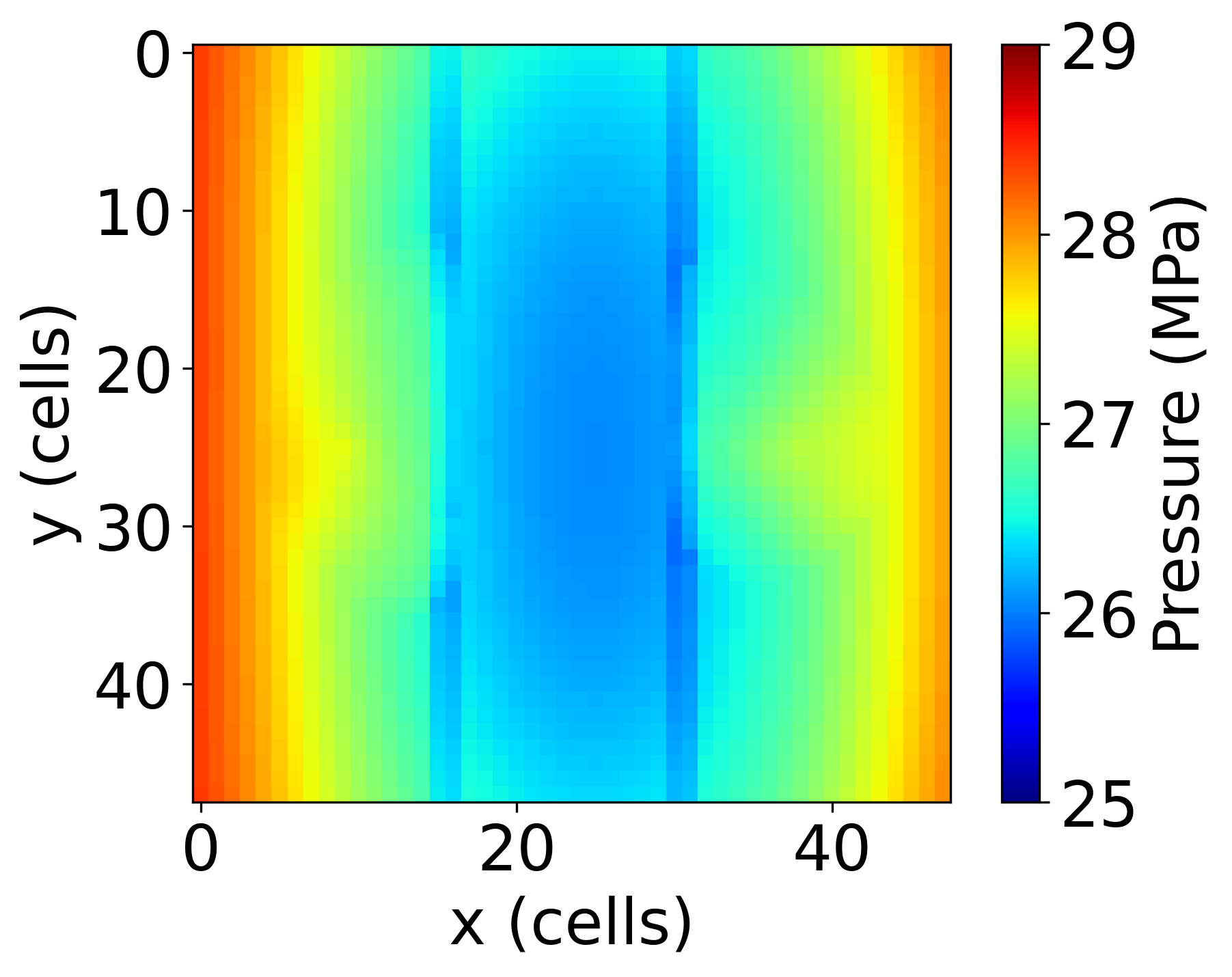}}
\\[1ex]
\subfloat[Realization 4 (surr)]{\label{surr_p_1_50_years}\includegraphics[width=58mm]{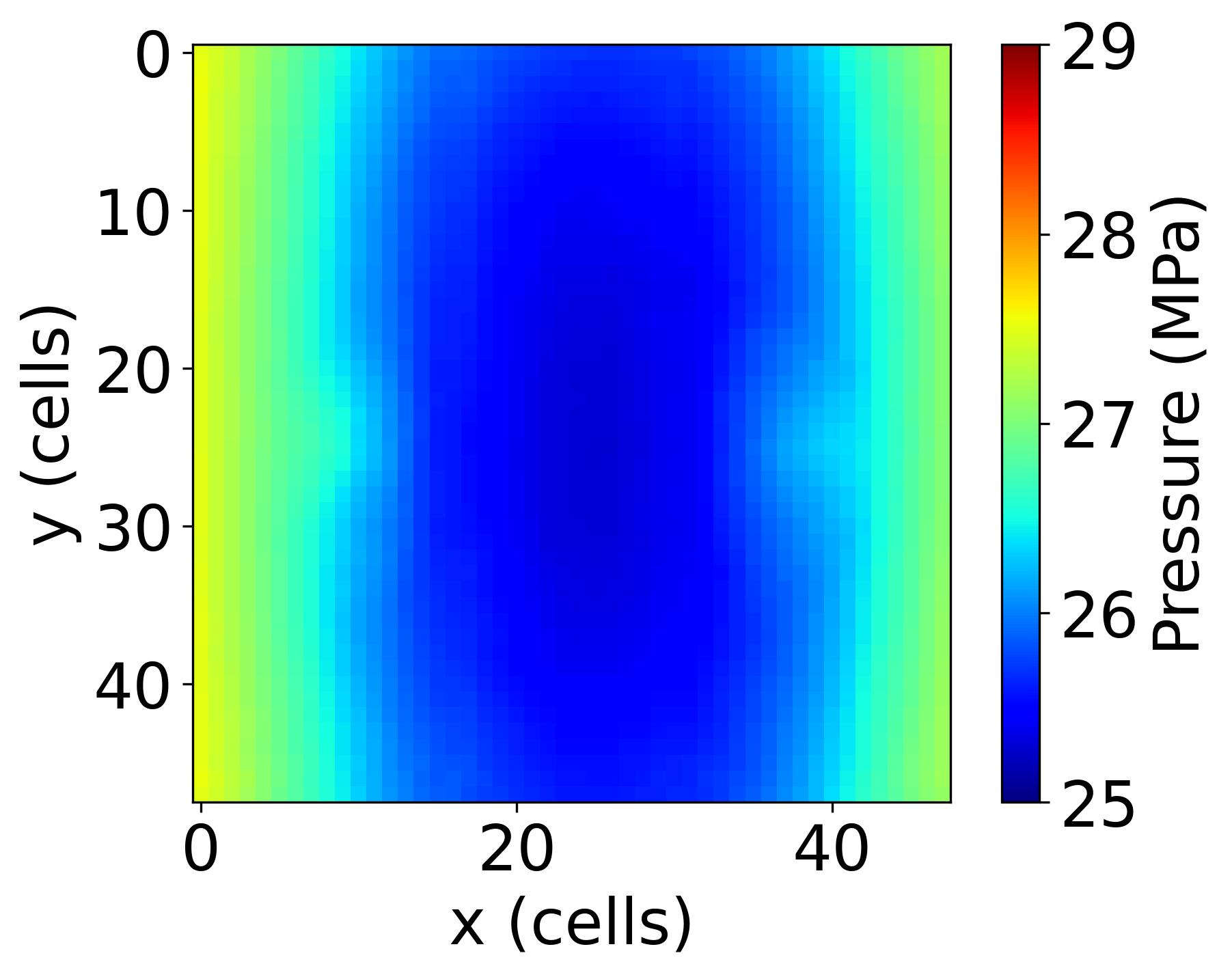}}
\hspace{2mm}
\subfloat[Realization 5 (surr)]{\label{surr_p_2_50_years}\includegraphics[width=58mm]{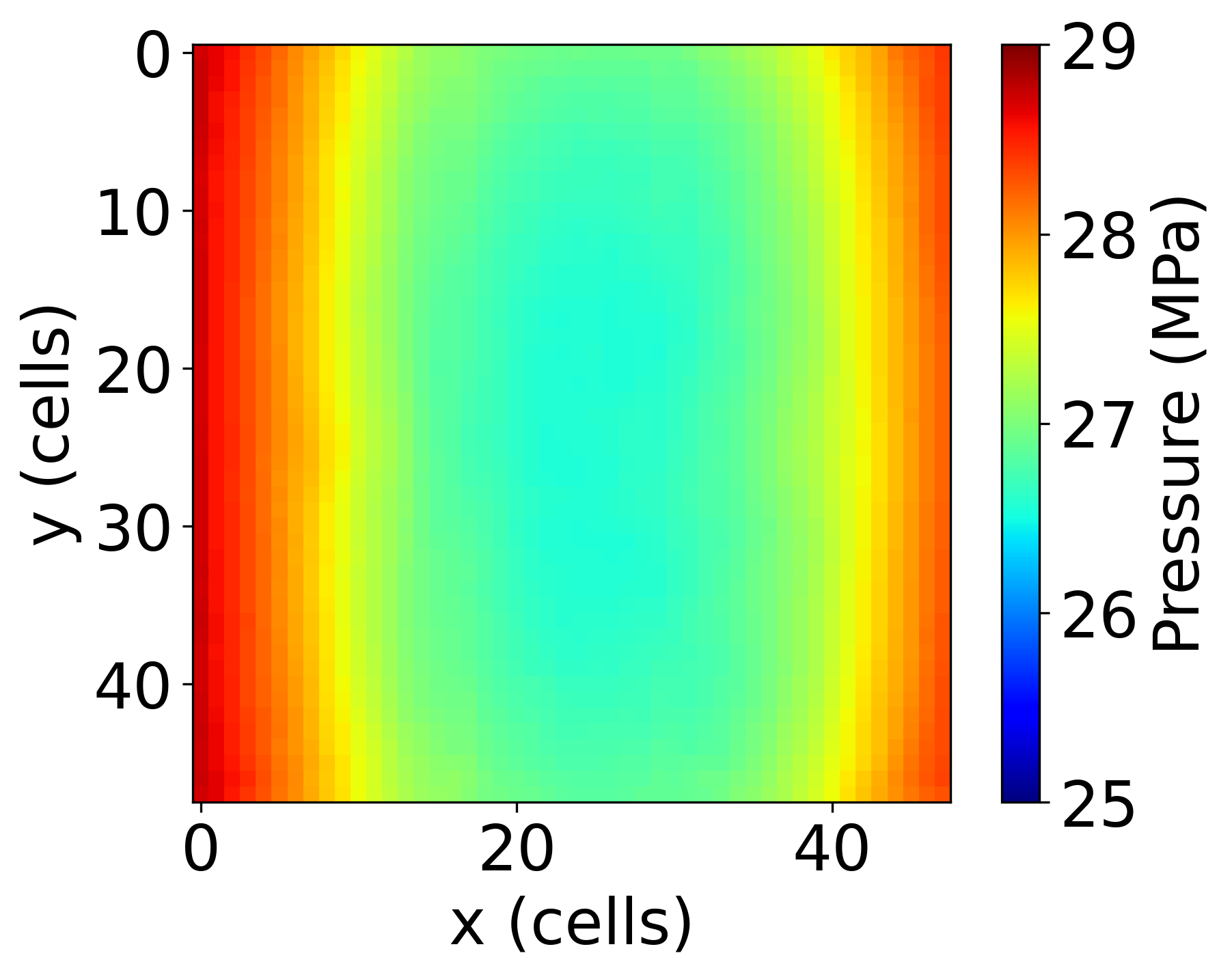}}
\hspace{2mm}
\subfloat[Realization 6 (surr)]{\label{surr_p_3_50_years}\includegraphics[width=58mm]{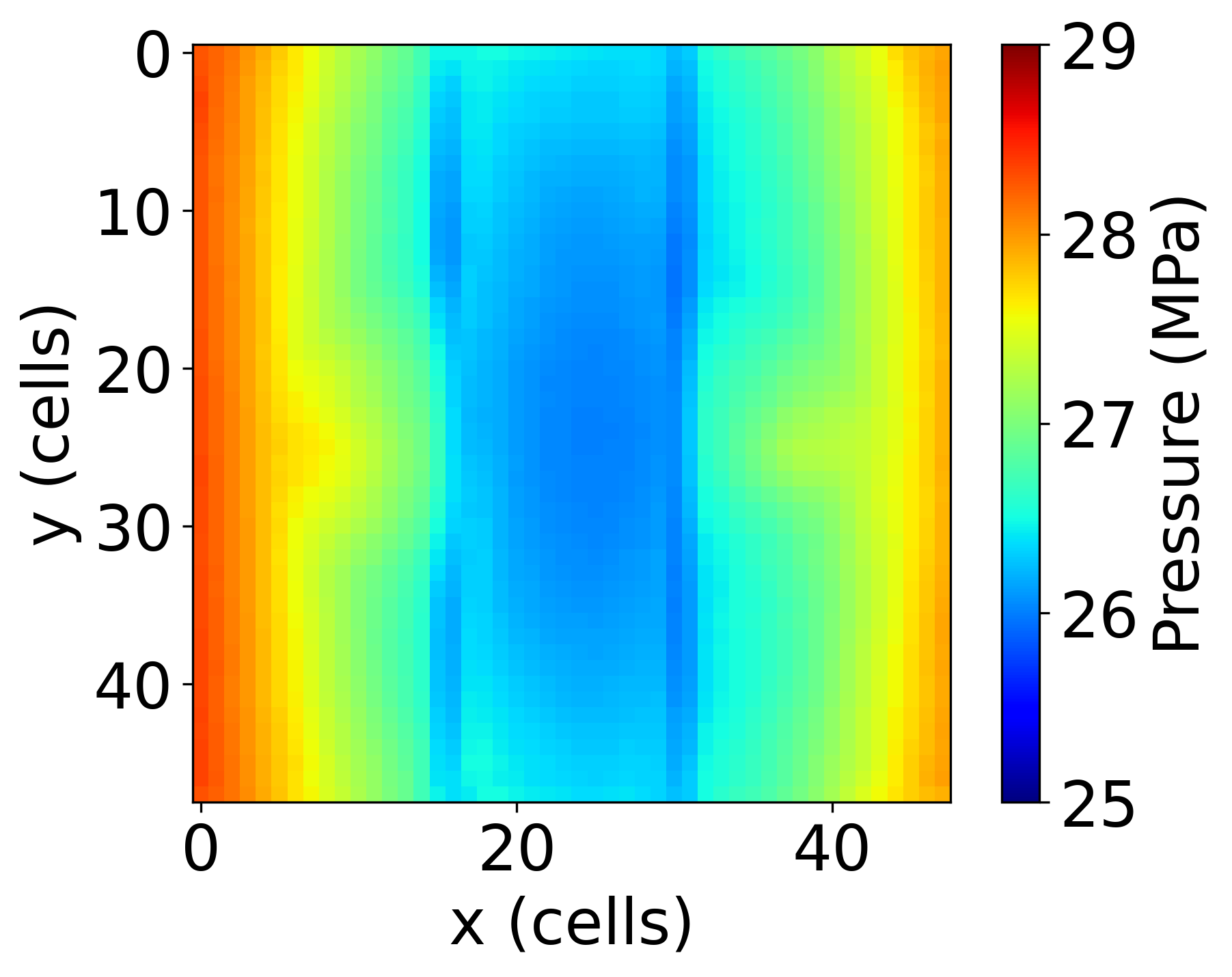}}
\caption{Pressure at the top layer of the target aquifer at 50~years from flow simulation (upper row) and the recurrent transformer U-Net surrogate model (lower row) for three new geomodel realizations.}
\label{S:pressure_50_years}
\end{figure}

Pressure fields from simulation and surrogate model predictions for realizations~4, 5 and 6, for the top layer of the target aquifer at 50~years, are shown in Fig.~\ref{S:pressure_50_years}. We show pressure results as 2D maps because the 3D fields are not as illustrative. The pressure relative errors for realizations~4, 5 and 6 are 0.15\%, 0.09\% and 0.11\%, respectively, so they are again representative in terms of accuracy. In all three realizations, the pressure at the left and right boundaries of the target aquifer is higher due to the increased depth, as the target aquifer is folded (anticlinal structure). The lowest pressure buildup is observed for realization~4 (Fig.\ref{S:pressure_50_years}a and d). This is the case because this geomodel is characterized by relatively large $\mu_{\log k}$ and fault permeabilities. In realization~6 (Fig.\ref{S:pressure_50_years}c and f), we observe a degree of pressure discontinuity in the vicinity of the faults ($x=16$ and $x=31$) due to the low fault permeabilities. Importantly, the various behaviors associated with different fault and aquifer properties are accurately represented by the surrogate model.

\subsection{Global sensitivity analysis using surrogate model}
\label{sec:gsa}

For the SEAM CO$_2$ geomodel, \citet{yoon2024assessing} evaluated CO$_2$ leakage rate and fault reactivation risk under uncertain fault properties. They used a global sensitivity analysis to compute the importance of a set of uncertain fault parameters on the estimated area of fault activation. In this study, for the modified (flow-only) SEAM CO$_2$ geomodel, we perform a variance-based global sensitivity analysis~\citep{sobol2001global} using surrogate model predictions. The importance of the metaparameters and PCA latent variables on the size of the CO$_2$ footprint in the target aquifer, at different time steps, is quantified. Consistent with \citet{han2023surrogate}, where global sensitivity of pressure and saturation at observation well locations was computed for an idealized geomodel without faults, the PCA latent variables characterizing the geological realization of the target aquifer are collectively treated as a single variable $\bm{\xi}$. This treatment then gives a total of 12 uncertain variables, designated as $\textbf{u}$ and given by
\begin{equation} \label{variables}
\textbf{u} = [\boldsymbol{\uptheta}_{\mathrm{meta}}, \bm{\xi}] = [\mu_{\log k}, \sigma_{\log k}, a_r, d, e, k_m, k_u, k_{f1}^{tm}, k_{f1}^{mu}, k_{f2}^{tm}, k_{f2}^{mu}, \bm{\xi}].
\end{equation}

The size of the CO$_2$ footprint is quantified here as ratio of the volume of the box-shaped region that contains the entire saturation plume in the target aquifer to the total bulk volume of the target aquifer and its surrounding region. Various metrics are used for this quantity -- this one is consistent with that considered by \citet{tang2025graph}, and that paper can be consulted for more details. Only cells in the target aquifer where the surrogate model saturation prediction exceeds a threshold value of 0.02 are taken to contribute to the CO$_2$ footprint.  

The first-order sensitivity index~\citep{saltelli2010variance} is used to quantify the variance contribution of each uncertain variable $u_i$ ($i = 1, 2, \ldots, 12$). This metric (which differs from that used in our earlier work) captures the individual contribution of each uncertain variable to the model prediction variance. This sensitivity index, denoted $z_i$, is given by
\begin{equation} \label{first_order}
z_{i} = \frac{\mathrm{Var}_{u_i} \left(\mathrm{E}_{u_\sim i}\left(v|u_i\right)\right)}{\mathrm{Var}_{\textbf{u}}\left(v\right)},
\end{equation}
\noindent where $\textbf{u}_{\sim i}$ is the set of all uncertain variables except $u_i$, $v$ denotes the computed CO$_2$ footprint in the target aquifer using the surrogate model, $\mathrm{E}_{u_\sim i}\left(v|u_i\right)$ is the expectation of $v$ over all possible values of $\textbf{u}_{\sim i}$ (all combinations of other uncertain variables except $u_i$), conditioned to a fixed $u_i$. The variance of this expectation over all possible values of $u_i$ is expressed as $\mathrm{Var}_{u_i} \left(\mathrm{E}_{u_\sim i}\left(v|u_i\right)\right)$. The denominator $\mathrm{Var}_{\textbf{u}}\left(v\right)$ quantifies the total variation of $v$ over all samples of $\textbf{u}$ used for the sensitivity index computations.  

\begin{figure}[!ht]
\centering   
{\includegraphics[width = 152mm]{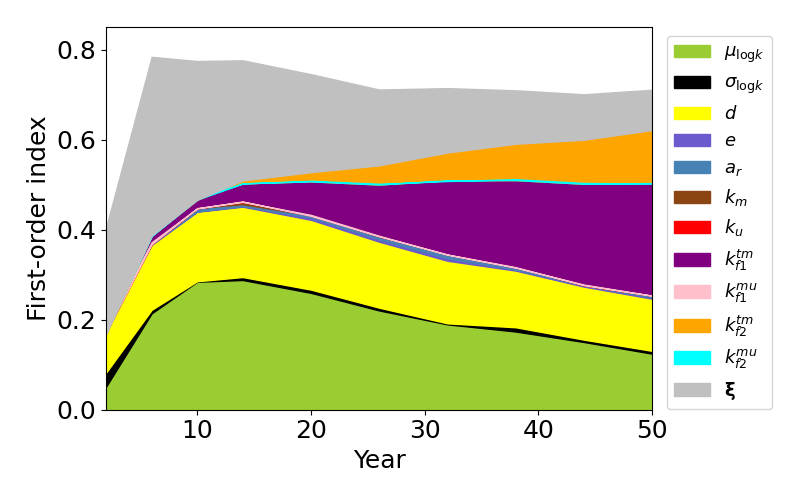}}
\caption{First-order sensitivity index, as a function of time, for metaparameters and PCA latent variables on the CO$_2$ footprint ratio in the target aquifer.}
\label{gsa}
\end{figure}

A total of 57,344 samples of the set of uncertain variables $\textbf{u}$ is generated using a quasi-Monte Carlo method based on Sobol sequences~\citep{saltelli2010variance}. The corresponding geomodel realizations are constructed, and the CO$_2$ footprint ratios evaluated using the surrogate model are used to compute the first-order sensitivity indices. This sample size is selected to ensure convergence of the sensitivity indices. In this work, the first-order indices are considered converged if the average absolute change across all uncertain variables is less than 0.01 for all time steps. 

Results for the first-order sensitivity indices at different time steps are shown in Fig.~\ref{gsa}. At early time (up to about 10~years), the CO$_2$ footprint ratio is most sensitive to $\mu_{\log k}$ in the target aquifer, the coefficient $d$ relating porosity and permeability ($\phi = d \cdot \log k_h + e$), and the geological realization of the target aquifer, $\bm{\xi}$. At later times, the fault permeabilities between the target and middle aquifers gain in importance. Interestingly, the CO$_2$ footprint ratio is more sensitive to $k_{f1}^{tm}$ than to $k_{f2}^{tm}$. This is presumably due to the locations of the injection wells relative to the faults. Footprint ratio in the target aquifer is not sensitive to the fault permeabilities between the middle and upper aquifers, the permeability anisotropy ratio, the coefficient $e$, or the permeabilities in the middle and upper aquifers. We emphasize that detailed global sensitivity analyses of this type are enabled by the surrogate model. Such computations would be prohibitively expensive using a high-fidelity simulator.

%% file: Section_5.tex
In this section, we use the surrogate model in conjunction with a hierarchical MCMC-based data assimilation procedure to evaluate the uncertainty reduction achieved for different monitoring strategies and types of monitoring data. We first describe the problem setup and then present the hierarchical MCMC-based procedure. This procedure differs from that applied in our previous studies since we now use multiple Markov chains. Detailed data assimilation results for a new synthetic ‘true’ model are presented. Results for two additional synthetic true models with qualitatively different fault leakage scenarios are provided in SI.

\subsection{Data assimilation problem setup}
\label{sec:setup}

The synthetic true model for the overall domain is denoted as $\textbf{m}_{\mathrm{true}}$. This model is generated by sampling the 11 metaparameters $\boldsymbol{\uptheta}_{\mathrm{meta}}$ from their prior ranges (given in Table~\ref{rock physics}) along with a set of PCA variables $\bm{\xi}$. This model is then simulated with the same specifications described in Section~\ref{sec:simulation_setup}, and the simulation results at the observation well locations comprise the observed data. The locations of the three vertical observation wells are shown in Fig.~\ref{well_location} and discussed in Section~\ref{sec:simulation_setup}. Pressure and saturation for cells in the target, middle and upper aquifers provide the observations; data in the caprock layers between these aquifers are not used.

The design of the monitoring strategy is an important issue in geological carbon storage operations. Cost and risk must be balanced, and this requires that many strategies be assessed and compared. This could be computationally prohibitive using high-fidelity simulation models, since each assessment requires a full data assimilation run. With surrogate models, however, this type of assessment can be readily accomplished.

Along these lines, we consider four monitoring strategies. In two of the cases observation wells penetrate all three aquifers, while in the other two (partial monitoring) cases these wells penetrate only the middle and upper aquifers. For each case we consider the use of either pressure data only, or the use of both pressure and saturation data. An advantage of the partial monitoring strategies is that, by avoiding penetration into the target aquifer where the actual injection occurs, the risk of CO$_2$ leakage through observation wells is reduced. For the full monitoring strategy using both pressure and saturation data, measurements are collected from the three observation wells in 37~layers, including all 25~layers of the target aquifer, and 6~layers each of the middle and upper aquifers. These observations are made at 2, 6, 10, 14, and 20~years after the start of injection. Thus there are a total of $5 \times 37 \times 3 = 555$ measurements each for saturation and pressure (a total of 1110~measurements). For the full monitoring case where only pressure is measured (at 37~layers), this results in 555~pressure measurements. For the partial monitoring strategy with both pressure and saturation measured at 12~layers (in the middle and upper aquifers), we have a total of $5 \times 12 \times 3 = 180$ measurements each for saturation and pressure (a total of 360~measurements). Finally, for the partial monitoring strategy with only pressure, we have just the 180~pressure measurements.

Errors resulting from the precision of the measurement device and data interpretation impact the observed data. Consistent with \citet{jiang2024history} and \citet{han2025accelerated}, the standard deviation of measurement error is set to 0.02 for saturation and 0.095~MPa for pressure. The observed data with measurement error, $\textbf{d}_{\mathrm{obs}}$, can be expressed as
\begin{equation} \label{noise}
\textbf{d}_{\mathrm{obs}} = \textbf{d}_{\mathrm{true}} + \bm{\epsilon} = f(\textbf{m$_{\mathrm{true}}$}) + \bm{\epsilon},
\end{equation}
\noindent where $f(\textbf{m$_{\mathrm{true}}$})$ = $\textbf{d}_{\mathrm{true}} \in \mathbb{R}^{n_m}$ represents the true data, in our case simulation results at the observation locations for true model $\textbf{m}_{\mathrm{true}}$, $n_m$ is the total number of measurements (which varies between monitoring strategies), and $\bm{\epsilon} \sim {\mathcal N}(\textbf{0}, C_{\mathrm{D}})$ is the measurement error with diagonal covariance $C_{\mathrm{D}}  \in \mathbb{R}^{n_m \times n_m}$.

\subsection{Hierarchical MCMC using multiple Markov chains}
\label{sec:mcmc}

The hierarchical MCMC sampling method used in this work is based on the noncentered preconditioned Crank-Nicolson within Gibbs algorithm introduced by \citet{chen2018dimension} and implemented by \citet{han2023surrogate, han2025accelerated} for hierarchical problems in geological carbon storage settings. In our previous studies, only one Markov chain was used and a convergence criterion based on the average relative change of the bin-wise posterior probability density was applied. Here we use multiple parallel Markov chains to accelerate the computations, and we evaluate convergence using the Gelman–Rubin scale reduction factor~\citep{gelman1992inference}.

From Bayes' theorem, the posterior probability density function for both the metaparameters and PCA latent variables, $p({\boldsymbol{\uptheta}_{\mathrm{meta}}, \bm{\xi}|\bm{\mathrm{d}}_{\mathrm{obs}}})$, conditioned to observation data $\bm{\mathrm{d}}_{\mathrm{obs}}$, is given by
\begin{equation} \label{Bayes}
  p({\boldsymbol{\uptheta}_{\mathrm{meta}}, \bm{\xi} | \bm{\mathrm{d}}_{\mathrm{obs}}}) = \frac{p(\boldsymbol{\uptheta}_{\mathrm{meta}}, \bm{\xi}) p(\bm{\mathrm{d}}_{\mathrm{obs}}|{\boldsymbol{\uptheta}}_{\mathrm{meta}}, \bm{\xi})} {p(\bm{\mathrm{d}}_{\mathrm{obs}})}.
  \end{equation}
\noindent Here $p(\boldsymbol{\uptheta}_{\mathrm{meta}}, \bm{\xi})$ is the prior probability density function, ${p(\bm{\mathrm{d}}_{\mathrm{obs}})}$ is a normalization constant, and $p(\bm{\mathrm{d}}_{\mathrm{obs}} | {\boldsymbol{\uptheta}}_{\mathrm{meta}}, \bm{\xi})$ is the likelihood function, computed as
\begin{equation} \label{likelihood}
  p(\bm{\mathrm{d}}_{\mathrm{obs}} | {\boldsymbol{\uptheta}}_{\mathrm{meta}}, \bm{\xi}) = \\
  c \exp\left({{-\frac{1}{2}  \left(\bm{\mathrm{d}}_{\mathrm{obs}}-\hat{f}\left(\textbf{m}_f\left({\boldsymbol{\uptheta}}_{\mathrm{meta}}, \bm{\xi}\right)\right) \right)}^T C_{\mathrm{tot}}^{-1} \left(\bm{\mathrm{d}}_{\mathrm{obs}}-\hat{f}\left(\textbf{m}_f\left({\boldsymbol{\uptheta}}_{\mathrm{meta}}, \bm{\xi}\right)\right) \right)} \right).
\end{equation} 
\noindent In Eq.~\ref{likelihood}, $c$ is a normalization constant, $\hat{f}\left(\textbf{m}_f\left({\boldsymbol{\uptheta}}_{\mathrm{meta}},  \bm{\xi}\right)\right)$ represents the surrogate model prediction for pressure and saturation at the observation locations, and $C_{\mathrm{tot}} = C_{\mathrm{D}} + C_{\mathrm{surr}}$ is the total error covariance. Here $C_{\mathrm{surr}}$ represents the covariance of the model error arising from surrogate model inaccuracy relative to the high-fidelity simulator. This model error covariance matrix is computed based on surrogate model errors over the test set, following the approach in \citet{han2025accelerated}. Note that $C_{\mathrm{surr}}$ (and thus $C_{\mathrm{tot}}$) is a nondiagonal matrix, which is required to capture correlations in the spatial and temporal model errors. A different $C_{\mathrm{surr}}$ matrix is used for each monitoring strategy.

The dimension-robust hierarchical MCMC sampling method involves simulating $m$ Markov chains (here we use $m = 3$) to generate sequences of samples to approximate the posterior distributions of the metaparameters. The Metropolis-within-Gibbs~\citep{chen2018dimension} approach used here requires sampling metaparameters and PCA latent variables independently. For each Markov chain, the initial metaparameters are randomly sampled from their prior distributions, and each component of the initial PCA latent variables is independently sampled from the standard normal distribution \(\mathcal{N}(0, 1)\). 

At each iteration of the sampling process, new sets of PCA latent variables and metaparameters are sampled independently for each Markov chain. Specifically, for chain $j$ ($j = 1, \dots, m$) at iteration $k + 1$, a new sample of the PCA latent variable $\bm{\xi}_j^{'}$ is proposed by adding a random perturbation $\boldsymbol{\epsilon}$ to the latent variable $\bm{\xi}_j^{k}$ from the previous iteration $k$. This is expressed as
\begin{equation} \label{xi}
\bm{\xi}_j^{'} = (1 - \beta^2) \bm{\xi}_j^{k} + \beta \boldsymbol{\epsilon}, 
\end{equation}
\noindent where $\beta$ is a coefficient controlling the update magnitude, and $\boldsymbol{\epsilon} \in \mathbb{R}^{n_d}$ is a random vector with each component sampled independently from $\mathcal{N}$(0, 1). The proposed PCA latent variable is accepted with a probability based on the Metropolis–Hastings criterion~\citep{hastings1970monte}. For the detailed implementation, please refer to \citet{han2023surrogate}. Here we set $\beta=0.05$ for the full monitoring strategy using both pressure and saturation data, which gives an appropriate acceptance rate. 

A similar procedure is then applied with the metaparameters. Specifically, for chain \( j \) at iteration \( k + 1 \), a new metaparameter sample $({\boldsymbol{\uptheta}}_{\mathrm{meta}})_j^{'}$ is drawn from the multivariate Gaussian proposal distribution centered on the current sample $({\boldsymbol{\uptheta}}_{\mathrm{meta}})_j^{k}$. This is expressed as $({\boldsymbol{\uptheta}}_{\mathrm{meta}})_j^{'} \sim \mathcal{N}$($({\boldsymbol{\uptheta}}_{\mathrm{meta}})_j^{k}$, $C_{\uptheta}$), where $C_{\uptheta} \in \mathbb{R}^{11 \times 11}$ is the covariance matrix of the proposal distribution for the 11 metaparameters. Specifically, for metaparameter $i$ ($i = 1, \dots, 11$), we set the standard deviation of the proposal distribution as $\sigma_i=(({{\uptheta}}_{\mathrm{meta}})_{i, \mathrm{max}}-({{\uptheta}}_{\mathrm{meta}})_{i, \mathrm{min}})/80$, where $({{\uptheta}}_{\mathrm{meta}})_{i, \mathrm{max}}$ and $({{\uptheta}}_{\mathrm{meta}})_{i, \mathrm{min}}$ are the maximum and minimum values of the prior range. The proposed sets of metaparameters for each chain are accepted (independently) with a probability analogous to that used for the PCA latent variable.

We evaluate the convergence of metaparameters based on the Gelman--Rubin scale reduction factor. Convergence is evaluated for each metaparameter $i$ ($i = 1, \dots, 11$) using the potential scale reduction factor $\hat{R}_i$, which is given by
\begin{equation}
\hat{R}_i = \sqrt{ \frac{\hat{V}_i}{W_i} }.
\end{equation}
\noindent Here $W_i$ is the within-chain variance and $\hat{V}_i$ denotes the estimated marginal posterior variance for metaparameter $i$. The within-chain variance $W_i$ is given by
\begin{equation}
W_i = \frac{1}{m} \sum_{j=1}^{m} \mathrm{Var}{\left( \left({{\uptheta}}_{\mathrm{meta}}\right)_i^j\right)},
\end{equation}
\noindent where $\mathrm{Var}{\left( \left({{\uptheta}}_{\mathrm{meta}}\right)_i^j\right)}$ represents the variance of $n$ post–burn-in samples for metaparameter $i$ in chain $j$. Here $n = k - n_{\mathrm{burn}}$, where $k$ is the iteration counter and $n_{\mathrm{burn}}$ is the burn-in period (here we set $n_{\mathrm{burn}}=5000$~iterations). The estimated marginal posterior variance $\hat{V}_i$ for metaparameter $i$ is computed as
\begin{equation}
\hat{V}_i = \left(1 - \frac{1}{n} \right) W_i + \frac{1}{n} B_i.
\end{equation}
\noindent Here $B_i$ is the between-chain variance for metaparameter $i$, expressed as
\begin{equation}
B_i = \frac{n}{m - 1} \sum_{j=1}^{m} \left( \mathrm{E}{\left( \left({{\uptheta}}_{\mathrm{meta}}\right)_i^j\right)} - \mathrm{E}\left(({{\uptheta}}_{\mathrm{meta}})_i^{(\cdot)}\right) \right)^2.
\end{equation}
\noindent where $\mathrm{E}{\left( \left({{\uptheta}}_{\mathrm{meta}}\right)_i^j\right)}$ is the mean of metaparameter $i$ computed from the post–burn-in samples in chain $j$, and $\mathrm{E}\left(({{\uptheta}}_{\mathrm{meta}})_i^{(\cdot)}\right)$ denotes the corresponding mean across all three chains. Values of $\hat{R}_i$ approaching 1 indicate convergence. Consistent with \citet{gelman2004bayesian}, we use a conservative convergence criterion by requiring $\hat{R}_i < 1.1$ for all 11~metaparameters.

\subsection{Results for true model~1}
\label{sec:true model 1}

True model~1 corresponds to (test-case) realization~4 discussed in Section~\ref{sec:surrogate_individuals}.
The two faults in true model~1 are relatively permeable, resulting in leakage to the middle and upper aquifers. The true model~1 saturation fields at 20~years and at the end of injection (50~years) are shown in Figs.~\ref{S:saturation_20_years}a and \ref{S:saturation_50_years}a, respectively, and the pressure field at the end of injection is shown in Fig.~\ref{S:pressure_50_years}a. 

For the full monitoring strategy using both pressure and saturation data, the hierarchical MCMC data assimilation procedure requires 43,900 iterations to achieve convergence in the posterior distributions of the metaparameters, using the criterion described above. With three chains and the requirement of two function evaluations at each iteration (for $\boldsymbol{\uptheta}_{\mathrm{meta}}$ and $\bm{\xi}$), this corresponds to a total of 263,400~function evaluations. A total of 27,635 sets of metaparameters and their corresponding geomodel realizations are accepted during the post–burn-in period over the three chains. These are taken to be the final posterior samples. For the full monitoring strategy using only pressure data, convergence is achieved after 158,406 function evaluations, resulting in 15,435 posterior samples. The partial monitoring strategy with both pressure and saturation data requires 544,206 function evaluations, from which 72,552 posterior samples are collected. Finally, under the partial monitoring strategy using only pressure data, convergence requires 387,006 function evaluations, resulting in 49,635 posterior samples.

Each surrogate model evaluation for saturation or pressure (individually) requires about 0.09~seconds on a single Nvidia A100 GPU. Surrogate model evaluations for the three chains at each iteration are performed in parallel, though we do not achieve full ($3\times$) parallel speedup due to memory limitations in the A100 GPU. The total elapsed time for a typical data assimilation run is about 12~hours. This includes surrogate model evaluations as well as geomodel construction using PCA, hierarchical MCMC execution, likelihood computation, and convergence diagnostics. The sequential hierarchical MCMC-based data assimilation procedure would not be practical using high-fidelity flow simulations, as each GEOS run requires about 10~minutes with 32 AMD EPYC-7543 CPU cores.

\begin{figure}[!ht]
\centering 
\subfloat[$\mu_{\mathrm{log}k}$ (full; $S$, $p$)]{\includegraphics[width = 43mm]{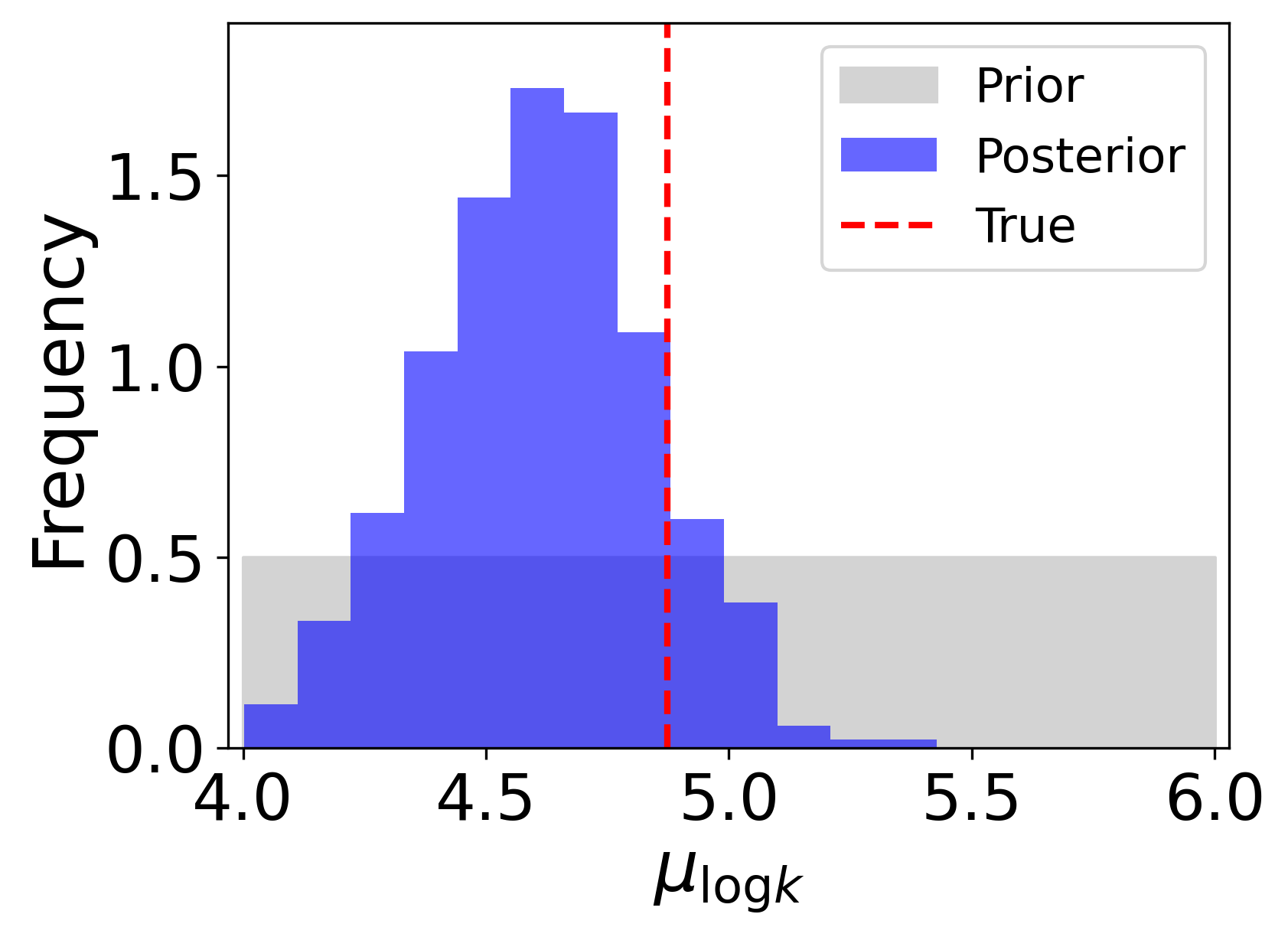}}
\hspace{2mm}
\subfloat[$\mu_{\mathrm{log}k}$ (partial; $S$, $p$)]{\includegraphics[width = 43mm]{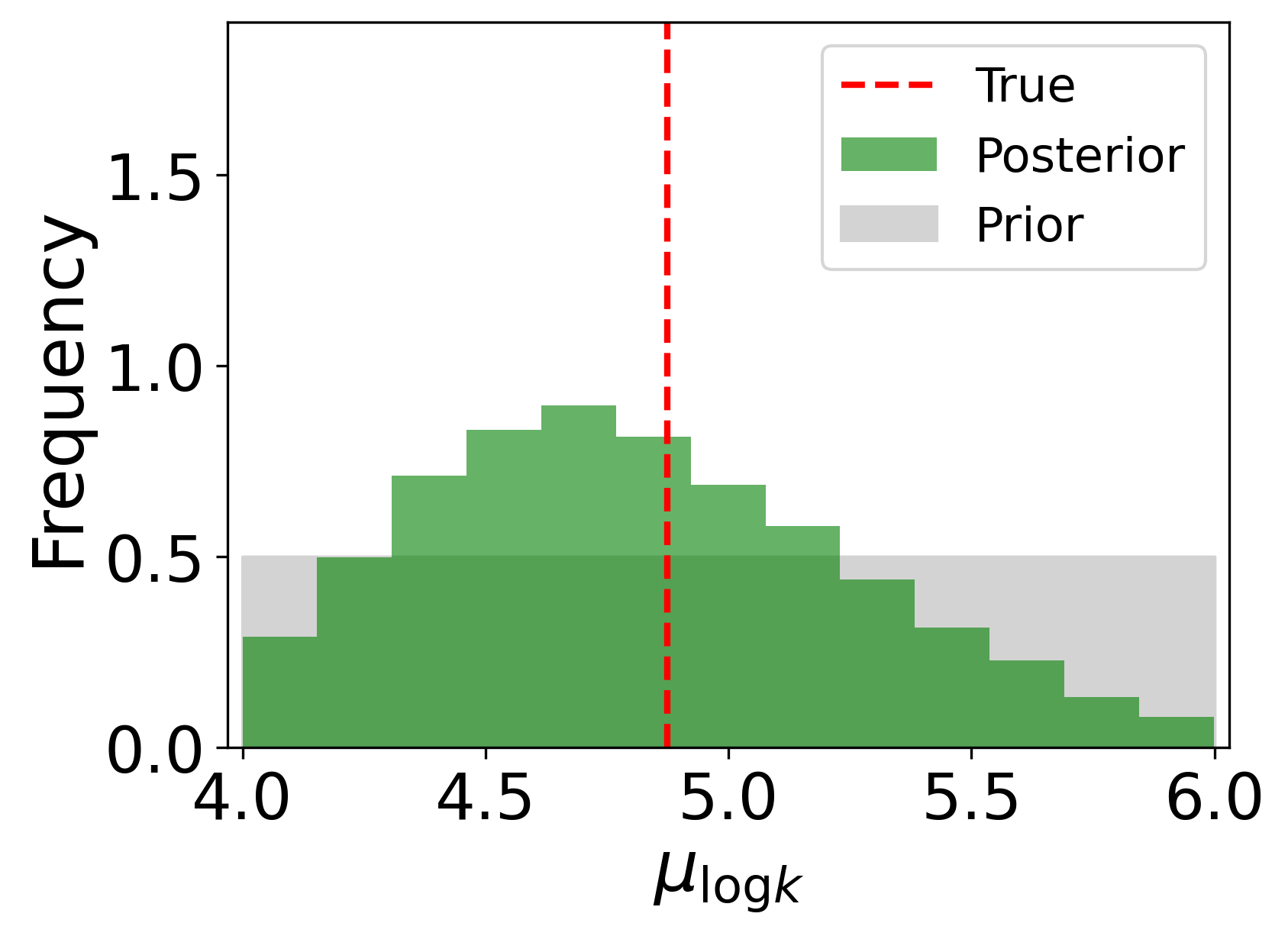}}
\hspace{2mm}
\subfloat[$\mu_{\mathrm{log}k}$ (full; $p$)]{\includegraphics[width = 43mm]{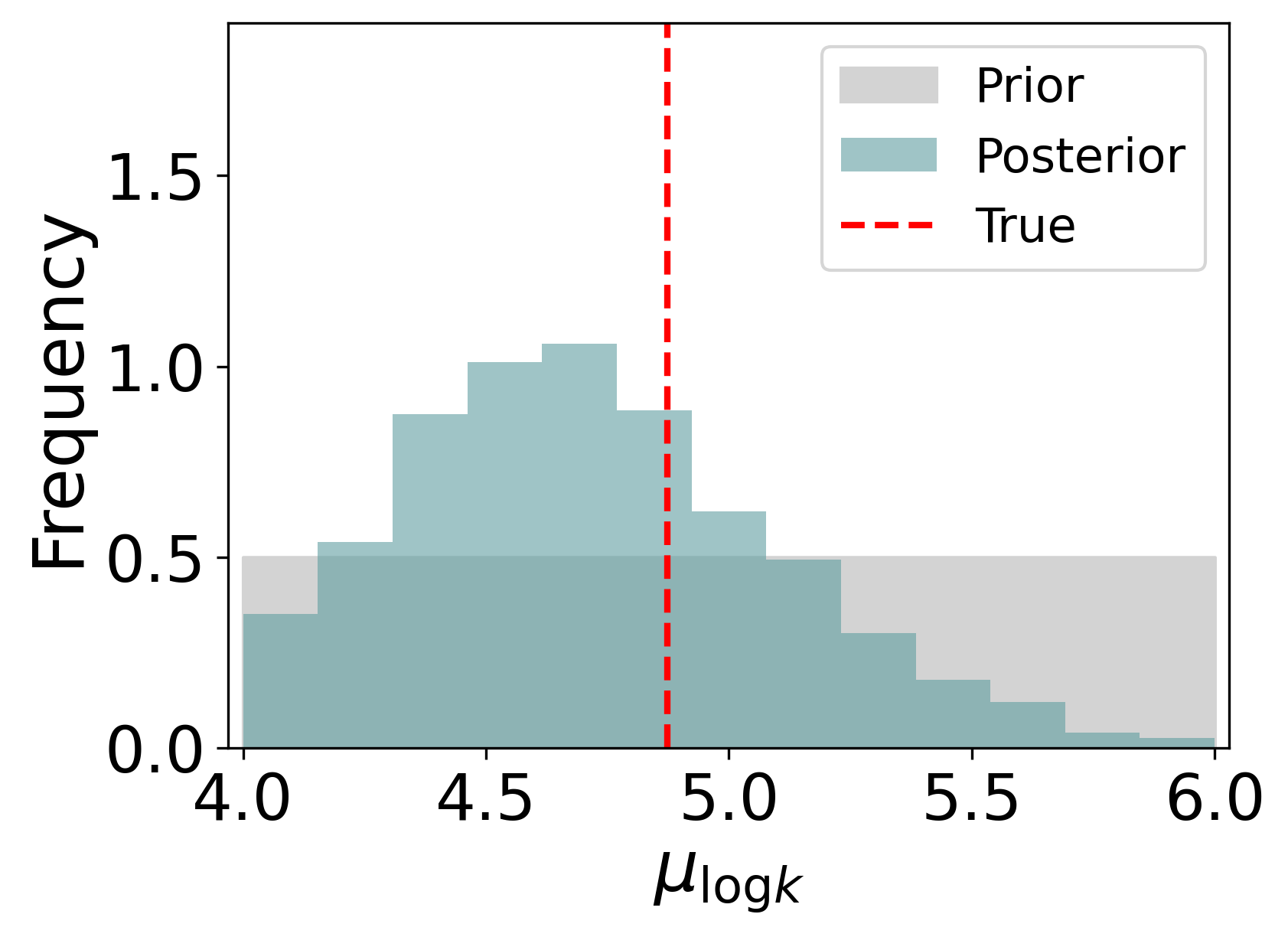}}
\hspace{2mm}
\subfloat[$\mu_{\mathrm{log}k}$ (partial; $p$)]{\includegraphics[width = 43mm]{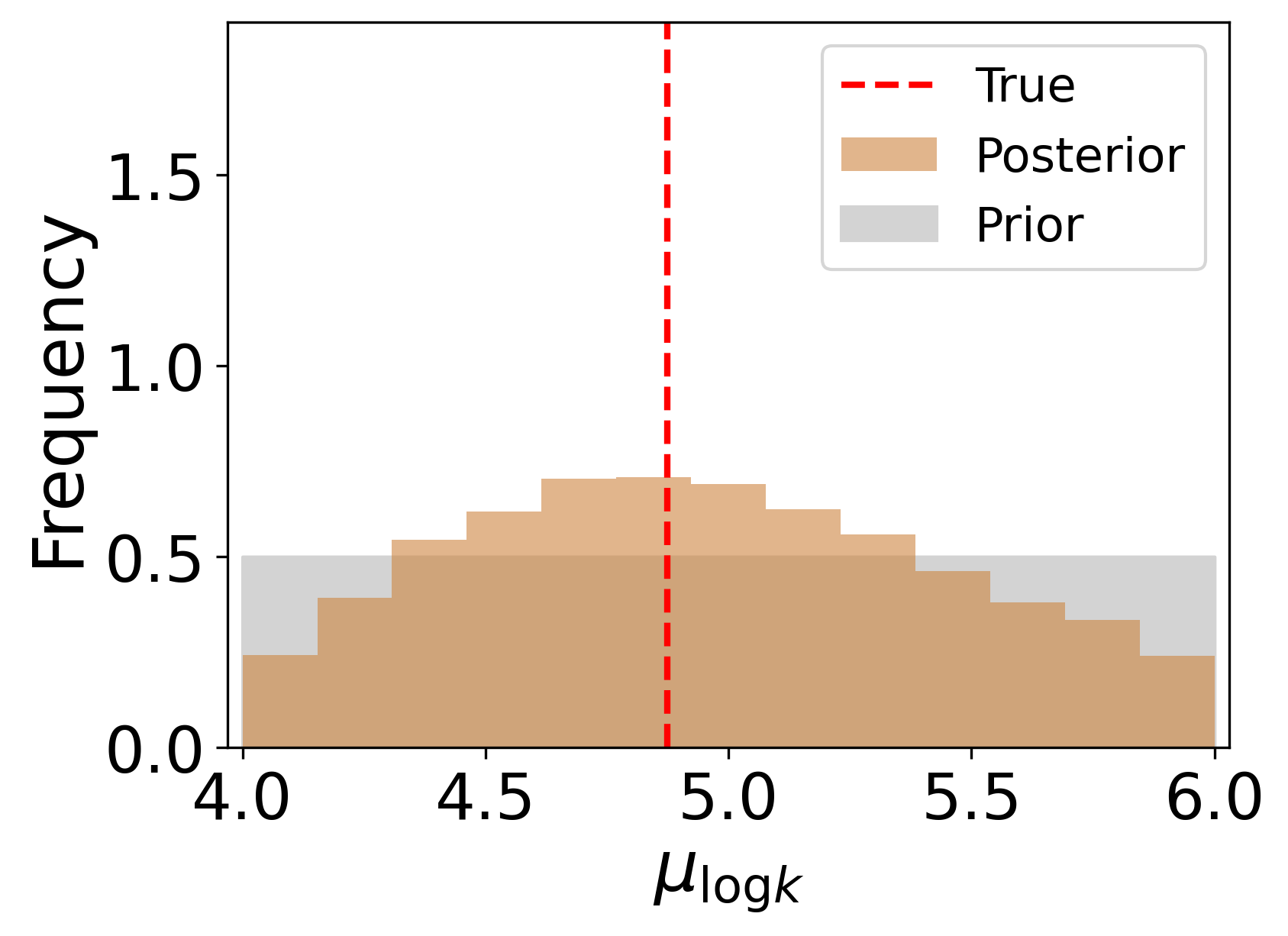}}\\
\subfloat[$\sigma_{\mathrm{log}k}$ (full; $S$, $p$)]{\includegraphics[width = 43mm]{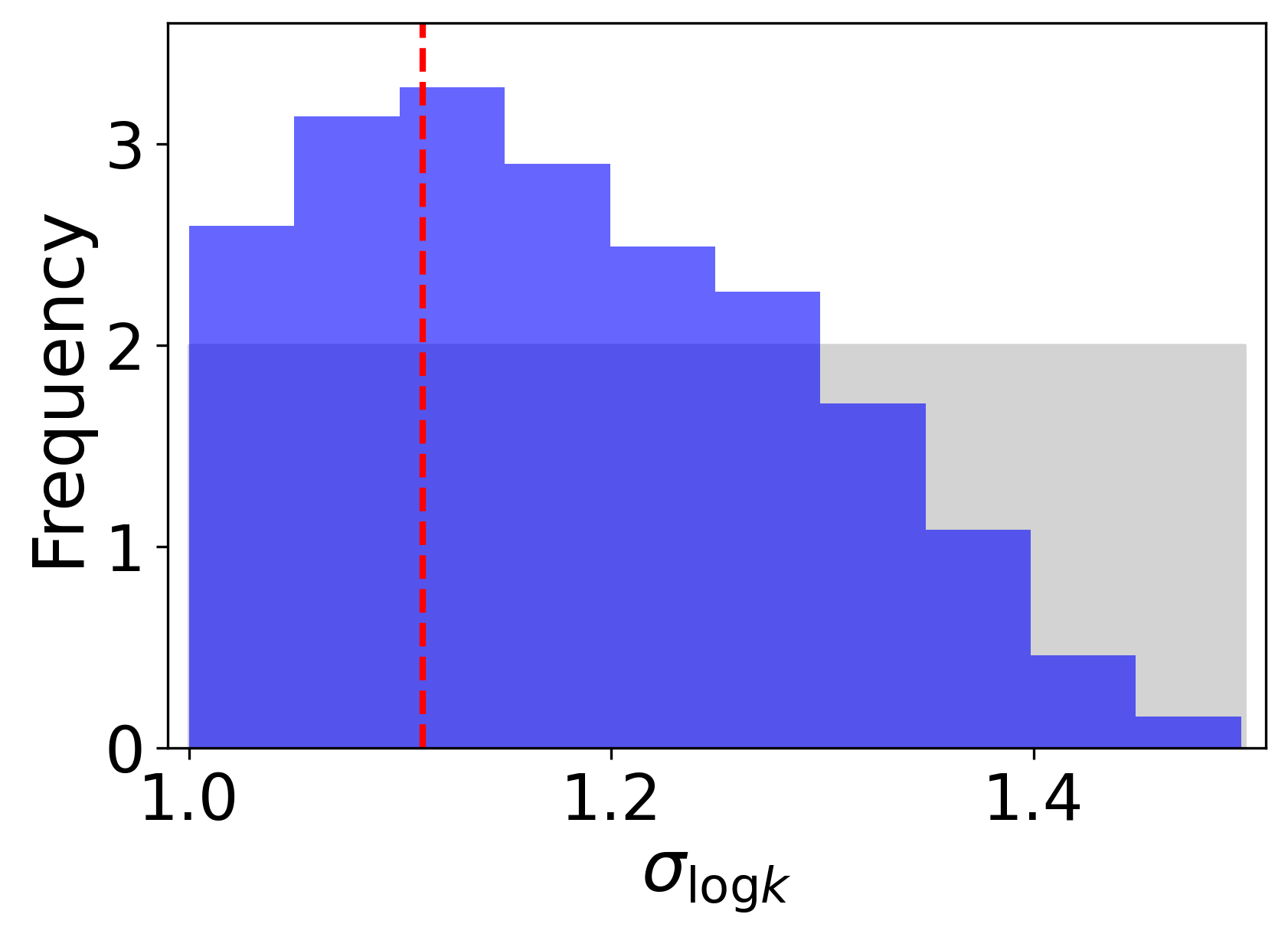}}
\hspace{2mm}
\subfloat[$\sigma_{\mathrm{log}k}$ (partial; $S$, $p$)]{\includegraphics[width = 43mm]{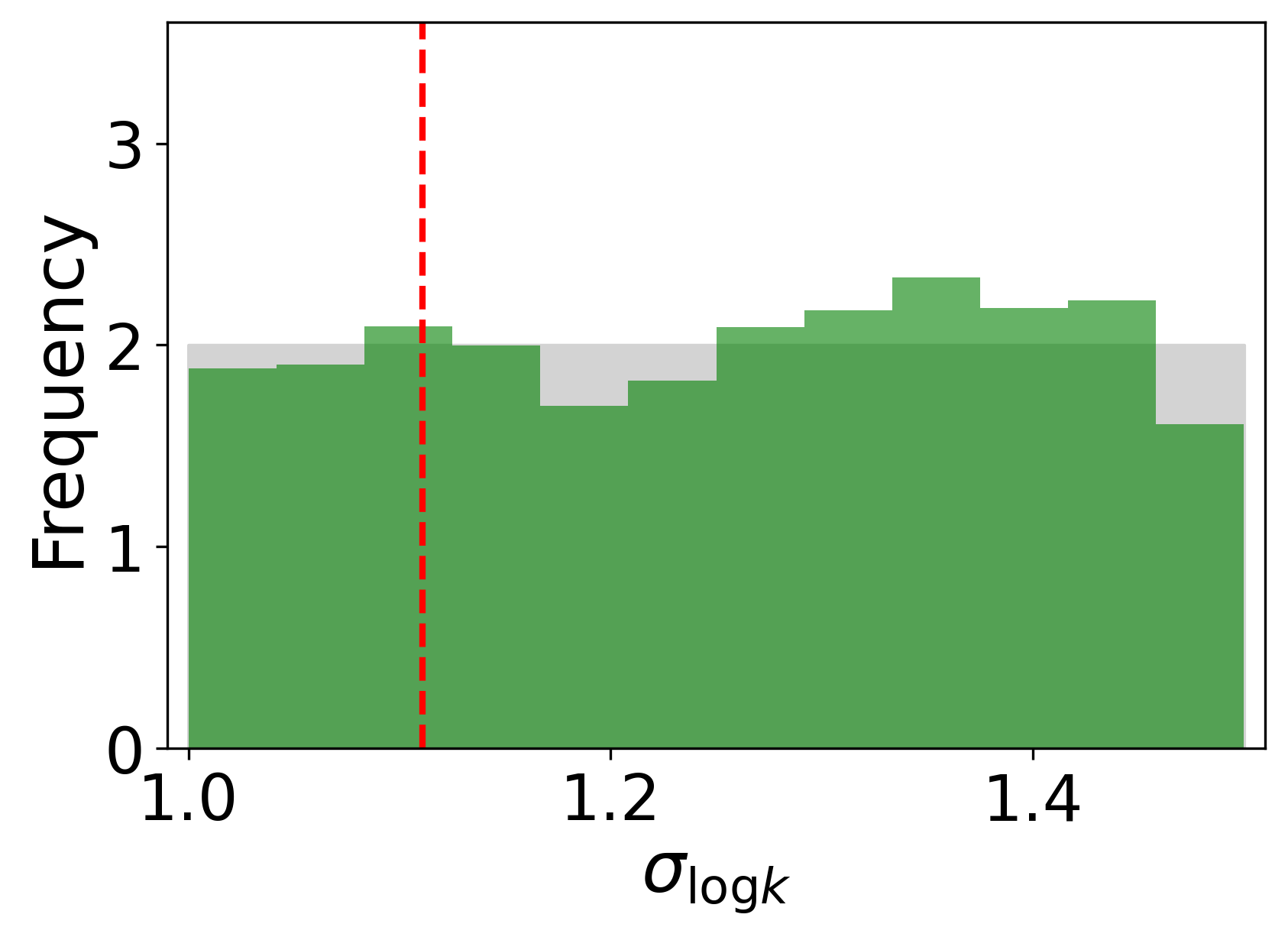}}
\hspace{2mm}
\subfloat[$\sigma_{\mathrm{log}k}$ (full; $p$)]{\includegraphics[width = 43mm]
{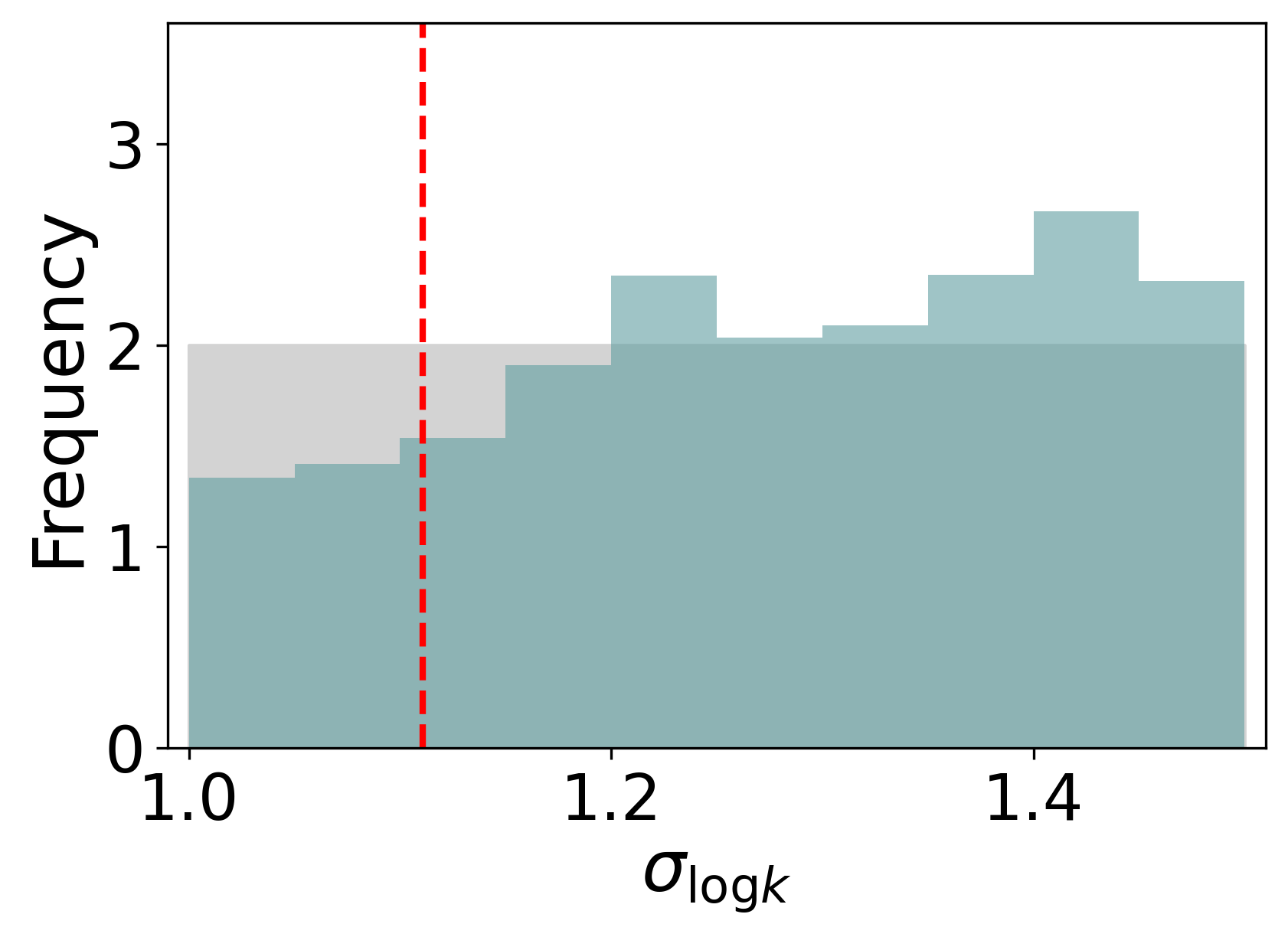}}
\hspace{2mm}
\subfloat[$\sigma_{\mathrm{log}k}$ (partial; $p$)]{\includegraphics[width = 43mm]{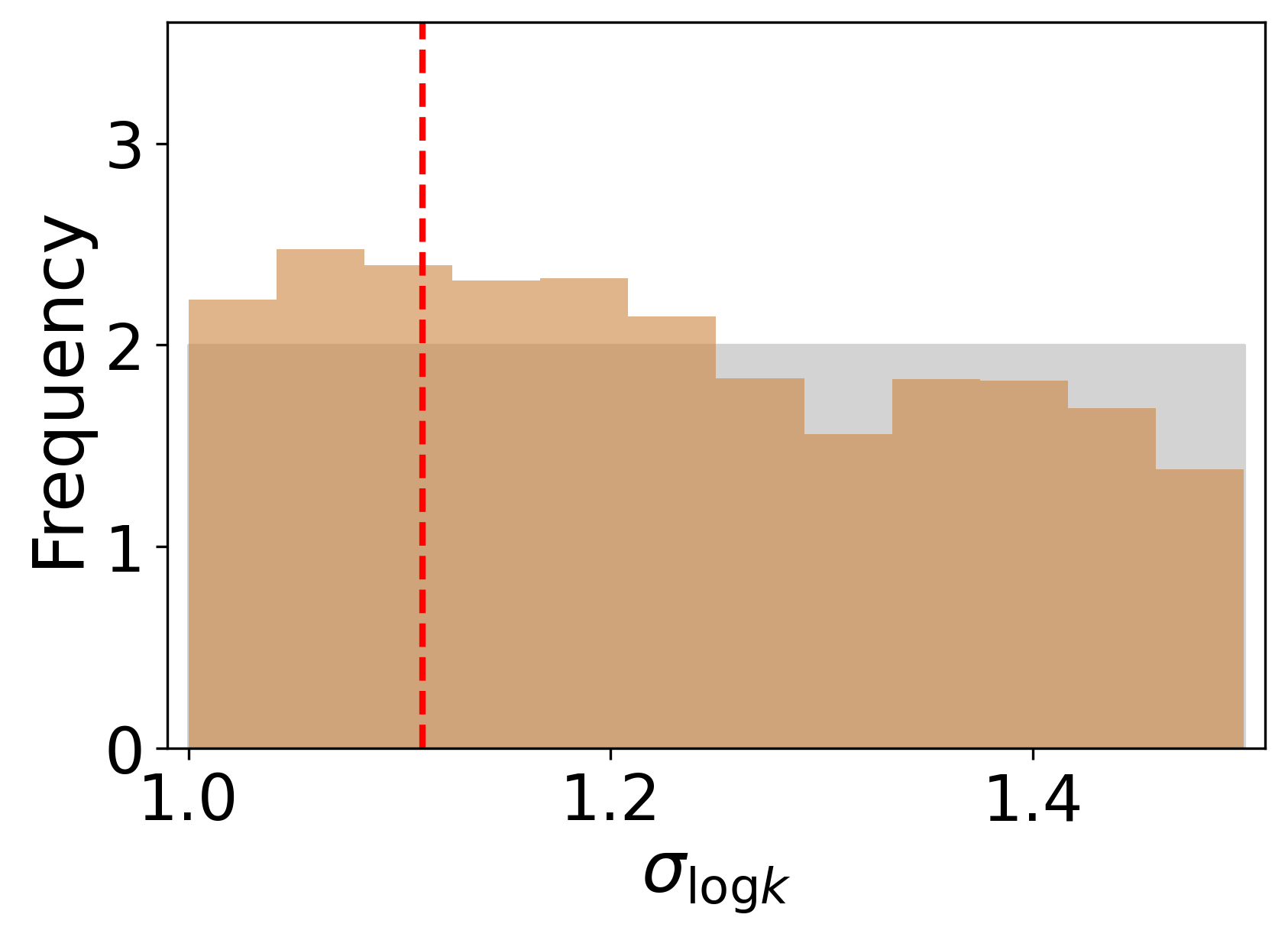}}\\
\subfloat[$a_r$ (full; $S$, $p$)]{\includegraphics[width = 43mm]{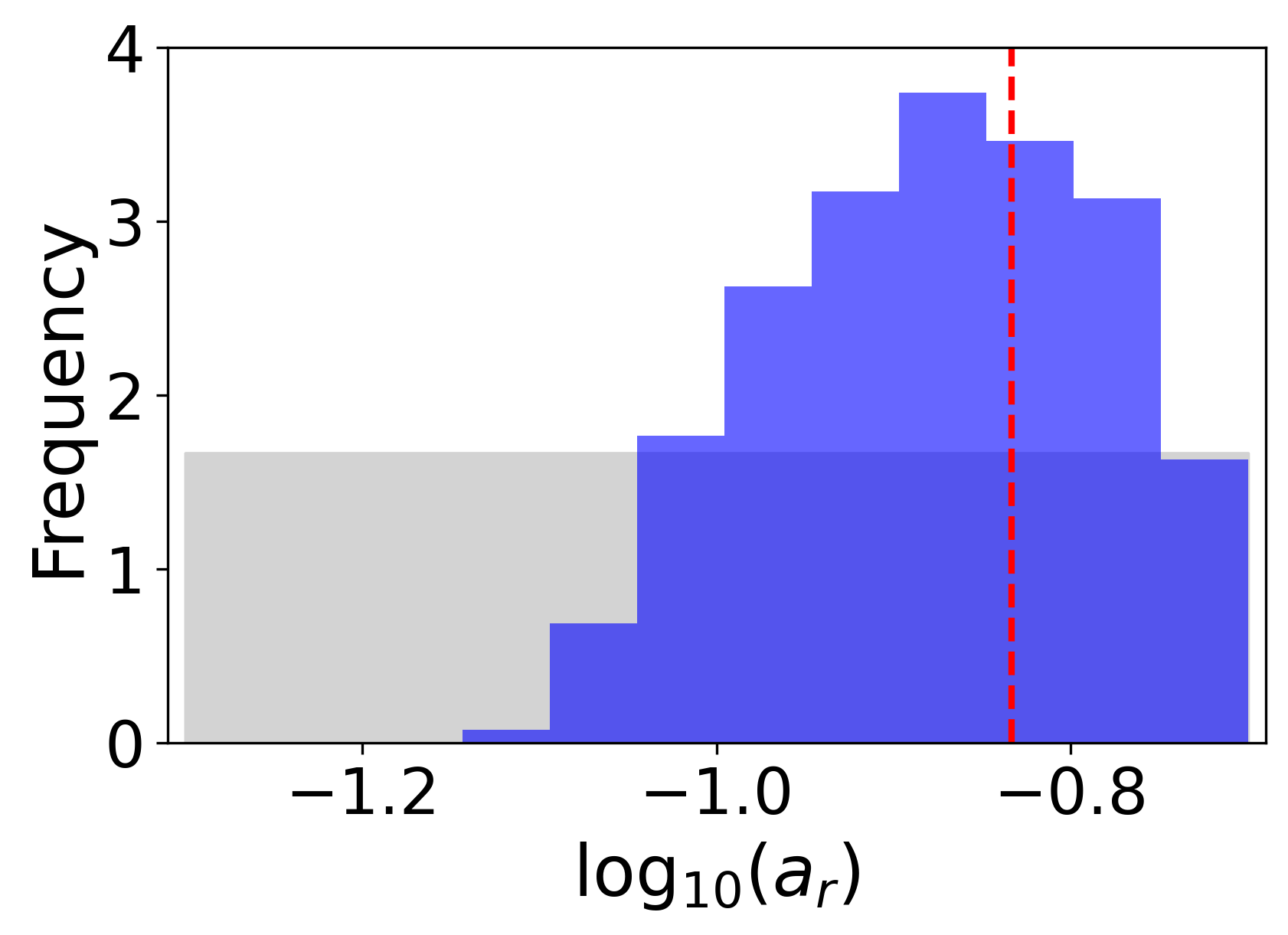}}
\hspace{2mm}
\subfloat[$a_r$ (partial; $S$, $p$)]{\includegraphics[width = 43mm]{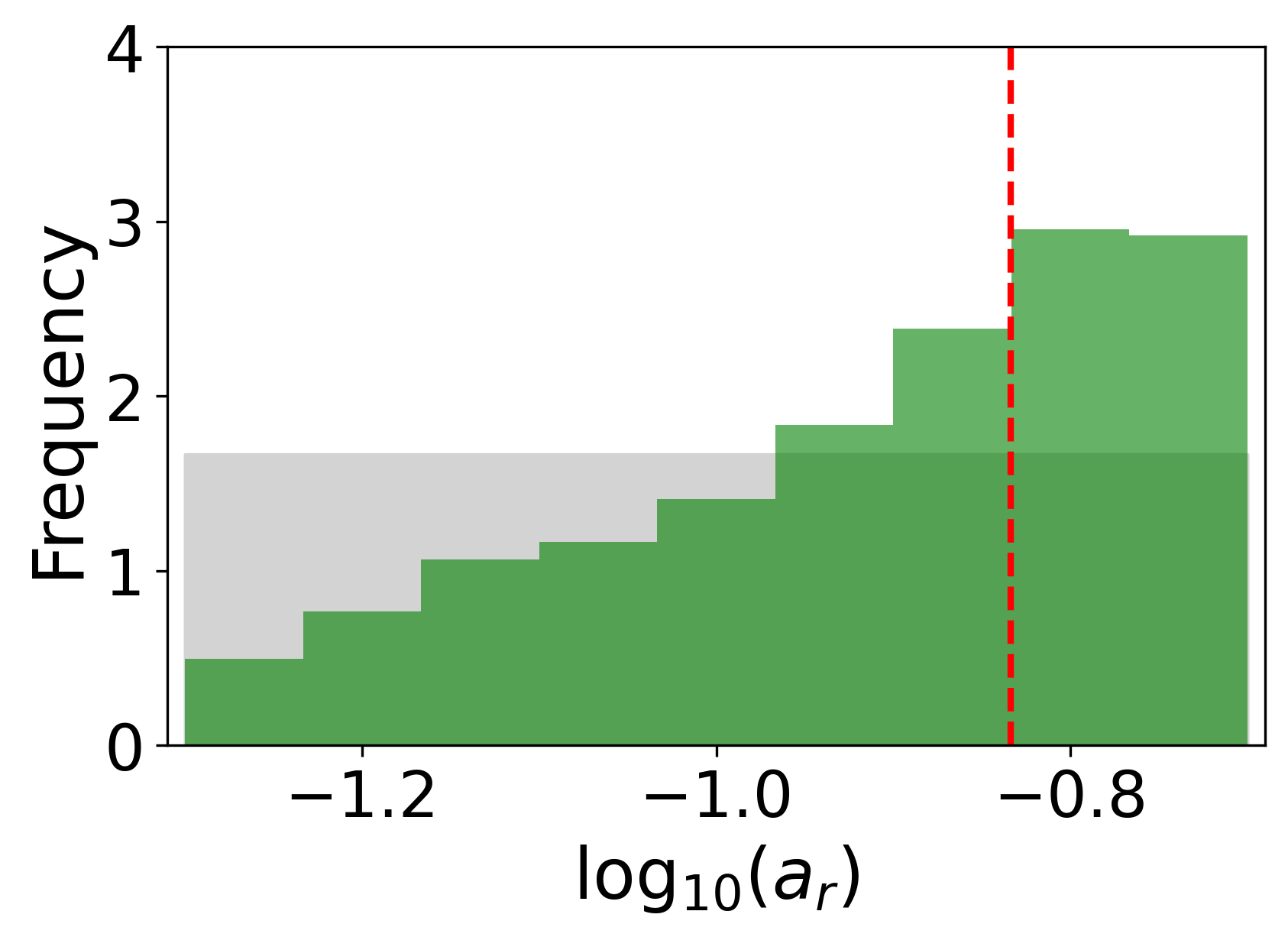}}
\hspace{2mm}
\subfloat[$a_r$ (full; $p$)]{\includegraphics[width = 43mm]{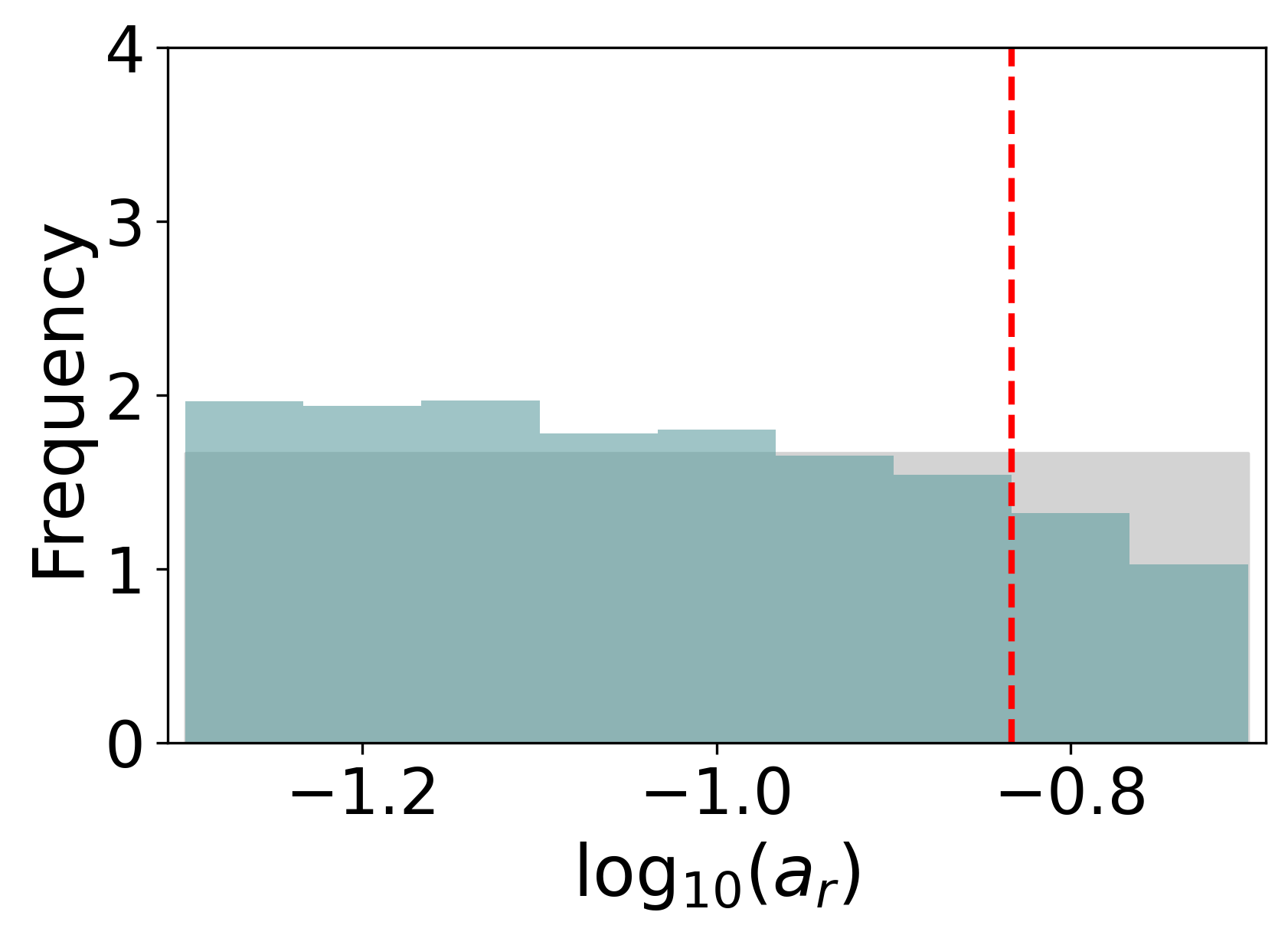}}
\hspace{2mm}
\subfloat[$a_r$ (partial; $p$)]{\includegraphics[width = 43mm]{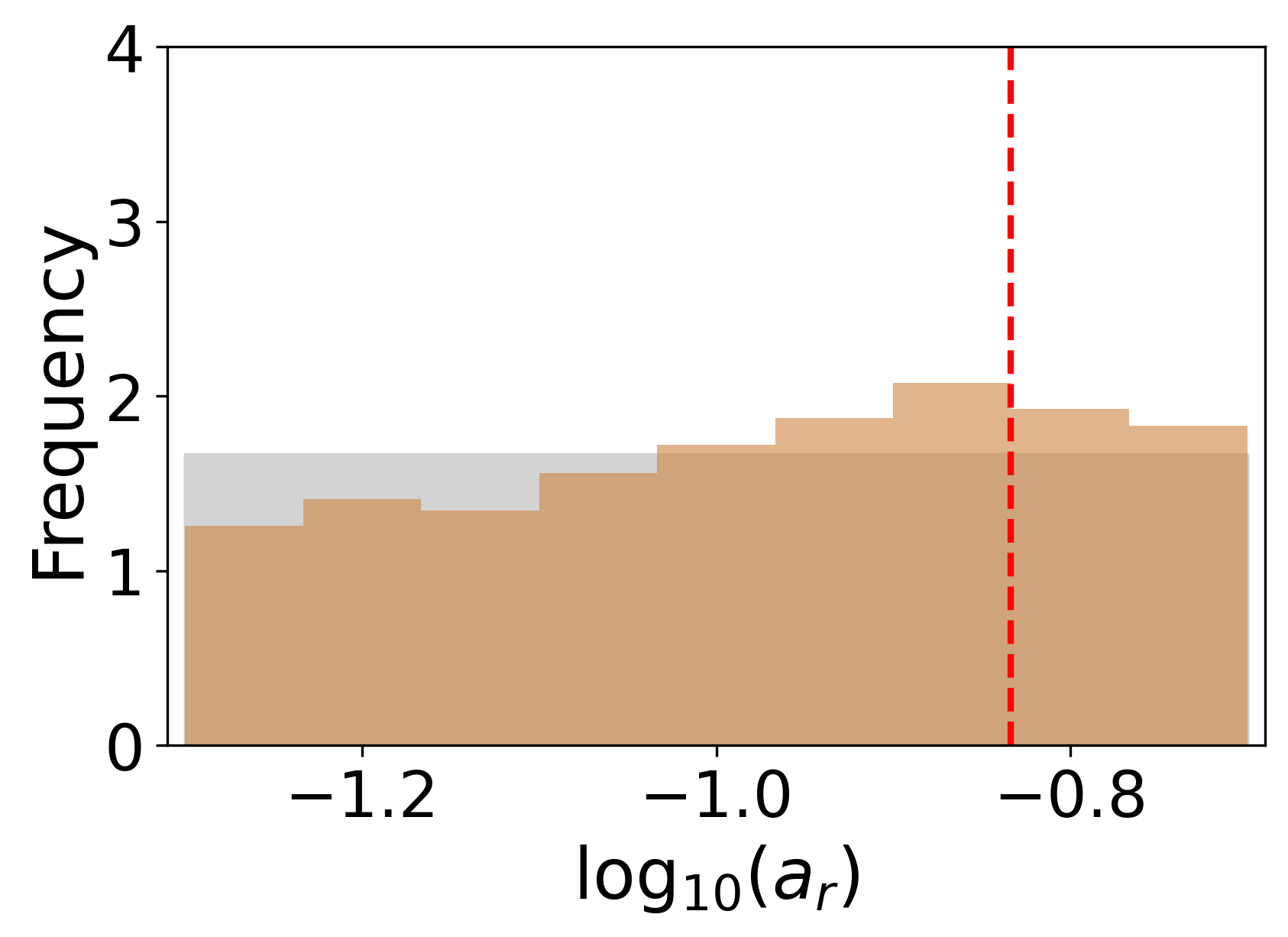}}\\
\caption{Data assimilation results for the metaparameters $\mu_{\mathrm{log}k}$, $\sigma_{\mathrm{log}k}$, and $\log_{10}(a_r)$, for true model~1, using four different monitoring strategies. Gray regions represent prior distributions, blue histograms (first column) are posterior distributions for the full monitoring strategy using both pressure and saturation data, green histograms (second column) represent posterior distributions for the partial monitoring strategy using both pressure and saturation data, cyan histograms (third column) are posterior distributions for the full monitoring strategy using only pressure data, and brown histograms (fourth column) are posterior distributions for the partial monitoring strategy using only pressure data. Red vertical lines indicate true values.}
\label{meta_true_1}
\end{figure}

\begin{figure}[!ht]
\centering 
\subfloat[$k_{f1}^{tm}$ (full; $S$, $p$)]{\includegraphics[width = 43mm]{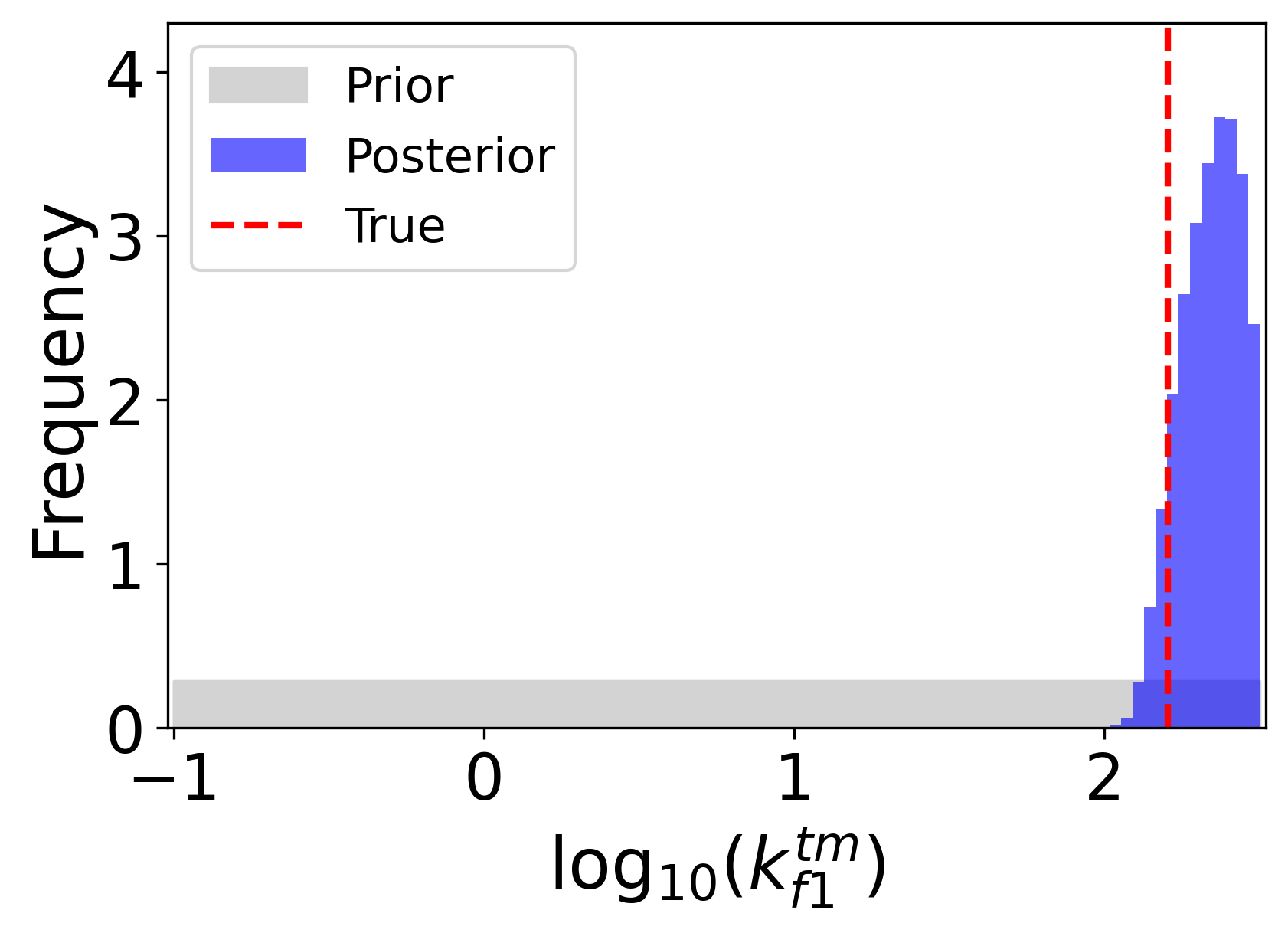}}
\hspace{2mm}
\subfloat[$k_{f1}^{tm}$ (partial; $S$, $p$)]{\includegraphics[width = 43mm]{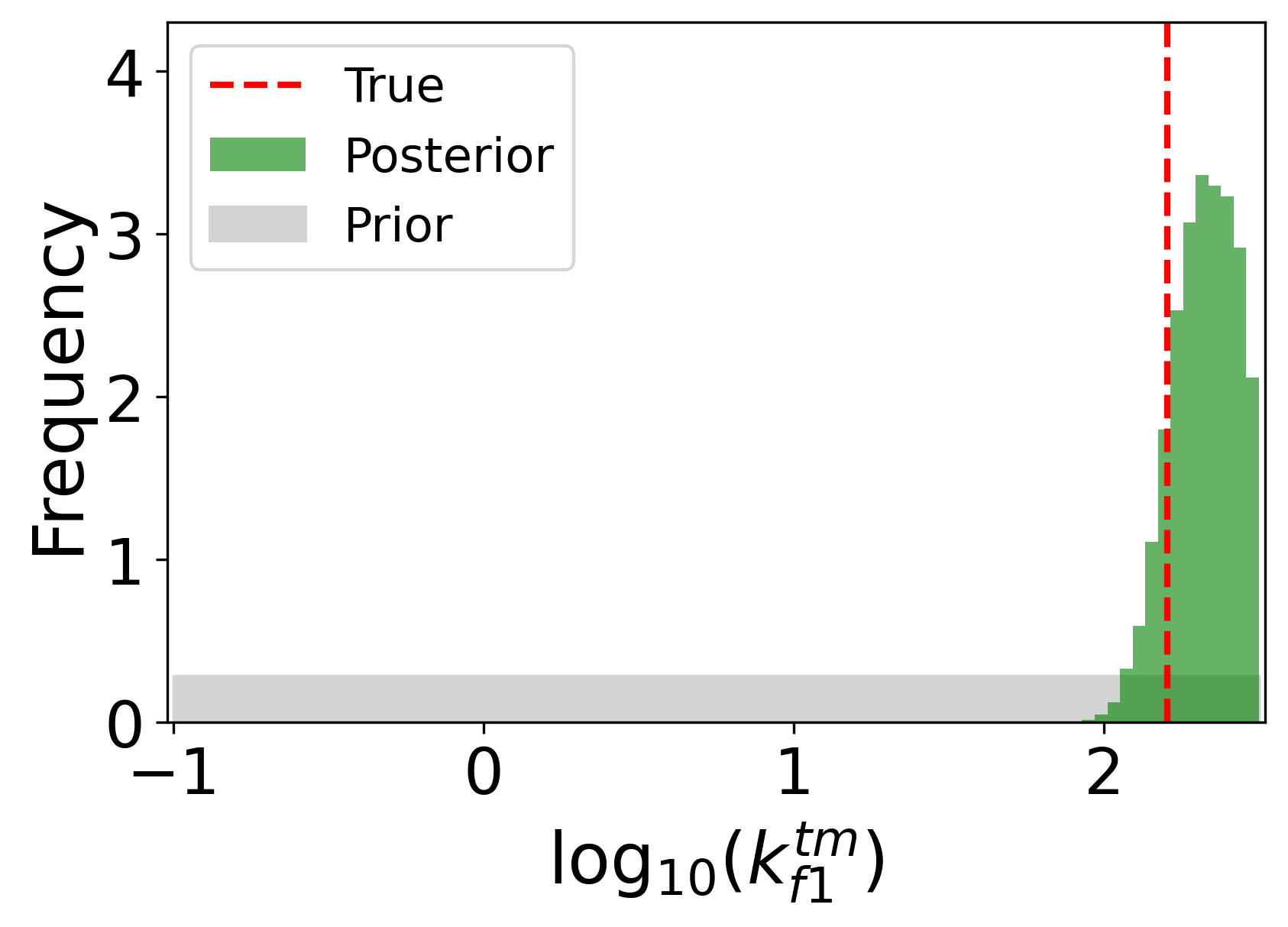}}
\hspace{2mm}
\subfloat[$k_{f1}^{tm}$ (full; $p$)]{\includegraphics[width = 43mm]{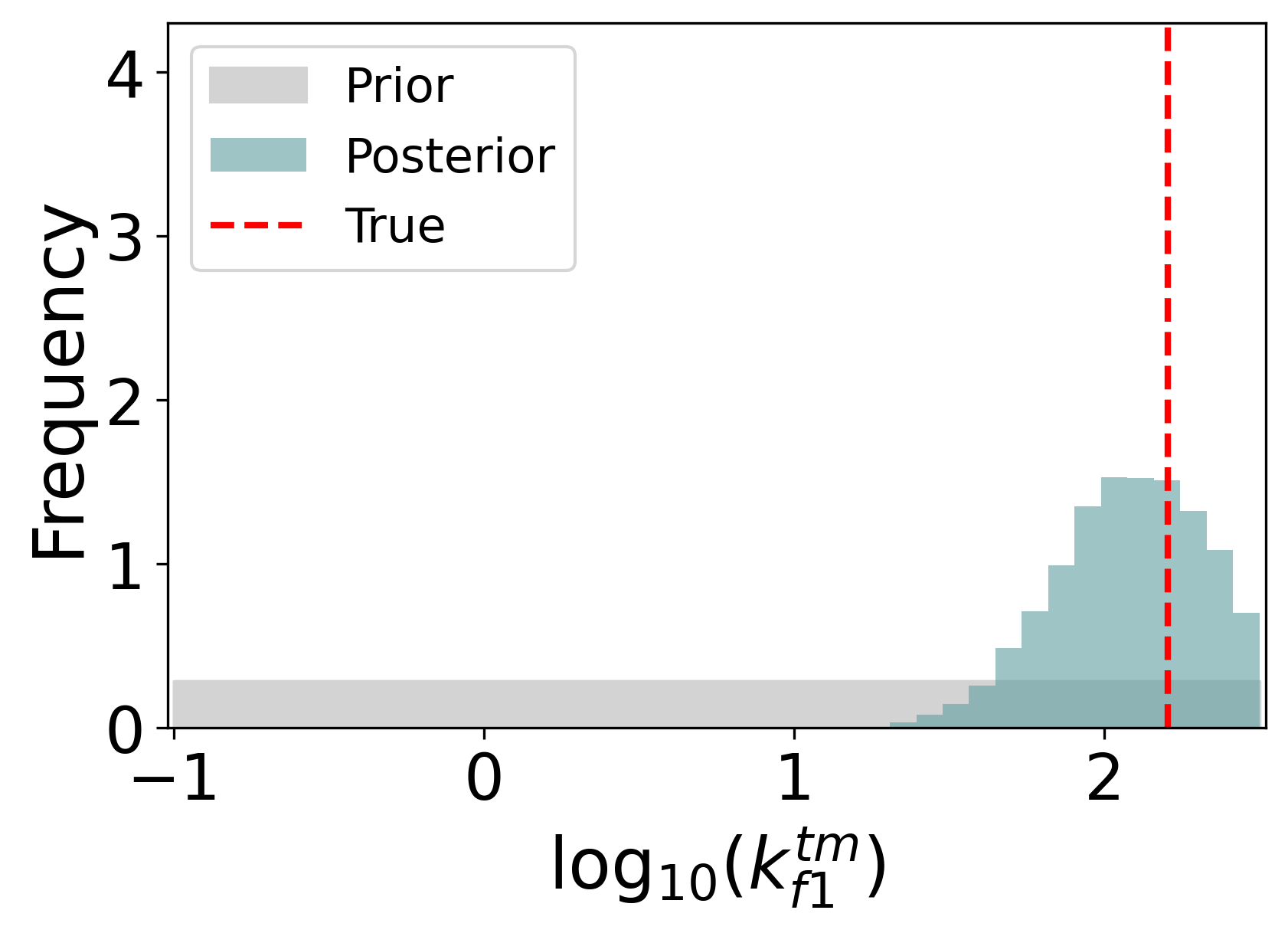}}
\hspace{2mm}
\subfloat[$k_{f1}^{tm}$ (partial; $p$)]{\includegraphics[width = 43mm]{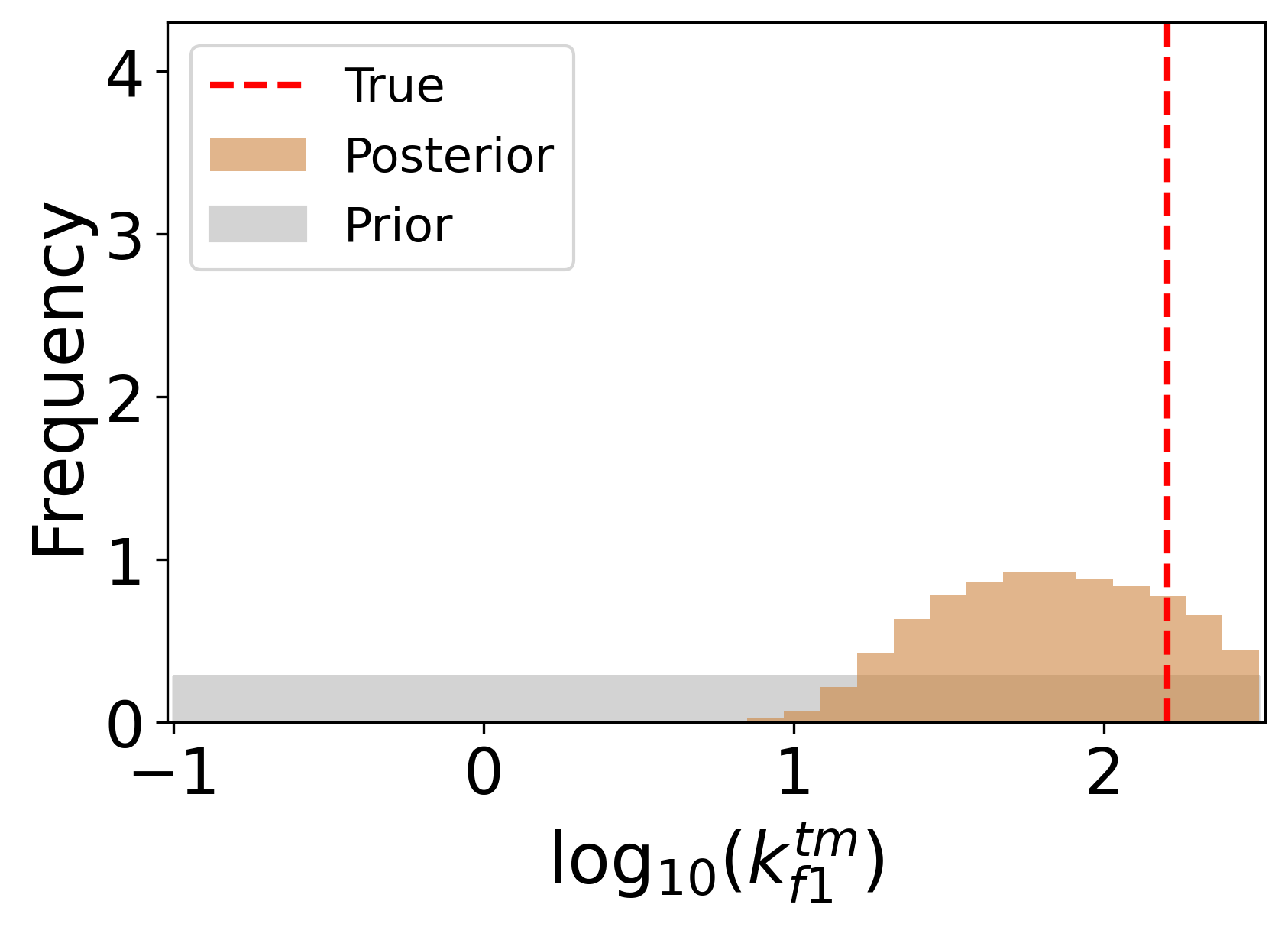}}\\
\subfloat[$k_{f1}^{mu}$ (full; $S$, $p$)]{\includegraphics[width = 43mm]{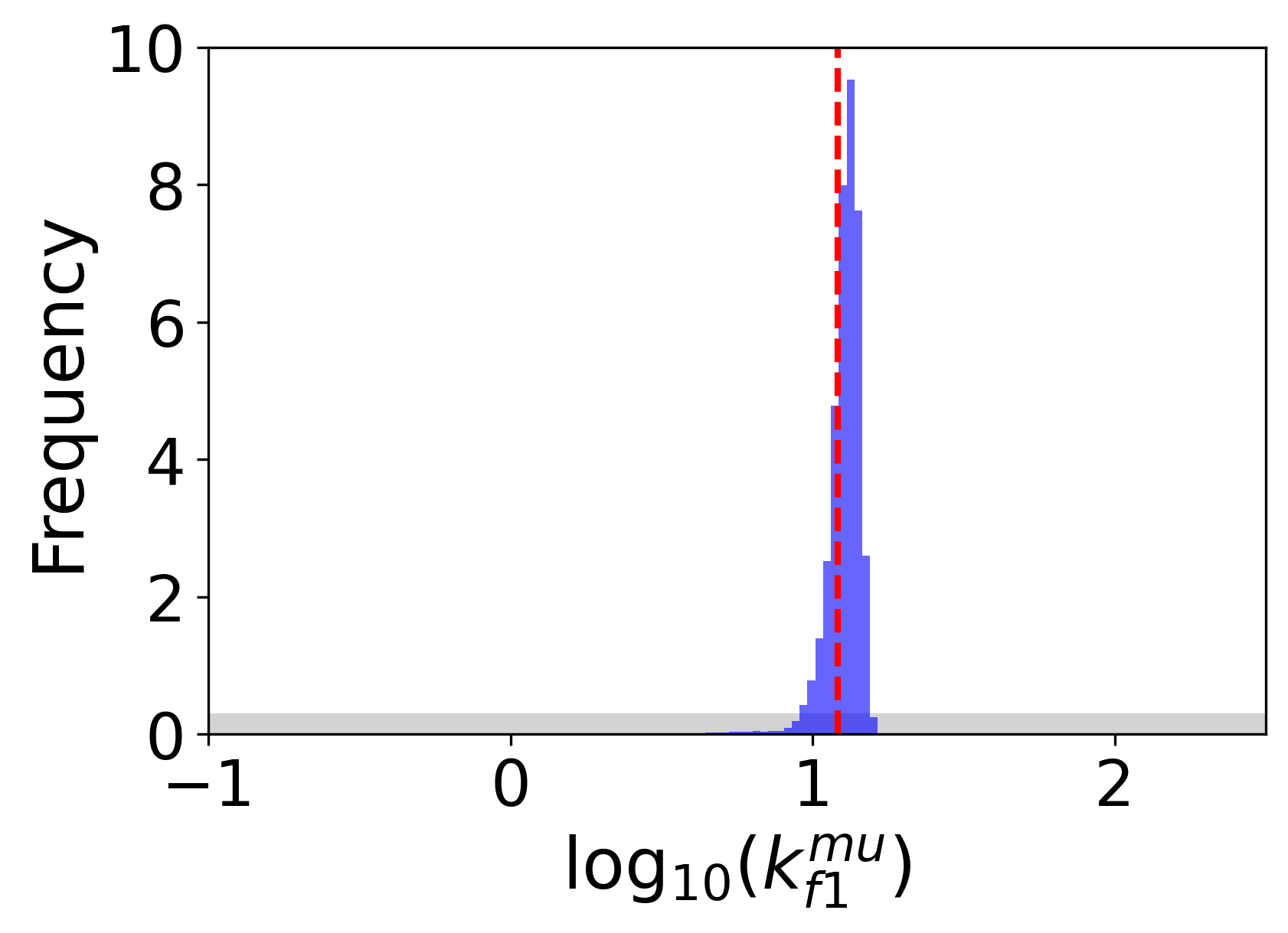}}
\hspace{2mm}
\subfloat[$k_{f1}^{mu}$ (partial; $S$, $p$)]{\includegraphics[width = 43mm]{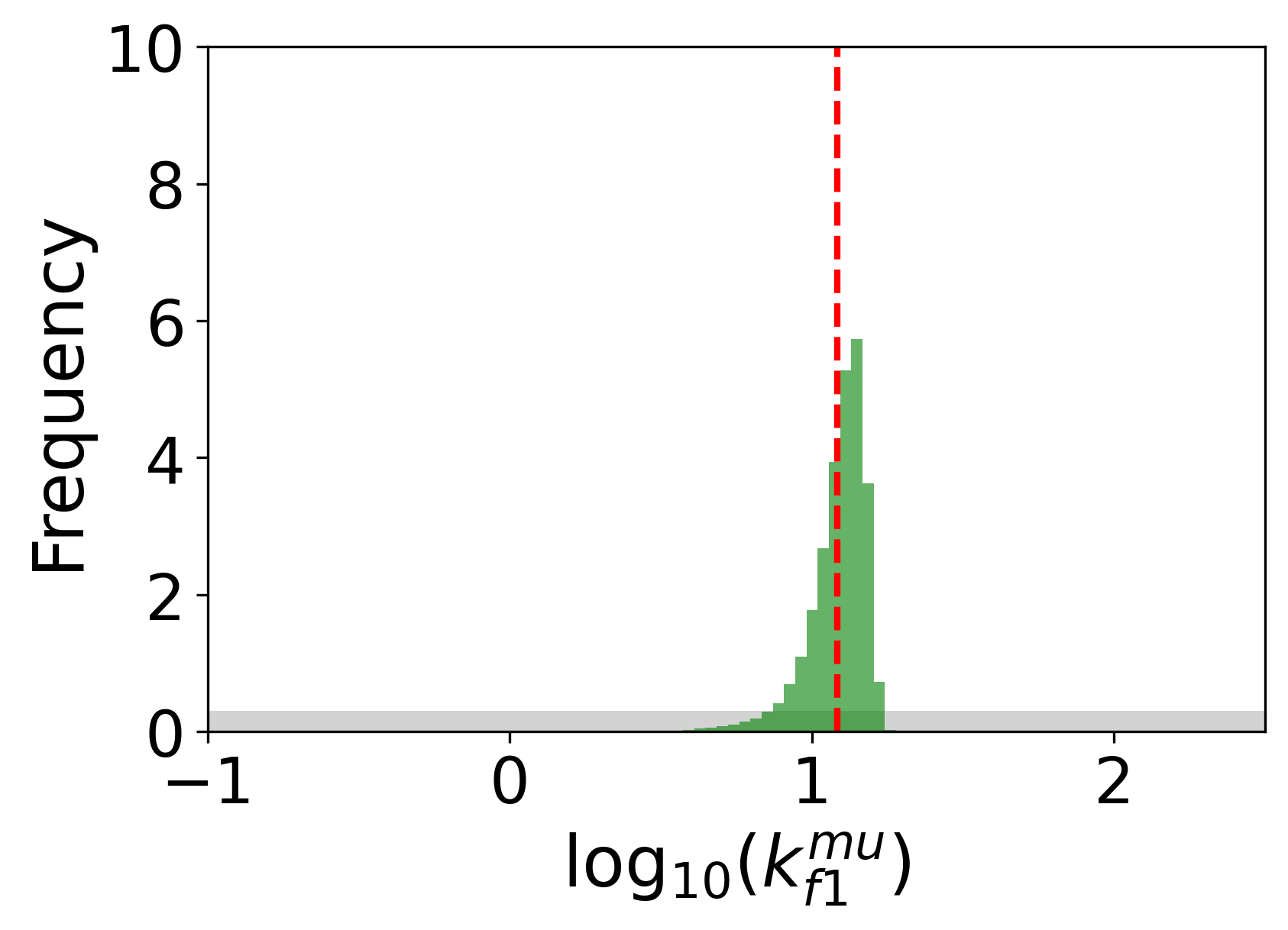}}
\hspace{2mm}
\subfloat[$k_{f1}^{mu}$ (full; $p$)]{\includegraphics[width = 43mm]{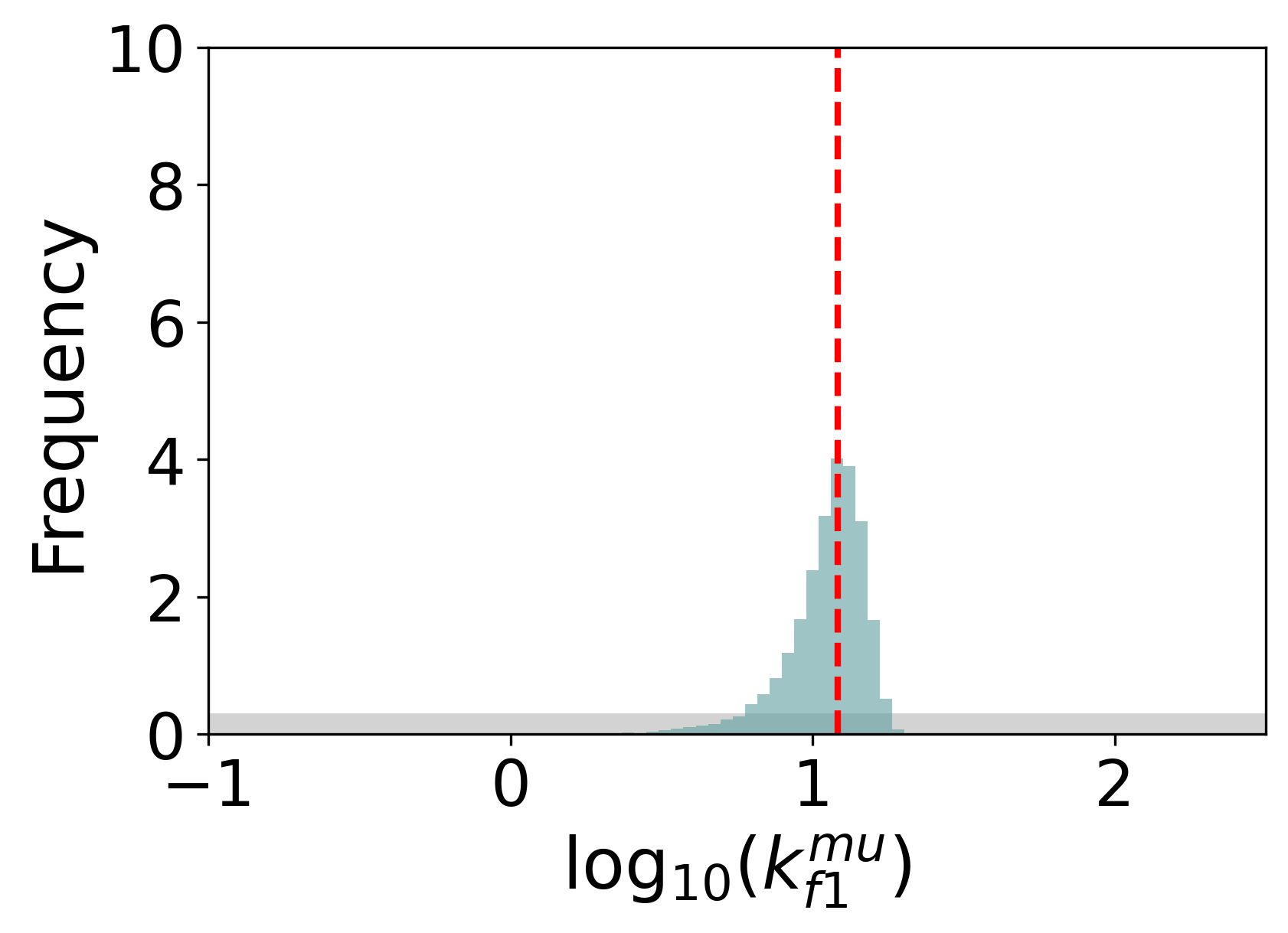}}
\hspace{2mm}
\subfloat[$k_{f1}^{mu}$ (partial; $p$)]{\includegraphics[width = 43mm]{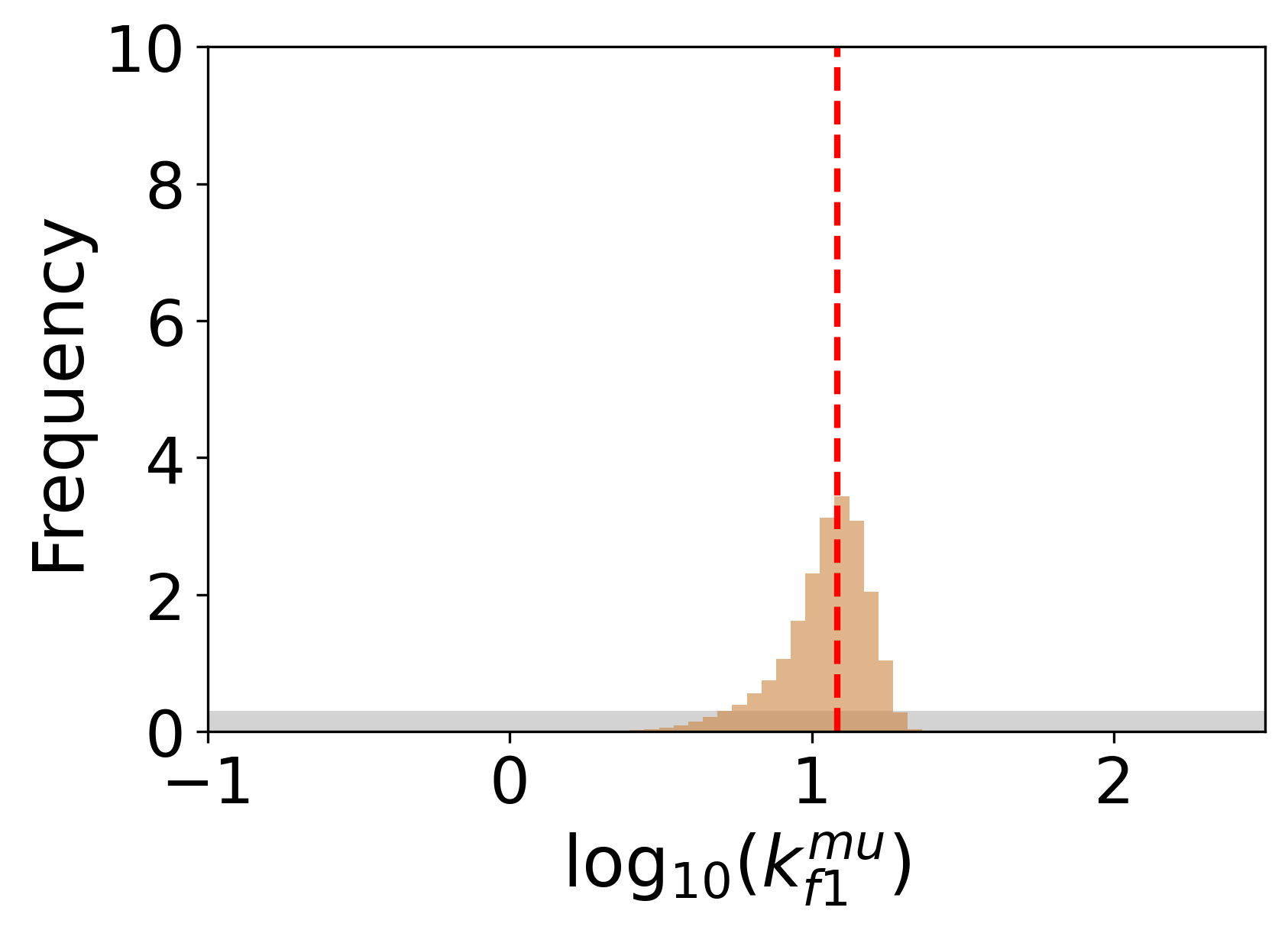}}\\
\subfloat[$k_{f2}^{tm}$ (full; $S$, $p$)]{\label{k2_tm_full}\includegraphics[width = 43mm]{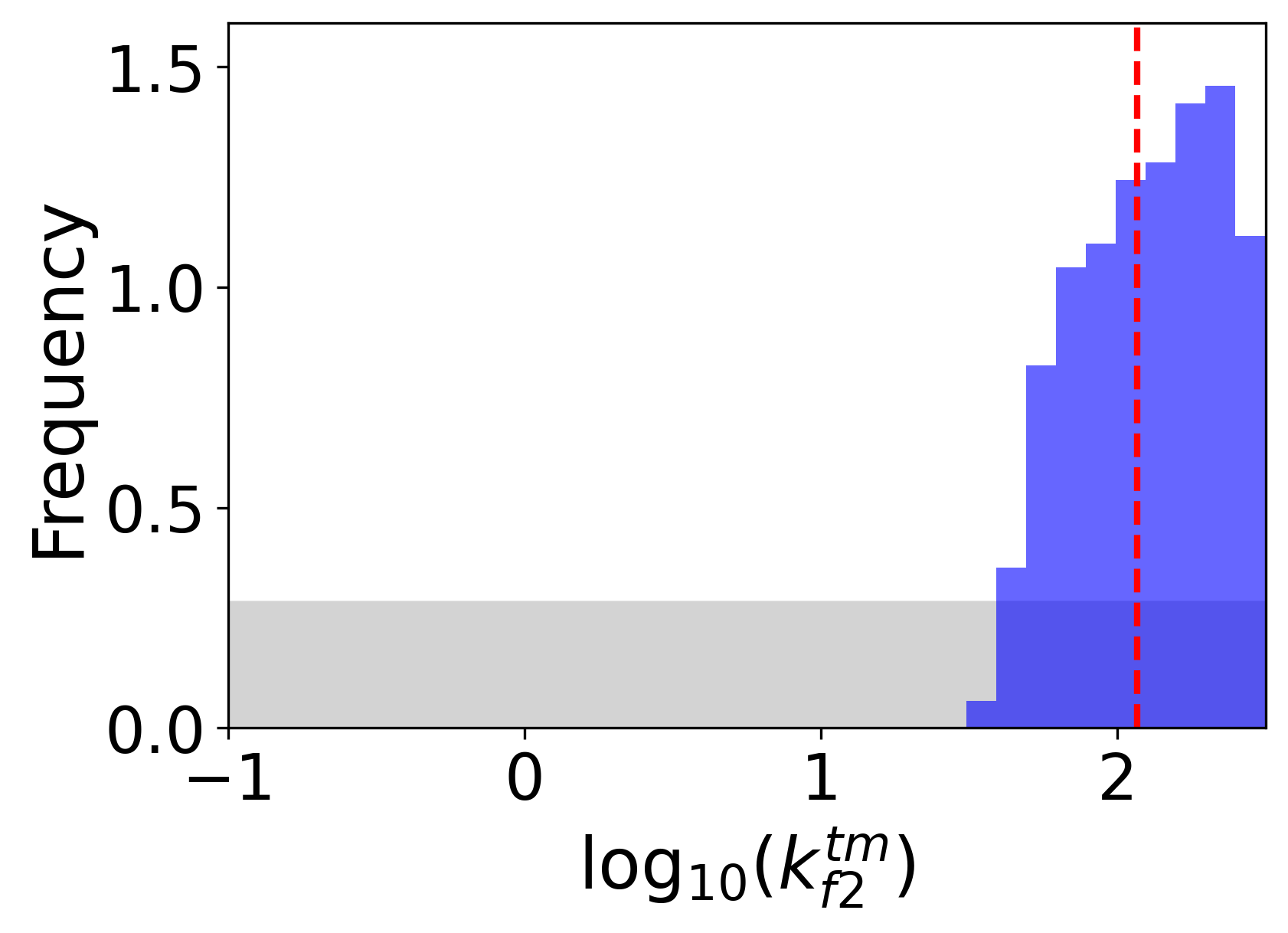}}
\hspace{2mm}
\subfloat[$k_{f2}^{tm}$ (partial; $S$, $p$)]{\includegraphics[width = 43mm]{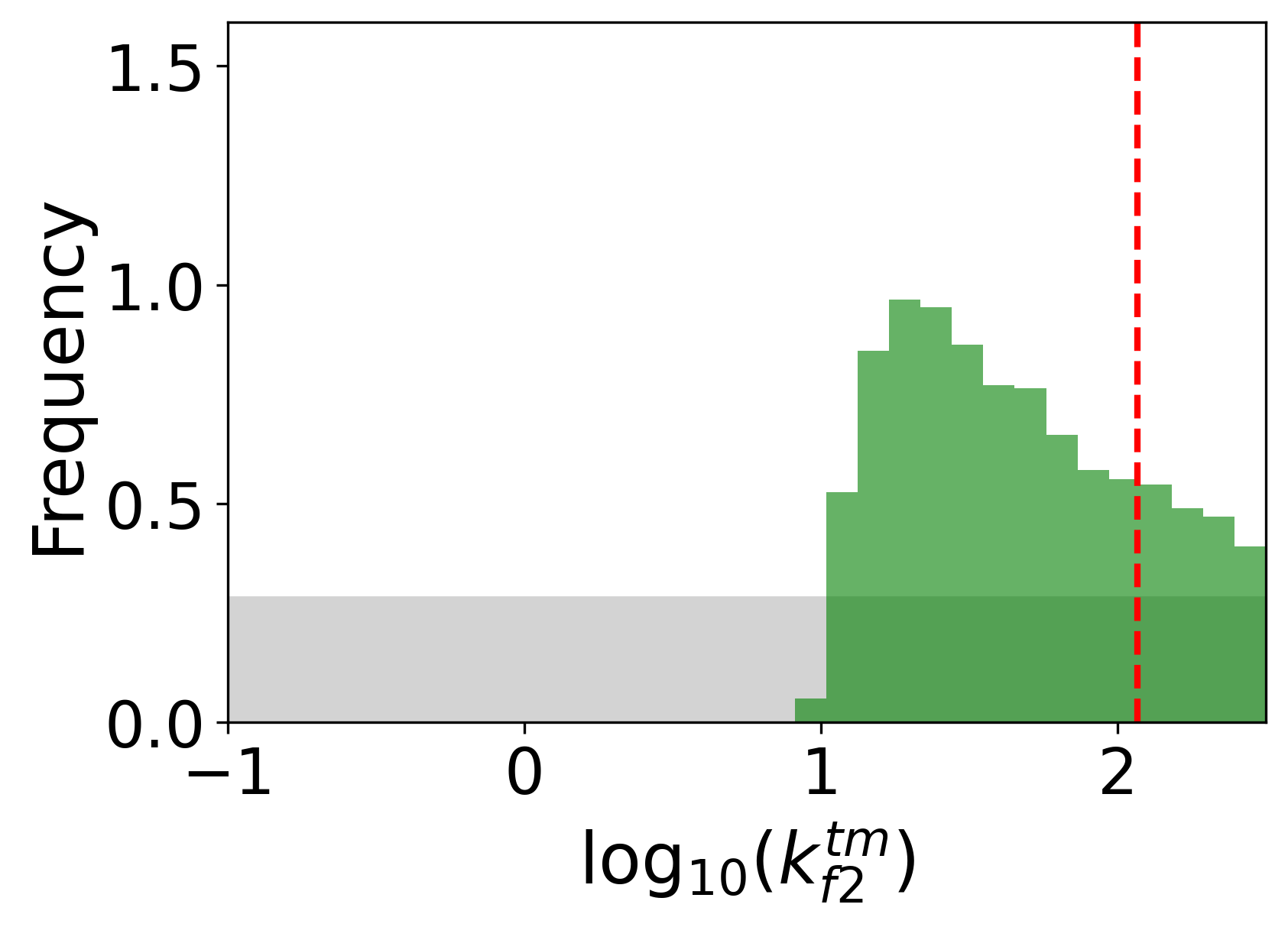}}
\hspace{2mm}
\subfloat[$k_{f2}^{tm}$ (full; $p$)]{\includegraphics[width = 43mm]{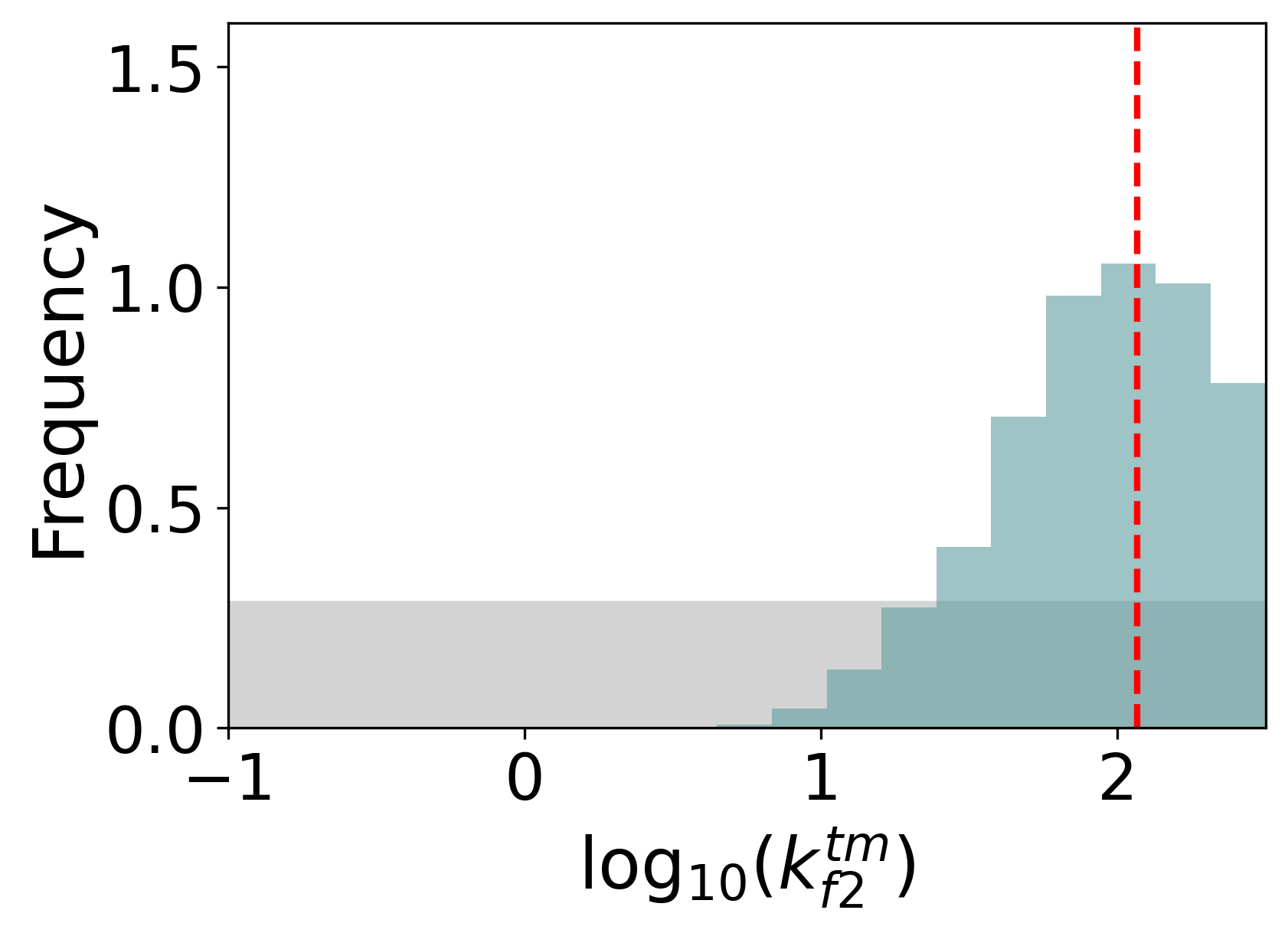}}
\hspace{2mm}
\subfloat[$k_{f2}^{tm}$ (partial; $p$)]{\includegraphics[width = 43mm]{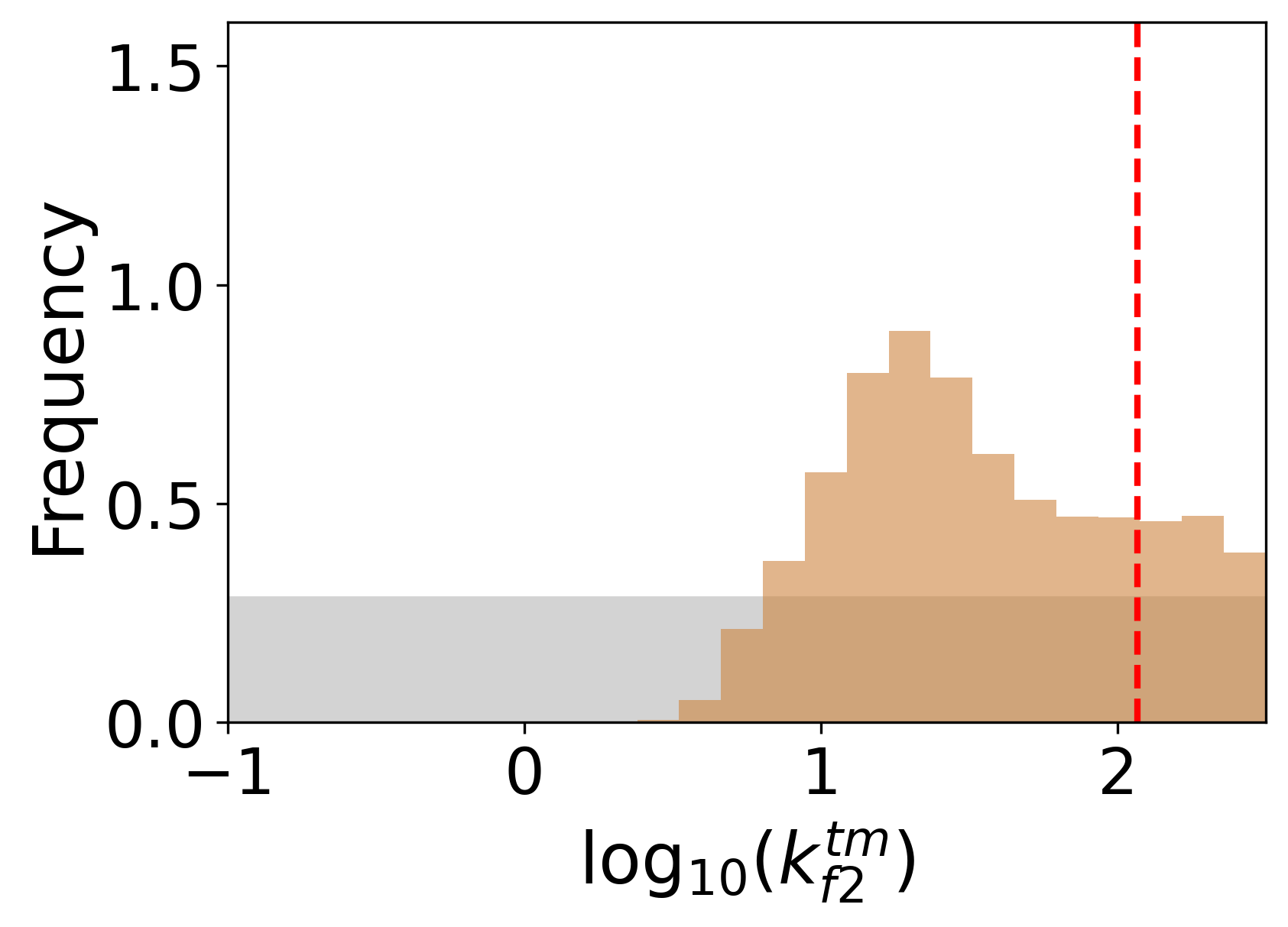}}\\
\subfloat[$k_{f2}^{mu}$ (full; $S$, $p$)]{\includegraphics[width = 43mm]{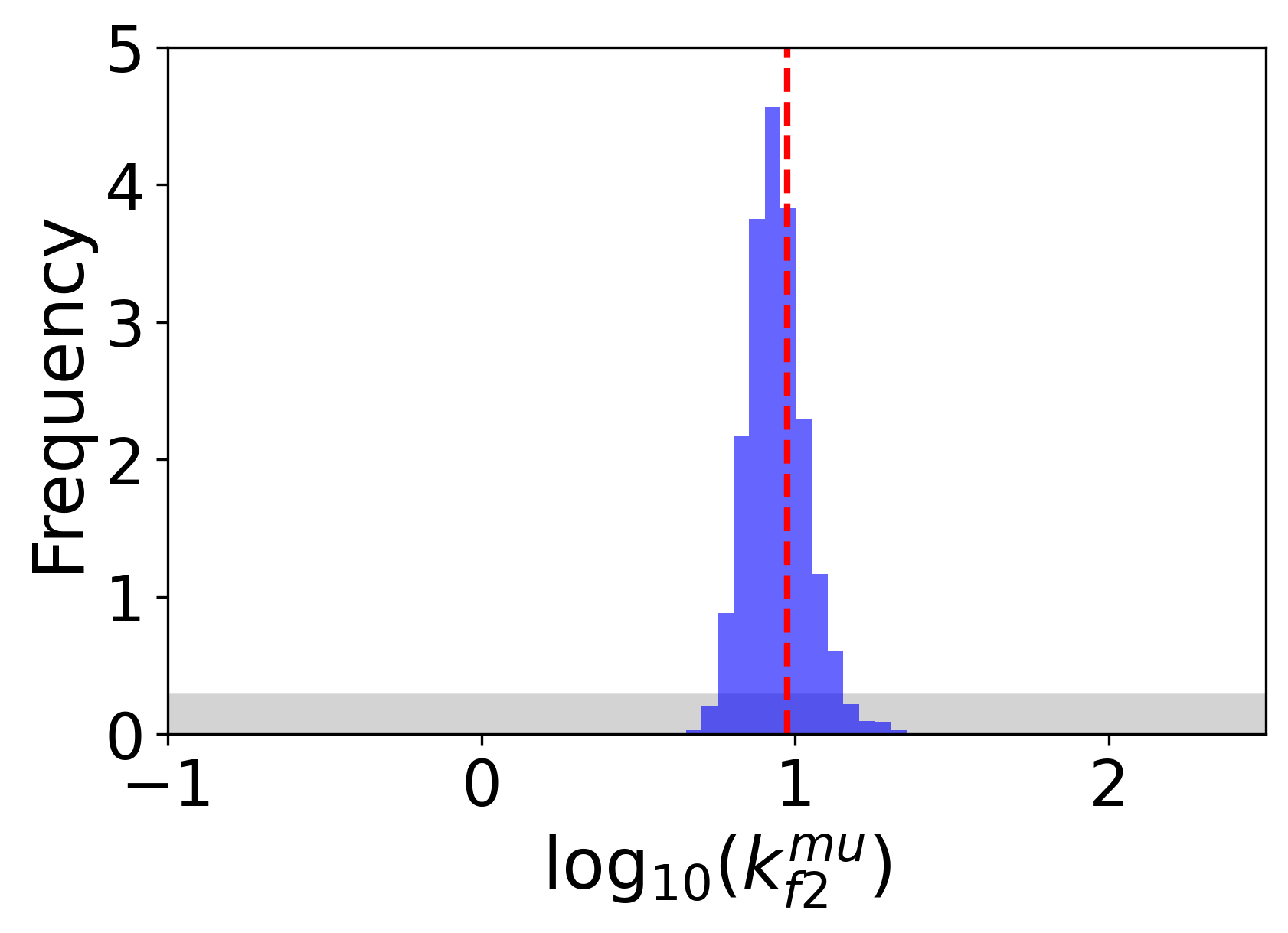}}
\hspace{2mm}
\subfloat[$k_{f2}^{mu}$ (partial; $S$, $p$)]{\includegraphics[width = 43mm]{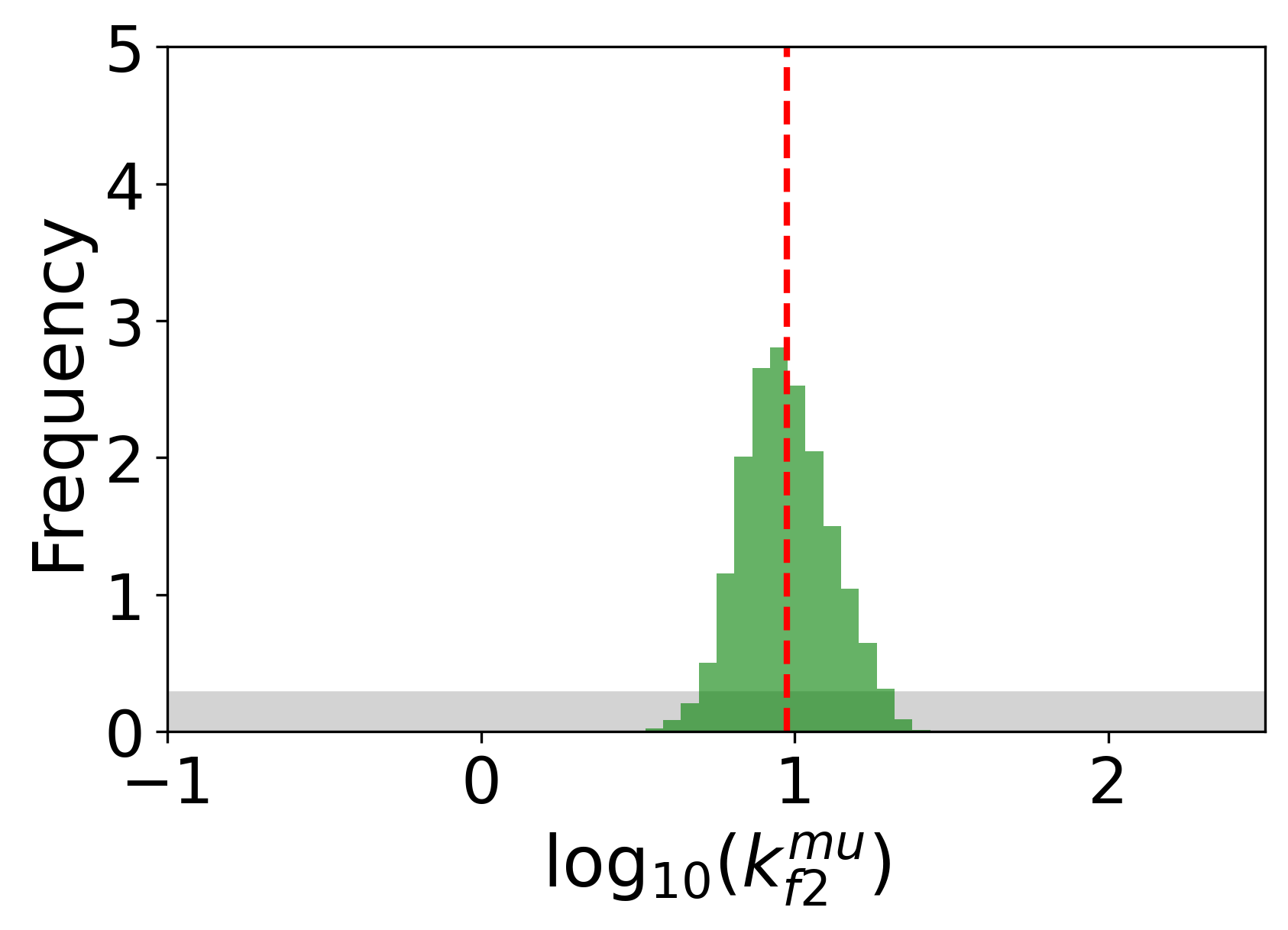}}
\hspace{2mm}
\subfloat[$k_{f2}^{mu}$ (full; $p$)]{\includegraphics[width = 43mm]{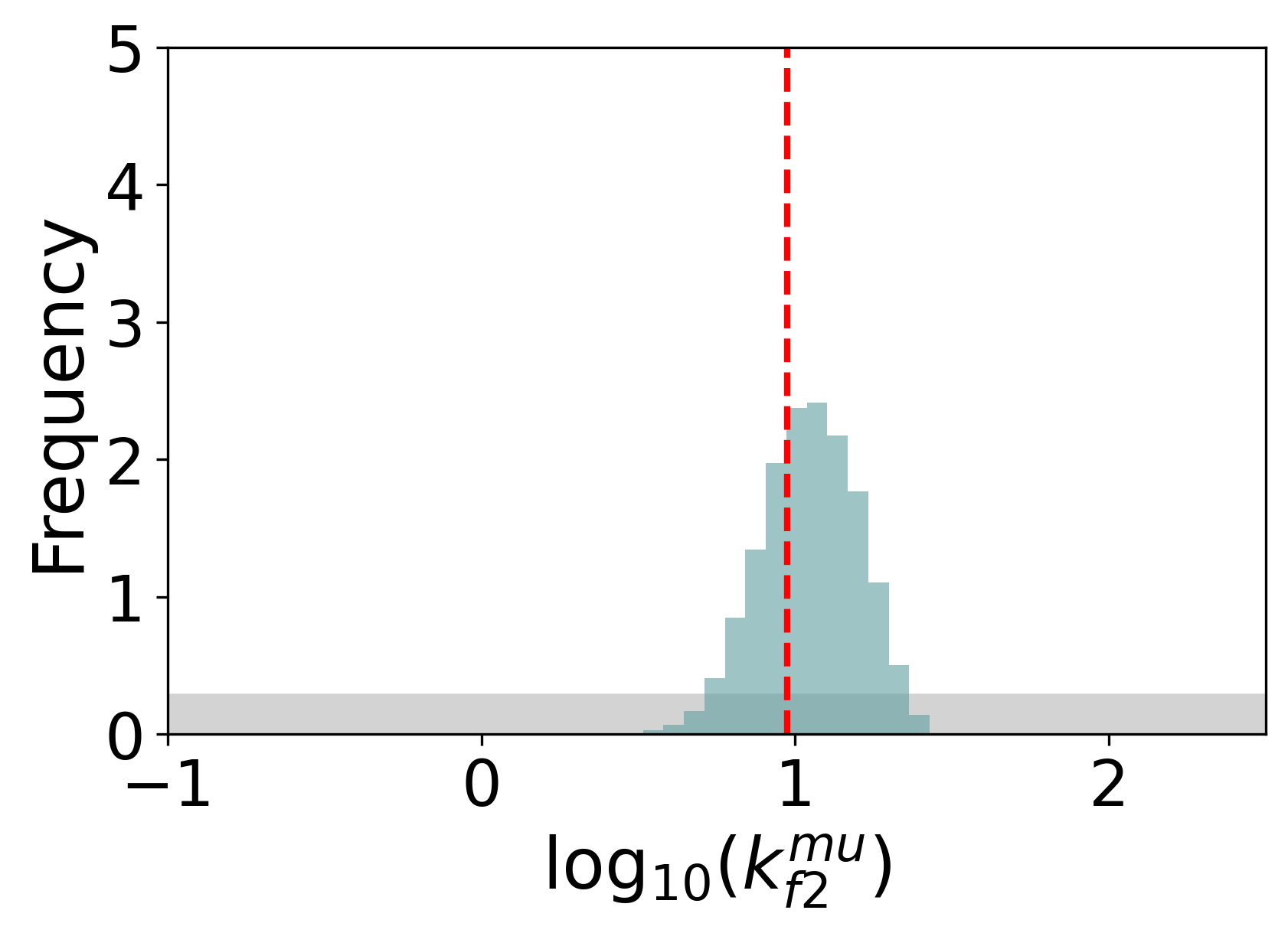}}
\hspace{2mm}
\subfloat[$k_{f2}^{mu}$ (partial; $p$)]{\includegraphics[width = 43mm]{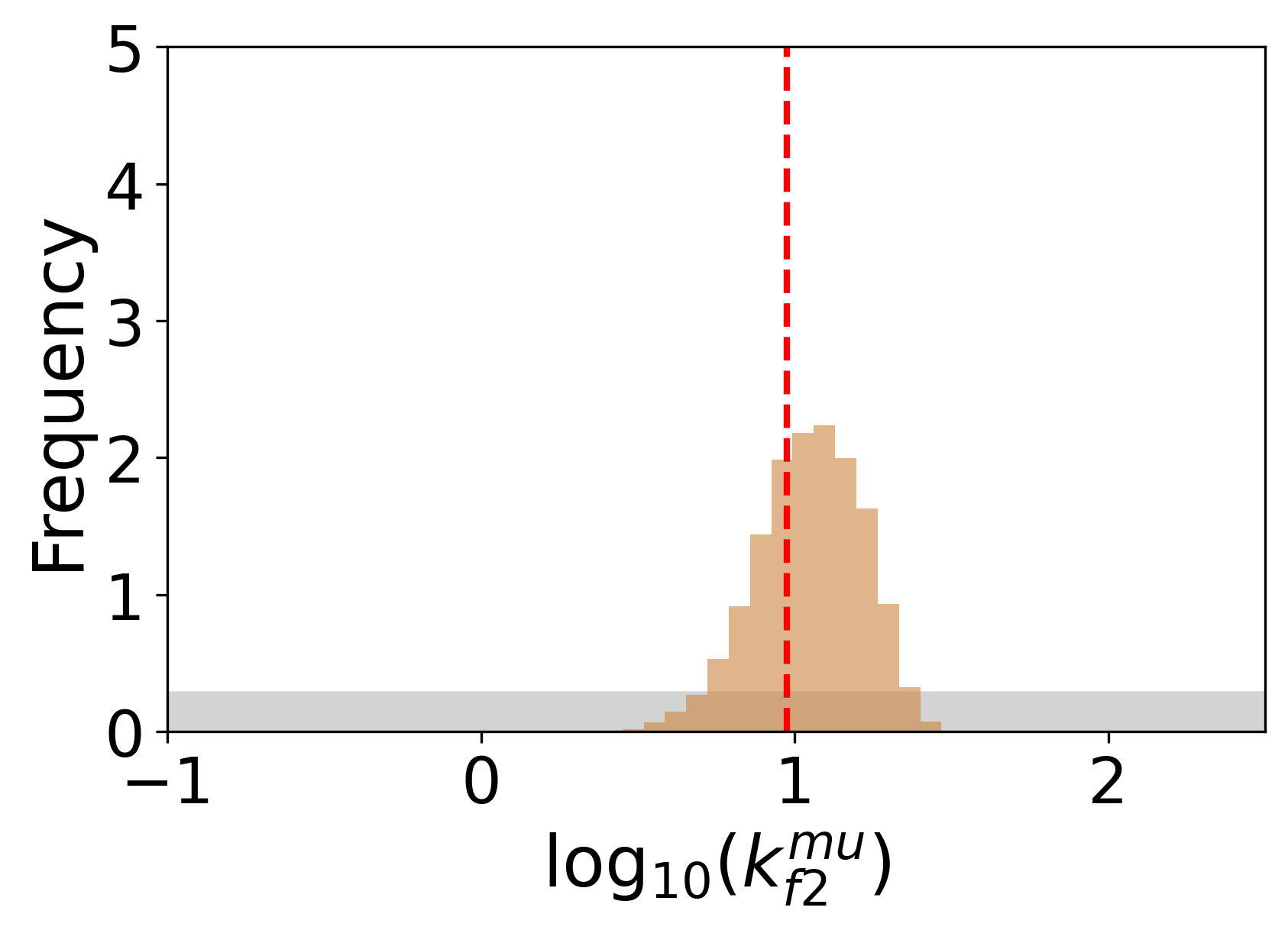}}\\
\caption{Data assimilation results for the metaparameters $k_{f1}^{tm}$, $k_{f1}^{mu}$, $k_{f2}^{tm}$, and $k_{f2}^{mu}$ for true model~1, using four different monitoring strategies. Gray regions represent prior distributions, blue histograms (first column) are posterior distributions for the full monitoring strategy using both pressure and saturation data, green histograms (second column) represent posterior distributions for the partial monitoring strategy using both pressure and saturation data, cyan histograms (third column) are posterior distributions for the full monitoring strategy using only pressure data, and brown histograms (fourth column) are posterior distributions for the partial monitoring strategy using only pressure data. Red vertical lines indicate true values.}
\label{meta_1_true_1}
\end{figure}

Data assimilation results for key metaparameters, using the four different monitoring strategies, are presented in Figs.~\ref{meta_true_1} and \ref{meta_1_true_1}. The gray regions in all subfigures display the uniform prior distributions and the red vertical lines denote the true value of the metaparameter. Results in the first column (blue histograms) show the posterior metaparameter distributions using the full monitoring strategy, in which both pressure and saturation data are measured in all three aquifers at the observation wells. The second column (green histograms) corresponds to posterior distributions for the partial monitoring strategy (monitoring wells in only the middle and upper aquifers), with both pressure and saturation data. The third column (cyan histograms) displays posterior distributions for the full monitoring strategy using only pressure data, while the fourth column (brown histograms) shows posterior distributions for the partial monitoring strategy using only pressure data.

The use of the full monitoring strategy with both pressure and saturation data provides the most uncertainty reduction for all metaparameters. This is consistent with expectations, since this scenario involves the largest amount of observed data. We also see that the true metaparameter values consistently fall within the posterior distributions (this is the case with all monitoring strategies). By comparing results in the first and third columns, we can see the impact of saturation data on the posterior distributions with the full monitoring strategy. It is evident that saturation data are informative for all metaparameters, though these data especially impact $\mu_{\mathrm{log}k}$, $a_r$, and the fault-related quantities. 

Results of the type shown in Figs.~\ref{meta_true_1} and \ref{meta_1_true_1} (and later figures) are useful for designing monitoring plans as they quantify the amount of uncertainty reduction that can be expected. Along these lines, if the fault permeability quantities are of particular interest, the results in Fig.~\ref{meta_1_true_1} (third and fourth columns) show that these can be estimated using only pressure data. Importantly, we see that even the partial monitoring strategy, with only pressure measured in the middle and upper aquifers, provides some uncertainty reduction for these quantities. This is likely because pressure communication from the target aquifer to the middle and upper aquifers is strongly impacted by the fault permeabilities.

\begin{figure}[!ht]
\centering   
\subfloat[Prior 1]{\includegraphics[width=34mm]{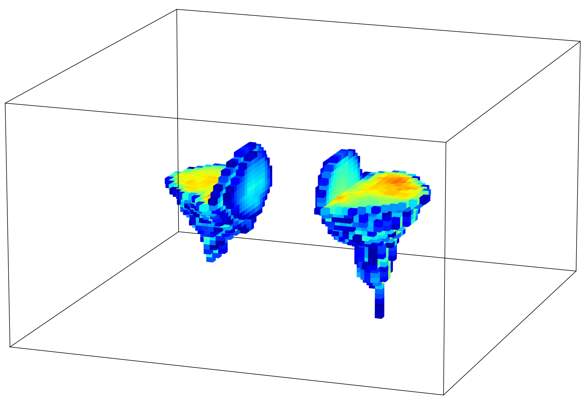}}
\hspace{0.2mm}
\subfloat[Prior 2]{\includegraphics[width=34mm]{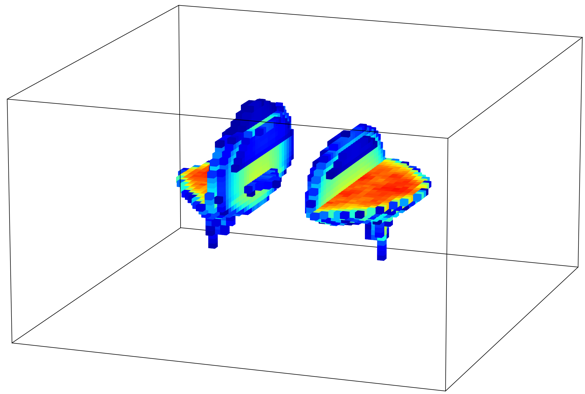}}
\hspace{0.2mm}
\subfloat[Prior 3]{\includegraphics[width=34mm]{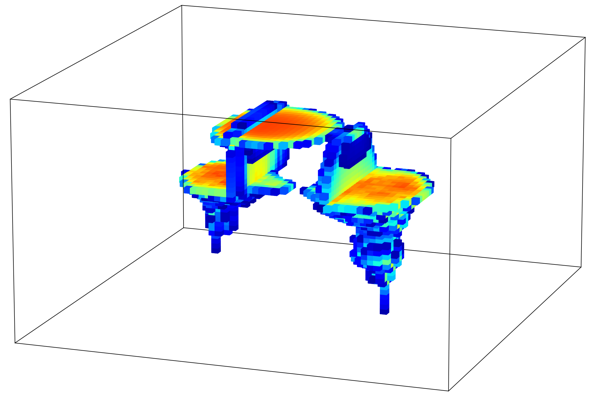}}
\hspace{0.2mm}
\subfloat[Prior 4]{\includegraphics[width=34mm]{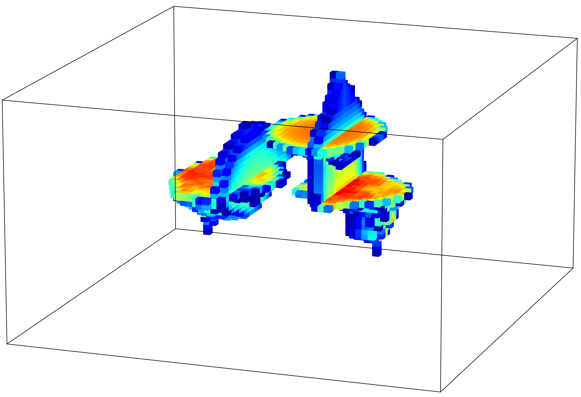}}
\hspace{0.2mm}
\subfloat[Prior 5]{\includegraphics[width=34mm]{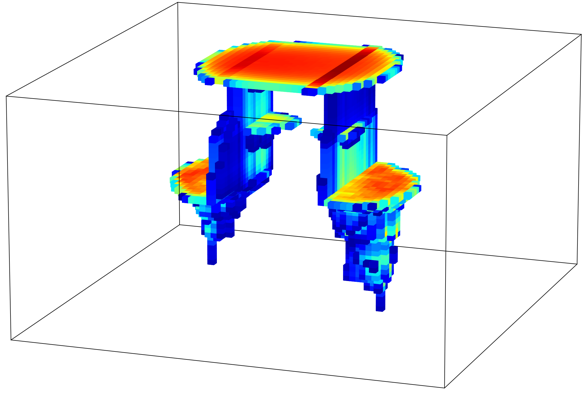}}
\includegraphics[width=5.8mm]{Sw_Scale.png}\\[1ex]
\subfloat[Posterior 1]{\includegraphics[width=34mm]{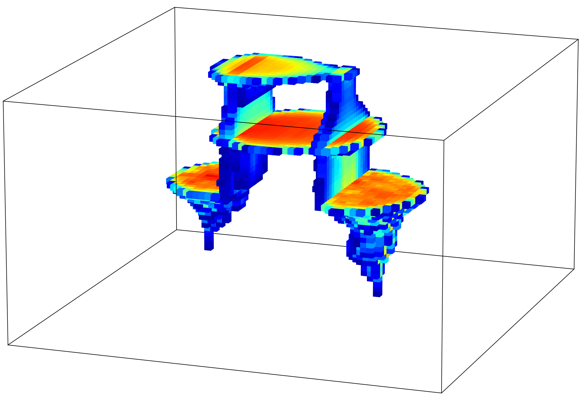}}
\hspace{0.2mm}
\subfloat[Posterior 2]{\includegraphics[width=34mm]{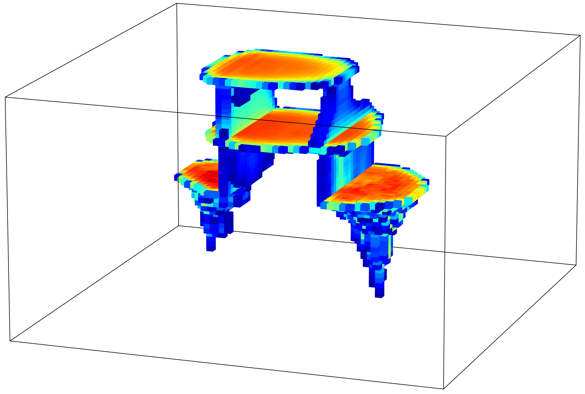}}
\hspace{0.2mm}
\subfloat[Posterior 3]{\includegraphics[width=34mm]{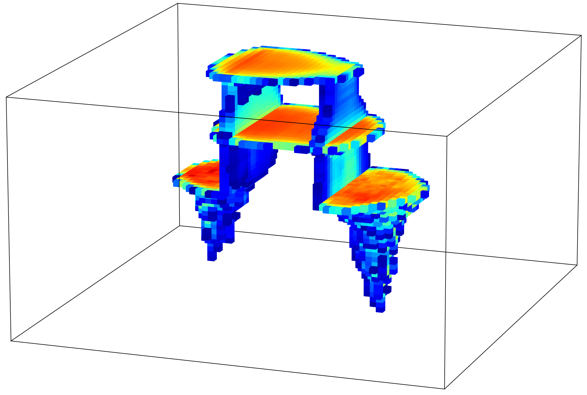}}
\hspace{0.2mm}
\subfloat[Posterior 4]{\includegraphics[width=34mm]{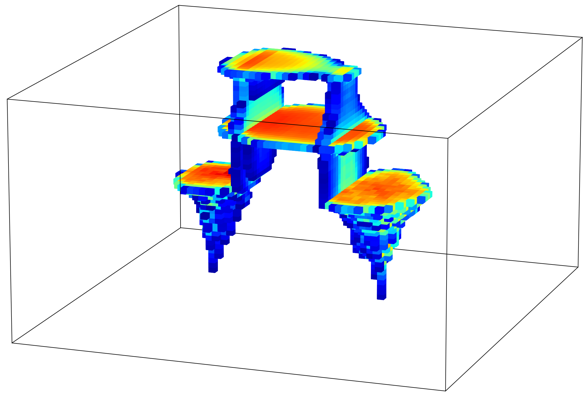}}
\hspace{0.2mm}
\subfloat[Posterior 5]{\includegraphics[width=34mm]{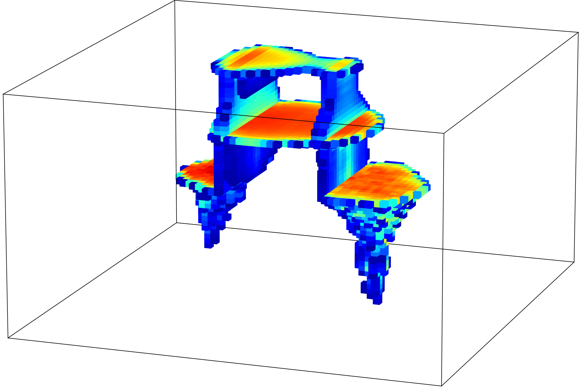}}
\includegraphics[width=5.8mm]{Sw_Scale.png}\\[1ex]
\caption{Representative CO$_2$ saturation fields at 50~years, from prior geomodels (upper row) and posterior geomodels (lower row), for true model~1. True CO$_2$ saturation field at 50~years for true model~1 is shown in Fig.~\ref{S:saturation_50_years}a.}
\label{Prior_Posterior:Saturation_Plume}
\end{figure}

\begin{figure}[!ht]
\centering   
\subfloat[Prior 1]{\includegraphics[width=35mm]{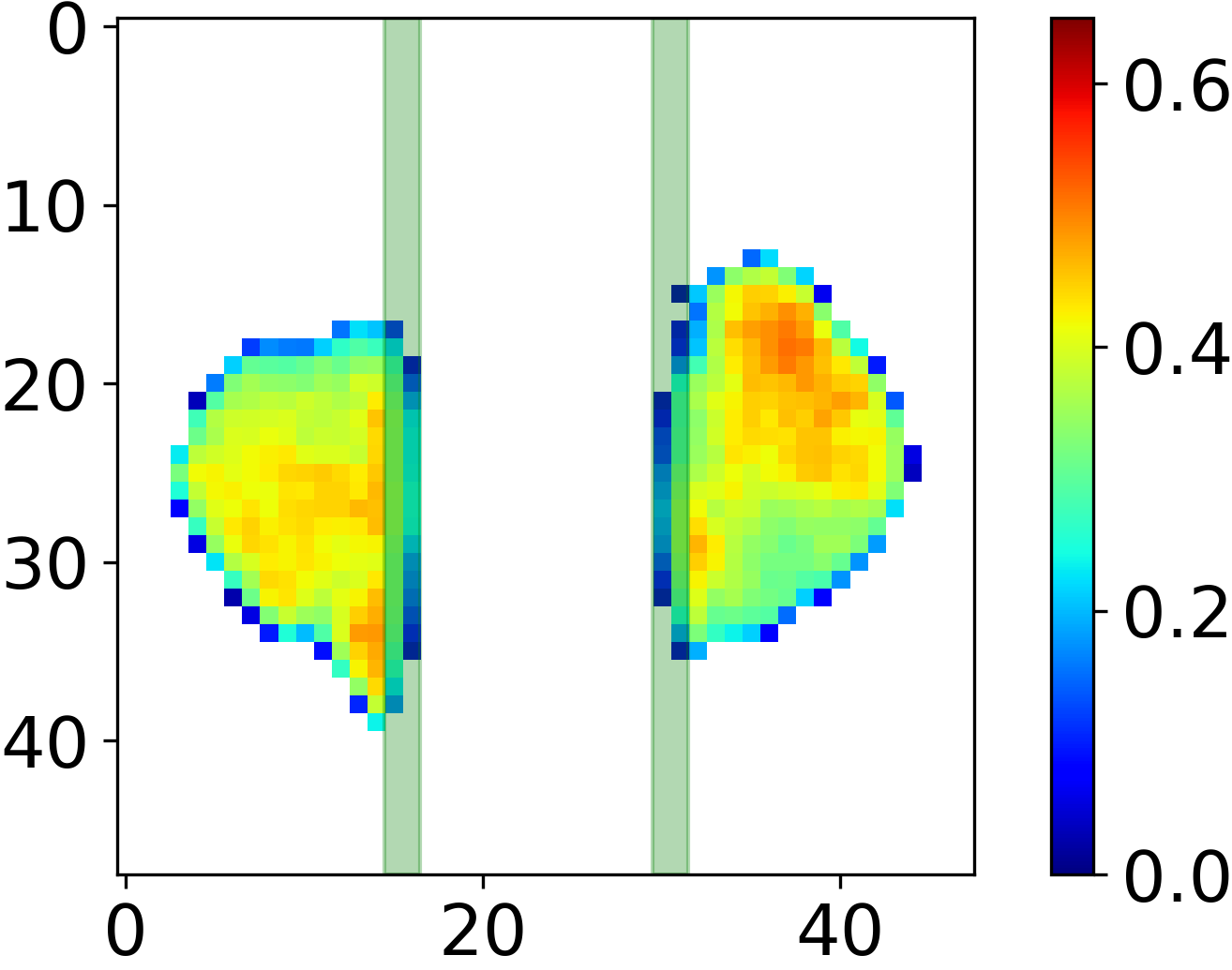}}
\hspace{0.2mm}
\subfloat[Prior 2]{\includegraphics[width=35mm]{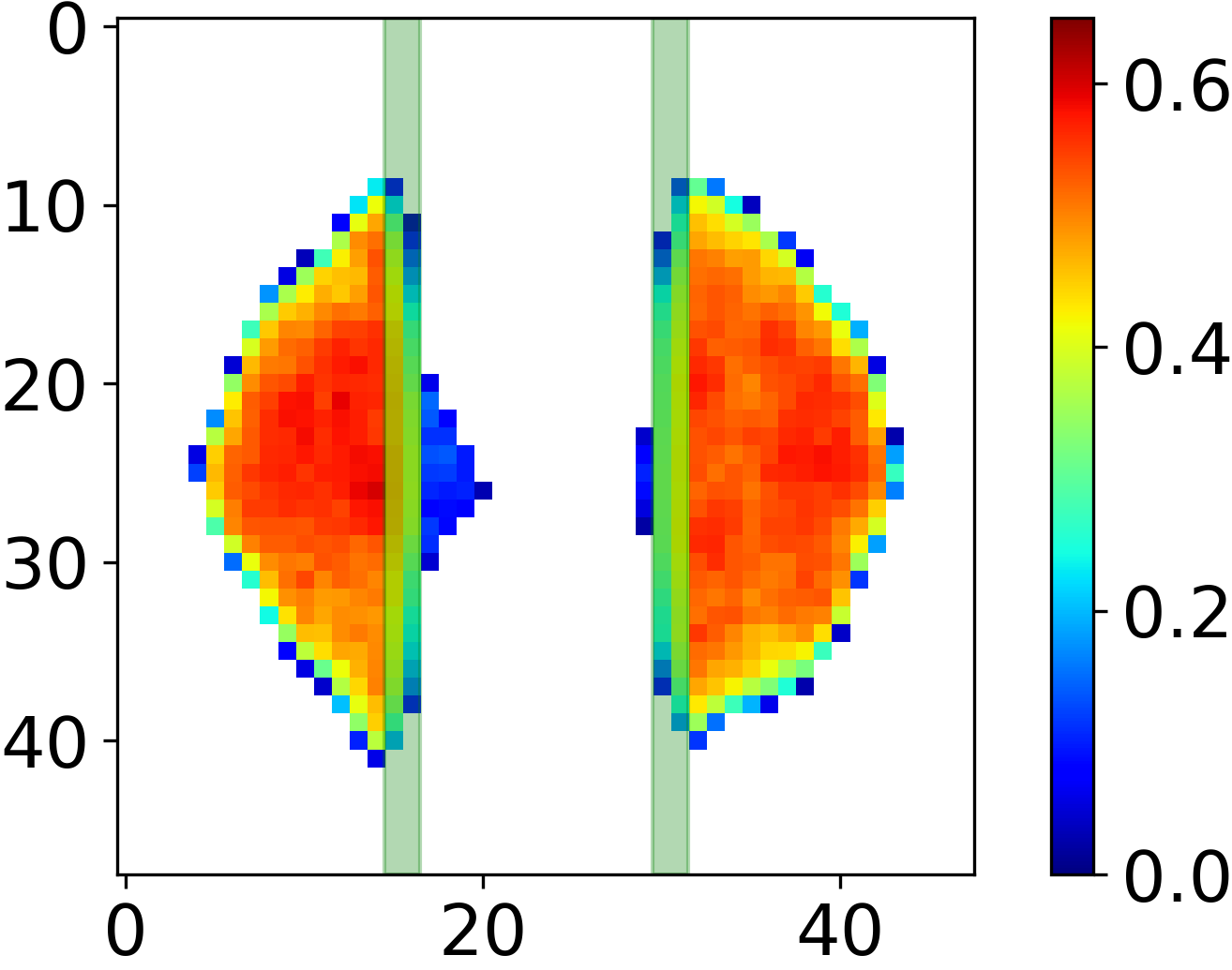}}
\hspace{0.2mm}
\subfloat[Prior 3]{\includegraphics[width=35mm]{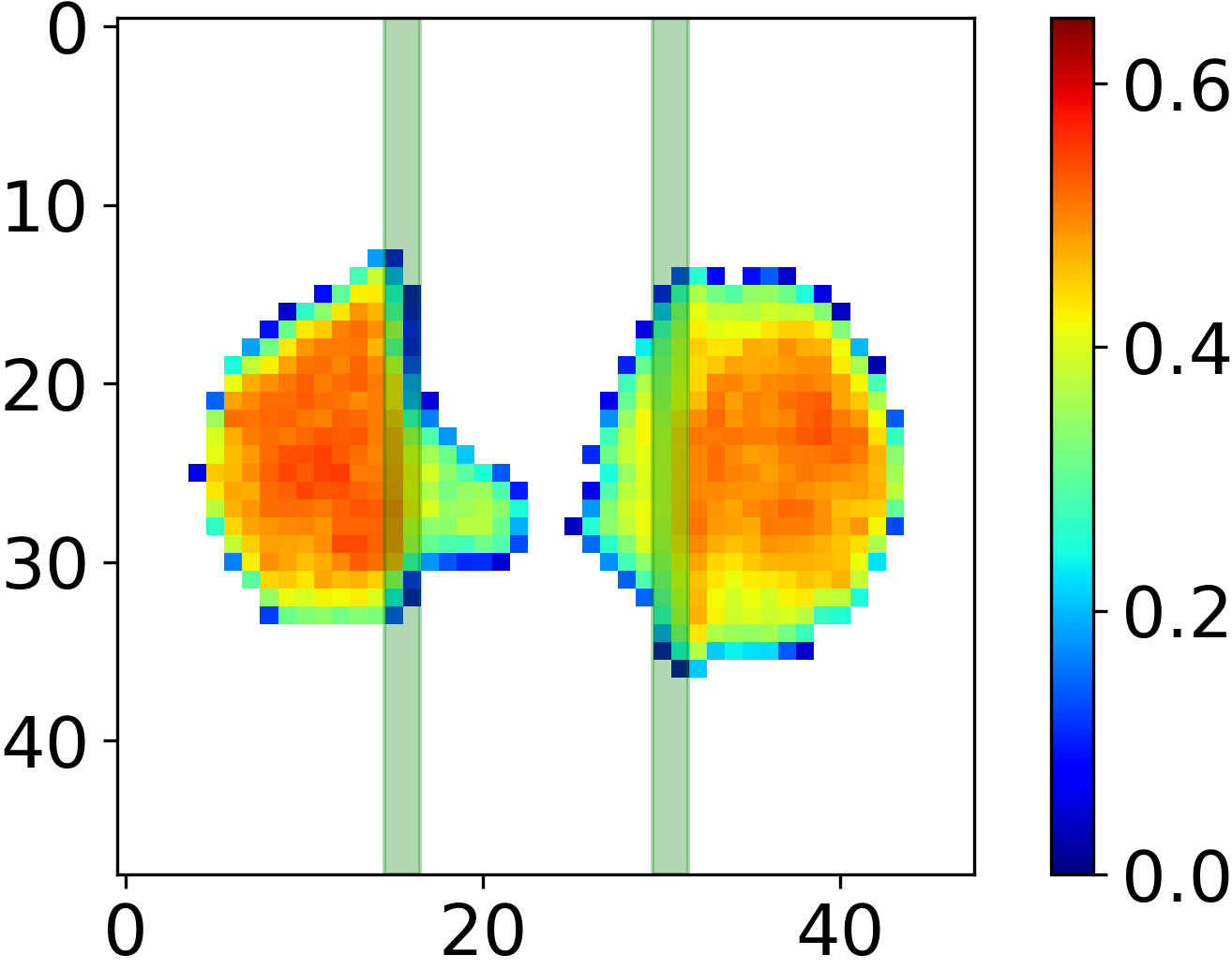}}
\hspace{0.2mm}
\subfloat[Prior 4]{\includegraphics[width=35mm]{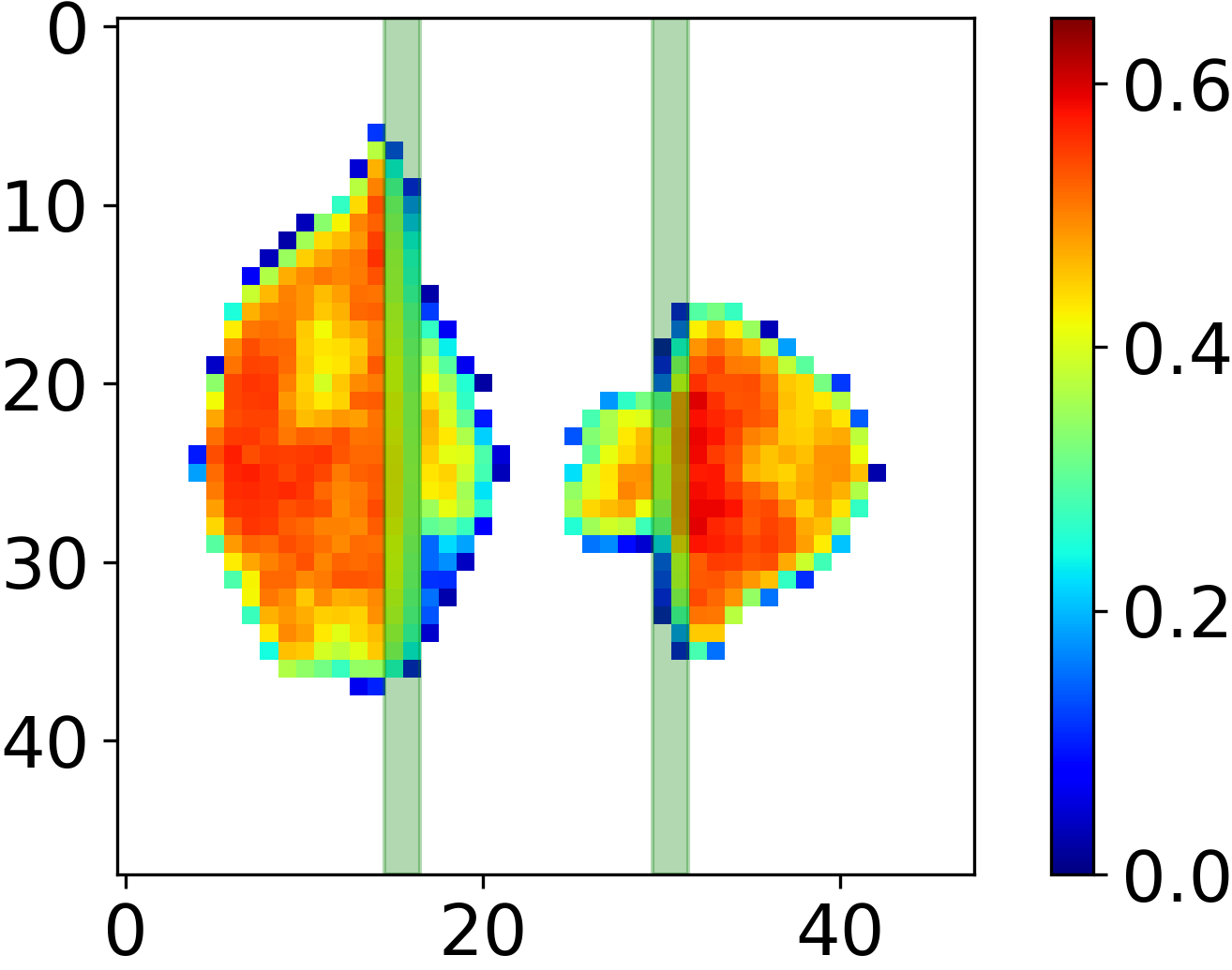}}
\hspace{0.2mm}
\subfloat[Prior 5]{\includegraphics[width=35mm]{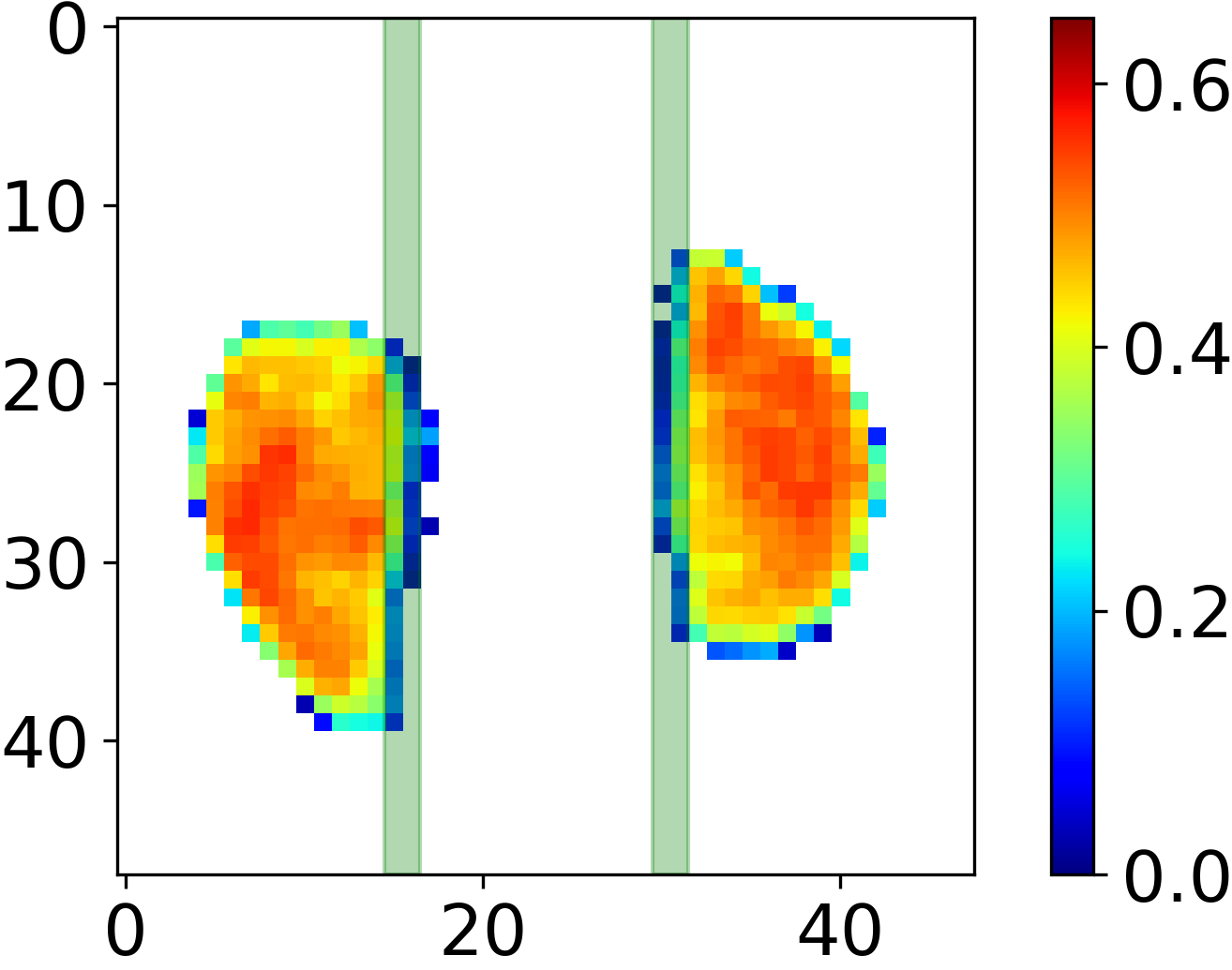}}\\[1ex]
\subfloat[True]{\includegraphics[width=35mm]{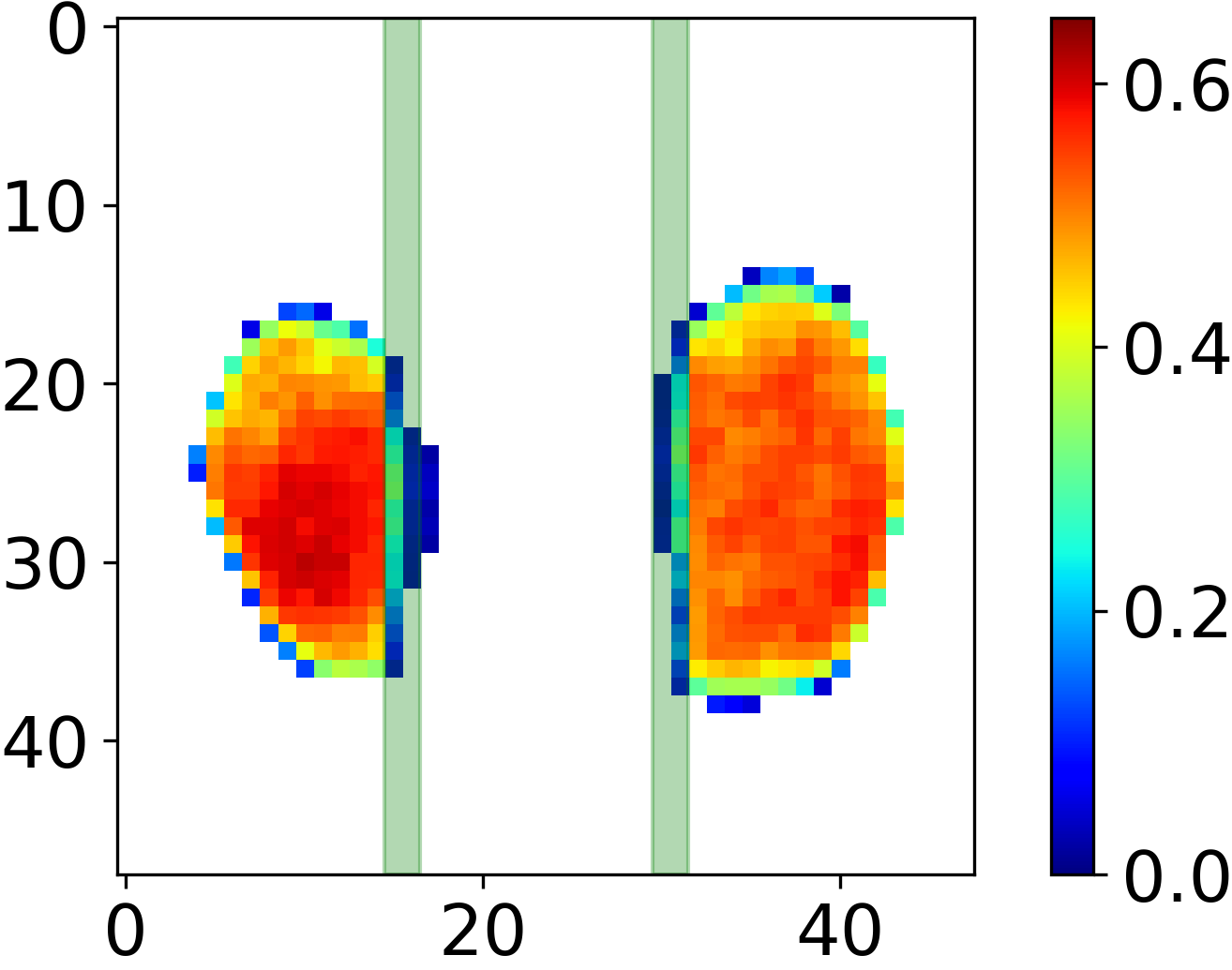}}
\hspace{0.2mm}
\subfloat[Posterior 1]{\includegraphics[width=35mm]{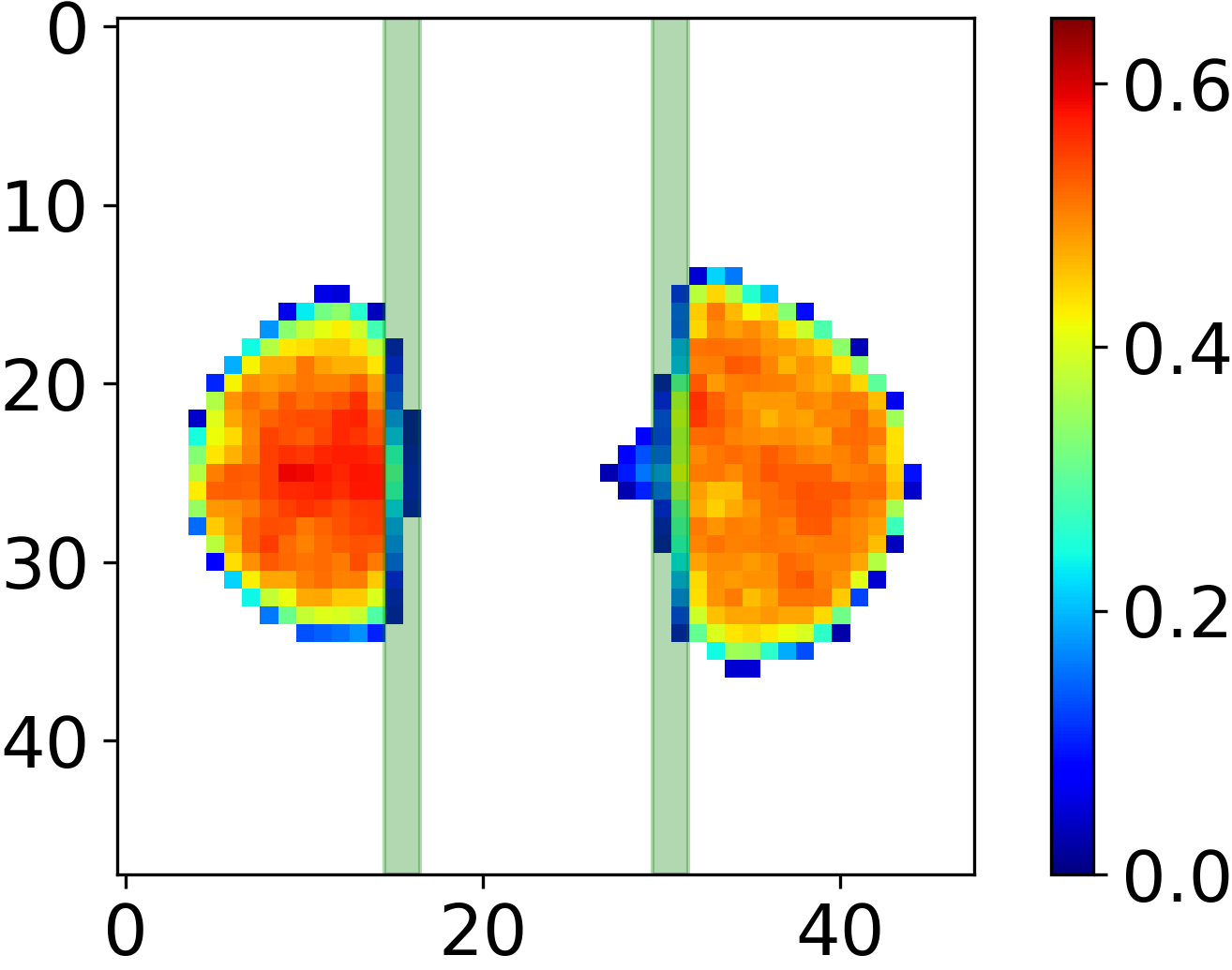}}
\hspace{0.2mm}
\subfloat[Posterior 2]{\includegraphics[width=35mm]{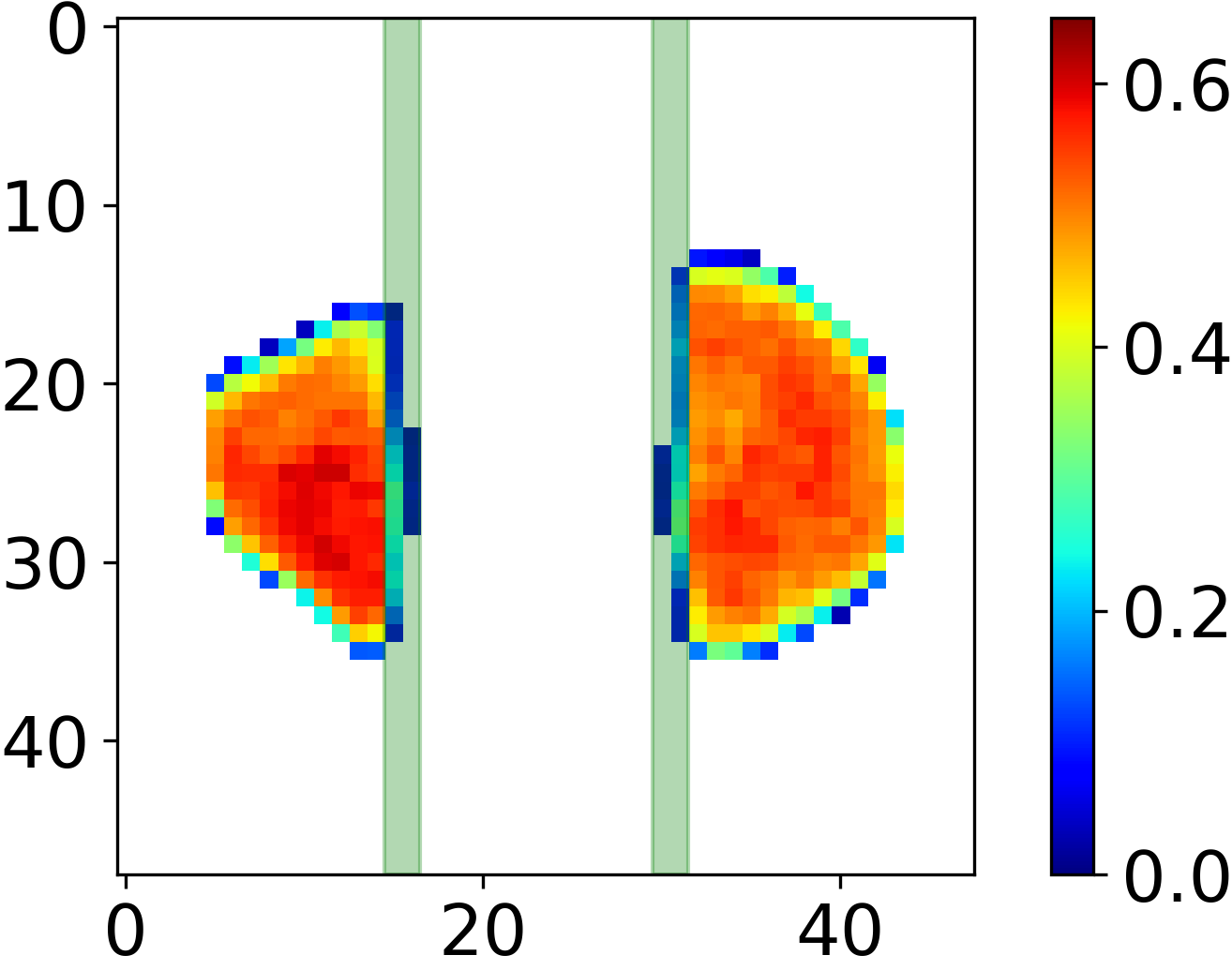}}
\hspace{0.2mm}
\subfloat[Posterior 3]{\includegraphics[width=35mm]{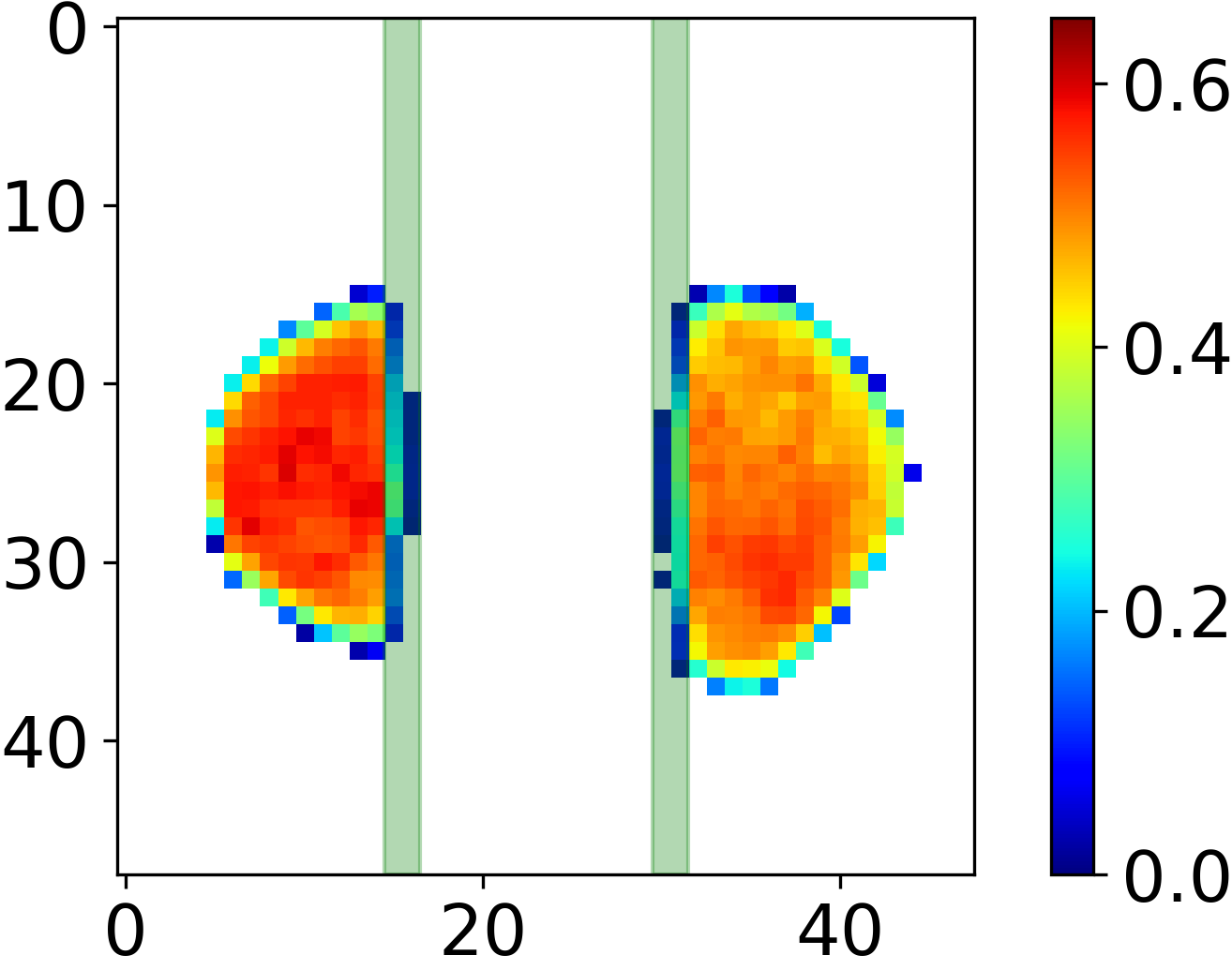}}
\hspace{0.2mm}
\subfloat[Posterior 4]{\includegraphics[width=35mm]{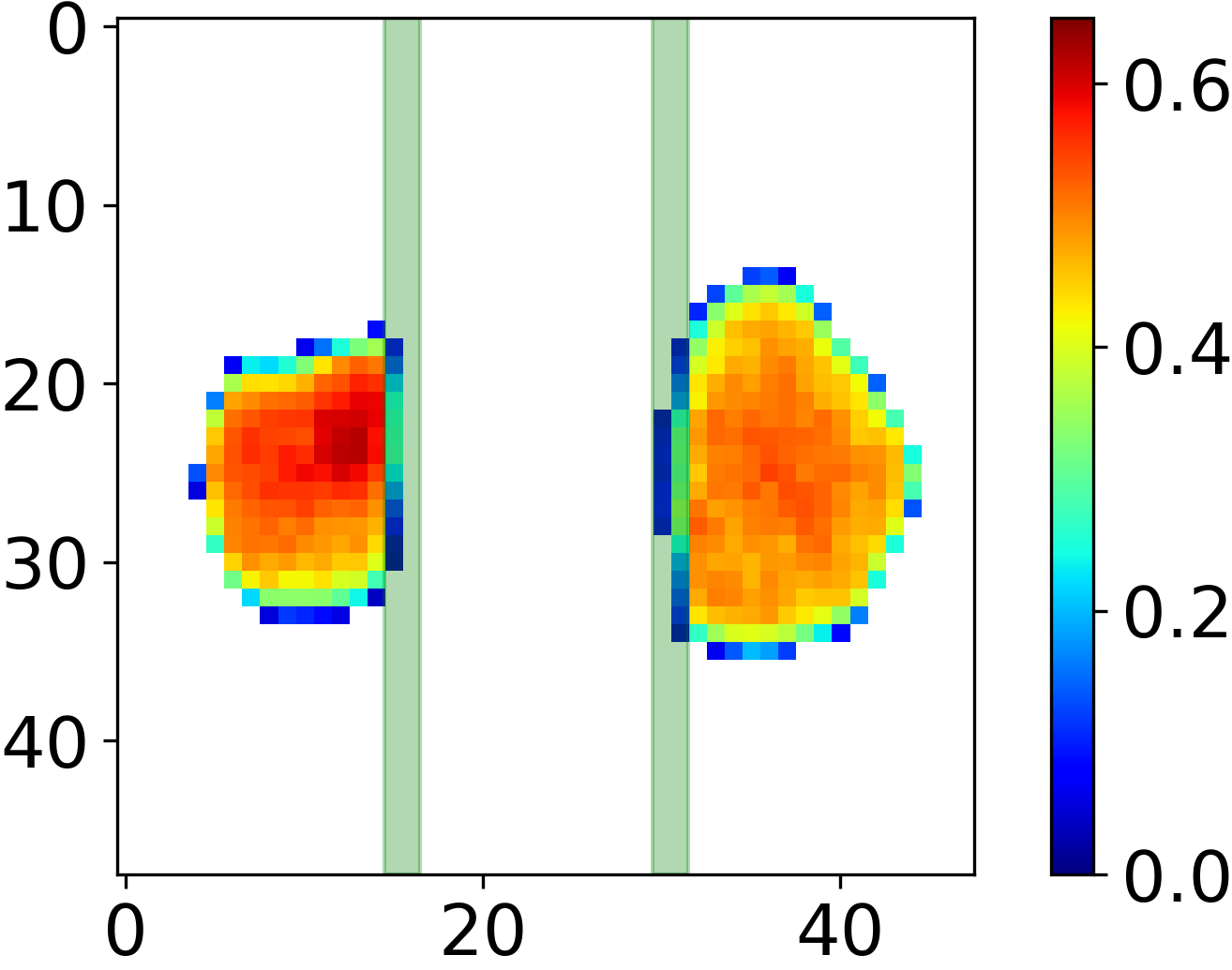}}
\caption{Saturation maps at 50~years for the top layer of the target aquifer for true model~1. These results correspond to the 3D saturation fields in Fig.~\ref{Prior_Posterior:Saturation_Plume}. True saturation map shown in Fig.~\ref{Prior_Posterior:Saturation_Top}f.}
\label{Prior_Posterior:Saturation_Top}
\end{figure}

\begin{figure}[!ht]
\centering   
\subfloat[Prior 1]{\includegraphics[width=35.3mm]{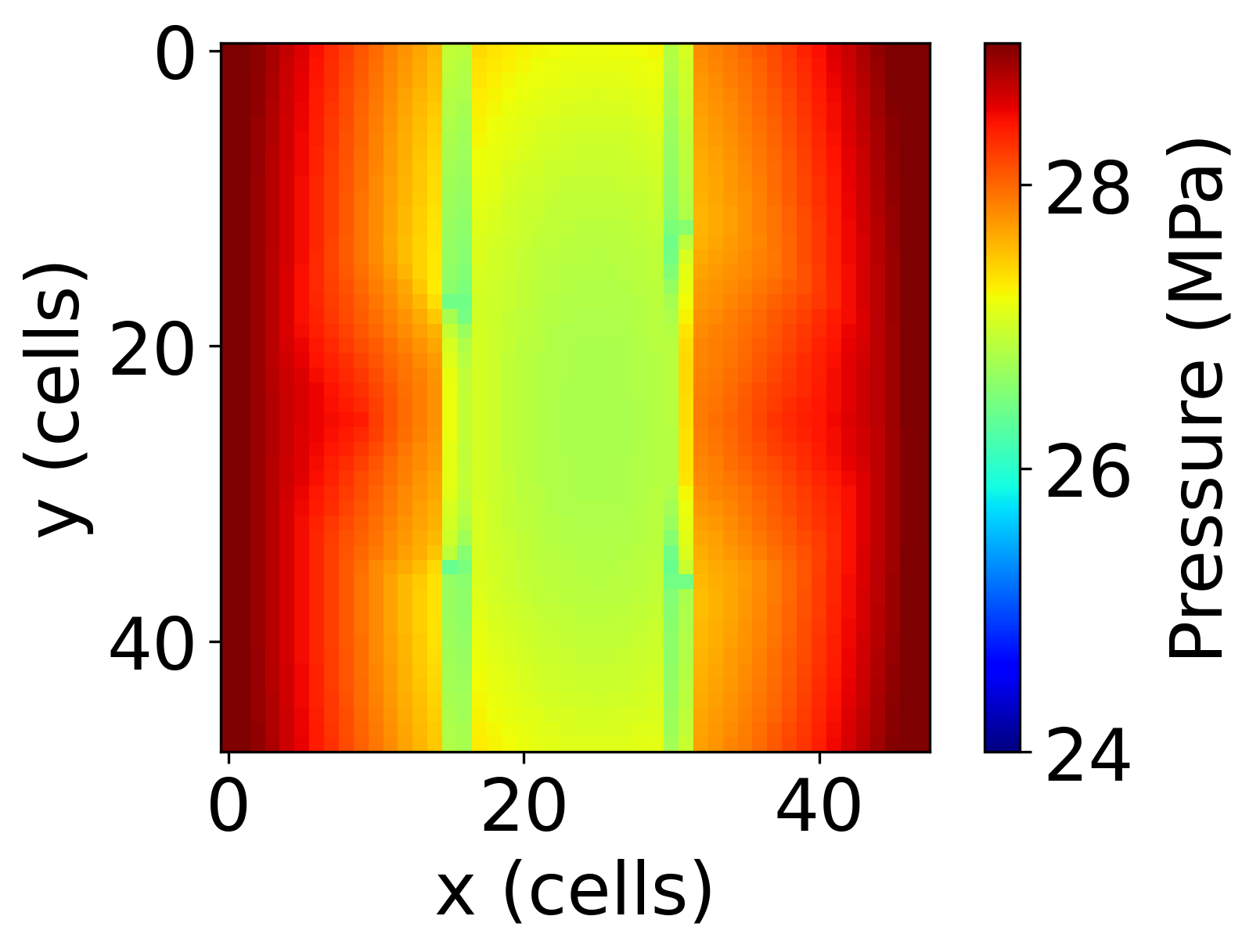}}
\hspace{0.1mm}
\subfloat[Prior 2]{\includegraphics[width=35.3mm]{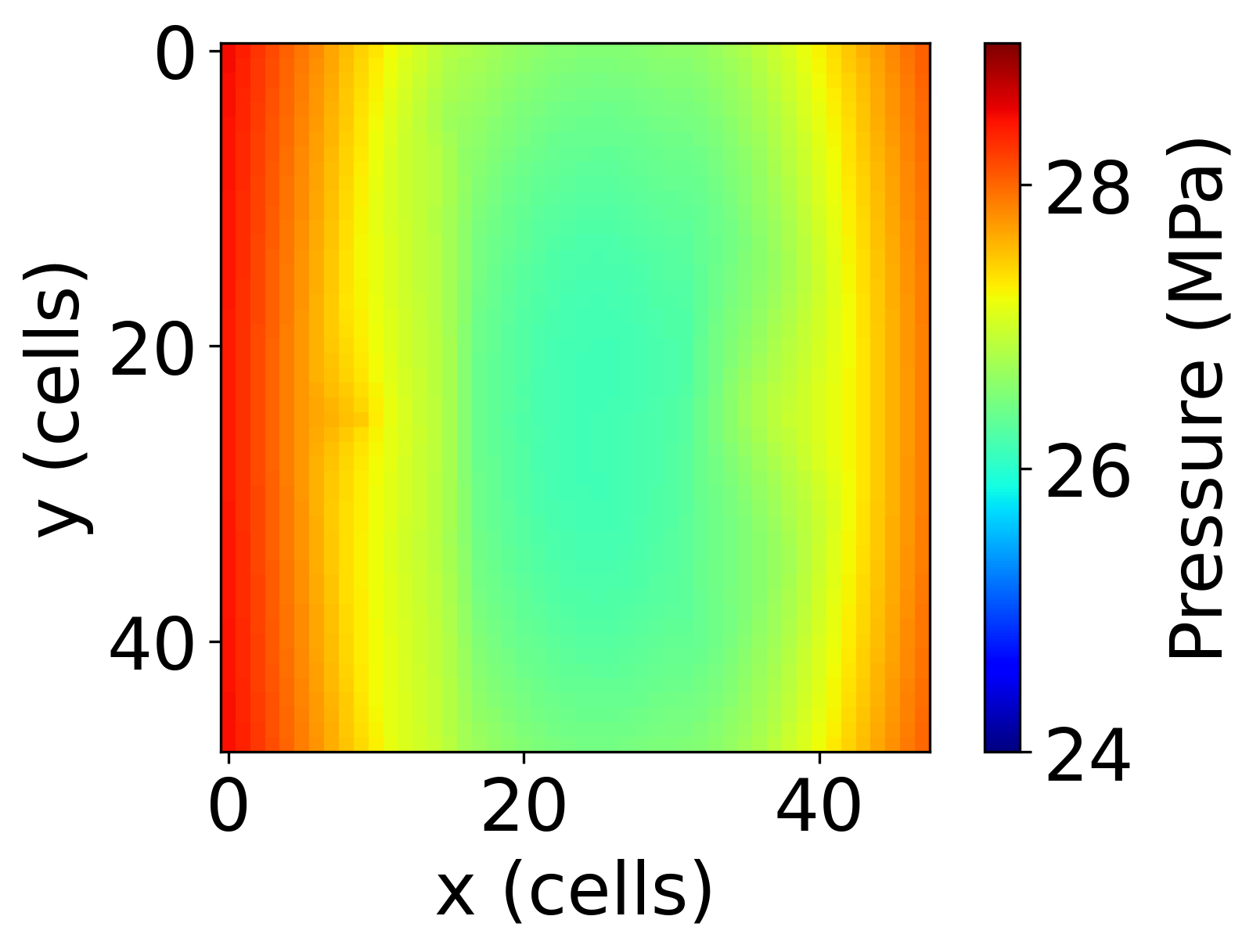}}
\hspace{0.1mm}
\subfloat[Prior 3]{\includegraphics[width=35.3mm]{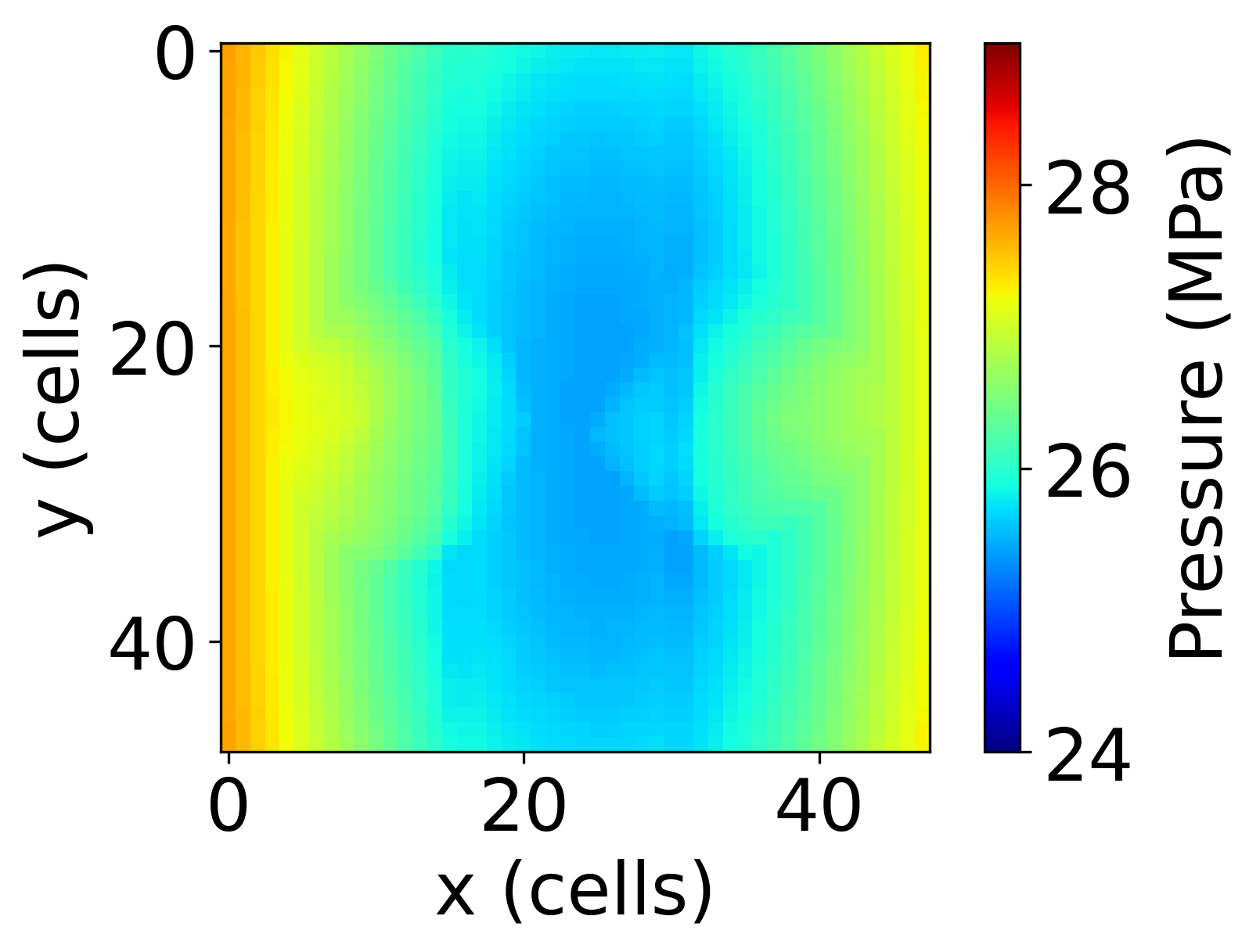}}
\hspace{0.1mm}
\subfloat[Prior 4]{\includegraphics[width=35.3mm]{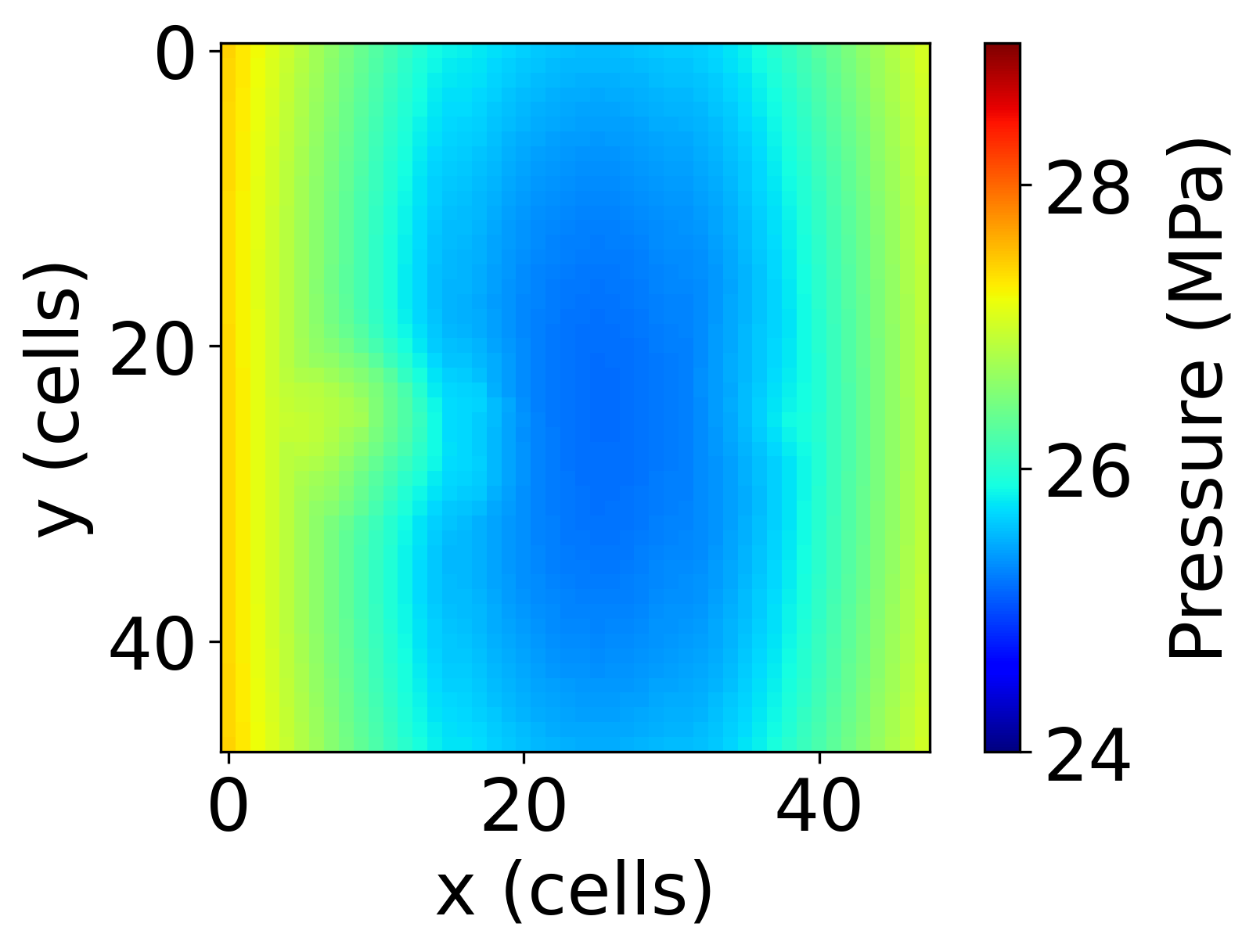}}
\hspace{0.1mm}
\subfloat[Prior 5]{\includegraphics[width=35.3mm]{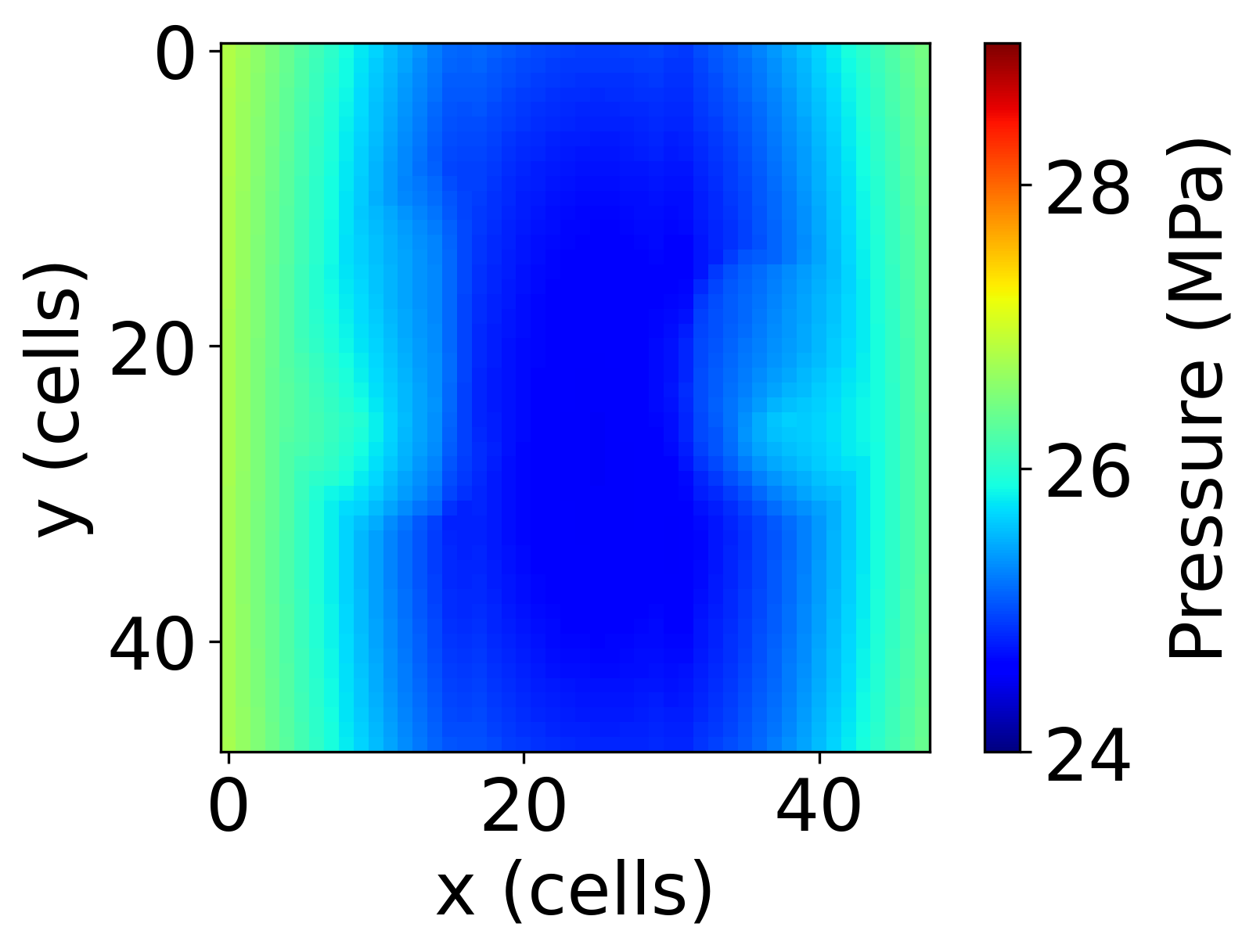}}\\[1ex]
\subfloat[True]{\includegraphics[width=35.3mm]{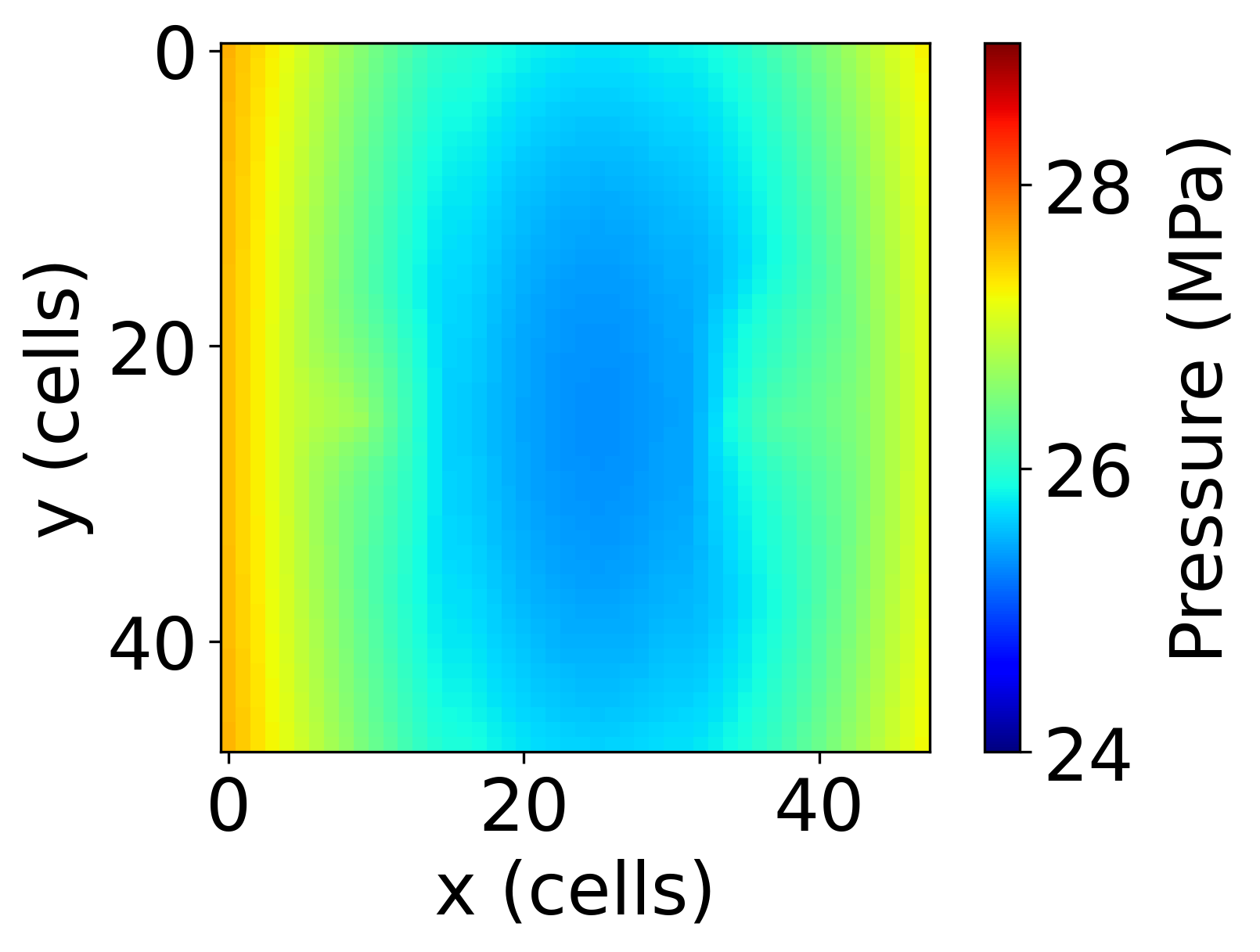}}
\hspace{0.1mm}
\subfloat[Posterior 1]{\includegraphics[width=35.3mm]{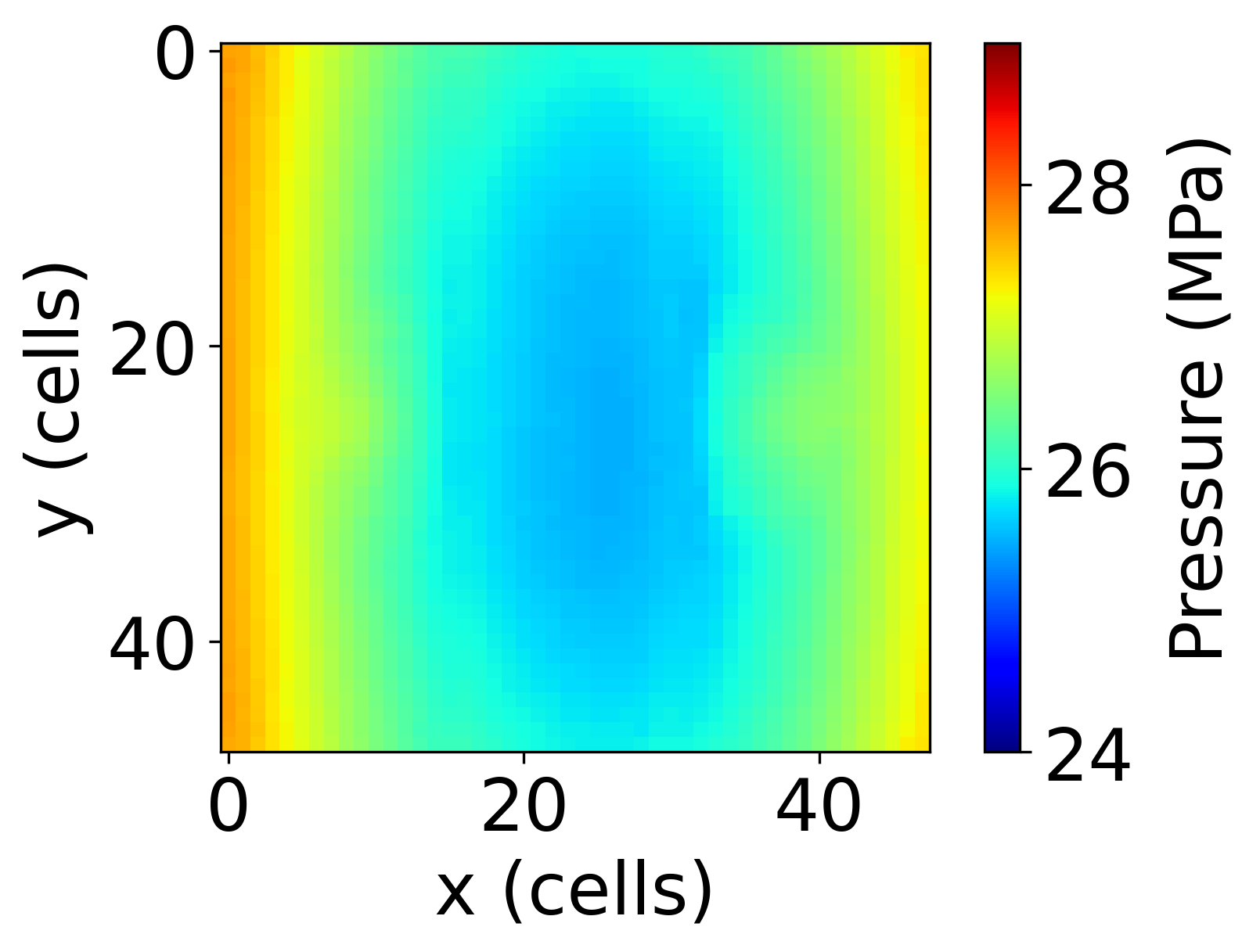}}
\hspace{0.1mm}
\subfloat[Posterior 2]{\includegraphics[width=35.3mm]{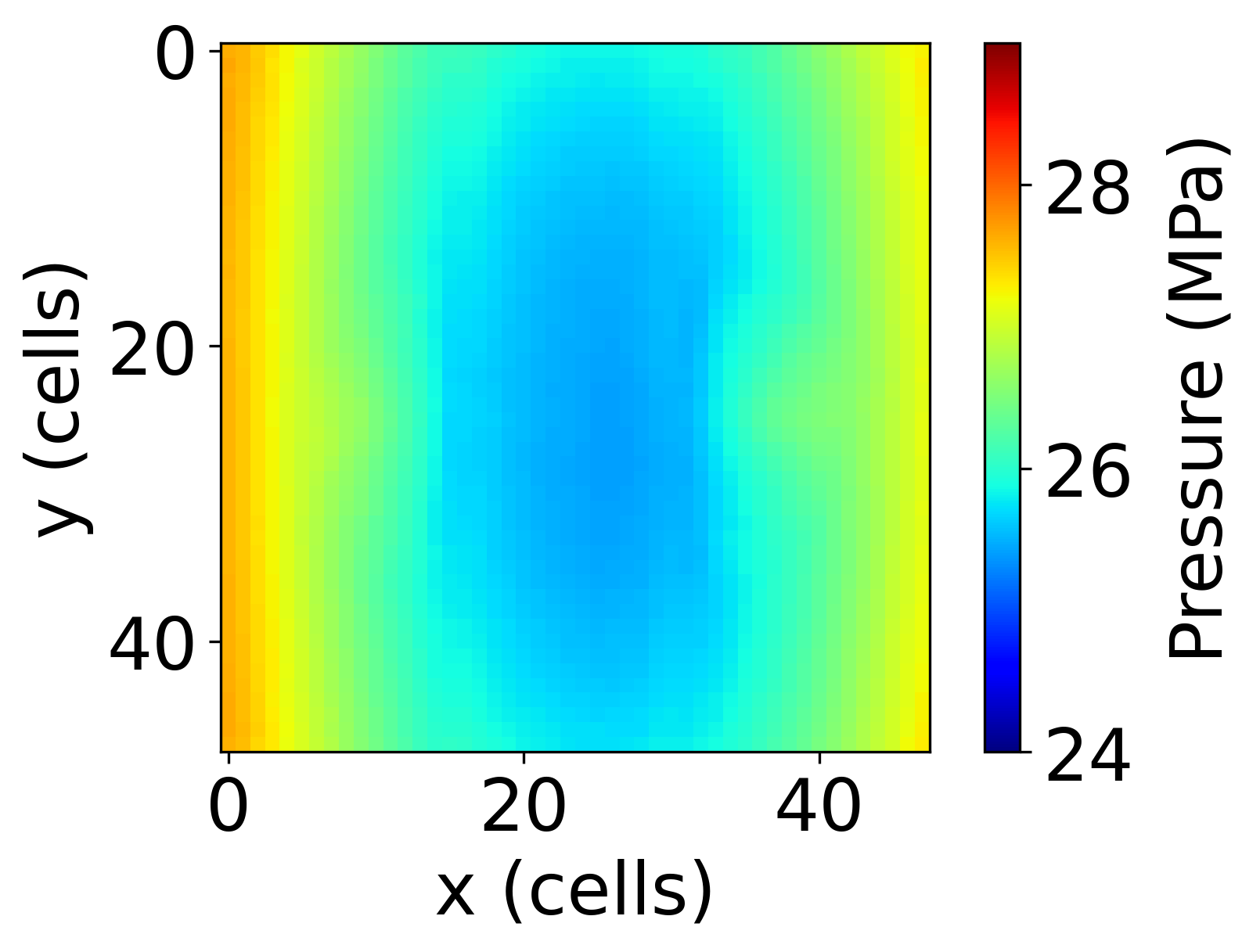}}
\hspace{0.1mm}
\subfloat[Posterior 3]{\includegraphics[width=35.3mm]{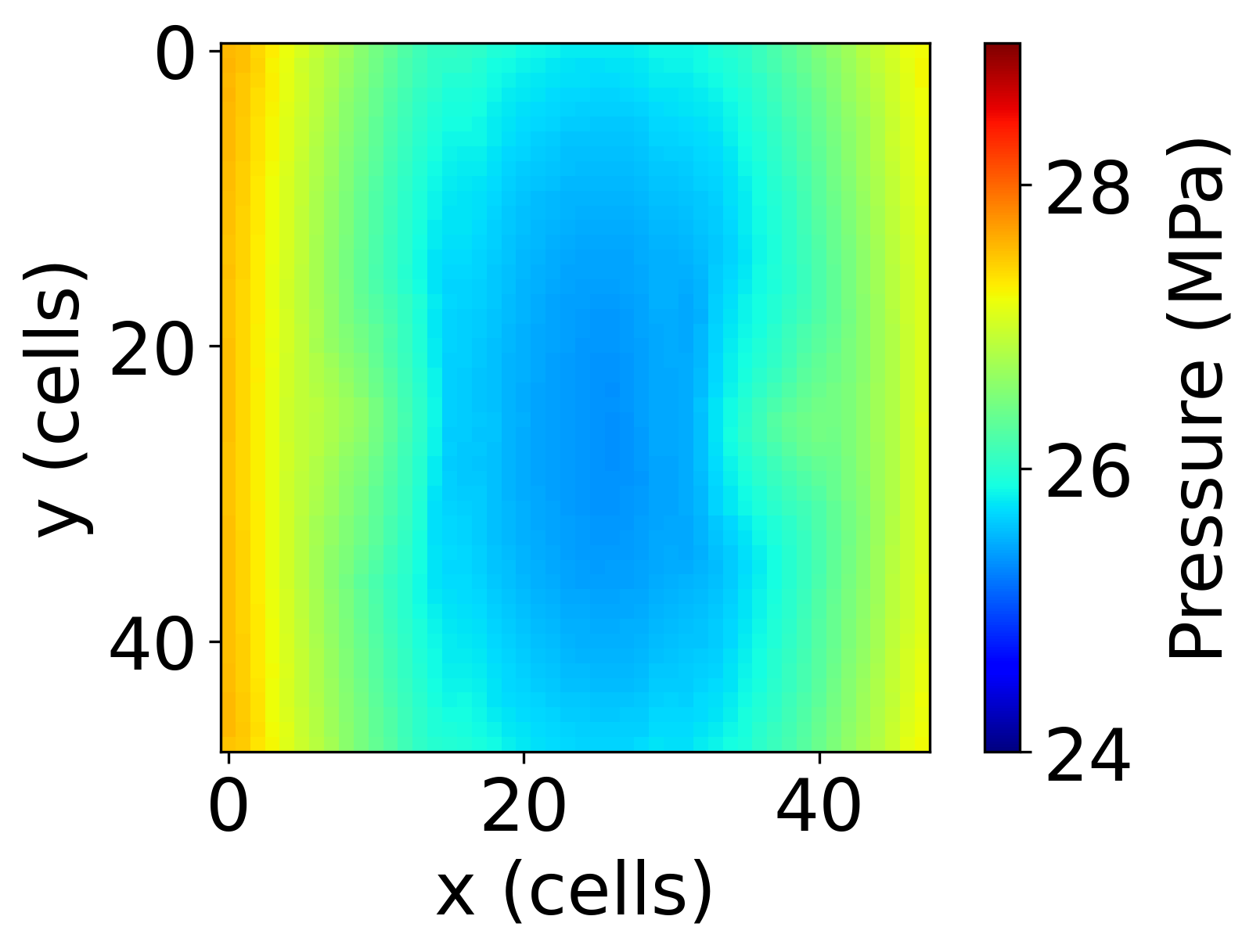}}
\hspace{0.1mm}
\subfloat[Posterior 4]{\includegraphics[width=35.3mm]{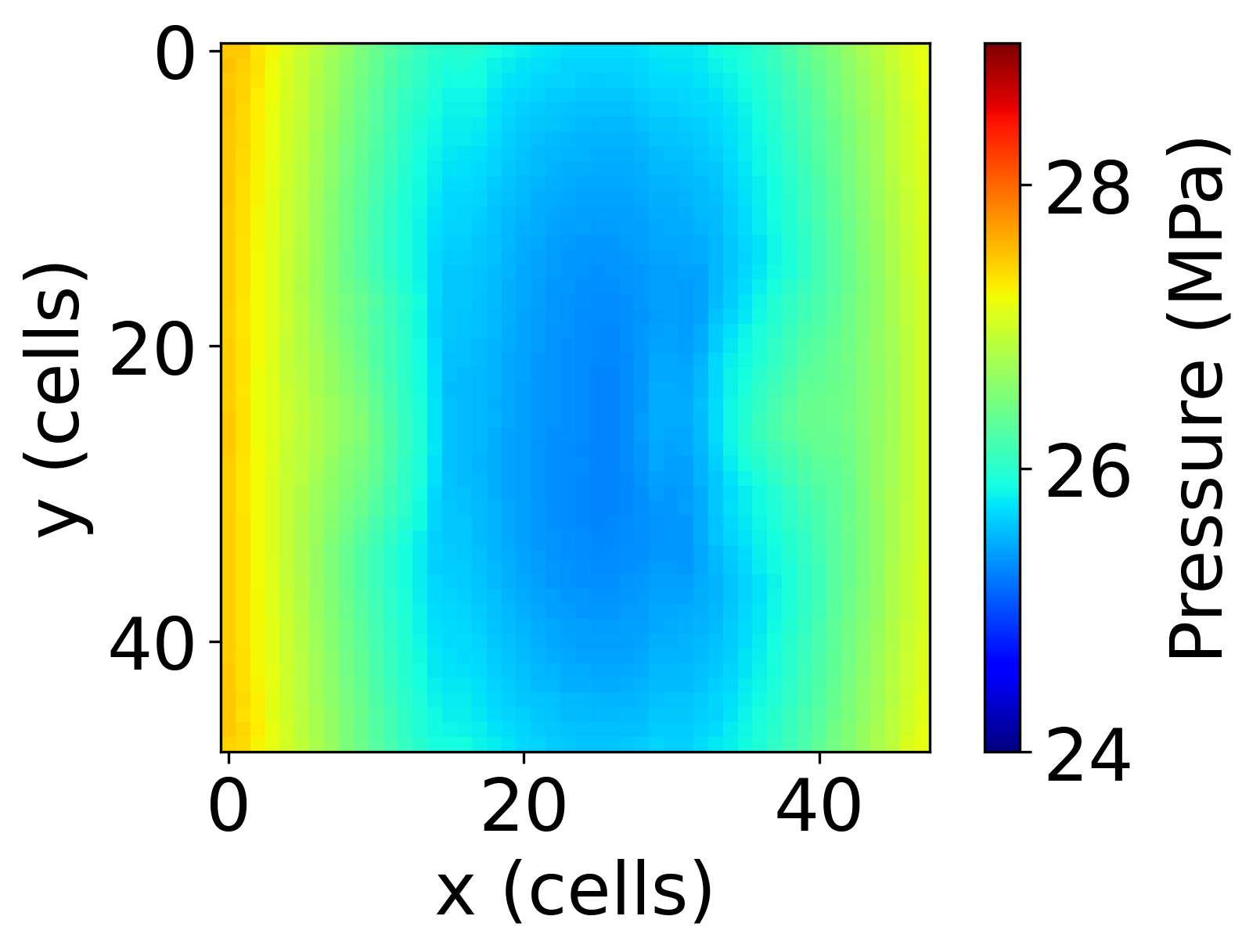}}
\caption{Representative pressure maps for the top layer of the target aquifer at 50~years, for prior geomodels (upper row) and posterior geomodels (lower row), for true model~1. True pressure map shown in Fig.~\ref{Prior_Posterior:Pressure}f.}
\label{Prior_Posterior:Pressure}
\end{figure}

Representative prior and posterior saturation fields are shown in Figs.~\ref{Prior_Posterior:Saturation_Plume} (3D views) and \ref{Prior_Posterior:Saturation_Top} (top-layer views). Analogous results for pressure appear in Fig.~\ref{Prior_Posterior:Pressure} (top-layer views). The posterior results in these figures correspond to the use of the full monitoring strategy with both data types (i.e., the left columns in Figs.~\ref{meta_true_1} and \ref{meta_1_true_1}). The surrogate model is used to generate saturation and pressure predictions in each figure. Representative prior and posterior fields are determined using the procedure described in \citet{han2023surrogate}. Specifically, we apply $k$-means clustering to a large set of models and then select the medoid of each cluster. The prior fields are the cluster medoids determined from the $n_e$ = 500 test cases, and the posterior results are cluster medoids from 500 (randomly selected) post–burn-in accepted samples. Results at the end of injection (at 50~years) are shown in all cases. The true 3D saturation field at the end of injection appears in Fig.~\ref{S:saturation_50_years}a. 

The five representative prior saturation fields exhibit qualitatively different fault-leakage behaviors. These scenarios and flow behaviors range from very little leakage (Fig.~\ref{Prior_Posterior:Saturation_Plume}a), to leakage into the middle aquifer (Fig.~\ref{Prior_Posterior:Saturation_Plume}b and c), to initial migration to the upper aquifer through Fault~2 (Fig.~\ref{Prior_Posterior:Saturation_Plume}d), and to more leakage into the upper aquifer relative to the middle aquifer (Fig.~\ref{Prior_Posterior:Saturation_Plume}e). It is important to note that the flow behavior in the true solution, where substantial leakage into both the middle and upper aquifers occurs, is not observed in the representative priors. The posterior saturation fields in Fig.~\ref{Prior_Posterior:Saturation_Plume}, by contrast, all show this behavior. The posterior results display some variability in, e.g., the shapes of the plumes in the storage aquifer, though they all clearly resemble the true solution in Fig.~\ref{S:saturation_50_years}a. 

Saturation maps for the top layer of the target aquifer, corresponding to the 3D fields in Fig.~\ref{Prior_Posterior:Saturation_Plume}, are displayed in Fig.~\ref{Prior_Posterior:Saturation_Top}. The 2D maps for the five representative prior saturation fields (at 50~years) are shown in the upper row of Fig.~\ref{Prior_Posterior:Saturation_Top}. Significant cross-fault flow is observed in Fig.~\ref{Prior_Posterior:Saturation_Top}c and d. The true 2D map, shown in Fig.~\ref{Prior_Posterior:Saturation_Top}f, shows very little cross-fault flow. This is in close correspondence with the four representative posterior saturation maps displayed in the lower row of Fig.~\ref{Prior_Posterior:Saturation_Top}.

Representative prior and posterior 3D pressure fields are selected using the same clustering procedure. The corresponding 2D pressure maps for the top layer of the target aquifer (which are more illustrative than the 3D fields) are shown in Fig.~\ref{Prior_Posterior:Pressure}. We again observe considerable variability in the prior solutions. The prior field in Fig.~\ref{Prior_Posterior:Pressure}a, for example, exhibits pressure discontinuities across the faults. The posterior pressure fields resemble the true map in Fig.~\ref{Prior_Posterior:Pressure}f, again demonstrating the accuracy and uncertainty reduction achieved through our workflow.

\begin{figure}[!ht]
\centering   
\subfloat[CO$_2$ footprint ratio in the target aquifer]{\label{CO2_footprint_true_model_1}\includegraphics[width = 85mm]{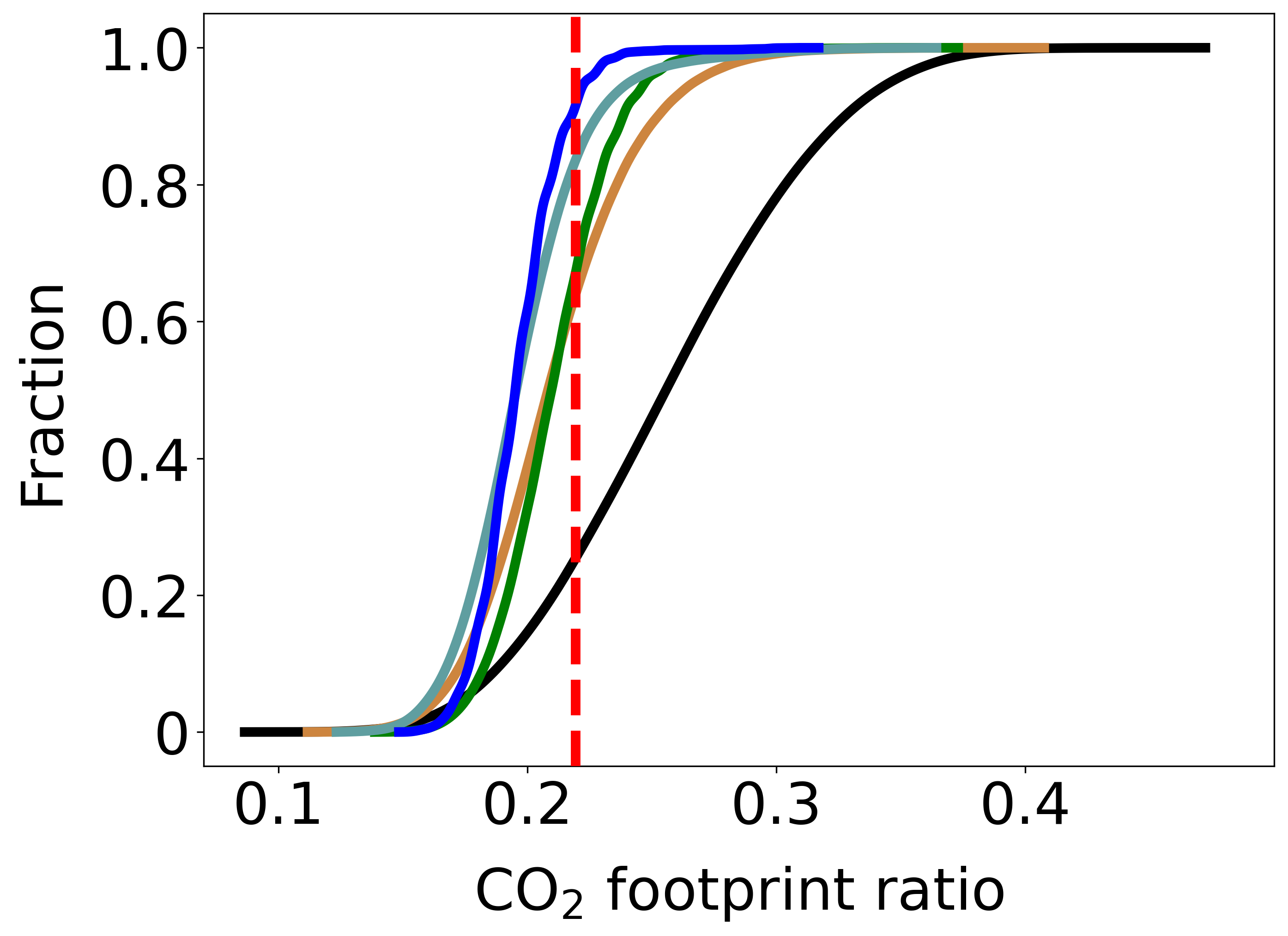}} \\
\subfloat[CO$_2$ leakage volume in the middle aquifer]{\label{CO2_volume_middle_true_model_1}\includegraphics[width = 85mm]{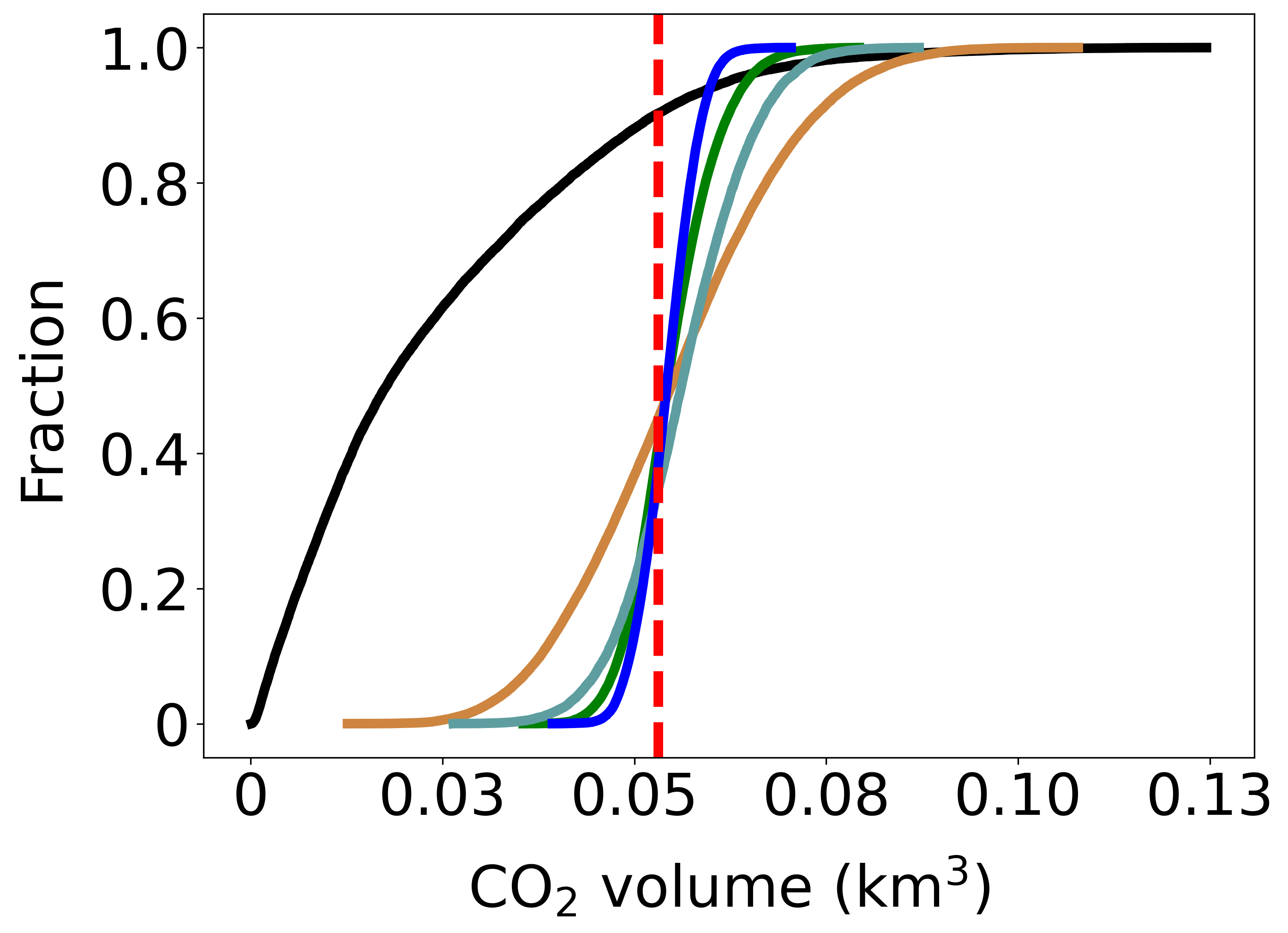}}
\hspace{4mm}
\subfloat[CO$_2$ leakage volume in the upper aquifer]
{\label{CO2_volume_upper_true_model_1}\includegraphics[width = 85mm]{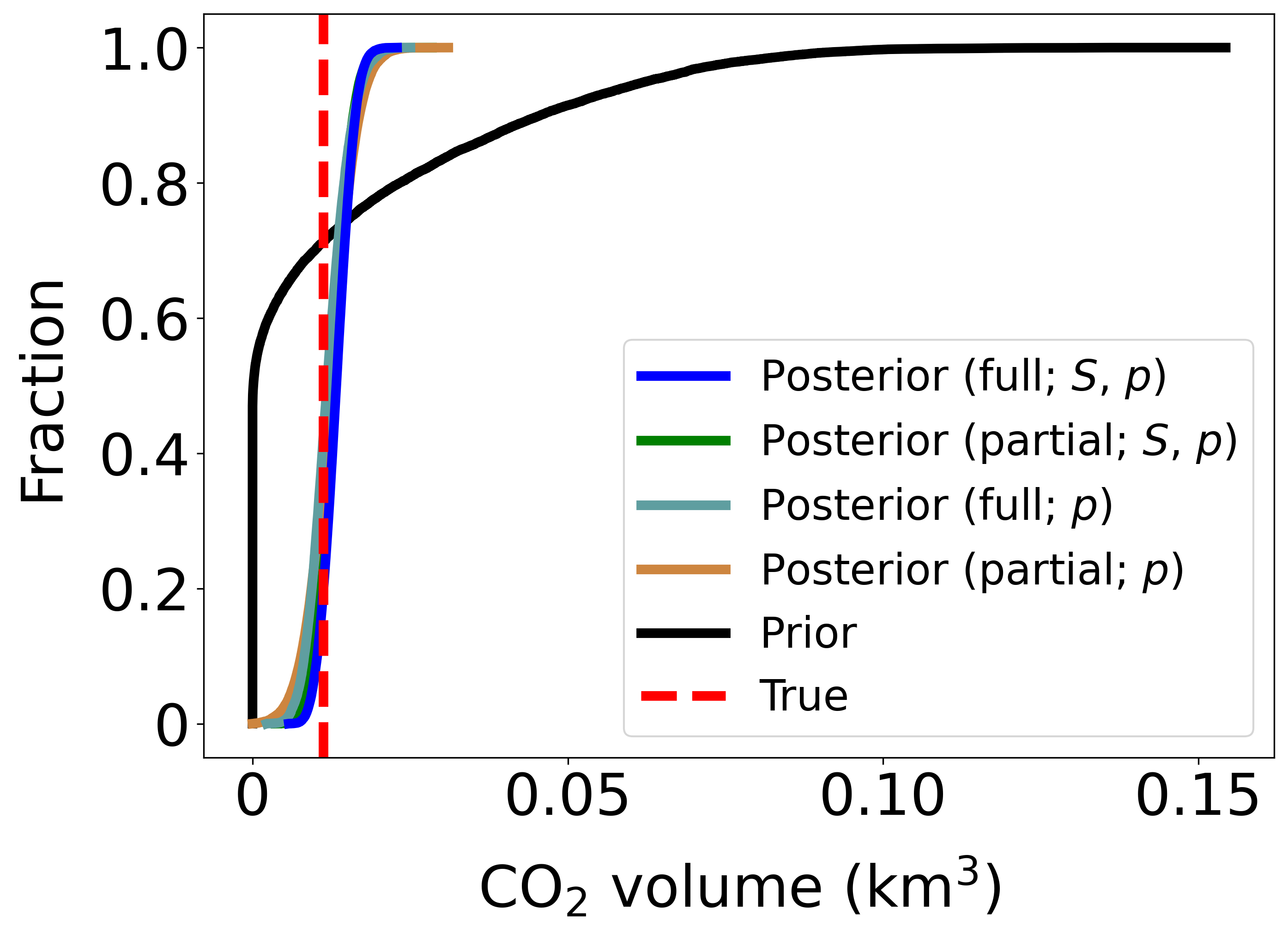}}
\caption{Cumulative density functions for CO$_2$ footprint ratio in the target aquifer and for CO$_2$ leakage volume into the middle and upper aquifers at the end of injection for true model~1. Black curves show the prior distributions and blue, green, cyan, and brown curves correspond to posterior distributions using the four monitoring strategies. Red vertical dashed lines display the true values. Legend in (c) applies to all subplots.}
\label{quantity_of_interest}
\end{figure}

In some cases, integrated or large-scale quantities such as the CO$_2$ footprint (or footprint ratio) in the target aquifer, and the volume of leakage into the middle and upper aquifers, are of interest. Data assimilation results for the cumulative density functions for these quantities, evaluated at the end of injection using different monitoring strategies, are displayed in Fig.~\ref{quantity_of_interest}. The black curves show the prior distributions and the blue, green, cyan, and brown curves display the posterior distributions using the four monitoring strategies. The red vertical dashed lines depict the true values. Interestingly, significant uncertainty reduction is observed for all monitoring strategies -- even for partial monitoring  using only pressure data. The full monitoring strategy using both pressure and saturation data provides the narrowest posterior distributions, with notable uncertainty reduction in the tails of the CDFs. Thus, this type of monitoring strategy could be used if extreme scenarios are of concern. Note finally that over half of the prior models do not show leakage into the upper aquifer (Fig.~\ref{quantity_of_interest}c), though all four monitoring strategies clearly show that this indeed occurs for true model~1.

%% file: Section_6.tex
In this study, we developed a new recurrent transformer U-Net surrogate model to predict the temporal evolution of CO$_2$ saturation and pressure fields in faulted subsurface aquifer systems characterized by uncertain flow parameters. We considered a synthetic but realistic geomodel, which was constructed based on the SEAM CO$_2$ model. The system includes a target aquifer with a surrounding region, two overlying aquifers, caprock between the aquifers, and two extensive faults that intersect the three aquifers. Depending on their properties, the faults can act as seals or as leakage pathways. In the latter case, the faults enable pressure communication and CO$_2$ migration between the target aquifer and the overlying aquifers. We introduced hierarchical uncertainty into the geomodel. This entailed uncertain (multi-Gaussian) cell-by-cell log-permeability and porosity fields that are themselves characterized by uncertain metaparameters (e.g., average and standard deviation of log-permeability). Fault permeabilities and permeabilities in the two overlying aquifers were also treated as uncertain. Supercritical CO$_2$ was injected into the target (lowest) aquifer through two vertical wells. Each well injected at a rate of 1~Mt CO$_2$ per year for 50~years.

We computed error statistics with varying numbers of training models (corresponding to heterogeneous realizations characterized by metaparameters sampled from broad property distributions) for both the recurrent transformer U-Net surrogate model and an existing recurrent residual U-Net model. The recurrent transformer U-Net model consistently provided predictions with lower overall saturation MAE and pressure relative error. The new model also demonstrated advantages in challenging cases involving flow across faults. With 4000 training samples (which was the maximum number considered), the median saturation MAE and pressure relative error for the recurrent transformer U-Net model were 0.025 and 0.12\%, respectively. The model was shown to capture a range of qualitatively different leakage scenarios and flow behaviors associated with different fault and aquifer properties. The surrogate model was then used within a global sensitivity analysis to quantify the importance of the metaparameters and geomodel realizations on the CO$_2$ footprint in the target aquifer. This analysis showed that the importance of a given parameter can vary substantially in time. 

We then performed data assimilation using a hierarchical MCMC procedure, with the recurrent transformer U-Net surrogate model applied for all function evaluations. Data provided by four different monitoring strategies were considered. These strategies involved collecting both saturation and pressure data versus collecting only pressure data, and placing monitoring wells in all three aquifers versus having them in only the middle and upper aquifers. Data assimilation results for the different monitoring strategies showed that the most uncertainty reduction in the metaparameters was achieved by collecting both saturation and pressure in all three aquifers, as would be expected. With this monitoring strategy, posterior predictions for the 3D saturation fields were shown to be in close correspondence with the true model. If a subset of properties such as fault permeabilities are of primary interest, our results showed that a more limited (and less expensive) monitoring strategy may suffice. In any event, results of the type provided in this study may be very useful in practice, as they quantify the degree of uncertainty reduction for various parameters or flow responses that can be achieved with different monitoring strategies. Please note that additional results for two other true models are provided in SI.

There are a number of useful directions for future research in this area. The SEAM-based geomodel and the recurrent transformer U-Net surrogate can be used (with modification as required) for other investigations involving data assimilation and optimization. The consideration of different types of geological systems, corresponding to different hierarchical uncertainties, would provide the framework with wider applicability, as would the inclusion of additional uncertain variables such as relative permeability and capillary pressure functions. Geomodels gridded with non-neighbor connections, as discussed in Section~\ref{sec:limitations}, should also be treated. Our focus in this study was on the flow aspects of faults. It will be of interest to extend our modeling to coupled flow-geomechanics systems, which will allow us to investigate both CO$_2$ leakage and fault reactivation risks. This will entail more complicated and time-consuming simulations and will require the development of a more comprehensive surrogate model. Finally, testing on actual faulted systems should also be conducted.

%% file: SI_Section_2.tex
\begin{figure}[!ht]
\centering   
\subfloat[Posterior 1]{\label{posterior_sw_1_true_2}\includegraphics[width=34mm]{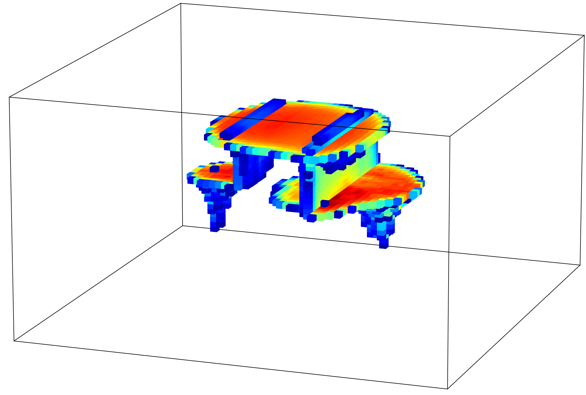}}
\hspace{0.2mm}
\subfloat[Posterior 2]{\label{posterior_sw_2_true_2}\includegraphics[width=34mm]{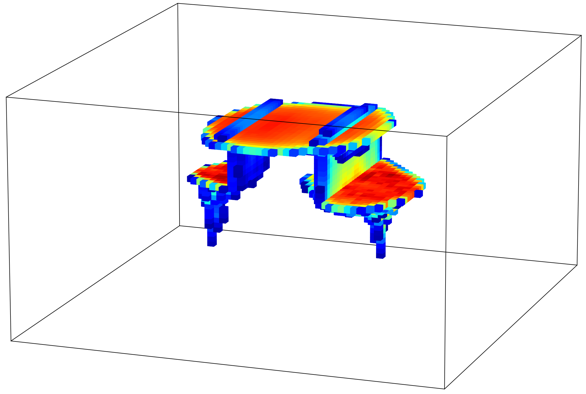}}
\hspace{0.2mm}
\subfloat[Posterior 3]{\label{posterior_sw_3_true_2}\includegraphics[width=34mm]{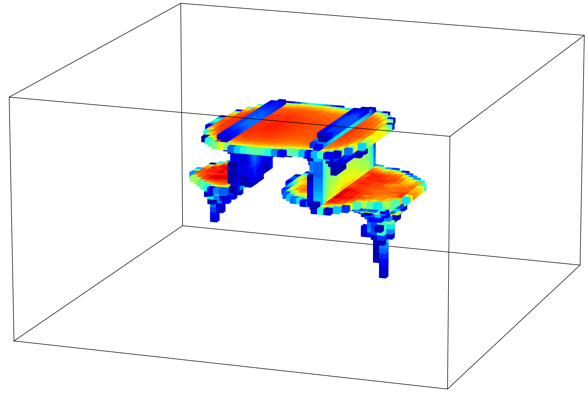}}
\hspace{0.2mm}
\subfloat[Posterior 4]{\label{posterior_sw_4_true_2}\includegraphics[width=34mm]{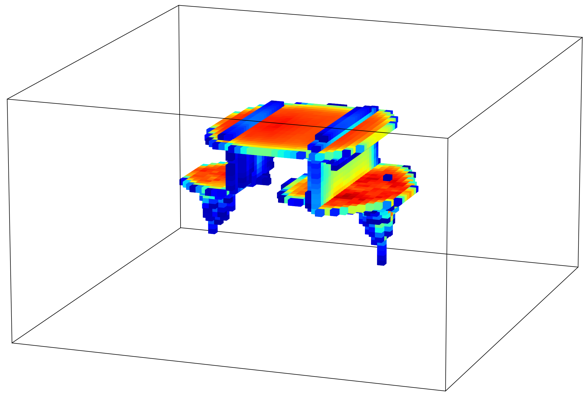}}
\hspace{0.2mm}
\subfloat[Posterior 5]{\label{posterior_sw_5_true_2}\includegraphics[width=34mm]{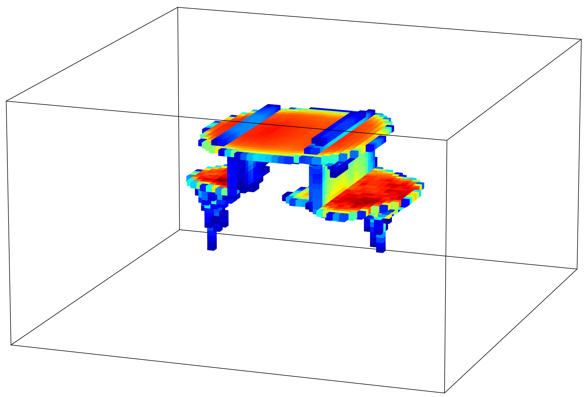}}
\includegraphics[width=5.8mm]{Sw_Scale.png}\\[1ex]
\caption{Representative CO$_2$ saturation fields at 50~years from posterior geomodels for true model~2. True saturation field at 50~years for true model~2 is shown in Fig.~10b in the main paper.}
\label{Prior_Posterior:Saturation_Plume_True_2}
\end{figure}

The true saturation fields at 20~years and 50~years for true model~2 are shown in Figs.~9b and~10b in the main paper (true model~2 corresponds to realization~5 in the main paper). This case exhibits CO$_2$ leakage into the middle aquifer but no leakage from the middle aquifer to the upper aquifer (the true values of the fault permeabilities are $k_{f1}^{tm} = 266.3$~md, $k_{f1}^{mu} = 0.15$~md, $k_{f2}^{tm} = 18.8$~md, $k_{f2}^{mu} = 0.57$~md). For the full monitoring strategy using both pressure and saturation data, the hierarchical MCMC procedure requires a total of 112,206 function evaluations to achieve convergence. A total of 10,078 sets of metaparameters are collected from the three chains as posterior samples for this scenario. The representative prior saturation fields are the same as for true model~1 (displayed in the upper row in Fig.~15 in the main paper). The representative posterior saturation fields at 50~years for the full monitoring strategy, using both pressure and saturation data, are shown in Fig.~\ref{Prior_Posterior:Saturation_Plume_True_2}. We again observe reduced variability in the posterior saturation plumes and close agreement with the true solution, i.e., significant CO$_2$ migration into the middle aquifer but none into the upper aquifer.

\begin{figure}[!ht]
\centering   
\subfloat[True]{\includegraphics[width=35mm]{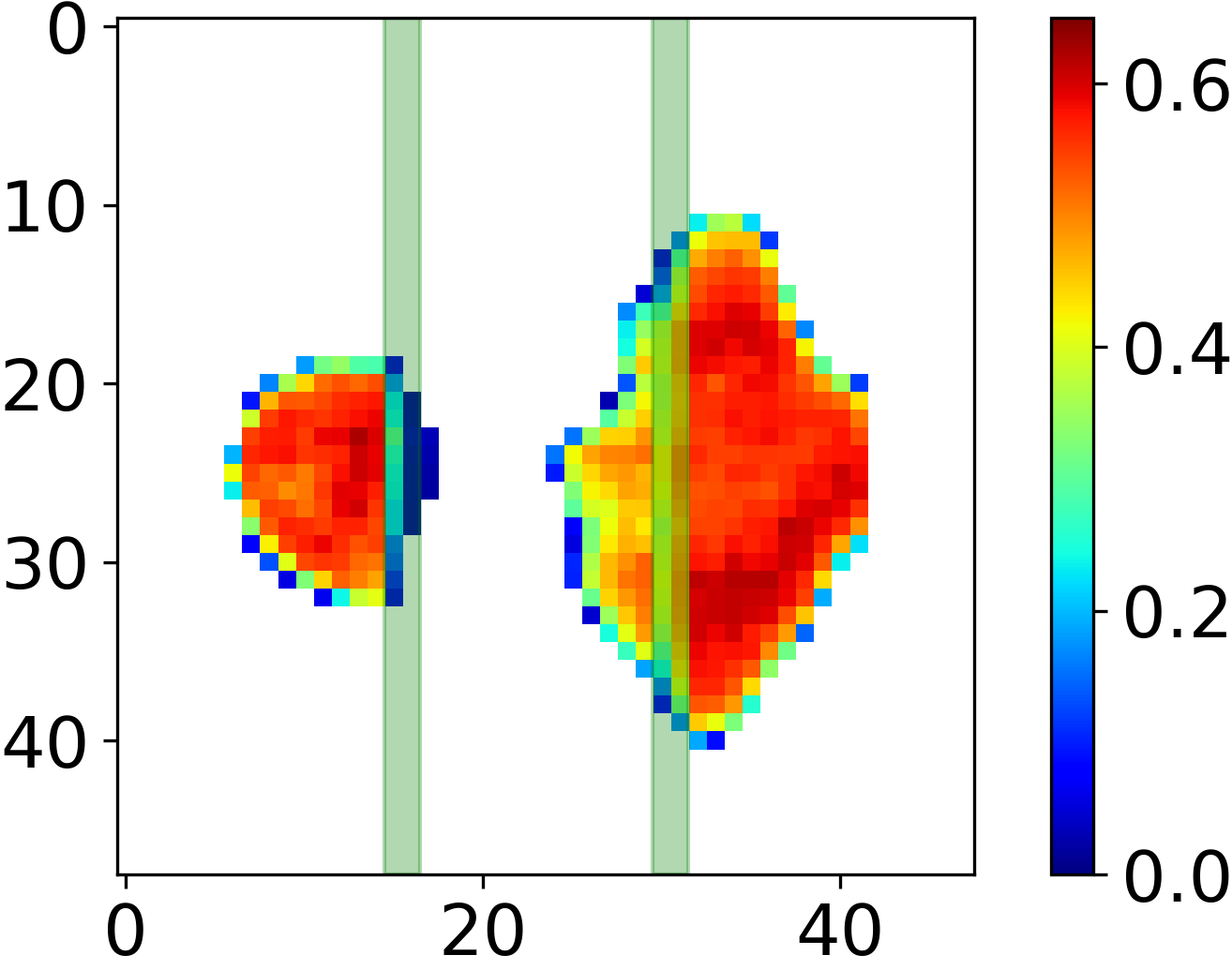}}
\hspace{0.2mm}
\subfloat[Posterior 1]{\includegraphics[width=35mm]{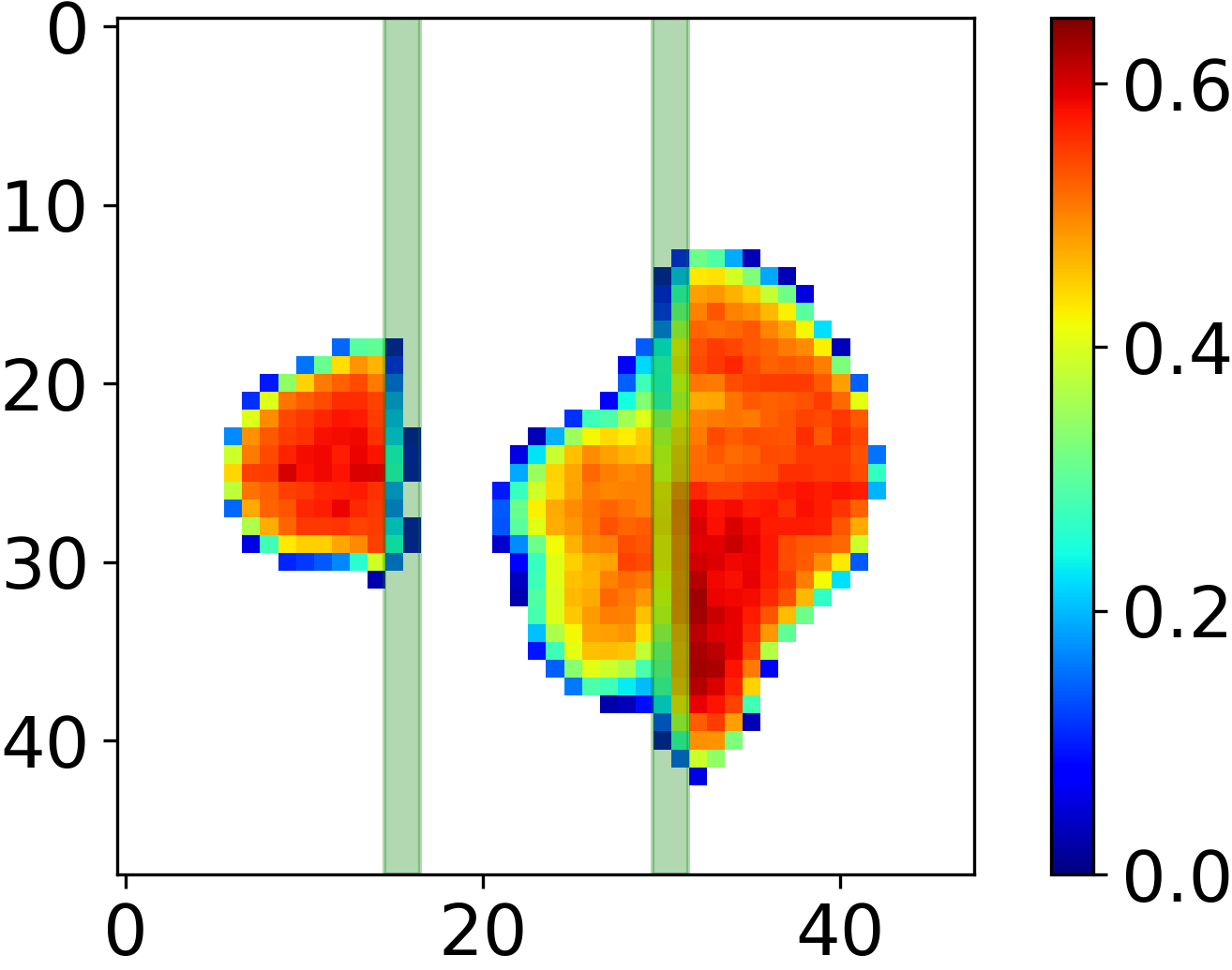}}
\hspace{0.2mm}
\subfloat[Posterior 2]{\includegraphics[width=35mm]{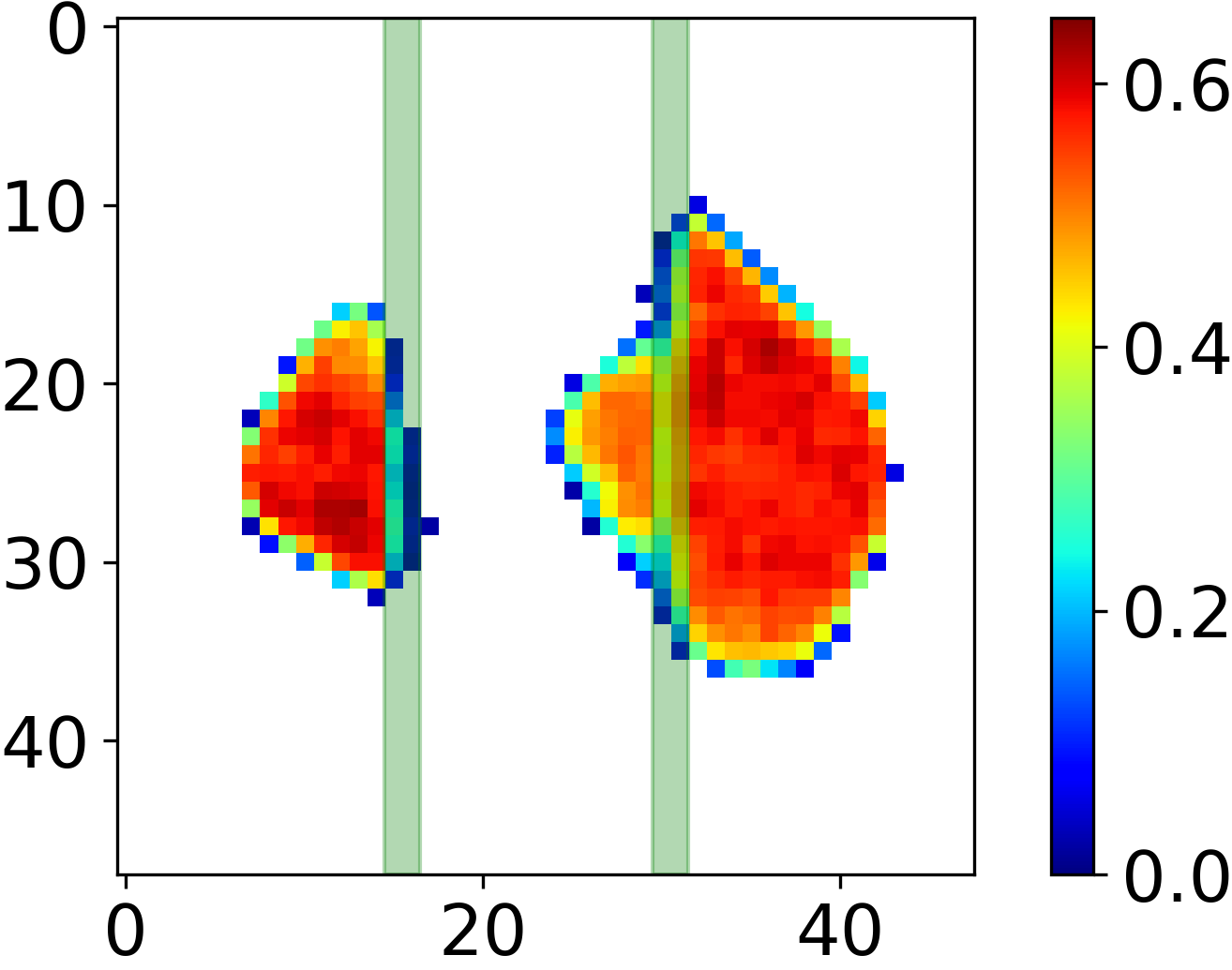}}
\hspace{0.2mm}
\subfloat[Posterior 3]{\includegraphics[width=35mm]{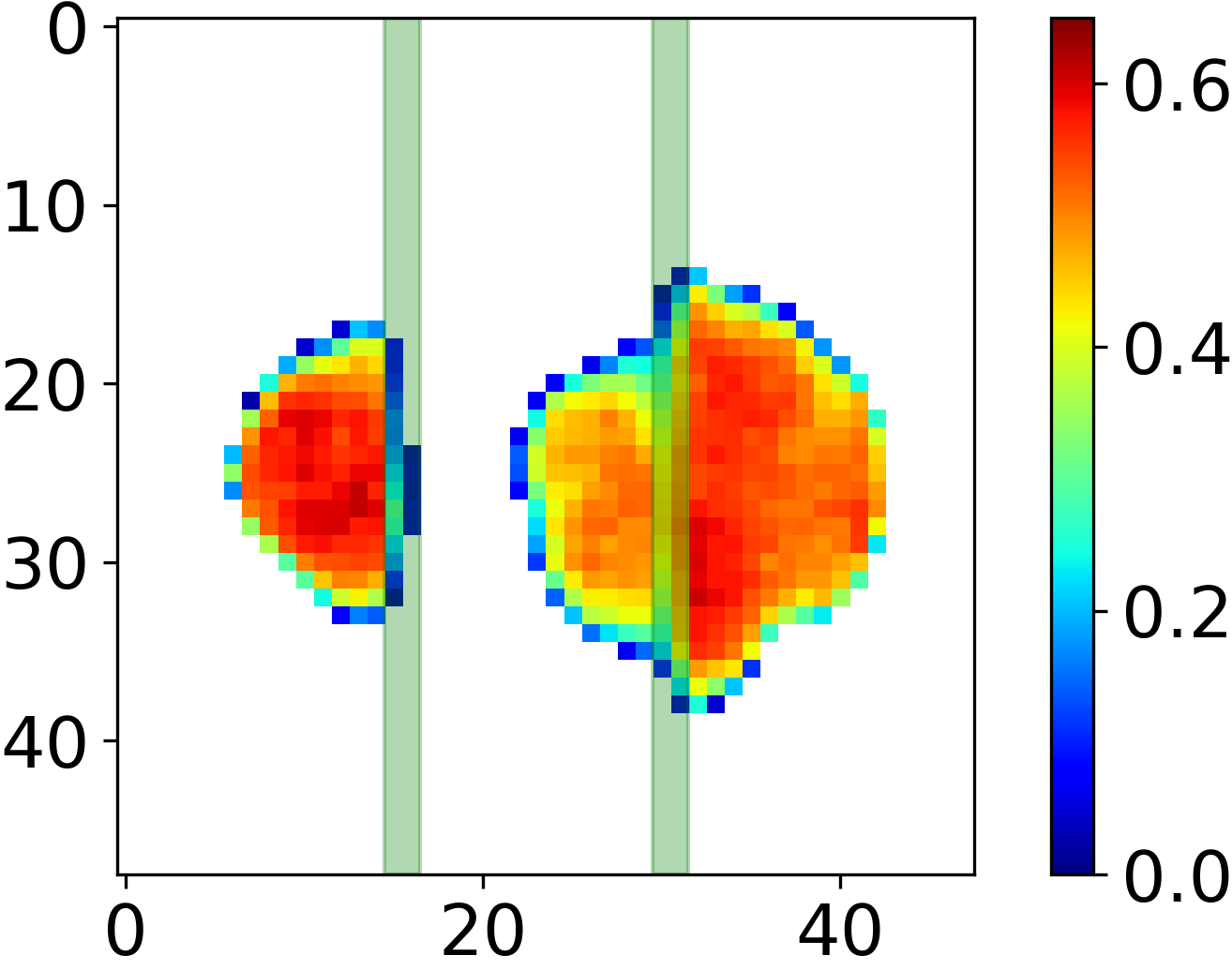}}
\hspace{0.2mm}
\subfloat[Posterior 4]{\includegraphics[width=35mm]{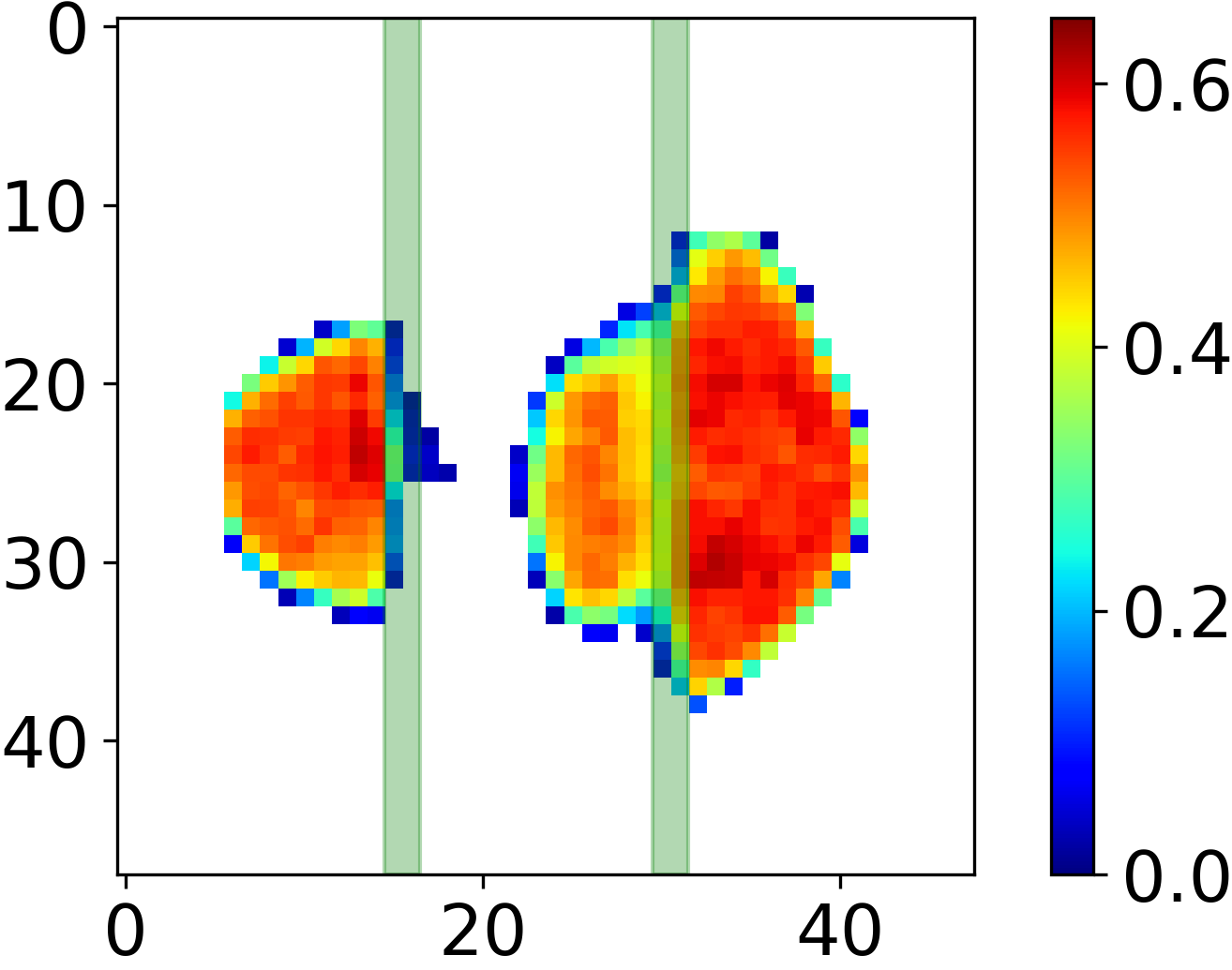}}
\caption{Saturation maps for the top layer of the target aquifer corresponding to the 3D saturation fields in Fig.~\ref{Prior_Posterior:Saturation_Plume_True_2}. True saturation map at 50~years for true model~2 is shown in Fig.~\ref{Prior_Posterior:Saturation_True_2_Top}a.}
\label{Prior_Posterior:Saturation_True_2_Top}
\end{figure}

The true top-layer saturation map, along with maps for the first four posterior saturation fields appearing in Fig.~\ref{Prior_Posterior:Saturation_Plume_True_2}, are displayed in Fig.~\ref{Prior_Posterior:Saturation_True_2_Top}. The prior saturation maps are the same as for true model~1 (these are shown in the upper row of Fig.~16 in the main paper). Consistent with the true model, the posterior maps all show a larger plume toward the right in each case, with significant leakage across the nearby fault. This differs from true model~1 (Fig.~16f in the main paper), where we saw very little leakage across the faults.

\begin{figure}[!ht]
\centering   
\subfloat[CO$_2$ footprint ratio in the target aquifer]{\includegraphics[width = 85mm]{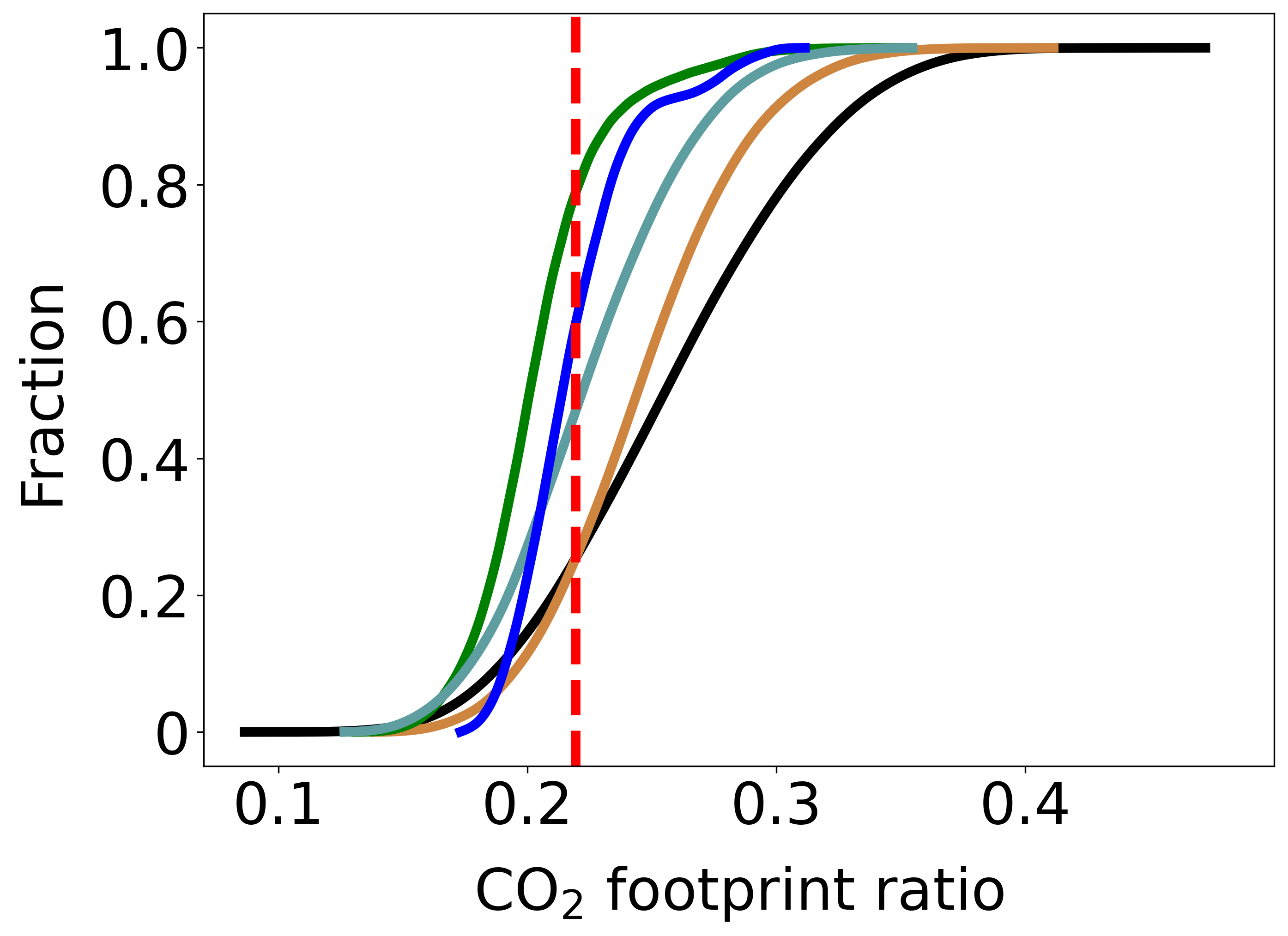}} \\
\subfloat[CO$_2$ leakage volume in the middle aquifer]{\includegraphics[width = 85mm]{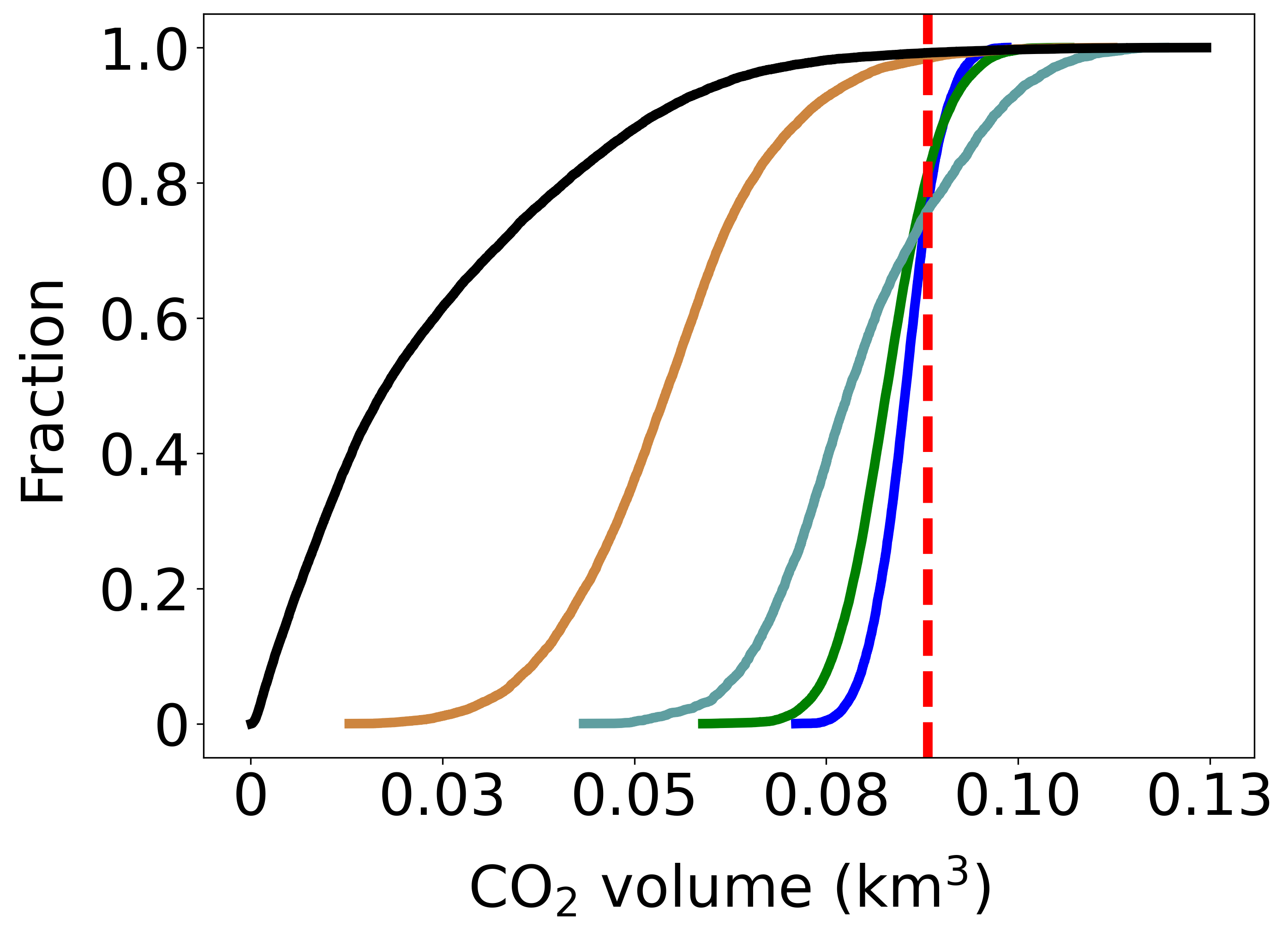}}
\hspace{4mm}
\subfloat[CO$_2$ leakage volume in the upper aquifer]
{\includegraphics[width = 85mm]{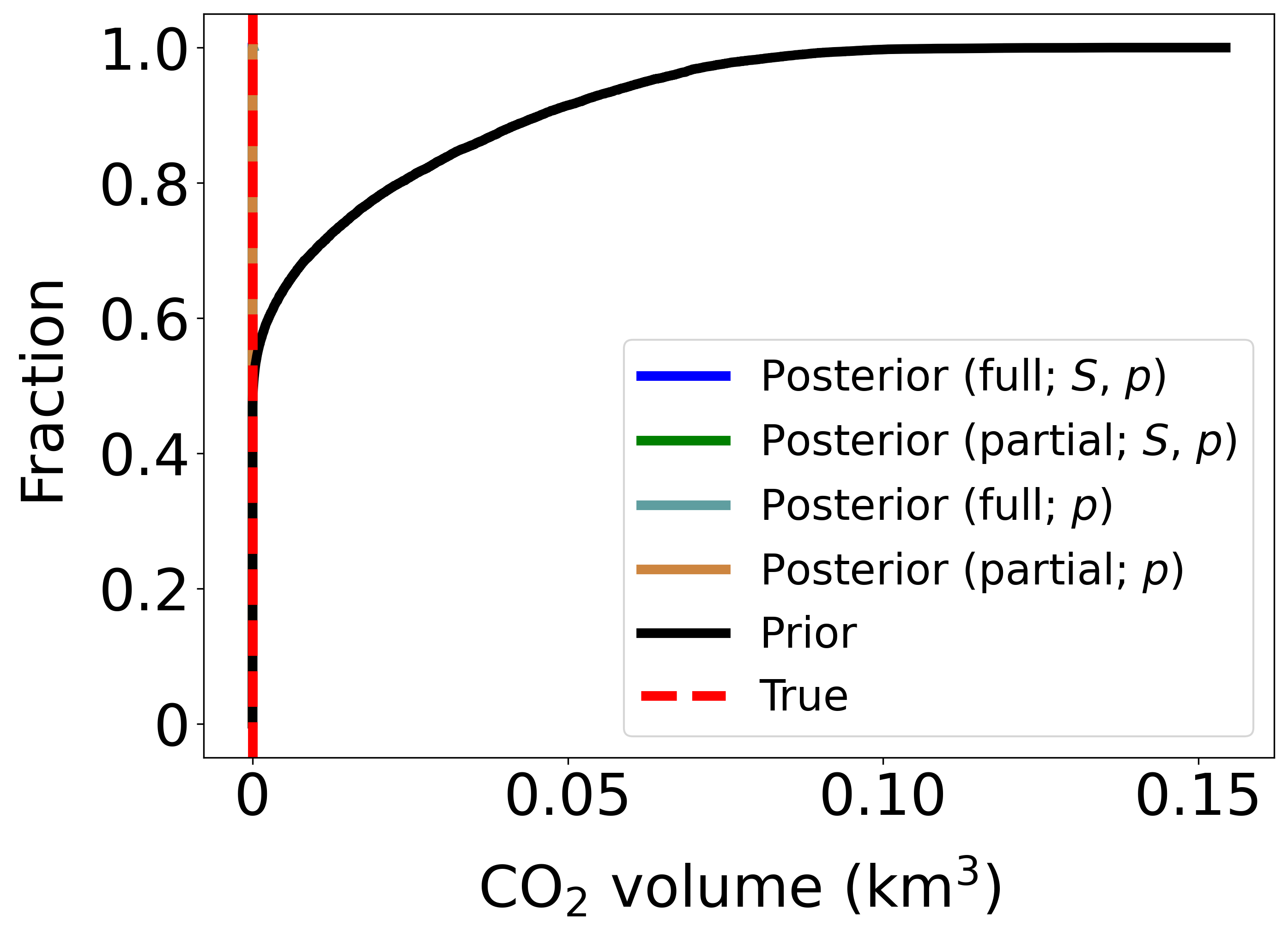}}
\caption{Cumulative density functions for CO$_2$ footprint ratio in the target aquifer and for CO$_2$ leakage volume into the middle and upper aquifers at the end of injection for true model~2. Black curves show the prior distributions and blue, green, cyan, and brown curves correspond to posterior distributions using the four monitoring strategies. Red vertical dashed lines display the true values. Legend in (c) applies to all subplots.}
\label{quantity_of_interest_true_model_2}
\end{figure}

Data assimilation results for footprint ratio and leakage volumes, using the four different monitoring strategies, are shown in Fig.~\ref{quantity_of_interest_true_model_2}. There we observe the narrowest posterior distributions for the full monitoring strategy with both data types (blue CDFs). The partial monitoring strategy with both data types achieves larger uncertainty reduction than the full monitoring strategy using only pressure data for this case. All four monitoring strategies correctly indicate that there is no leakage into the upper aquifer.

%% file: SI_Section_3.tex
\begin{figure}[!ht]
\centering   
\subfloat[Posterior 1]{\includegraphics[width=34mm]{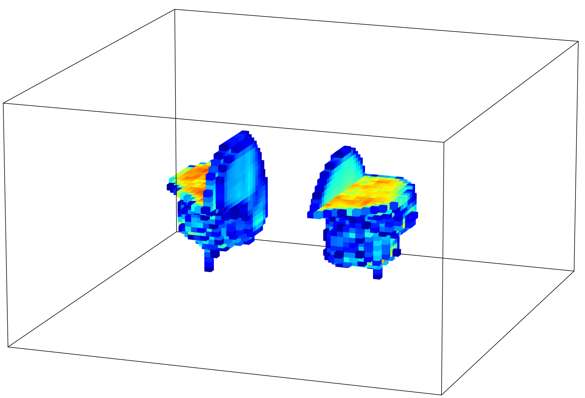}}
\hspace{0.2mm}
\subfloat[Posterior 2]{\includegraphics[width=34mm]{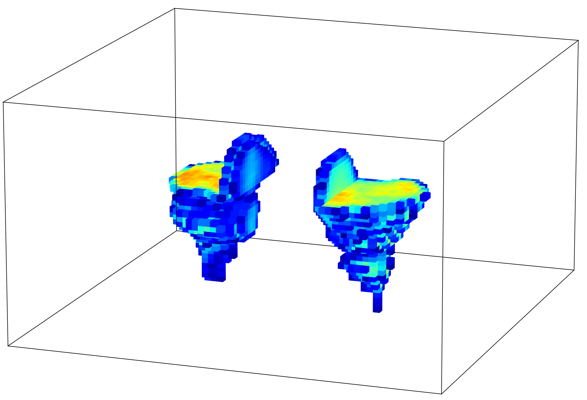}}
\hspace{0.2mm}
\subfloat[Posterior 3]{\includegraphics[width=34mm]{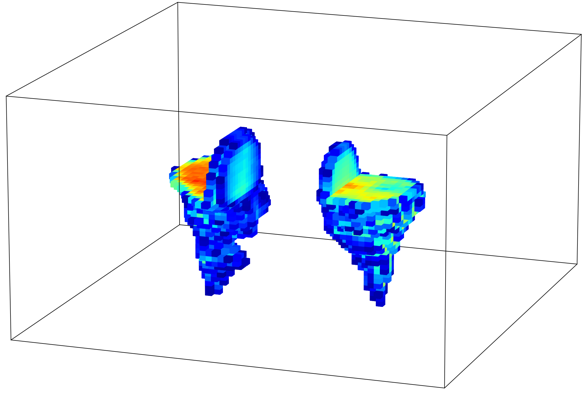}}
\hspace{0.2mm}
\subfloat[Posterior 4]{\includegraphics[width=34mm]{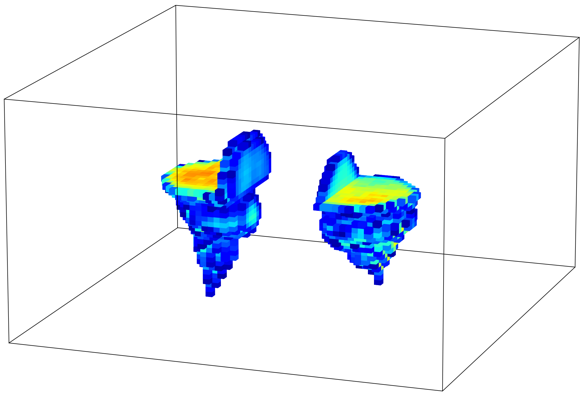}}
\hspace{0.2mm}
\subfloat[Posterior 5]{\includegraphics[width=34mm]{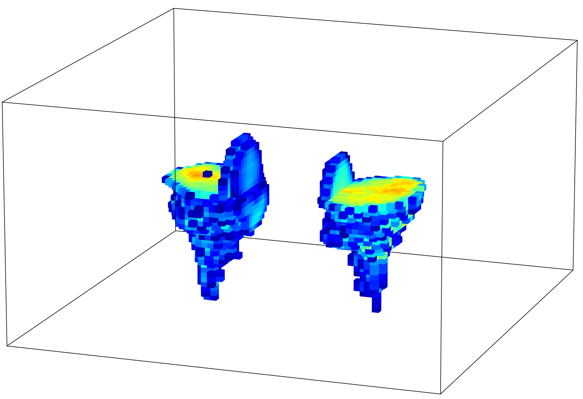}}
\includegraphics[width=5.8mm]{Sw_Scale.png}\\[1ex]
\caption{Representative CO$_2$ saturation fields at 50~years from posterior geomodels for true model~3. True CO$_2$ saturation field at 50~years for true model~3 is shown in Fig.~10c in the main paper.}
\label{Prior_Posterior:Saturation_Plume_True_3}
\end{figure}
\begin{figure}[!ht]
\centering   
\subfloat[True]{\includegraphics[width=35mm]{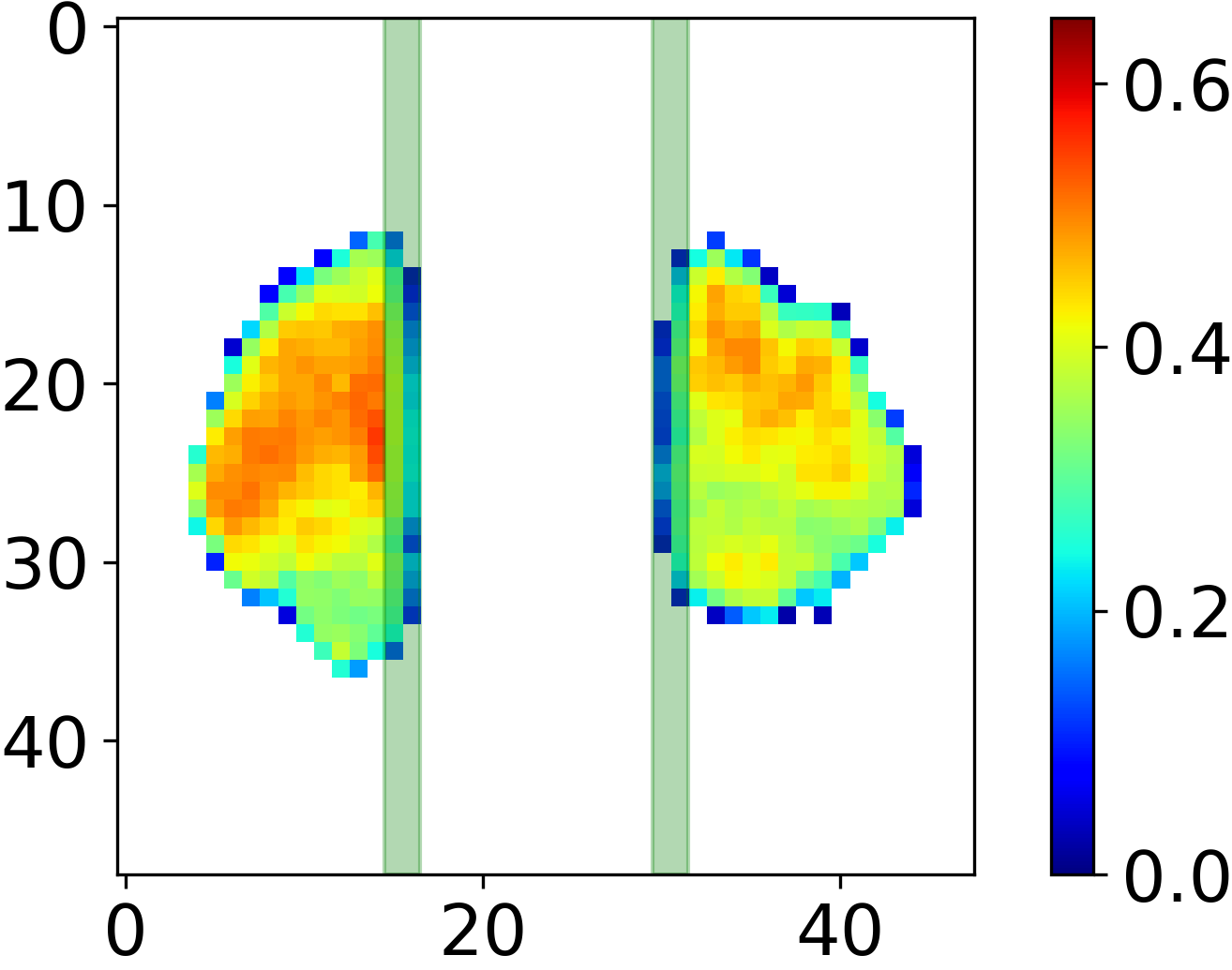}}
\hspace{0.2mm}
\subfloat[Posterior 1]{\includegraphics[width=35mm]{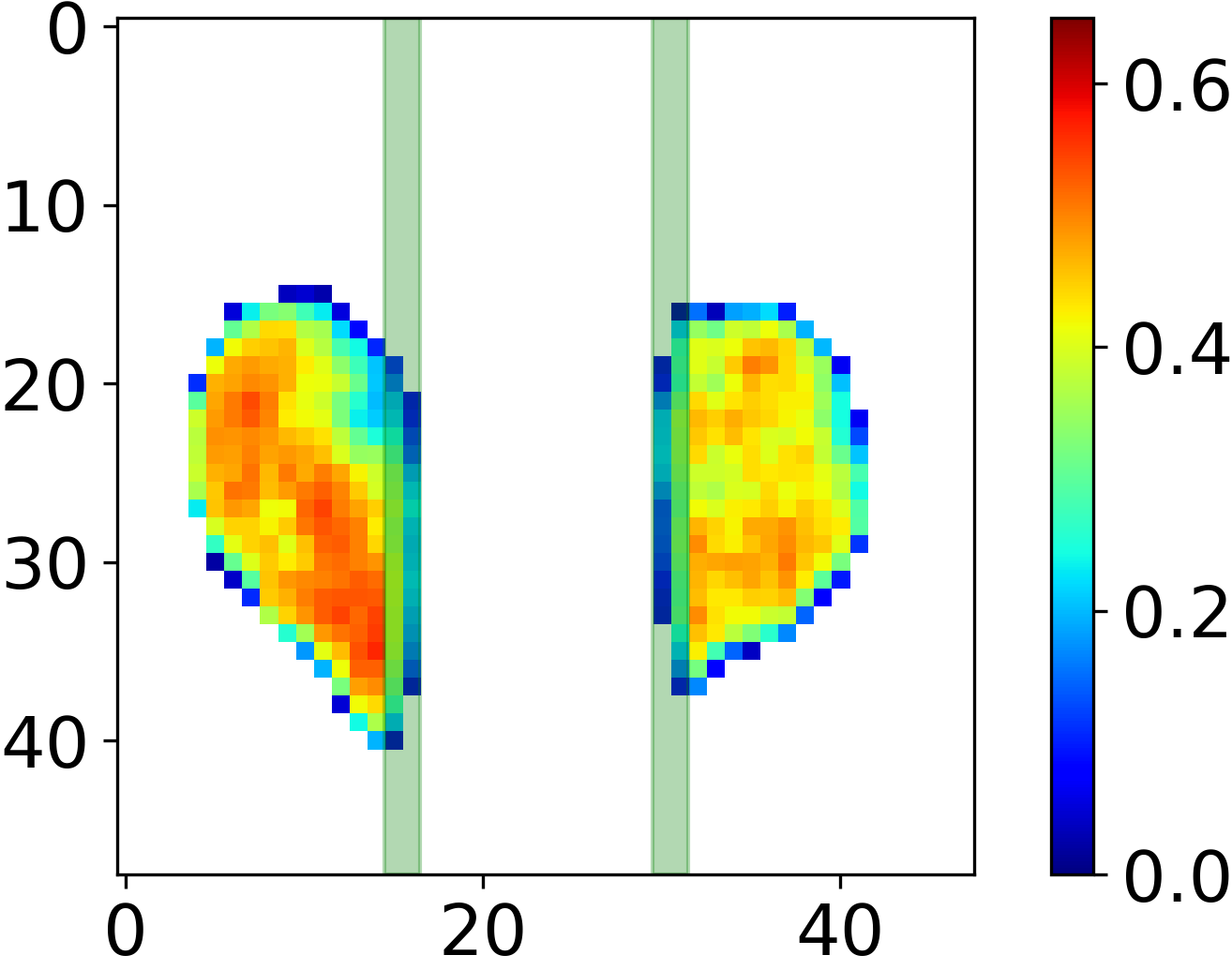}}
\hspace{0.2mm}
\subfloat[Posterior 2]{\includegraphics[width=35mm]{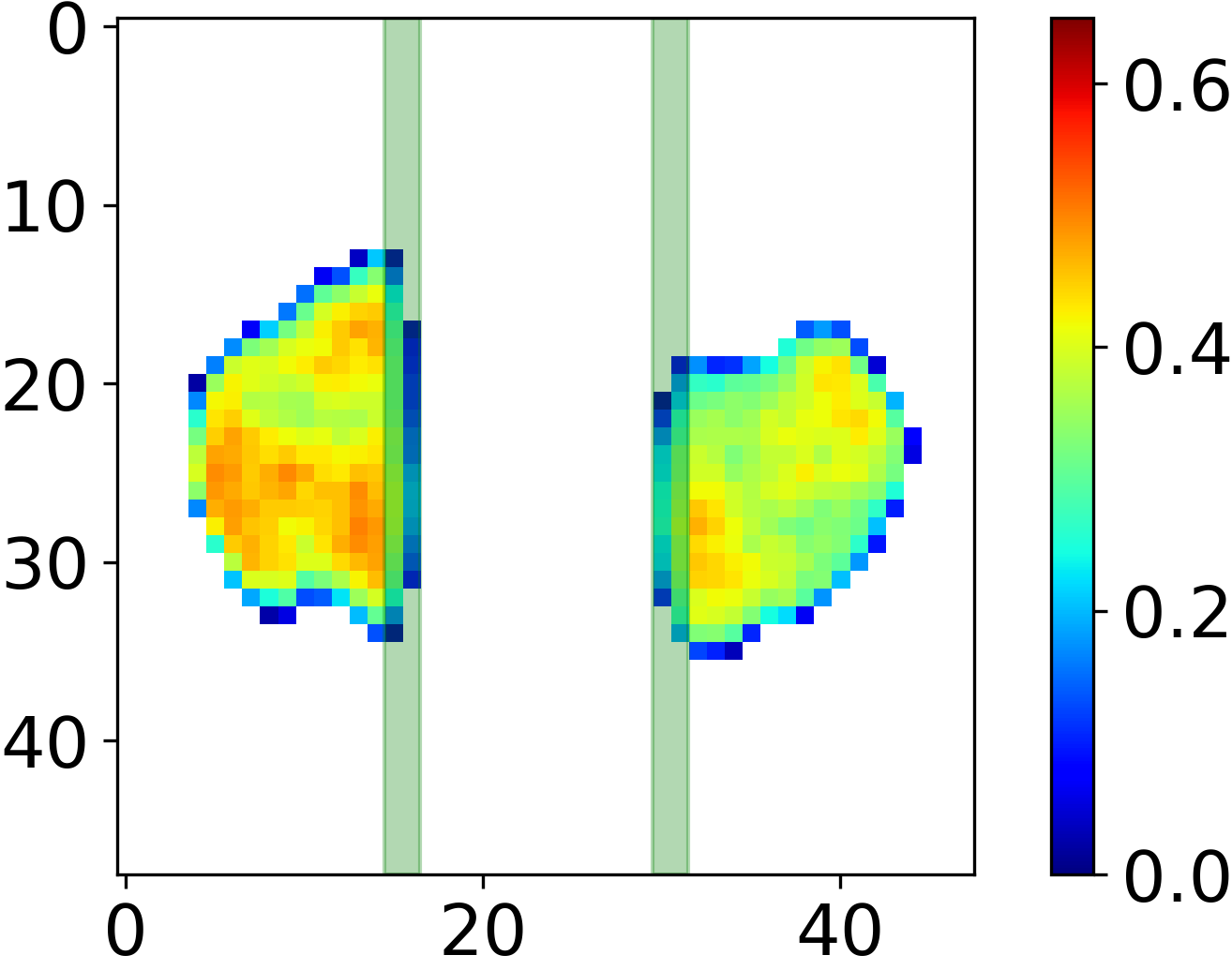}}
\hspace{0.2mm}
\subfloat[Posterior 3]{\includegraphics[width=35mm]{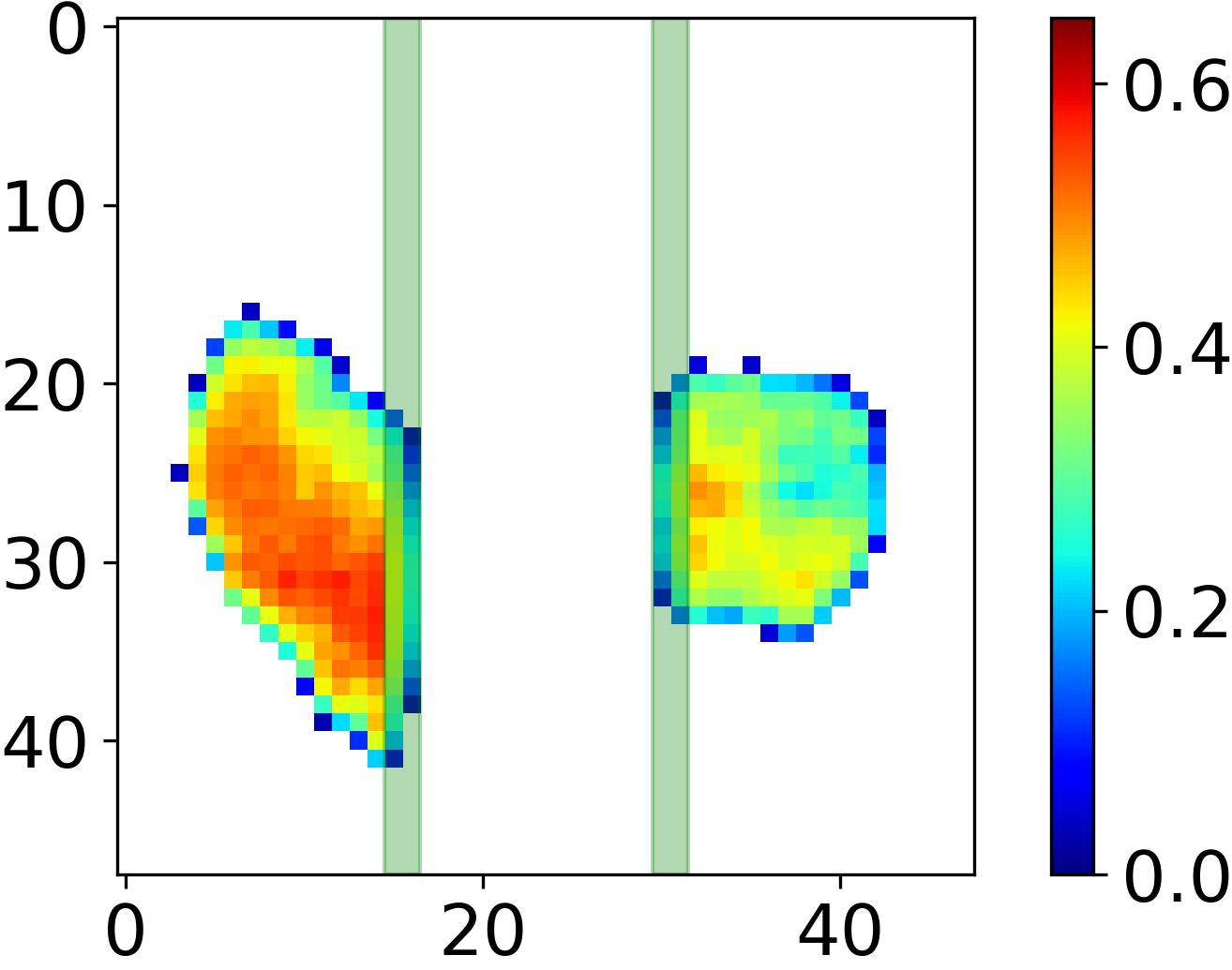}}
\hspace{0.2mm}
\subfloat[Posterior 4]{\includegraphics[width=35mm]{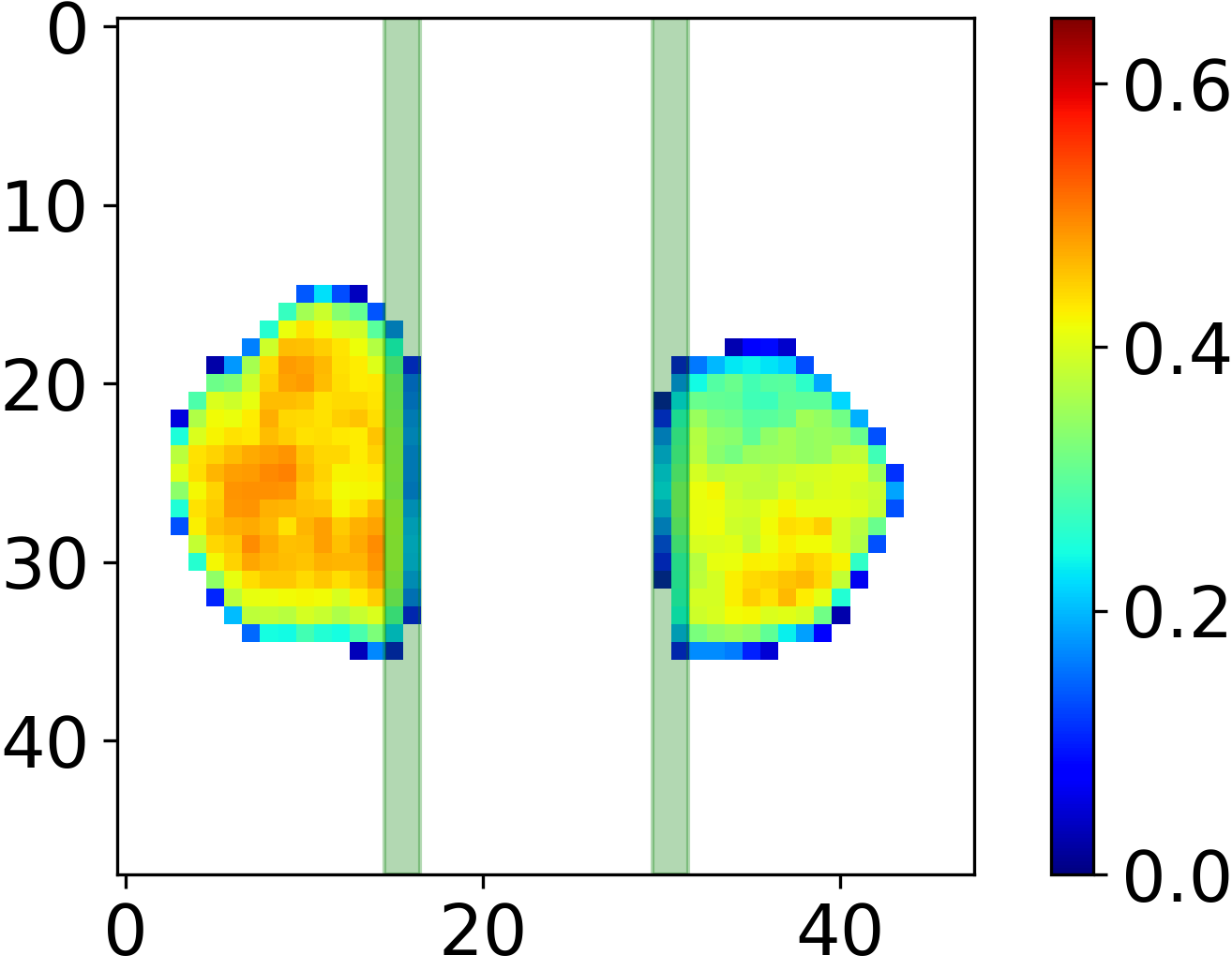}}
\caption{Saturation maps for the top layer of the target aquifer corresponding to the 3D saturation fields in Fig.~\ref{Prior_Posterior:Saturation_Plume_True_3}. True saturation map at 50~years for true model~3 is shown in Fig.~\ref{Prior_Posterior:Saturation_True_3_Top}a.}
\label{Prior_Posterior:Saturation_True_3_Top}
\end{figure}

In true model~3 (realization~6 in the main text), CO$_2$ migrates into the faults, but it does not flow into the middle or upper aquifer over the time frame of the simulation. The hierarchical MCMC procedure requires 343,806 function evaluations to achieve convergence for the full monitoring strategy using both pressure and saturation data. A total of 36,262 sets of metaparameters comprise the posterior samples. The true saturation fields at 20~years and 50~years are shown in Figs.~9c and 10c in the main paper. Representative posterior saturation fields at 50~years for the full monitoring strategy, using both pressure and saturation data, are shown in Fig.~\ref{Prior_Posterior:Saturation_Plume_True_3}. The corresponding saturation maps at the top layer of the target aquifer are presented in Fig.~\ref{Prior_Posterior:Saturation_True_3_Top}. We again observe reduced variability in the posterior results and consistency with the true saturation results.

\begin{figure}[!ht]
\centering   
\subfloat[CO$_2$ footprint ratio in the target aquifer]{\includegraphics[width = 85mm]{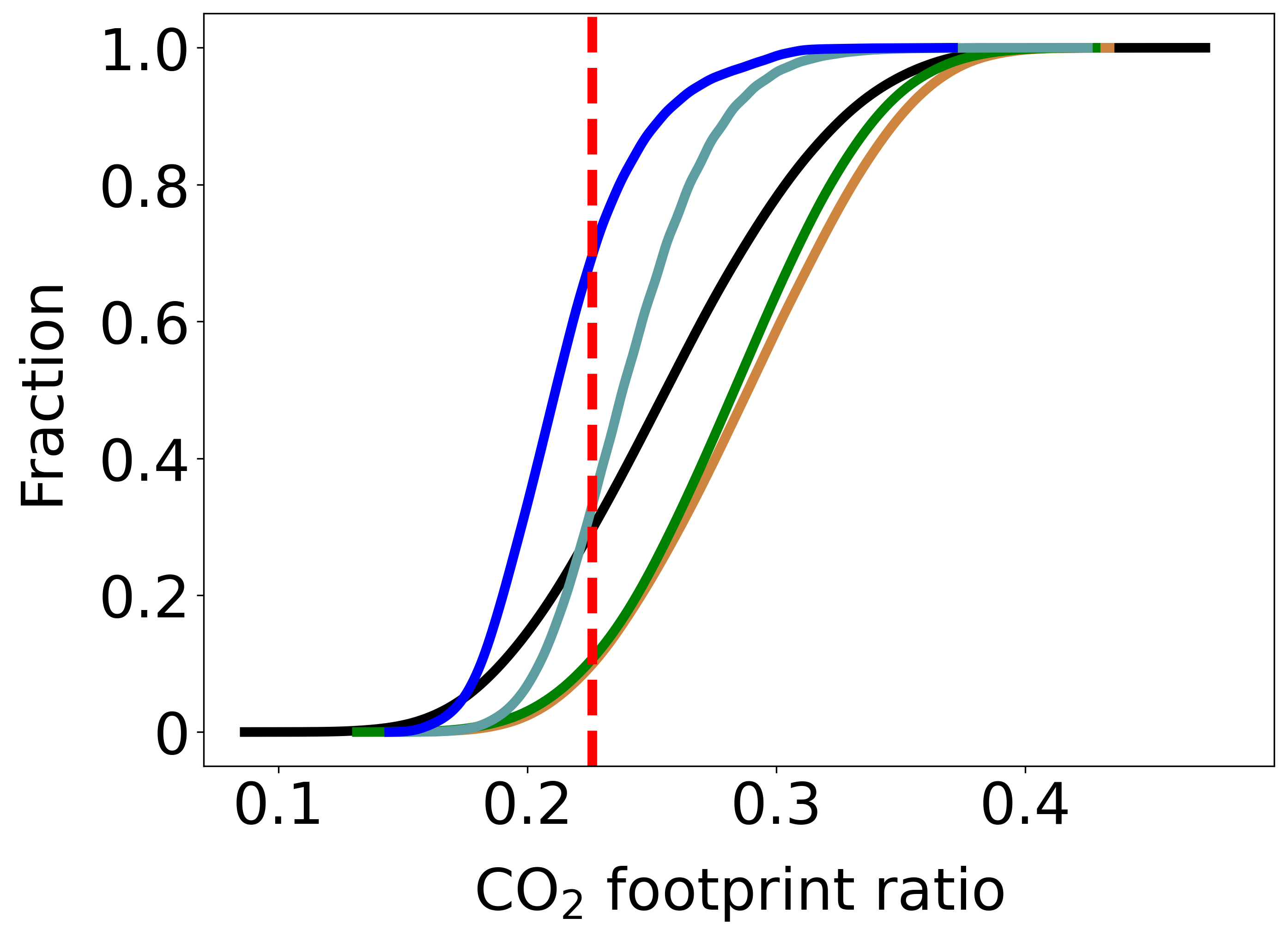}} \\
\subfloat[CO$_2$ leakage volume in the middle aquifer]{\includegraphics[width = 85mm]{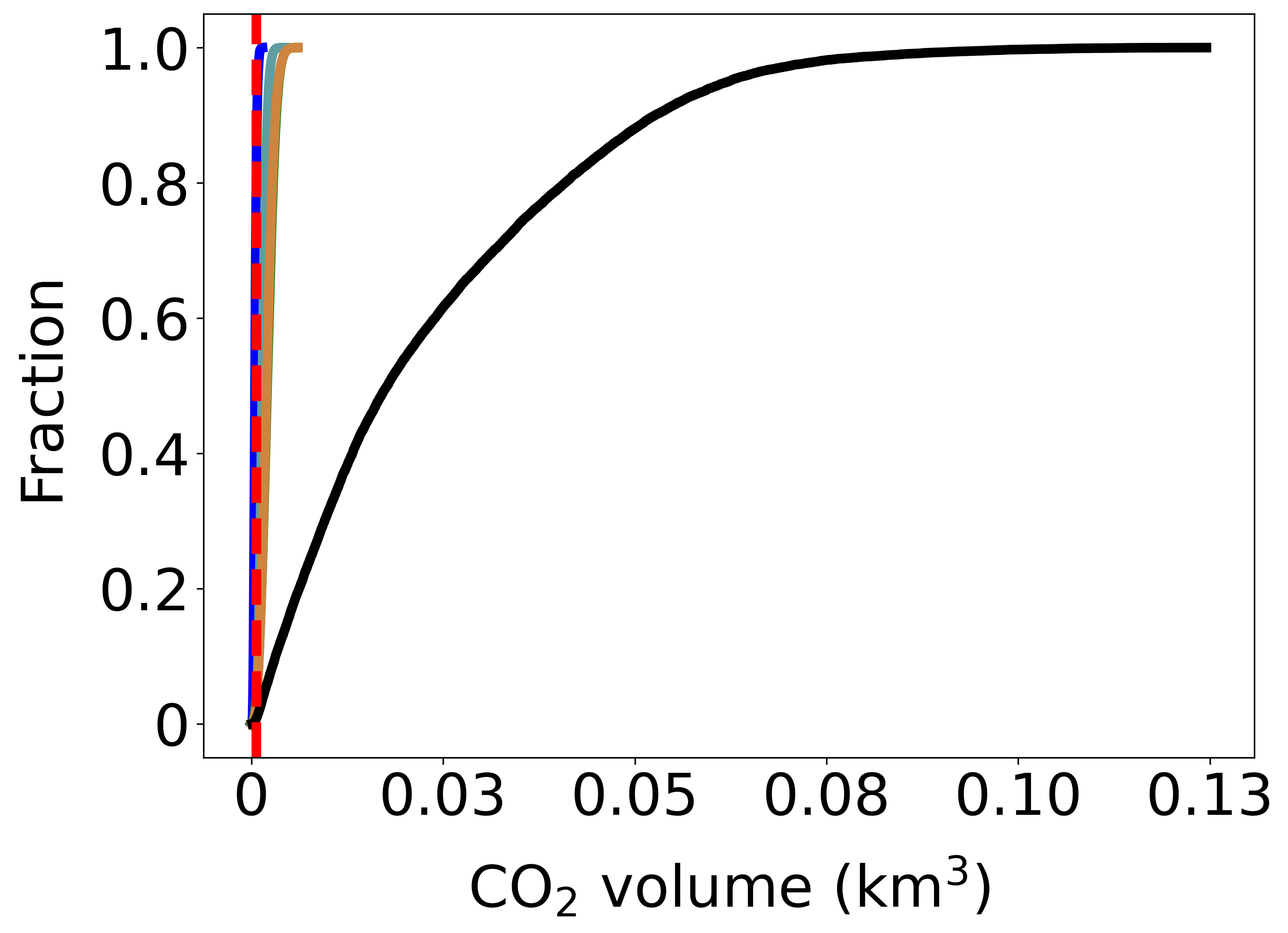}}
\hspace{4mm}
\subfloat[CO$_2$ leakage volume in the upper aquifer]
{\includegraphics[width = 85mm]{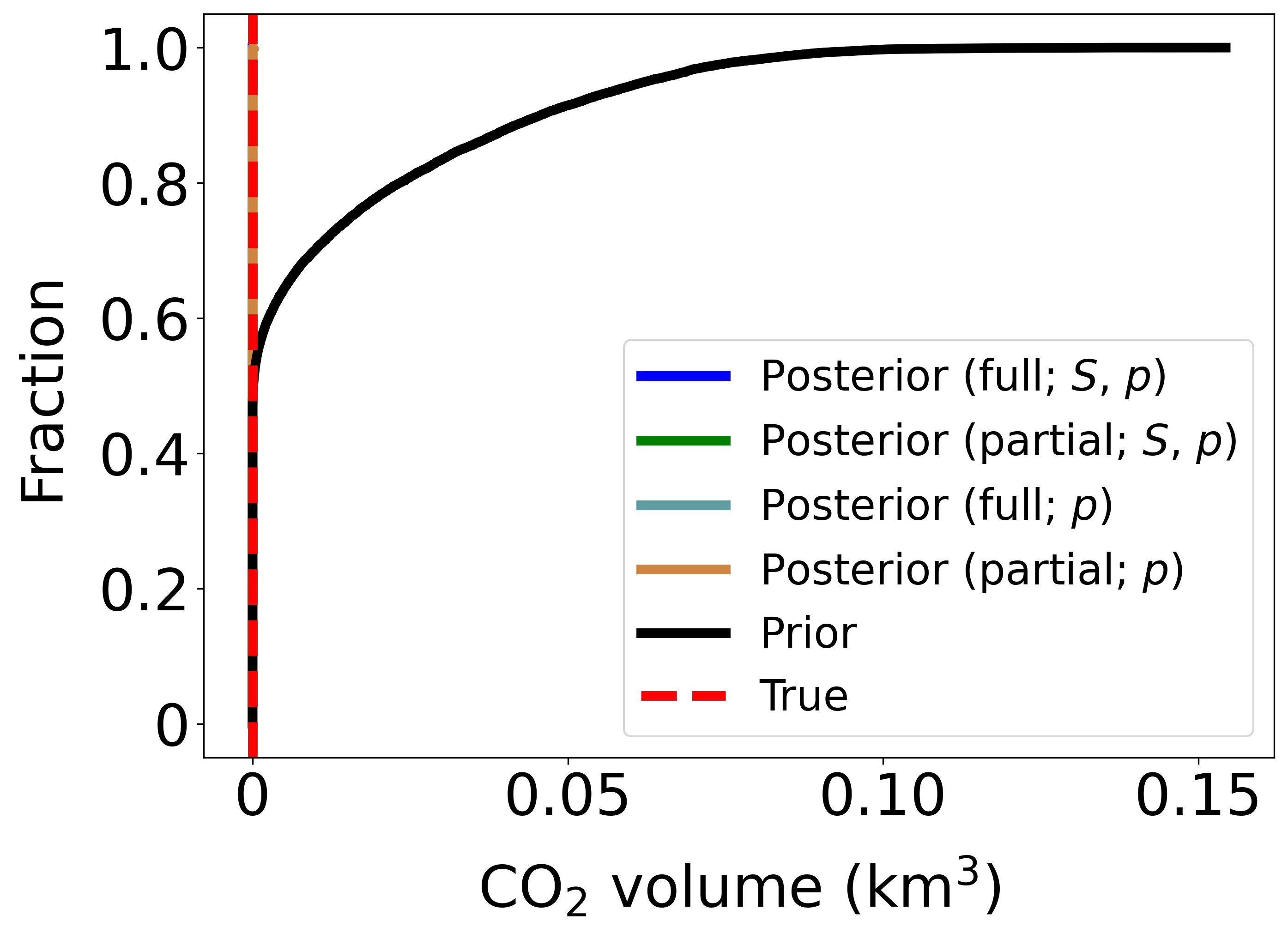}}
\caption{Cumulative density functions for CO$_2$ footprint ratio in the target aquifer and for CO$_2$ leakage volume into the middle and upper aquifers at the end of injection for true model~3. Black curves show the prior distributions and blue, green, cyan, and brown curves correspond to posterior distributions using the four monitoring strategies. Red vertical dashed lines display the true values. Legend in (c) applies to all subplots.}
\label{quantity_of_interest_1_true_model_3}
\end{figure}

Footprint and leakage CDFs for true model~3 are shown in Figs.~\ref{quantity_of_interest_1_true_model_3}. For this case the partial monitoring strategies do not provide a large amount of uncertainty reduction in CO$_2$ footprint ratio in the target aquifer. This is likely because, for true model~3, CO$_2$ does not migrate into the middle and upper aquifers. We see that all monitoring strategies correctly predict zero (or very little) leakage into both the middle and upper aquifers.

\begin{figure}[H]
\centering   
\subfloat[Pressure difference across fault~1]{\includegraphics[width = 85mm]{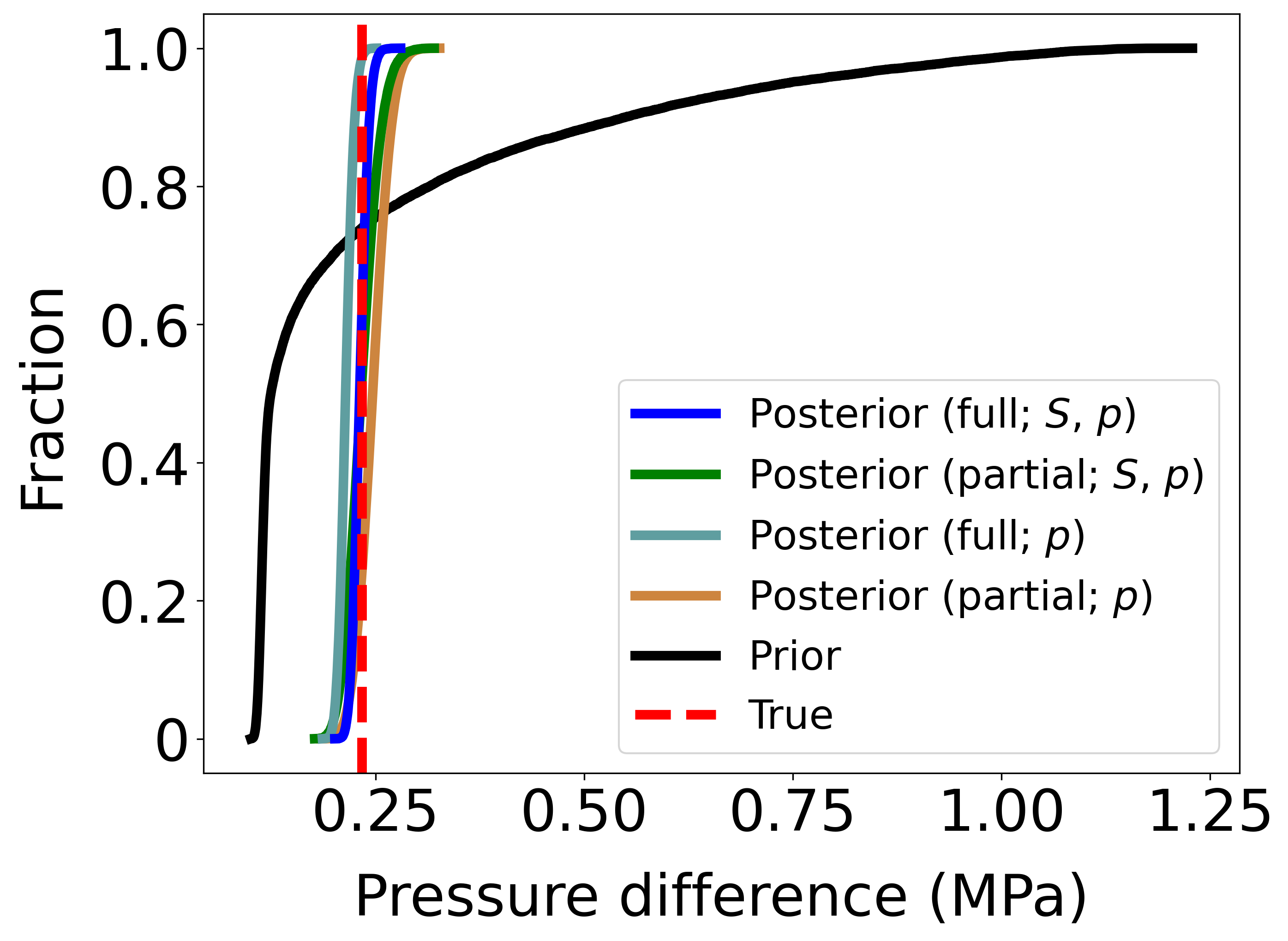}}
\hspace{4mm}
\subfloat[Pressure difference across fault~2]
{\includegraphics[width = 85mm]{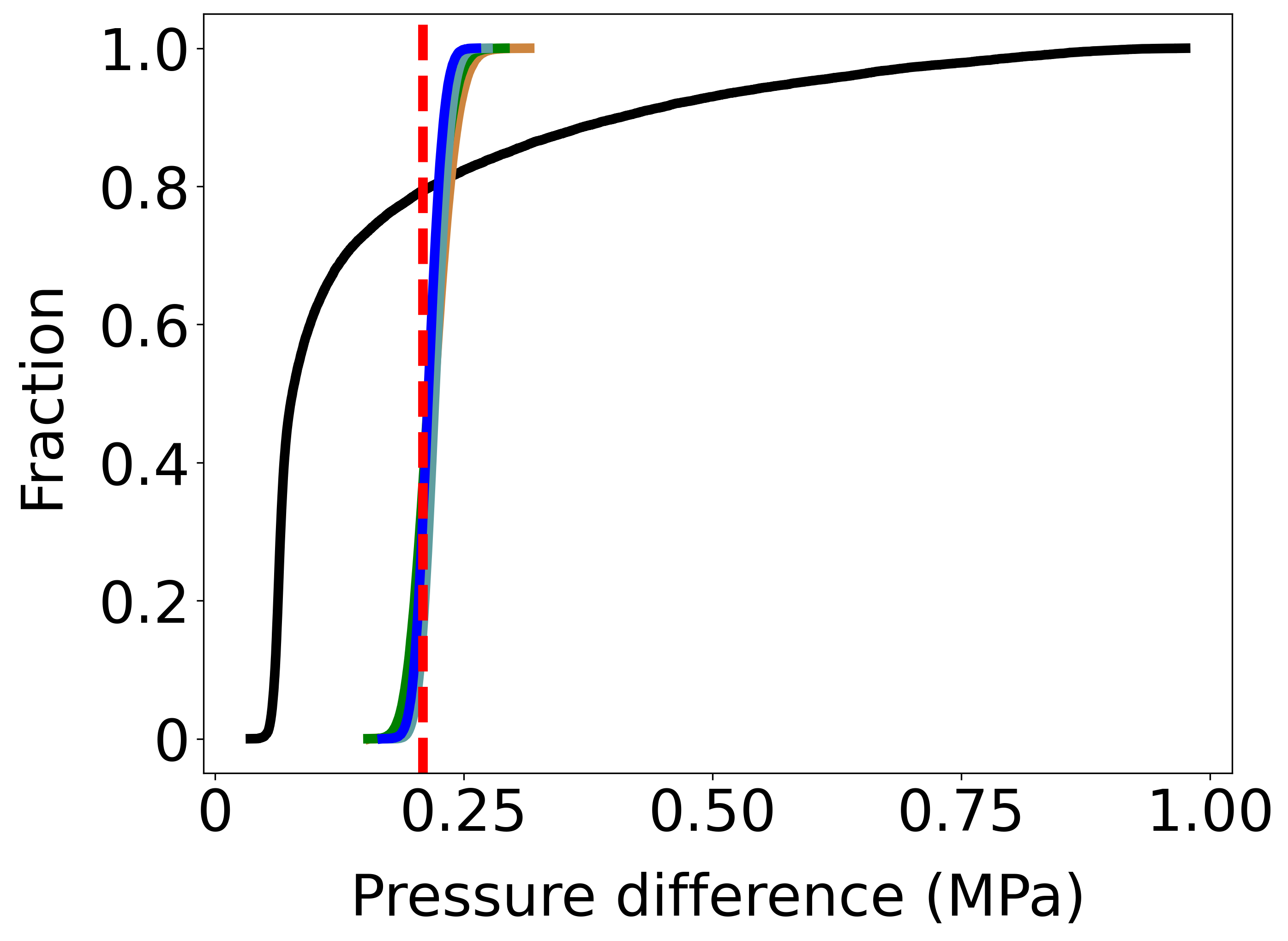}}
\caption{Cumulative density functions for average pressure difference across the two faults in the target aquifer region at the end of injection for true model~3. Black curves show the prior distributions and blue, green, cyan, and brown curves correspond to posterior distributions using the four monitoring strategies. Red vertical dashed lines display the true values. Legend in (a) applies to both subplots.}
\label{quantity_of_interest_2_true_model_3}
\end{figure}

For true model~3 there is a noticeable pressure effect in the vicinity of the faults due to the low fault permeabilities. This is evident in Fig.~11c in the main paper. To assess whether this effect can be predicted for this case, we construct prior and posterior CDFs for the average pressure difference across each fault in the target aquifer region at 50~years. These results, for the four different monitoring strategies, are shown in Fig.~\ref{quantity_of_interest_2_true_model_3}. It is noteworthy that, even though 50\% or more of the prior models show little or no pressure difference across each fault, the pressure difference in true model~3 is captured by all four monitoring strategies. 

